\documentclass[numbers]{article}

\usepackage[preprint]{neurips_2025}

\usepackage{natbib} 

\usepackage[utf8]{inputenc} % allow utf-8 input
\usepackage[T1]{fontenc}    % use 8-bit T1 fonts
\usepackage{hyperref}       % hyperlinks
\usepackage{url}            % simple URL typesetting
\usepackage{booktabs}       % professional-quality tables
\usepackage{amsfonts}       % blackboard math symbols
\usepackage{nicefrac}       % compact symbols for 1/2, etc.
\usepackage{microtype}      % microtypography
\usepackage{times}
\usepackage{latexsym}
\usepackage[table,xcdraw]{xcolor}
\usepackage{cuted}

\usepackage{makecell}
\usepackage{multirow}
\usepackage{hyperref}
\usepackage{url}
\usepackage{subcaption}
\usepackage{algorithm}
\usepackage{algorithmic}
\usepackage{amssymb}
\usepackage[most]{tcolorbox}
\usepackage{pifont}
\usepackage{color}
\usepackage{algorithm}
\usepackage{amsmath}
\usepackage{algorithmic}
\usepackage{makecell}
\usepackage{newfloat}
\usepackage{listings}
\usepackage{afterpage}
\usepackage{colortbl} 
\usepackage{wrapfig}
\usepackage{soul}  
\usepackage{threeparttable}
\usepackage{array}    
\usepackage{ragged2e}  
\usepackage{longtable}

\newcommand{\ie}{\textit{i}.\textit{e}.}
\newcommand{\eg}{\textit{e}.\textit{g}.}

\usepackage{rotating}

\definecolor{c1}{HTML}{003371}
\definecolor{c2}{HTML}{057748}

\tcbuselibrary{breakable}
\NewDocumentCommand{\cdp}
{ mO{} }{\textcolor{blue}{\textsuperscript{\textit{Dong ping}}\textsf{\textbf{\small[#1]}}}}
\NewDocumentCommand{\yinuo}
{ mO{} }{\textcolor{blue}{\textsuperscript{\textit{Yi Nuo}}\textsf{\textbf{\small[#1]}}}}

\title{When LLMs Learn to be Students: The SOEI Framework for Modeling and Evaluating Virtual Student Agents in Educational Interaction}

\author{\textbf{Yiping Ma}$^{1,2*}$\hspace{9pt} 
\textbf{Shiyu Hu}$^{3*}$\hspace{9pt} 
\textbf{Xuchen Li}$^{4,5}$ \hspace{9pt}
\textbf{Yipei Wang}$^{6}$ \hspace{9pt}
\textbf{Yuqing Chen}$^{1,2}$ \hspace{9pt}\\
\textbf{Shiqing Liu}$^{7\dagger}$ \hspace{9pt}
\textbf{Kang Hao Cheong}$^{3,8\dagger}$ \hspace{9pt}\\
($*$ Equal Contributions \quad $\dagger$ Corresponding Authors)\\
\textsuperscript{1}Lab of Artificial Intelligence for Education, East China Normal University\\
\textsuperscript{2}School of Computer Science and Technology, East China Normal University\\
\textsuperscript{3}School of Physical and Mathematical Sciences, Nanyang Technological University\\
\textsuperscript{4}Institute of Automation, Chinese Academy of Sciences\\
\textsuperscript{5}School of Artificial Intelligence, University of Chinese Academy of Sciences\\
\textsuperscript{6}Institute of Automation, Southeast University\\
\textsuperscript{7}Department of Education, East China Normal University\\
\textsuperscript{8}College of Computing and Data Science, Nanyang Technological University\\
\tt\small 52275901020@stu.ecnu.edu.cn \quad shiyu.hu@ntu.edu.sg \\
\tt\small lixuchen2024@ia.ac.cn \quad  230248984@seu.edu.cn \quad 52265901026@stu.ecnu.edu.cn\\
\tt\small sqliu@dedu.ecnu.edu.cn  \quad kanghao.cheong@ntu.edu.sg
}

\begin{document}

\maketitle

\begin{abstract}
Recent advances in large language models (LLMs) have enabled intelligent tutoring systems, yet the development of LLM-based Virtual Student Agents (LVSAs)—LLM-based agents designed to emulate student-like behavior—remains underexplored. Such agents are essential for teacher-facing applications, where simulating diverse learner traits can support adaptive instruction and pedagogical skill development. However, current methods lack principled personality modeling, scalable evaluation of behavioral consistency, and empirical validation in interactive teaching settings.
We propose the \textbf{SOEI} framework, a structured pipeline comprising \textbf{S}cene, \textbf{O}bject, \textbf{E}valuation, and \textbf{I}nteraction, for constructing and evaluating personality-aligned LVSAs in classroom scenarios. Leveraging Chinese language instruction as a cognitively and emotionally rich testbed, we generate five LVSAs based on Big Five traits through LoRA fine-tuning and expert-informed prompt design. Their behavioral realism and personality coherence are assessed using hybrid human–GPT-4 evaluation and a multi-dimensional annotation protocol.
Through controlled experiments with real pre-service teachers, we demonstrate that LVSAs can elicit adaptive teaching strategies and maintain trait-consistent behavior across multi-turn dialogues. Our results provide:
(1) an educationally and psychologically grounded generation pipeline for LLM-based student agents;
(2) a hybrid, scalable evaluation framework for behavioral realism; and
(3) empirical insights into the pedagogical utility of LVSAs in shaping instructional adaptation.
By embedding LVSAs into both generative modeling and human-in-the-loop teaching, SOEI bridges AI for Education (AI4Edu) and Education for AI (Edu4AI), positioning classroom interaction as a rigorous testbed for controllability, personality alignment, and human-likeness in large language models.
\end{abstract}

\section{Introduction}

\begin{figure*}[t]
\centering
  \includegraphics[width=\textwidth]{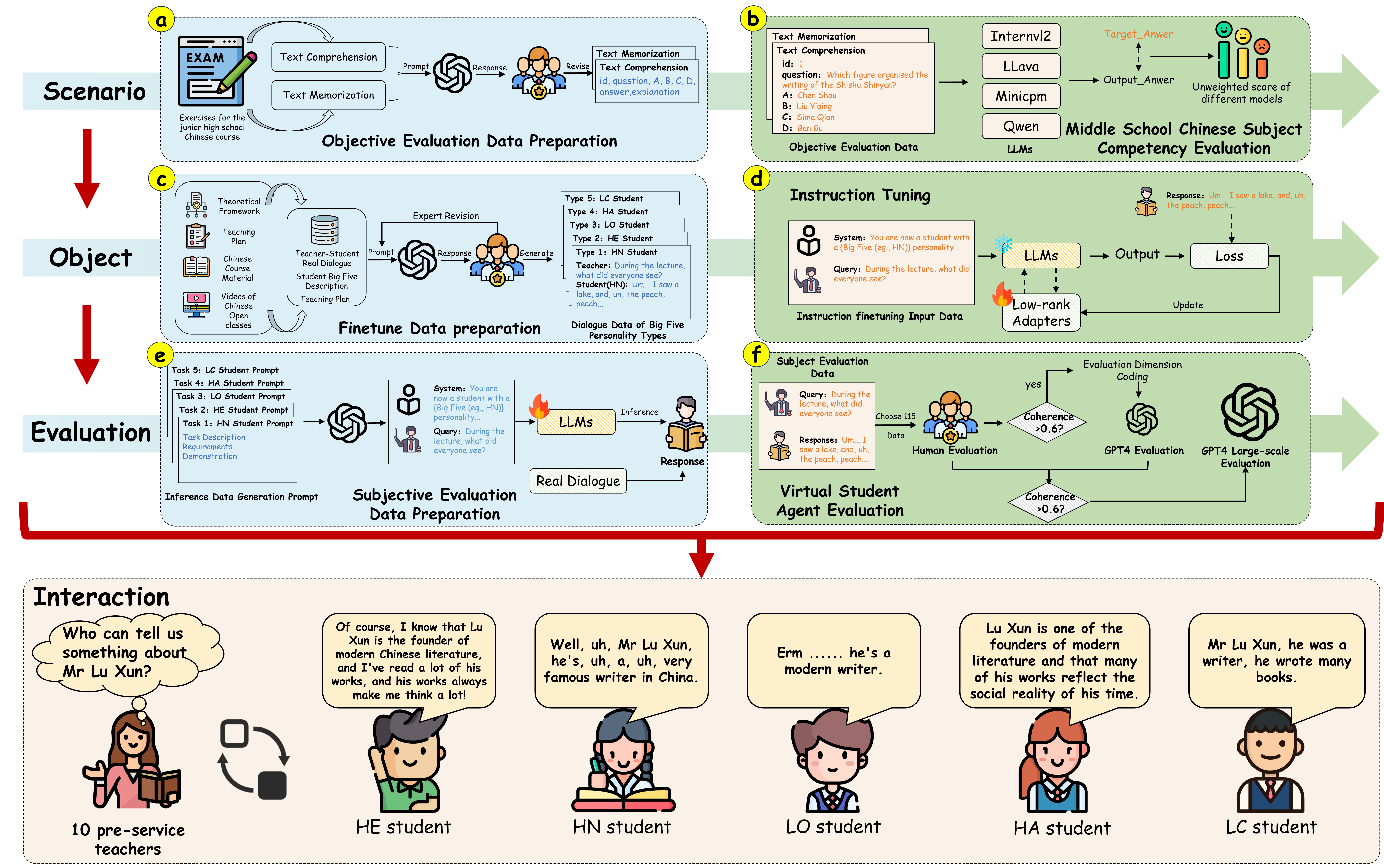}
    \vspace{-5pt}
  \caption{
  Overview of the SOEI pipeline, which supports the structured modeling and evaluation of virtual student agents. 
  The framework consists of four modules—\textbf{Scene}, \textbf{Object}, \textbf{Evaluation}, and \textbf{Interaction}—each contributing to data preparation, agent construction, behavior assessment, and dialogue-based validation. 
  HE, HN, LO, HA, and LC correspond to High Extraversion, High Neuroticism, Low Openness, High Agreeableness, and Low Conscientiousness, respectively.
  }
  \label{fig:overview}
    \vspace{-10pt}
\end{figure*}

Recent advances in large language models (LLMs) have propelled educational AI, particularly in developing virtual tutors that support personalized learning, content generation, and automated feedback~\cite{ge2024openagi,liu2024make,hong2024knowledge,pavic2024conceptual,yasmin2022ai}. In contrast, the design of LLM-based virtual student agents (LVSAs)—essential for teacher-facing applications—remains underexplored.

To simulate authentic student behavior, LVSAs must exhibit cognitive diversity and behavioral variability~\cite{dotger2015exploring,shaughnessy2018uncovering,lugrin2019towards}. However, existing methods primarily rely on rule-based scripting or prompt engineering~\cite{zhang2024simulating,lee23generative,markel2023gpteach}, lacking grounding in educational psychology~\cite{piaget1952origins,nippold1993developmental,costa1992neo}, scalable evaluation strategies, and empirical classroom validation~\cite{christensen2011simschool,delamarre2021interactive,badiee2015design}.

To address these limitations, we propose the \textbf{SOEI framework}—a structured and verifiable pipeline for building personality-conditioned LVSAs. As illustrated in Fig.~\ref{fig:overview}, it comprises four modules: Scene, which defines cognitively rich tasks grounded in junior high Chinese instruction (Sec.~\ref{sec:scene}); Object (Sec.~\ref{sec:object}), which constructs LVSAs based on Big Five traits via LoRA fine-tuning and in-context learning; Evaluation (Sec.~\ref{sec:evaluation}), which combines expert annotation with large-scale GPT-4 scoring; and Interaction (Sec.~\ref{sec:interaction}), which assesses the stability of personality traits and their impact on teacher strategy across multi-turn dialogues.

SOEI enables closed-loop validation of personality-driven agent behaviors. In Turing-style evaluations, fine-tuned LVSAs are rated by both human experts and GPT-4 as behaviorally indistinguishable from real students, with consistent trait expression across five personality dimensions. Interaction experiments further demonstrate that these traits persist throughout 50+ dialogue turns, validating the controllability and stability of personality-conditioned generation in naturalistic teaching scenarios.

Our contributions are threefold:
(1) a unified framework integrating generation and interactive validation of personality-aligned LVSAs;
(2) a scalable behavioral evaluation protocol combining human and GPT-based scoring;
(3) empirical evidence that LVSAs exhibit stable trait-specific behaviors and induce pedagogical adaptation in authentic teaching interactions.

By embedding LVSAs into generative modeling and human-in-the-loop teaching, SOEI bridges controllability and instructional effectiveness, contributing to both AI4Edu and Edu4AI agendas.

\section{Related Work}

The development of virtual student agents spans three intersecting areas: teacher-facing simulation systems, personality-driven generation, and interactive modeling of instructional adaptation. However, these efforts remain fragmented, lacking a unified framework that enables end-to-end construction and evaluation of pedagogically grounded student agents.

\textbf{Virtual student systems for teacher training.}
Early approaches employed role-playing simulations and scripted models~\cite{shapira2015simulation,dalgarno2016practising}, offering realism but limited scalability. Rule-based virtual students~\cite{delamarre2021interactive,kelleci2021using} improved automation but lacked flexibility. Recent LLM-based agents~\cite{achiam2023gpt,li2024leveraging} offer greater diversity, yet rely heavily on prompt engineering and struggle to express cognitively grounded student traits. Systems like Evelyn AI~\cite{bhowmik2024evaluation} and SimClass~\cite{zhang2024simulating} support multi-turn, persona-conditioned interactions, but their evaluations are often constrained to scripted tasks or subjective surveys.

\textbf{Personality-driven agent modeling.}
The Big Five personality theory~\cite{costa1992neo} provides a foundation for simulating behavioral variation in educational dialogue~\cite{piaget1952origins,nippold1993developmental}. While recent methods adjust surface-level linguistic cues~\cite{zhang2024simulating,lee23generative}, few adopt structured modeling of cognitive-affective traits. Maintaining persona fidelity across multi-turn dialogue remains a challenge~\cite{molchanova2025exploring,wang2025evaluating}, particularly in emotionally sensitive contexts. Advances such as persona-aware contrastive learning~\cite{ji2025enhancing} and psychologically scaffolded prompting~\cite{joshi2025improving} improve consistency and interpretability without fine-tuning.

\textbf{Modeling teacher adaptation in human–AI interaction.}
Compared to student-facing applications, teacher adaptation modeling remains underexplored~\cite{wang2024survey,lee23generative}. Some tools~\cite{Duolingo,Khan,mathGPT} emphasize content delivery rather than responsive instruction. Although immersive platforms exist~\cite{kelleci2021using,garcia2024review,yue2024mathvc}, few support fine-grained modeling of teacher–student dynamics. Recent work explores AI-augmented reflective practices~\cite{zheng2025cognitive} and culturally responsive content adaptation~\cite{trajkovski2025ai}, though typically under limited or isolated validation settings.

\textbf{Summary.}
While prior research has advanced individual components—virtual students, personality conditioning, and classroom simulation—it lacks integration into a cohesive, interaction-centered pipeline. To bridge this gap, we propose the \textbf{SOEI framework}, which unifies generation, evaluation, and interaction to support teacher-facing, empirically grounded educational agents. See App.~\ref{sec:more-relate-work} for extended references.

\section{Scene}
\label{sec:scene}

Modeling virtual students requires a pedagogically grounded environment that enables cognitively and linguistically rich behavior simulation. While prior work has focused on structured tasks such as math problem-solving~\cite{satpute2024can, wu2024mathchat}, it often neglects open-ended classroom settings—crucial for capturing natural language use, emotional expressiveness, and learner variability.
We select \textbf{junior high school Chinese instruction} as the target domain, given its strong alignment with personality-grounded language generation. As discussed in App.~\ref{subsubsec:unique-value}, this scenario offers: (1) expressive linguistic outputs for modeling stylistic variation, (2) cognitively and emotionally salient stimuli aligned with adolescent traits, and (3) a staged instructional structure supporting prompt consistency and modular composition. Although we focus on Chinese in this study, the framework is extensible to other structured subjects such as mathematics and history.

\subsection{Task Definition and Scenario Instantiation}

To enable consistent prompt generation across the SOEI pipeline, we formalize each instructional task unit as a structured tuple:
$\mathcal{T} = \langle C, \Phi, Q, L, \pi \rangle,$
where $C$ is the course content (e.g., textbook text), $\Phi$ denotes the instructional phase (e.g., introduction, practice), $Q$ is the question type (closed- or open-ended), $L$ specifies linguistic constraints (e.g., sentence length, hesitation markers), and $\pi$ indicates the target personality trait, selected from the Big Five dimensions: \textbf{HE}, \textbf{HN}, \textbf{LO}, \textbf{HA}, or \textbf{LC}. We adopt a single-trait activation strategy, where only one trait is varied per instance while others remain neutral, to enhance interpretability and control.
We further define a mapping function:
$\mathcal{F}_{\text{scene}} : \mathcal{T} \rightarrow \{ D_S, D_O, D_E, D_I \},$
which partitions scene-based task instances into datasets for foundation model validation ($D_S$), personality-aligned agent modeling ($D_O$), behavioral evaluation ($D_E$), and interactive teacher–student simulation ($D_I$). This unified schema ensures semantic alignment across all SOEI modules while supporting both one-shot and multi-turn learning contexts.

\subsection{$D_S$: Validating Foundation Model Adaptability}
\label{subsec:BCUAE}

To verify that $\mathcal{T}$ defines an instructional structure interpretable by foundation LLMs, we construct $D_S$ as a diagnostic dataset. Rather than being used for training or personality conditioning, $D_S$ serves as a scene-level feasibility check to ensure that downstream modeling (in $D_O$, $D_E$, and $D_I$) proceeds from scenarios within the LLMs’ comprehension and recall capabilities.

We compile the \textbf{Basic Chinese Understanding Ability Evaluation Dataset (BCUAE)}, adapted from the National Smart Education Platform\footnote{\href{https://basic.smartedu.cn/}{National Smart Education Platform}}, maintained by the Chinese Ministry of Education. BCUAE includes 1,051 items, with 613 for text comprehension (e.g., structural analysis, emotional inference) and 438 for memorization (e.g., passage reproduction). Items were constructed through expert-curated transformation and prompt engineering, and reviewed by certified junior high school language teachers.
As such, $D_S$ serves as a controlled benchmark to test whether foundation models can execute the core components of $\mathcal{T}$, laying the groundwork for personality-grounded generation and interaction modeling within the SOEI framework.

\begin{wraptable}{r}{0.36\linewidth}
\vspace{-10pt}
\centering
\caption{Accuracy scores of four foundation models on Chinese comprehension and memorization tasks.}
% \vspace{3pt}
\small
\setlength{\tabcolsep}{2pt}
\begin{tabular}{lccc}
\toprule
\textbf{Model} & \textbf{Comp.} & \textbf{Memo.} & \textbf{Avg.} \\
\midrule
InternVL~\cite{chen2024internvl}  & 0.736 & 0.758 & 0.747 \\
MiniCPM~\cite{hu2024minicpm}      & 0.736 & 0.664 & 0.700 \\
Qwen~\cite{bai2023qwen}           & 0.584 & 0.614 & 0.599 \\
LLaVa~\cite{li2024llava}          & 0.491 & 0.397 & 0.444 \\
\bottomrule
\end{tabular}
\label{tab:BCUAE-result}
% \vspace{-10pt}
\end{wraptable}

\subsection{Evaluation of Foundation Models}

We evaluate four foundation models—InternVL, MiniCPM, LLaVa, and Qwen (see App.~\ref{subsec:relate-work-fundation-llm})—on $D_S$ using comprehension accuracy, memorization performance, and question type adaptability (closed/open). As shown in Tab.~\ref{tab:BCUAE-result}, InternVL (0.747) and MiniCPM (0.700) outperform Qwen (0.599) and LLaVa (0.444). The top-performing models also demonstrate stable performance across comprehension (0.736) and memorization (0.758 and 0.664).
These results confirm that $\mathcal{T}$ provides a valid and executable instructional schema, and that $D_S$ serves as a reliable scene-level diagnostic tool for grounding the subsequent components in the SOEI pipeline.

\begin{figure}[t!]
  \includegraphics[width=\columnwidth]{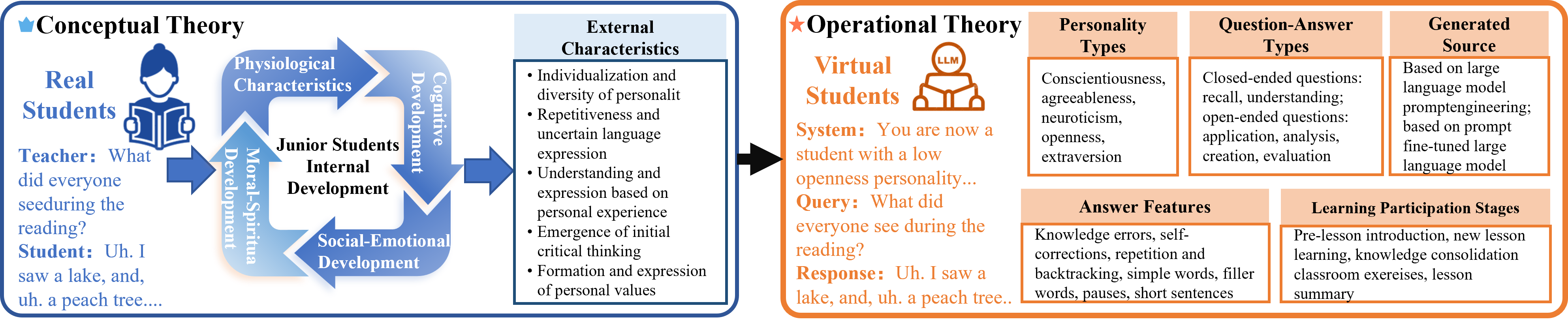}
    % \vspace{-15pt}
  \caption{Mapping conceptual theory to operational modeling for LVSA: Real student behaviors (left) are grounded in physiological, cognitive, social-emotional, and moral-spiritual development. These are transformed into operational dimensions (right)—personality traits, question-answer types, generation sources, linguistic styles, and learning stages—to construct controllable LVSAs.}
  \label{fig:LVSA}
    % \vspace{-10pt}
\end{figure}

\section{Object}
\label{sec:object}

Modeling personality-aligned virtual students poses key challenges in LSVA construction. Adolescent learners exhibit personality-driven cognitive–emotional traits~\cite{hall1905adolescence, piaget1952origins} that are difficult for foundation LLMs to emulate. Existing methods, often based on prompt engineering, struggle to maintain long-term behavioral consistency and controllability~\cite{zhang2024simulating}. To address these limitations, the SOEI framework integrates Big Five personality theory with educational psychology to guide agent construction. This section details our theoretical foundation, data preparation process, and fine-tuning strategy for generating personality-conditioned LSVAs.

\begin{wrapfigure}{r}{0.65\columnwidth}
% \vspace{-30pt}
\centering
\includegraphics[width=0.65\columnwidth]{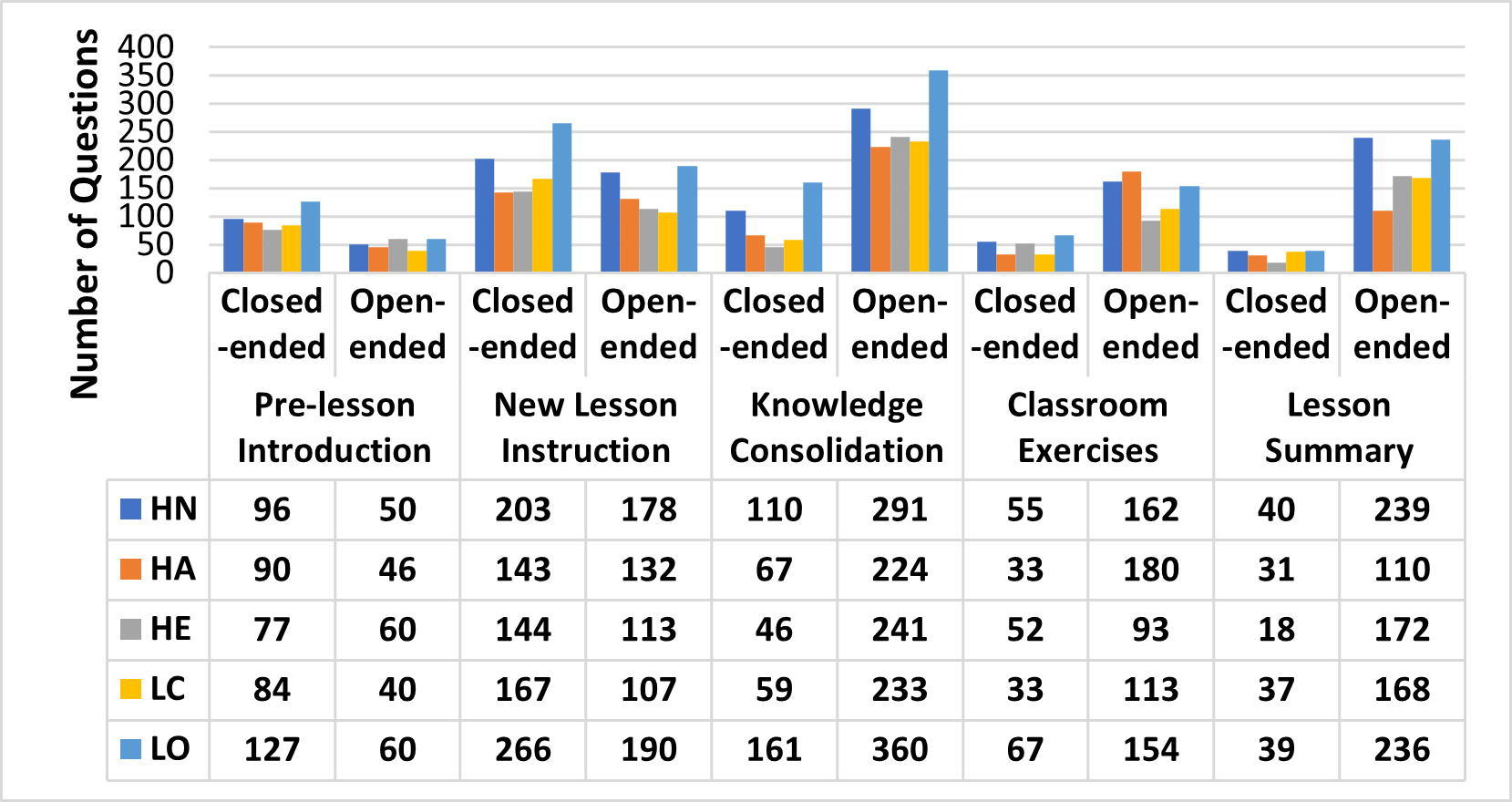}
% \vspace{-5pt}
\caption{Statistical information of $D_O$.}
\label{fig:fine-tuning-data}
% \vspace{-8pt}
\end{wrapfigure}

\subsection{Theoretical Framework}

To model personalized LVSA behavior, we draw on developmental psychology to connect theoretical insights with implementable design. Early adolescence (ages 10–15) is marked by cognitive, emotional, and social transitions~\cite{hall1905adolescence, blakemore2006development}, often manifesting as hesitancy, emotional fluctuation, and peer-sensitive reasoning. We encode these traits through five aligned dimensions: personality (Big Five)~\cite{costa1992neo}, question types~\cite{chin2006classroom}, instructional phases~\cite{hu2024teaching}, linguistic style~\cite{pennebaker1999linguistic}, and generation strategies—including Few-shot ICL~\cite{brown2020language}, Chain of Thought~\cite{wei2022chain}, and LoRA fine-tuning~\cite{hu2021lora}—to enhance realism and controllability. Fig.~\ref{fig:LVSA} summarizes this theory-to-implementation mapping, showing how real-world adolescent behaviors are structurally encoded for LVSA construction (see App.~\ref{subsubsec:conceptual-theory} and App.~\ref{subsubsec:operational-theory}).

\begin{wrapfigure}{r}{0.65\columnwidth}
\vspace{-20pt}
\centering
\includegraphics[width=0.65\columnwidth]{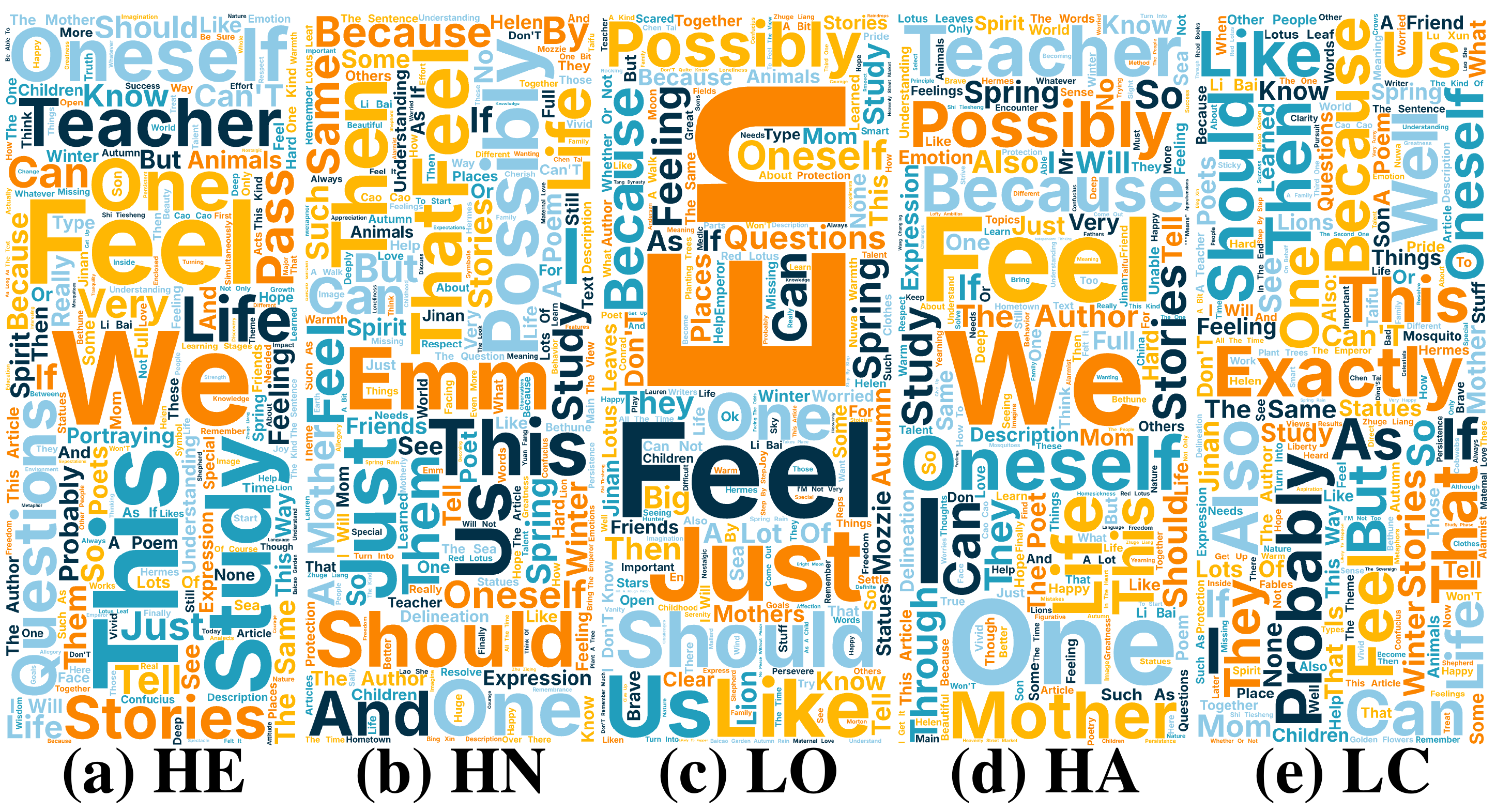}
% \vspace{-5pt}
\caption{Word cloud visualization of the Big Five personality fine-tuning dataset.}
\label{fig:wordcloud-en}
% \vspace{-5pt}
\end{wrapfigure}

\subsection{$D_O$: Fine-Tuning Dataset}

We define the personality set $\mathcal{P} = {\textbf{HE}, \textbf{HN}, \textbf{HA}, \textbf{LC}, \textbf{LO}}$, representing High Extraversion, High Neuroticism, High Agreeableness, Low Conscientiousness, and Low Openness, respectively. For each $\pi \in \mathcal{P}$, we construct a personality-conditioned subset $D_{O,\pi}$ based on the task structure $\mathcal{T}$, yielding the overall fine-tuning dataset:
$D_O = \bigcup_{\pi \in \mathcal{P}} D_{O,\pi}$.
Built upon the BCUAE dataset (§~\ref{subsec:BCUAE}), $D_O$ incorporates textbook materials, real classroom transcripts, and instructional plans (App.~\ref{subsubsec:ft-data}). Prompt–response pairs are generated via $\mathcal{T}$ and refined to match trait-specific linguistic styles (App.~\ref{subsubsec:ft-construction}). Personality-specific statistics are shown in Fig.~\ref{fig:fine-tuning-data}.
To assess stylistic alignment, we visualize word usage distributions for each $D_{O,\pi}$ using word clouds (Fig.~\ref{fig:wordcloud-en}). The results reveal clear personality–style patterns: \textbf{HE} exhibits expressiveness (“we,” “feel”), \textbf{HN} hesitancy (“um,” “I”), \textbf{LO} factuality (“because,” “questions”), \textbf{HA} empathy (“together,” “oneself”), and \textbf{LC} vagueness (“should,” “exactly”). These distinctions confirm the controllability and construct validity of $D_O$.

\subsection{Fine-Tuning Foundation Models}

To generate personalized LVSAs, we fine-tune foundation models on each $D_{O,\pi}$, obtaining personality-specific parameters $\Theta_\pi$. Fine-tuning employs the SWIFT framework\footnote{\href{https://github.com/modelscope/ms-swift}{SWIFT: Scalable lightweight infrastructure for fine-tuning}}, with independently optimized hyperparameters per trait (see App.~\ref{subsec:ft-configuration} and Fig.~\ref{fig:overview} d).
Post-finetuning evaluation reveals improved linguistic alignment with intended traits, along with enhanced personalization and coherence. Content analyses further verify that responses generated by $\Theta_\pi$ reflect personality-consistent styles, completing the modeling pathway from task configuration to personality dataset to model parameters ($\mathcal{T} \rightarrow D_{O,\pi} \rightarrow \Theta_\pi$).

\begin{figure*}[t!]
% \vspace{-50pt}
\centering
\includegraphics[width=\columnwidth]{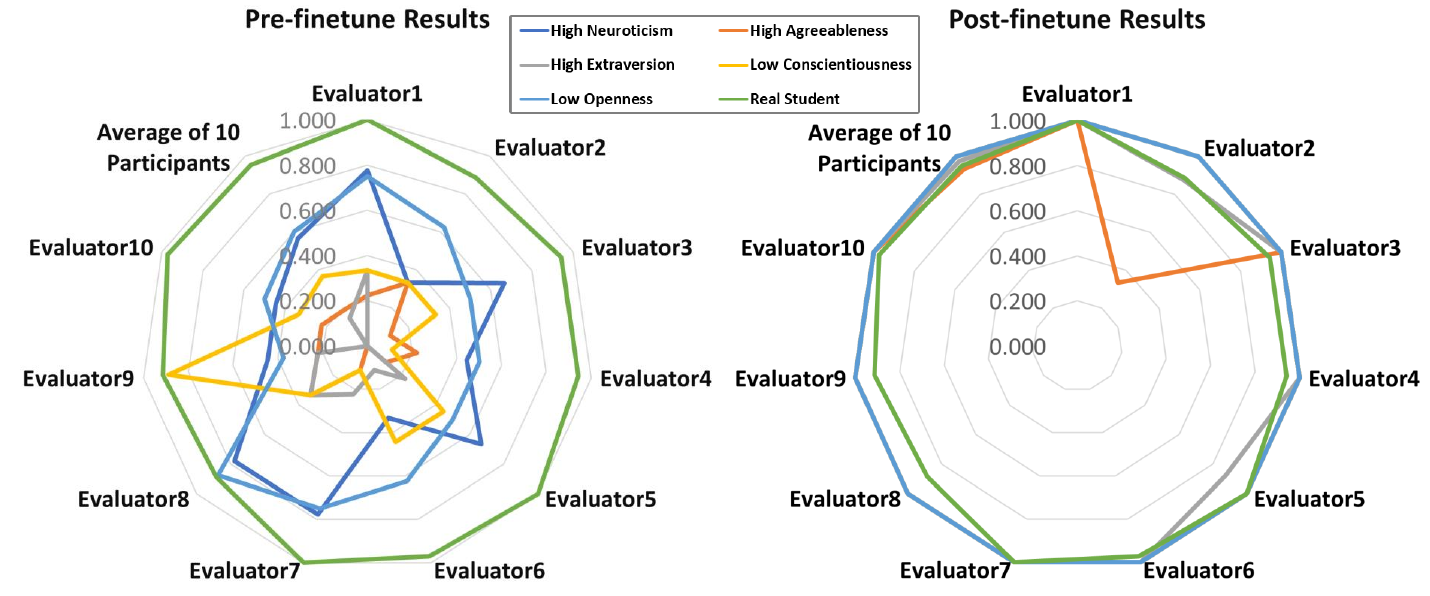}
% \vspace{-5pt}
\caption{Human evaluation results using radar chart visualization. Fleiss's Kappa = 0.6917 indicates substantial inter-rater agreement across personality types (see App.~\ref{subsubsec:human evaluation agreement}).}
\label{fig:radar-he}
% \vspace{-10pt}
\end{figure*}

\section{Evaluation}
\label{sec:evaluation}

Evaluating virtual students is inherently complex: traditional NLP metrics primarily assess surface-level quality~\cite{guo2023evaluating}, whereas authentic educational behavior encompasses personalized cognition, emotion, and interaction. Human evaluations, although reliable, are expensive and difficult to scale. To address this, we develop \textbf{a hybrid evaluation approach within SOEI}, integrating expert-based human-likeness judgments, GPT-4-assisted ratings, and reference-free objective metrics. The evaluation process employs responses $\Theta_\pi(\mathcal{T})$ generated by the standardized task structure $\mathcal{T}$, forming the evaluation dataset $\mathcal{D}_E$. We further define two subsets: $\mathcal{D}_E^{\text{human}}$ (120 human-scored samples) and $\mathcal{D}_E^{\text{GPT-4}}$ (12,232 GPT-4-rated samples), detailed in App.~\ref{subsec:info-sed}.

\subsection{Subjective Evaluation: Human and GPT-4}

\begin{wraptable}{r}{0.55\linewidth}
% \vspace{-8pt}
\centering
\caption{Human and GPT-4 evaluation scores across personality types (Pre/Post).}
\label{tab:human-gpt4-evaluation}
\small
\setlength{\tabcolsep}{2pt}
\begin{tabular}{lcccccc}
\toprule
\textbf{Evaluator} & \textbf{HN} & \textbf{HA} & \textbf{HE} & \textbf{LC} & \textbf{LO} & \textbf{RS} \\
\midrule
Ave-E (Pre)  & 0.567 & 0.189 & 0.144 & 0.367 & 0.600 & \multirow{2}{*}{0.951} \\
Ave-E (Post) & 1.000 & 0.933 & 0.975 & 1.000 & 1.000 & \\
\midrule
GPT-4 (Pre)  & 0.444 & 0.111 & 0.111 & 0.444 & 0.750 & \multirow{2}{*}{0.914} \\
GPT-4 (Post) & 1.000 & 1.000 & 1.000 & 1.000 & 0.889 & \\
\midrule
\textbf{Fleiss's Kappa} & \multicolumn{6}{c}{0.6806 (substantial agreement)} \\
\bottomrule
\end{tabular}
% \vspace{-5pt}
\begin{tablenotes}
\item[] \small Note: Ave-E = average human score; RS = real student baseline.
\end{tablenotes}
% \vspace{-5pt}
\end{wraptable}

\textbf{Human Evaluation.}  
We assess LVSA human-likeness using a Turing test-inspired approach, with real student (RS) responses serving as controls. Rather than comparing to fixed ground-truth answers, evaluators judged whether each response \textbf{effectively embodied student roles} within educationally relevant interactions. Each sample in $\mathcal{D}_E^{\text{human}}$ underwent a three-step evaluation: questionnaire, think-aloud protocol, and semi-structured interview (see App.~\ref{subsec:human-questionnaire}). Coders annotated behavioral features including hesitation, emotional tone, and social appropriateness, with interviews further refining evaluation criteria for future virtual student modeling.

Results (Fig.~\ref{fig:radar-he}) indicate substantial agreement among 10 evaluators (Fleiss's Kappa = 0.6917; see App.~\ref{subsubsec:human evaluation agreement}). The fine-tuned LVSAs achieved average recognition probabilities exceeding 0.9 across personality traits, indicating that human evaluators often failed to distinguish virtual students from real ones. One-way and two-way ANOVA analyses confirmed significant differences before and after fine-tuning ($p<0.001$), but no significant differences between fine-tuned LVSAs and RS responses ($p>0.05$). These findings validate the LVSAs' capacity to authentically simulate personality traits and demonstrate human-like behaviors in instructional interactions.

\textbf{Human–GPT Comparison Validation.}  
To scale evaluations while maintaining rigor, we prompted GPT-4 using Chain-of-Thought (CoT) format informed by expert-designed rubrics. Two trained evaluators conducted two-level qualitative coding using ATLAS.ti on real student responses, producing 4 primary and 15 secondary criteria covering emotional integration, cognitive levels, psychological states, and verbal expressions. This coding achieved high reliability (0.876; see App.~\ref{subsubsec:coding-interview}), validating the scoring schema. The codes were embedded into GPT-4’s CoT prompts, enabling structured, human-aligned step-by-step evaluations (see App.~\ref{subsubsec:gpt4-prompt-design}).
Tab.~\ref{tab:human-gpt4-evaluation} compares GPT-4 evaluations with human scores across five LVSA personalities. GPT-4 achieved an average score of 0.978 post-tuning and perfectly aligned (score = 1.000) with human ratings for the \textbf{HN} and \textbf{LC} traits. An overall Fleiss's Kappa of 0.6806 indicates substantial agreement, confirming GPT-4’s efficacy as a scalable evaluator for educational interaction behavior (see App.~\ref{subsubsec:human-gpt4-evaluation}).

\begin{table*}[h!]
\centering
\caption{Performance comparison (Pre/Post) across five personality types.}
\label{tab:result-diff-student-type}
\small
\setlength{\tabcolsep}{2pt}
\resizebox{\textwidth}{!}{
\begin{tabular}{lccccccccccc}
\toprule
\textbf{Trait} 
& \multicolumn{2}{c}{\textbf{InternVL}} 
& \multicolumn{2}{c}{\textbf{LLaVa}} 
& \multicolumn{2}{c}{\textbf{MiniCPM}} 
& \multicolumn{2}{c}{\textbf{Qwen}} 
& \multicolumn{2}{c}{\textbf{Avg.}} 
& \textbf{Trait $p$} \\
\cmidrule{2-11}
& Pre & Post & Pre & Post & Pre & Post & Pre & Post & Pre & Post & \\
\midrule
\textbf{HN} & 58.19\% & 94.31\% & 16.89\% & 80.45\% & 54.96\% & 94.62\% & 49.86\% & 94.62\% & 44.98\% & 91.00\% & 0.005** \\
\textbf{HA} & 33.99\% & 81.19\% & 14.52\% & 66.46\% & 24.75\% & 73.93\% & 43.89\% & 80.86\% & 29.29\% & 75.61\% & $<$0.001*** \\
\textbf{HE} & 44.64\% & 73.88\% & 12.69\% & 44.40\% & 19.78\% & 60.82\% & 30.34\% & 72.76\% & 26.86\% & 62.97\% & 0.002** \\
\textbf{LC} & 54.96\% & 50.49\% & 18.69\% & 52.67\% & 30.16\% & 34.43\% & 30.82\% & 39.02\% & 33.66\% & 44.15\% & 0.294 \\
\textbf{LO} & 79.21\% & 91.33\% & 13.33\% & 92.33\% & 55.67\% & 88.00\% & 47.83\% & 83.67\% & 49.01\% & 88.83\% & 0.066* \\
\midrule
\textbf{Avg.} & 54.20\% & 78.24\% & 15.22\% & 67.26\% & 37.06\% & 70.36\% & 40.55\% & 74.19\% & 36.76\% & 72.51\% & 0.009** \\
\midrule
\textbf{Model $p$} 
& \multicolumn{2}{c}{0.058*} 
& \multicolumn{2}{c}{0.004**} 
& \multicolumn{2}{c}{0.013**} 
& \multicolumn{2}{c}{0.007**} 
& \multicolumn{2}{c}{0.006**} 
& \\
\bottomrule
\end{tabular}
}
\vspace{-3pt}
\begin{tablenotes}
\item[] \small Note: * $p < 0.1$, ** $p < 0.05$, *** $p < 0.001$.
\end{tablenotes}
\end{table*}

\begin{table*}[h!]
\centering
\caption{Performance comparison (Pre/Post) across instructional phases.}
\label{tab:results-diff-stage}
\small
\setlength{\tabcolsep}{4pt}
\resizebox{\textwidth}{!}{
\begin{tabular}{lccccccccccc}
\toprule
\textbf{Phase} 
& \multicolumn{2}{c}{\textbf{InternVL}} 
& \multicolumn{2}{c}{\textbf{LLaVa}} 
& \multicolumn{2}{c}{\textbf{MiniCPM}} 
& \multicolumn{2}{c}{\textbf{Qwen}} 
& \multicolumn{2}{c}{\textbf{Avg.}} 
& \textbf{Phase $p$} \\
\cmidrule{2-11}
& Pre & Post & Pre & Post & Pre & Post & Pre & Post & Pre & Post & \\
\midrule
\textbf{PI} & 64.59\% & 82.32\% & 20.06\% & 66.60\% & 48.45\% & 79.73\% & 49.48\% & 81.79\% & 45.65\% & 77.61\% & 0.012** \\
\textbf{NL} & 54.71\% & 80.06\% & 16.59\% & 68.04\% & 38.04\% & 69.63\% & 41.41\% & 75.15\% & 37.69\% & 73.22\% & 0.008** \\
\textbf{KC} & 49.71\% & 77.32\% & 11.86\% & 69.52\% & 37.29\% & 71.19\% & 37.76\% & 70.51\% & 34.16\% & 72.14\% & 0.011** \\
\textbf{CE} & 48.63\% & 77.43\% & 16.14\% & 67.36\% & 33.89\% & 64.09\% & 37.37\% & 70.47\% & 34.01\% & 69.84\% & 0.006** \\
\textbf{LS} & 55.00\% & 76.97\% & 12.25\% & 69.14\% & 32.92\% & 72.10\% & 39.50\% & 76.18\% & 34.92\% & 73.60\% & 0.012** \\
\midrule
\textbf{Avg.} & 54.53\% & 78.82\% & 15.38\% & 68.13\% & 38.12\% & 71.35\% & 41.10\% & 74.82\% & 37.28\% & 73.28\% & 0.009** \\
\midrule
\textbf{Model $p$} 
& \multicolumn{2}{c}{$<$0.001***} 
& \multicolumn{2}{c}{$<$0.001***} 
& \multicolumn{2}{c}{$<$0.001***} 
& \multicolumn{2}{c}{$<$0.001***} 
& \multicolumn{2}{c}{$<$0.001***} 
& \\
\bottomrule
\end{tabular}
}
\vspace{-2pt}
\begin{tablenotes}
\item[] \small
Note: PI = Pre-lesson Introduction, NL = New Lesson Instruction, KC = Knowledge Consolidation, CE = Class Exercises, LS = Lesson Summary.
** means $p<0.05$; *** means $p<0.001$.
\end{tablenotes}
\end{table*}

\begin{table*}[h!]
\centering
\caption{Performance comparison (Pre/Post) across question types.}
\label{tab:results-diff-question-type}
\small
\setlength{\tabcolsep}{2pt}
\resizebox{\textwidth}{!}{
\begin{tabular}{lccccccccccc}
\toprule
\textbf{Type} 
& \multicolumn{2}{c}{\textbf{InternVL}} 
& \multicolumn{2}{c}{\textbf{LLaVa}} 
& \multicolumn{2}{c}{\textbf{MiniCPM}} 
& \multicolumn{2}{c}{\textbf{Qwen}} 
& \multicolumn{2}{c}{\textbf{Avg.}} 
& \textbf{Type $p$} \\
\cmidrule{2-11}
& Pre & Post & Pre & Post & Pre & Post & Pre & Post & Pre & Post & \\
\midrule
\textbf{CQ} & 58.61\% & 78.38\% & 17.72\% & 67.08\% & 43.19\% & 72.91\% & 47.64\% & 75.13\% & 41.79\% & 73.38\% & 0.015** \\
\textbf{OQ} & 50.40\% & 79.23\% & 12.98\% & 69.22\% & 32.81\% & 69.67\% & 34.47\% & 74.51\% & 32.67\% & 73.16\% & 0.006** \\
\midrule
\textbf{Avg.} & 54.51\% & 78.81\% & 15.35\% & 68.15\% & 38.00\% & 71.29\% & 41.06\% & 74.82\% & 37.23\% & 73.27\% & 0.009** \\
\midrule
\textbf{Model $p$} 
& \multicolumn{2}{c}{0.117} 
& \multicolumn{2}{c}{0.041**} 
& \multicolumn{2}{c}{0.068*} 
& \multicolumn{2}{c}{0.117} 
& \multicolumn{2}{c}{0.078*} 
& \\
\bottomrule
\end{tabular}
}
\vspace{-2pt}
\begin{tablenotes}
\item[] \small
Note: CQ = Closed-ended Question; OQ = Open-ended Question. * $p<0.1$, ** $p<0.05$, *** $p<0.001$.
\end{tablenotes}
\end{table*}

\subsection{Subjective Evaluation: Task-Specific Analysis with GPT-4}

We conducted detailed evaluations of personality-aligned models $\Theta_\pi$ using GPT-4 on the dataset $\mathcal{D}_E^{\text{GPT-4}}$, across three dimensions: student personality ($\pi \in \mathcal{P}$), instructional phases ($\Phi$), and question types ($Q$). All ratings were generated by GPT-4 following step-wise scoring rubrics (see App.~\ref{subsubsec:gpt4-prompt-design}).

\textbf{Personality-Level Results.} Tab.~\ref{tab:result-diff-student-type} shows that after fine-tuning, GPT-4 scores significantly increased from 36.76\% to 72.51\% across five personality types. Among these, \textbf{HN}, \textbf{HA}, and \textbf{HE} saw the most remarkable improvements, rising from 44.98\%, 29.29\%, and 26.86\% to 91.00\%, 75.61\%, and 62.97\%, respectively ($p<0.001$). Among the four foundation models $\mathcal{M}_i \in \{\text{InternVL}, \text{MiniCPM}, \text{Qwen}, \text{LLaVa}\}$, InternVL (78.24\%) and Qwen (74.19\%) achieved the highest post-tuning scores. LLaVa, starting at the lowest baseline (15.22\%), improved notably to 67.26\%, yet remained the weakest due to its cross-lingual limitations. The \textbf{LC} personality exhibited minimal improvement, highlighting challenges in modeling low-conscientiousness traits. Overall, these results demonstrate that fine-tuned models $\Theta_\pi$ effectively capture personality-consistent behaviors, significantly enhancing LVSA realism and differentiation.

\textbf{Learning Stage Results.} Tab.~\ref{tab:results-diff-stage} shows significant performance gains across all instructional phases defined in task structure $\mathcal{T}$ after fine-tuning. InternVL (78.82\%) and Qwen (74.82\%) again achieved the highest post-tuning scores, with an average model improvement of around 36\%. Paired t-tests confirmed these enhancements were statistically significant ($p<0.001$).  
Fine-tuning proved particularly effective for structured learning stage tasks, while personality modeling was more challenging due to greater contextual variability. Insights from semi-structured interviews (see App.~\ref{subsubsec:coding-interview}) further supported these findings, suggesting LSVAs can adapt across different instructional phases and help teachers refine instructional strategies throughout the teaching process.

\textbf{Question Type Results.} Tab.~\ref{tab:results-diff-question-type} demonstrates significant improvements in LVSA performance on closed-ended (\textbf{CQ}) and open-ended (\textbf{OQ}) questions after fine-tuning, with average gains of 31.59\% and 40.49\%, respectively. Paired t-tests confirmed statistical significance in performance gains for closed ($p<0.05$), open ($p<0.05$), and overall question types ($p<0.05$). InternVL and Qwen again performed best post-tuning (78.81\% and 74.82\%), improving by 24.30\% and 33.77\%. LLaVa also showed significant improvements across all question types ($p<0.05$), indicating enhanced generalizability. Notably, larger gains on open-ended questions suggest that fine-tuning particularly improved expressive reasoning more than factual recall capabilities. These findings illustrate that LVSA models benefit from personality-aligned fine-tuning not only at the identity level but also in adapting to varying cognitive demands.

\subsection{Objective Evaluation and Error Analysis}

\begin{wraptable}{r}{0.4\linewidth}
% \vspace{-35pt}
\centering
\caption{Personality-wise metrics across four dimensions.}
\small
\setlength{\tabcolsep}{2pt}
\begin{tabular}{lcccc}
\toprule
\textbf{} & \textbf{Text} &  & & \textbf{Positive} \\
& \textbf{Token} &  \textbf{TTR}  & \textbf{Clarity}  & \textbf{Sentiment} \\
\midrule
\textbf{LC} & 0.24 & 0.00 & 0.28 & 0.00 \\
\textbf{LO} & 0.00 & 0.74 & 0.00 & 0.47 \\
\textbf{HN} & 0.69 & 0.38 & 0.23 & 0.72 \\
\textbf{HE} & 1.00 & 0.88 & 1.00 & 1.00 \\
\textbf{HA} & 0.47 & 1.00 & 0.48 & 0.69 \\
\bottomrule
\end{tabular}
\label{tab:objective evaluation}
% \vspace{-10pt}
\end{wraptable}

To complement subjective assessments, we computed four reference-free metrics on the dataset $\mathcal{D}_E$: text length (Text Token), Type-Token Ratio (TTR), inverse perplexity (as a proxy for linguistic clarity), and sentiment polarity. Tab.~\ref{tab:objective evaluation} illustrates results for each personality type. Traits \textbf{HE} (High Extraversion) and \textbf{HA} (High Agreeableness) scored highest in clarity and positive sentiment, reflecting their extroverted and cooperative characteristics. In contrast, \textbf{LC} (Low Conscientiousness) and \textbf{LO} (Low Openness) exhibited lower scores with more flattened and vague linguistic styles. These findings align with subjective evaluation results but also reveal inherent metric limitations: TTR tends to overestimate lexical diversity in short texts, and sentiment analysis is heavily influenced by keyword presence. Complete metric distributions are available in App.~\ref{subsec:objective experiment results}.

Despite overall performance improvements, several anomalies highlight challenges in fine-tuning LLMs. Trait \textbf{LC} showed minimal improvement primarily due to weak trait representation and frequent hallucination issues, such as repetitive phrasing and factual drift (see App.~\ref{subsubsec:badcase-lc}). These problems were compounded by content safety filters, further restricting realistic expression and complicating modeling efforts.
Additionally, models showed inconsistent gains across different question types; no model achieved statistically significant improvements for both closed-ended and open-ended formats simultaneously. This suggests varying cognitive demands affect the stability of fine-tuning outcomes (see App.~\ref{subsubsec:badcase-question}).
Lastly, the \textbf{LLaVa} model consistently underperformed across evaluations due to cross-lingual domain mismatch, stemming from its predominantly English-centric pretraining and subsequent fine-tuning on Chinese data (see App.~\ref{subsubsec:badcase-llava}). These findings clarify directions for future research, specifically improving trait representation, enhancing data diversity, and bolstering robustness across languages and task variations.

\begin{figure*}[t!]
% \vspace{-40pt}
\centering
\includegraphics[width=\columnwidth]{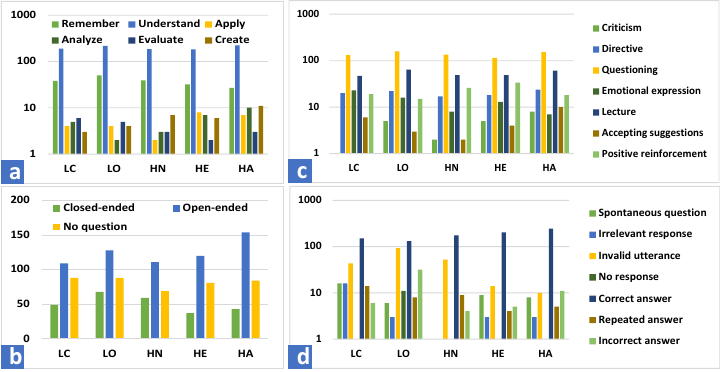}
% \vspace{-5pt}
\caption{Distribution of teacher–student interaction behaviors across LVSA personality types, categorized by (a) Bloom’s taxonomy ~\cite{bloom1956taxonomy}, (b) question types,  (c) teacher behavior types, and (d) student response types. The data are derived from annotated dialogue turns using a hybrid labeling framework combining GPT-4 prompt-based inference and expert correction. Results reveal trait-conditioned variations in questioning depth, cognitive engagement, and behavior management strategies.}
\label{fig:interaction-label-distribution}
% \vspace{-10pt}
\end{figure*}

\section{Interaction}
\label{sec:interaction}

% \subsection{Experimental Setup and Data Collection}

We conducted strictly controlled interaction experiments with 10 pre-service teachers, including five with prior classroom teaching experience and five without. Each teacher engaged in approximately 10-minute simulated lessons via a self-developed interaction interface (see Fig.~\ref{fig:interaction-ui}), interacting individually with five LVSAs, each representing one personality trait $\pi \in \mathcal{P}$. The interactions were structured according to the instructional task framework $\mathcal{T}$. All virtual students were instantiated using the fine-tuned InternVL model, which demonstrated the strongest performance in earlier experiments. We collected a total of 50 interaction sessions, comprising 1,288 dialogue turns and 11.47 hours of recorded data (see App.~\ref{subsec:interaction-setup}).

\subsection{Behavioral Labeling and Analysis}

The interaction dataset $\mathcal{D}_I$ was annotated along four behavioral dimensions, informed by the Flanders Interaction Analysis System and Bloom's taxonomy. Specifically, teacher utterances were labeled with: (1) cognitive level~\cite{bloom1956taxonomy} (e.g., remember, evaluate, create), (2) question type~\cite{chin2006classroom} (open-ended, closed-ended, or non-question), and (3) classroom behavior type~\cite{amatari2015instructional} (e.g., questioning, lecturing, praise). Student utterances were labeled with (4) response behavior~\cite{amatari2015instructional,nippold1999defining} (e.g., correct answer, spontaneous question, invalid utterance). Initial annotations were generated by GPT-4 through structured prompts and subsequently refined through manual review to ensure theoretical consistency (App.~\ref{subsec:annotation-details}).
Analysis of the 1,288 annotated dialogue turns across 10 teachers revealed distinct personality-conditioned interaction patterns (Fig.~\ref{fig:interaction-label-distribution}). \textbf{HA} and \textbf{HE} students elicited more open-ended questions and higher-order cognitive prompts (e.g., analysis, creation), while \textbf{LO} and \textbf{LC} students frequently produced invalid or irrelevant responses, prompting more directive and corrective teacher behaviors. Moreover, interactions with \textbf{HA}-type students were associated with increased praise and emotional engagement.
Further analyses, including subgroup comparisons between experienced and inexperienced teachers across annotation dimensions, are presented in App.~\ref{subsec:annotation-details}.

\subsection{Personality Identification through Teacher–Student Dialogue}
\label{subsec:personality-identification}

\begin{figure}[ht]
\centering
\includegraphics[width=\textwidth]{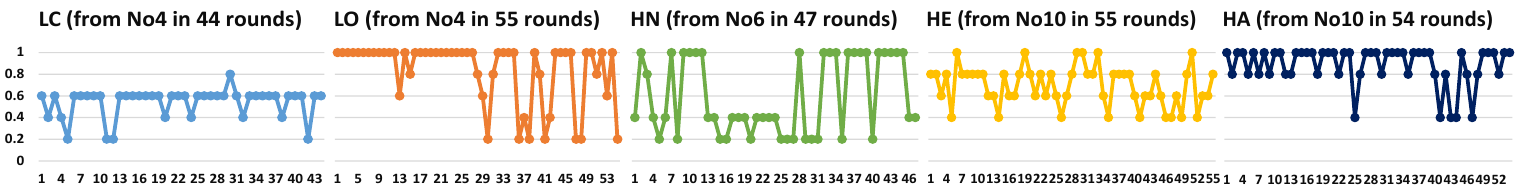}
% \vspace{-15pt}
\caption{
GPT-based personality score across dialogue turns for each personality type. Each subplot shows the interaction with the highest number of rounds for that personality (\eg, LC from No4, LO from No4, etc.). The vertical axis represents ranking score (1.0 = top prediction, 0 = not predicted); the horizontal axis is dialogue turn index.
}
\label{fig:personality-score-fluctuation}
% \vspace{-5pt}
\end{figure}

To assess whether personality traits can be effectively recognized from dialogue patterns, we prompted GPT-4 to rank the five candidate personality types ($\pi \in \mathcal{P}$) from most to least likely for each student turn. Accuracy was measured by whether the ground-truth personality appeared within the top-$k$ predictions. Over all 1,288 interaction turns, the LVSA achieved 44.2\% Top-1 accuracy, 65.5\% Top-2 accuracy, and 83.4\% Top-3 accuracy.
We also computed a position-weighted prediction score for each turn. As shown in Tab.~\ref{tab:personality-score-table}, \textbf{HA} achieved the highest score (0.931), followed by \textbf{LO} and \textbf{HN}, indicating that certain personality signals are more saliently reflected in dialogue.

\begin{wraptable}{r}{0.7\linewidth}
% \vspace{-10pt}
\centering
\caption{Average personality prediction scores per participant and personality type.}
\small
\setlength{\tabcolsep}{3pt}
\begin{tabular}{lccccccccccc}
\toprule
\textbf{} & No1 & No2 & No3 & No4 & No5 & No6 & No7 & No8 & No9 & No10 & \textbf{Ave.} \\
\midrule
\textbf{LC} & 0.48 & 0.66 & 0.61 & 0.54 & 0.61 & 0.62 & 0.45 & 0.58 & 0.56 & 0.63 & \textbf{0.57} \\
\textbf{LO} & 0.79 & 0.72 & 0.67 & 0.84 & 0.79 & 0.85 & 0.79 & 0.66 & 0.87 & 0.83 & \textbf{0.78} \\
\textbf{HN} & 0.66 & 0.91 & 0.81 & 0.86 & 0.62 & 0.60 & 0.87 & 0.84 & 0.72 & 0.84 & \textbf{0.77} \\
\textbf{HE} & 0.63 & 0.75 & 0.81 & 0.85 & 0.79 & 0.72 & 0.76 & 0.76 & 0.84 & 0.70 & \textbf{0.76} \\
\textbf{HA} & 0.87 & 0.97 & 0.94 & 0.97 & 0.93 & 0.93 & 0.93 & 0.91 & 0.98 & 0.89 & \textbf{0.93} \\
\bottomrule
\end{tabular}
\label{tab:personality-score-table}
% \vspace{-8pt}
\end{wraptable}

\textbf{Long-range stability analysis.}
To examine the consistency of GPT-4-based personality predictions during extended interactions, we visualized ranking score trajectories for participants with the longest sessions under each personality type (Fig.~\ref{fig:personality-score-fluctuation}). Results showed that \textbf{HA} and \textbf{HE} exhibited stable prediction scores across multiple turns, likely due to their pronounced social and emotional expressiveness, providing clear linguistic cues.
In contrast, personality types such as \textbf{LO} and \textbf{LC} showed higher volatility, frequently dipping below a score of 0.4 or failing to enter the Top-3 rankings. Crucially, such fluctuations should not be interpreted as modeling failures; rather, they authentically reflect the nuanced, context-dependent nature of real student behavior. Learners—especially those exhibiting low motivation or uncertain cognition—often display intermittent or ambiguous expressions, which are psychologically valid and pedagogically significant.
Consequently, we propose that variability should be considered a complementary metric to stability in evaluating virtual students. Stability verifies the controllable generation of personality traits, while variability captures human-like behavioral complexity. Importantly, despite temporary fluctuations, all five personality types maintained recognizable traits across extended dialogues, without significant decline over time. This confirms that LSVAs can sustain trait-consistent behaviors during prolonged interactions. Further analyses are detailed in App.~\ref{subsec:personality-eval}.

\textbf{Teacher reflections from post-interaction survey.}
To assess the pedagogical impact of LVSA-based interactions, we administered a post-experiment survey to all 10 pre-service teachers. All participants (100\%) reported increased adaptability to diverse student needs, and 80\% indicated improvements in classroom interaction skills.
Teachers generally perceived \textbf{HE} and \textbf{HA} students as the most engaging, whereas interactions with \textbf{LC} and \textbf{LO} students were reported as most challenging. Notably, 70\% of respondents indicated that they proactively adjusted their instructional strategies in response to student personality traits, reflecting emerging pedagogical sensitivity and flexibility.
Overall, participants viewed LVSAs as realistic simulations of classroom diversity, recognizing the interaction platform as a low-risk environment for practicing adaptive teaching strategies. Detailed quantitative and qualitative survey findings are presented in App.~\ref{subsec:survey-details}.

\section{Conclusion}
\label{sec:conclusion}

This study presents the SOEI framework, a systematic pipeline for constructing and evaluating personality-driven virtual student agents, integrating scenario modeling, personality-conditioned generation, behavioral evaluation, and interactive validation. For task design, we introduce a structured task tuple and single-trait activation mechanism to simulate cognitively diverse behaviors under pedagogical constraints. For evaluation, we develop a hybrid framework combining expert annotation and GPT-4-based scoring, balancing human alignment with scalability. For interaction, we conduct controlled studies with real teachers, demonstrating both the pedagogical impact of LVSAs and the long-range stability of their trait expression. SOEI fills a critical gap in simulating human-like student agents in classroom contexts and establishes a foundational paradigm for bridging AI4Edu and Edu4AI. Future work will explore broader subject domains, cross-cultural generalization, and long-term robustness, advancing trustworthy applications of LLM-powered educational agents.

\clearpage
% \section*{References}

\bibliographystyle{unsrt}
\bibliography{neurips_2025}

%%%%%%%%%%%%%%%%%%%%%%%%%%%%%%%%%%%%%%%%%%%%%%%%%%%%%%%%%%%%
\clearpage

\renewcommand{\thetable}{A\arabic{table}}
\renewcommand{\thefigure}{A\arabic{figure}}
\renewcommand{\thealgorithm}{A\arabic{algorithm}}
\setcounter{figure}{0}  
\setcounter{table}{0}  
\setcounter{algorithm}{0}

\appendix

\section{Comprehensive Related Works}
\label{sec:more-relate-work}

\subsection{Virtual Student Development and Related Technologies}
\label{subsec:relate-work-virtual-student}
Teachers, as a central resource for educational development, play a critical role in ensuring the quality of education. Pre-service teachers, who represent the future of the teaching workforce, depend heavily on practical training for their professional growth \cite{loewenberg2009work,zeichner2012turn,ward2018teaching}. Virtual students provide a promising tool for simulating classroom interactions, offering a cost-effective, highly interactive, and flexible platform for pre-service teacher training and feedback \cite{dotger2015exploring,dotger2010medicine,shaughnessy2018uncovering}. As technology in this domain has evolved, virtual students can be classified into three main categories: role-playing virtual students, programmatically pre-set virtual students, and virtual students powered by large language models (LLMs). Each of these approaches has distinct characteristics, technical underpinnings, and limitations.

\textbf{Role-playing virtual students} are among the earliest forms of simulated student behavior \cite{kersh1963classroom,colwell2013simulating,frederick2010teacher,shapira2015simulation,dalgarno2016practising}. In this approach, teachers or other adults imitate the cognitive and emotional responses of students, offering real-time interaction \cite{dieker2015tle,hume2012primary,teasdale2016instructional}. While this method allows for some degree of realism in mimicking student behaviors, it relies heavily on the participants' ability to accurately simulate a student’s responses. Consequently, this approach is resource-intensive, lacks scalability, and cannot provide the level of automation or flexibility required for modern teacher training systems \cite{sparks2011simulations}.

\textbf{Programmatically pre-set virtual students} use pre-defined algorithms and scripted rules to simulate student interactions, providing a degree of automation for classroom simulations \cite{christensen2011simschool,delamarre2021interactive,shernoff2018early,kelleci2021using}. These virtual students are typically powered by finite state machines or pre-scripted scenarios \cite{kervin2006classsim,lugrin2019towards}. For instance, a virtual student might respond to a teacher's question based on a pre-defined set of rules or dialogues. This approach is advantageous in its repeatability and operational control, which makes it useful for specific teaching tasks that require consistency \cite{carrington2011enhancing,hixon2009technology}. However, the rigidity of the pre-defined scripts often leads to uniform and mechanical responses, limiting their ability to adapt to the diverse and dynamic interactions that characterize real-world classrooms \cite{badiee2015design,rayner2014pre}.

Recently, \textbf{LLM-based virtual students} have gained attention with the development of models like GPT-4 \cite{achiam2023gpt,li2024leveraging,zhang2024simulating,lee23generative,markel2023gpteach}. These LLM-driven agents leverage advanced natural language processing capabilities to generate dynamic and contextually relevant student responses. Although LLMs introduce greater flexibility and linguistic variability into virtual student interactions, current research in this area remains nascent. Most existing studies focus on using prompt engineering to generate virtual student agents, yet the exchanging responses of LLMs inevitably faces significant challenges including role flipping, instruction repeating and fake replies \cite{qian2024chatdev}. In this way, these agents often struggle to replicate the full range of behaviors, cognitive abilities, and emotional nuances that real students exhibit. While LLMs have demonstrated strong capabilities in natural language generation and knowledge understanding \cite{jiang2022promptmaker,liu2023pre, park2023generative}, they have yet to fully mimic the language expression levels and cognitive challenges faced by real students. 

This study builds on this line of research by exploring whether LLMs can be leveraged to create human-like and personalized virtual student agents that better simulate student cognition and emotional responses. Our goal is to push beyond basic prompt-based LLM applications by constructing virtual students that can provide more authentic, dynamic, and adaptive learning interactions, thereby improving their applicability in pre-service teacher training.

\subsection{LLMs’ main capabilities to support AI4Edu}
\label{subsec:relate-work-llm-foundation}
In recent years, LLMs have demonstrated exceptional performance across a range of core language processing tasks \cite{brown2020language, achiam2023gpt}, including language understanding, generation, reasoning, knowledge transfer, and multimodal processing. Trained on vast amounts of textual data, these models have acquired the ability to comprehend complex linguistic structures and are widely applicable to various natural language processing tasks.

The \textbf{language understanding} capabilities of large language models are primarily reflected in their ability to infer meaning from context, capture complex syntactic structures, and accurately grasp the intent of a given text. In widely-used natural language understanding benchmarks such as GlUE \cite{wang2018glue}, SuperGLUE \cite{wang2019superglue} and SQuAD \cite{rajpurkar2018know}, models like GPT-4 \cite{achiam2023gpt}, LLaMA \cite{touvron2023llama}, and MiniCPM \cite{hu2024minicpm} have demonstrated performance that rivals, and in some cases surpasses, human levels \cite{AIbeathumans,li2023cmmlu}. For instance, GPT-4’s high scores on SuperGLUE indicate its ability to process extended text and maintain a deep understanding of contextual relationships \cite{achiam2023gpt,guo2023evaluating}. This is particularly important for building virtual students, as they must respond to teachers' questions with appropriate cognitive reactions, simulating the authentic thought processes of real students.

In benchmark tests like CommonGen \cite{lin2019commongen} and Wikitext-103 \cite{merity2017regularizing}, models such as GPT-4 and T5 \cite{raffel2020exploring} have demonstrated strong \textbf{language generation} capabilities, particularly in generating natural language paragraphs, dialogues, and long-form text \cite{li2024pre}. Wikitext-103, in particular, focuses on long-text generation tasks and evaluates the model's ability to maintain contextual consistency throughout the generation process. Current language models not only generate coherent responses in dialogue systems but also preserve logical consistency across multiple conversational turns \cite{zeng2023evaluating}. This provides a solid technical foundation for this study, where GPT-4 is used to generate teacher-student dialogue datasets under different personality traits.

In \textbf{language reasoning} benchmarks such as the ARC (AI2 Reasoning Challenge) \cite{clark2018think} and MNLI \cite{nangia2017repeval}, models are required to perform complex logical inference \cite{hu2020ocnli} and commonsense knowledge \cite{clark2018think} based on the given text. These tasks demand that the model not only understand the semantics of the questions but also use contextual information to provide appropriate answers. Current top-tier LLMs have demonstrated strong reasoning abilities in such tests \cite{zhong2023agieval, huang2024c}. For example, GPT-4 and PaLM \cite{chowdhery2023palm} have shown significant improvements in the ARC test compared to earlier models, enabling them to handle more complex reasoning tasks. This reasoning capability provides a solid technical foundation for virtual students to exhibit diverse reactions in the classroom, particularly in responding to open-ended questions, where the virtual student can demonstrate multi-step reasoning abilities.

\textbf{Language transfer} is a significant strength of large language models, enabling them to apply knowledge learned from one task to other domains or language tasks. In multilingual benchmarks such as Cross-Lingual Natural Language Inference (XNLI) \cite{conneau2018xnli} and TyDiQA \cite{clark2020tydi}, models have demonstrated strong cross-linguistic capabilities, effectively handling comprehension and reasoning tasks across various languages. Models like LLaMA, T5, Qwen \cite{bai2023qwen}, PaLM and GPT-4 have shown high generalization abilities in these tests, successfully addressing the challenges posed by different languages and tasks \cite{chowdhery2023palm,ahuja2023mega}. This ability is particularly crucial for developing multilingual virtual students, as teachers may interact with students in diverse linguistic and cultural contexts. By leveraging the knowledge transfer capabilities of large language models, virtual students can adapt to various language environments, providing flexible instructional feedback.

In recent years, \textbf{multimodal models} like Flamingo \cite{alayrac2022flamingo} and InternVL \cite{chen2024internvl} have become research hotspots, combining visual and textual processing capabilities and extending the application range of language models. In multimodal benchmarks such as Visual Question Answering (VQA) \cite{antol2015vqa,kafle2017visual}, models like InternVL have demonstrated their ability to generate accurate text responses based on visual inputs, showcasing their strong text generation and comprehension skills. These multimodal models offer greater potential for the future development of virtual students. For instance, by integrating both visual and textual generation abilities, virtual students can operate in more enriched multimodal teaching environments, handling more complex classroom tasks.

Among these models, GPT-4 and PaLM continue to excel in language understanding and reasoning tasks, particularly in handling complex language tasks. LLaMA demonstrates strong knowledge transfer capabilities in cross-linguistic tasks, making it especially suitable for building multilingual virtual students. MiniCPM performs exceptionally well in Chinese natural language processing tasks, making it ideal for non-English educational settings. Qwen exhibits powerful reasoning and cross-linguistic abilities, excelling particularly in language generation tasks. While InternVL is primarily a multimodal model, its text generation capabilities are also impressive, providing technical support for future virtual students to operate in multimodal learning environments.

\subsection{Representative AI4Edu Applications}
\label{subsec:relate-work-ai4edu}
Leading educational technology companies worldwide have pioneered the launch of various innovative products based on large language models (LLMs). Duolingo Max \cite{Duolingo} leverages LLMs to enhance language learning through generative dialogue and intelligent feedback, helping students improve their language skills via interactive exercises. Khan Academy’s Khanmigo \cite{Khan}, powered by LLMs, offers intelligent tutoring and personalized learning support, spanning from primary to higher education \cite{pavic2024conceptual,wulandari2024gamified,yasmin2022ai}. Google launched the Socratic platform \cite{socratic}, assisting students in solving problems across multiple disciplines such as science, mathematics, and history. Chinese edtech giants such as TAL Education, Youdao, iFLYTEK and Squirrel AI are also actively exploring LLM applications \cite{li2023adapting}. TAL’s MathGPT \cite{mathGPT} focuses on providing students with step-by-step guidance in mastering core mathematical problem-solving techniques. Youdao's “ZiYue” \cite{ziyue} prioritizes a “scenario-first” approach. iFLYTEK's “Spark Desk” \cite{spark} can conduct human-like interactive learning in various fields, including mathematics, English oral practice, essay correction, and more. Squirrel AI \cite{squirrelai} has developed a personalized learning system that leverages big data and AI to tailor individual learning paths, significantly improving students' learning efficiency and performance.

Internationally renowned universities are also contributing to the development and application of LLMs. MIT succeeds in achieving 81\% correctness on undergraduate-level maths problems through fresh-shot learning based on OpenAI Codex pre-trained models \cite{drori2022neural}, which combines natural language processing with code generation to assist students in advanced data analysis and programming tasks \cite{chen2021evaluating}. Similarly, East China Normal University has developed EduChat \cite{dan2023educhat}, which offers functionalities such as open Q\&A, essay grading, heuristic teaching, and emotional support.

To further validate and enhance the performance of LLMs in educational settings, several benchmark datasets and evaluation tools have been introduced. These datasets assess the models' knowledge comprehension and reasoning capabilities across various subjects. For instance, the C-Eval \cite{huang2024c} dataset is designed to evaluate the performance of LLMs across a range of subjects from primary to higher education. The SciQAG \cite{wan2024sciqag} dataset measures the model’s ability to answer scientific questions, testing its depth of domain knowledge. Mukea \cite{ding2022mukea} serves as a benchmark for multimodal knowledge extraction, challenging the models' ability to integrate information from diverse input sources. These benchmark evaluation tools not only drive the advancement of LLMs but also provide researchers with precise metrics to continually optimize the educational effectiveness of these models.

Moreover, leading artificial intelligence conferences such as NeurIPS \cite{denny2024generative} and AAAI \cite{ananda2024ai} have seen a growing number of research papers and workshops discussing the applications of LLMs in education. Topics such as supporting individualized learning paths \cite{hayat2023personalization, hwang2023towards, schmucker2023ruffle}, providing personalized feedback \cite{mcnichols2023automated,sahai2023improving}, and education dataset construction \cite{han2023exploring} are becoming increasingly popular, highlighting the critical role that LLMs play in advancing educational technologies.

Although significant progress has been made in using LLMs as domain-expert to improve student learning outcomes—particularly in areas such as problem-solving \cite{didolkar2024metacognitive}, knowledge explanation \cite{gao2024fine}, and automated grading \cite{lagakis2024evaai}—there remains a noticeable gap in research focused on enhancing teacher skills. To date, only one study \cite{lee23generative} in this area has been featured at NeurIPS, indicating that this field is still emerging and holds considerable potential for further exploration. However, despite the pivotal role teachers play in education, research on leveraging LLMs to support their diverse instructional needs remains notably scarce, underscoring an important area for future work.

This study addresses this gap by introducing a novel research direction: using virtual students to enhance teachers' pedagogical skills. Unlike previous efforts that mainly focused on constructing expert systems to assist student learning, this study aims to generate more human-like and emotionally responsive virtual students through LLMs. These virtual students simulate real classroom reactions, providing teachers with authentic interactive teaching environments. By engaging with these virtual students, teachers can practice, reflect on, and refine their teaching methods, ultimately enhancing their professional capabilities and preparing them for future classroom scenarios.

\clearpage

\section{Detailed Information for Theory and Technology Used in This Work}

\subsection{Foundation LLMs}
\label{subsec:relate-work-fundation-llm}

To evaluate the performance of LLMs in junior high school Chinese language tasks, this study selected four representative foundation models for objective assessment. All models used in this study are consistent with their originally intended purposes and comply with their respective licensing terms. Each model has demonstrated outstanding performance across various domains, with unique strengths in handling multimodal tasks or pure text generation. These models are pivotal for evaluating comprehension and recitation abilities, which are essential for constructing virtual student agents in future research.

\textbf{InternVL} \cite{chen2024internvl} is a multimodal large language model that integrates both visual and linguistic processing capabilities. It excels at cross-modal reasoning, particularly in tasks requiring the integration of textual information with images. Although primarily applied to multimodal scenarios, InternVL’s strong language processing skills make it suitable for the text comprehension tasks required in this study.

\textbf{LLaVa} \cite{li2024llava} is a lightweight yet powerful LLM based on the Mistral architecture. It specializes in text generation and language understanding, offering efficient reasoning capabilities even at smaller scales. This model is expected to perform well in medium-scale tasks, such as text comprehension and recitation, making it a flexible choice for resource-constrained environments.

\textbf{MiniCPM} \cite{hu2024minicpm} is optimized for Chinese language tasks, with deep capabilities in understanding and generating Chinese. Trained on a large corpus of Chinese data, MiniCPM excels in tasks such as essay correction and passage analysis, making it particularly well-suited for junior high school Chinese language comprehension and recitation tasks.

\textbf{Qwen} \cite{bai2023qwen} supports multimodal interactions, excelling in both visual-linguistic exchanges and pure text-based tasks. Its ability to integrate visual and linguistic information allows it to handle complex tasks, making it a valuable tool for both current text-based tasks and future applications involving multimodal inputs.

By comparing these models’ performance in junior high school Chinese scenarios, the study aims to assess their ability to process and understand Chinese text, specifically focusing on comprehension and recitation. As future research may extend to multimodal teaching environments, such as virtual students recognizing visual content presented by teachers, the inclusion of multimodal models ensures a foundation for future expansion into scenarios requiring both textual and visual processing.

\begin{table*}[h!]
    \centering
    \fontsize{8pt}{10pt}\selectfont
    \setlength{\tabcolsep}{3.5pt}
    \caption{The URLs of official foundation LLMs repository in this work.}
    \begin{tabular}{lllll}
    \toprule
      \textbf{Foundation LLMs} & \textbf{Creator} & \textbf{\#Parameters} & \textbf{Access} & \textbf{License}\\ 
      \midrule
       \textbf{InternVL2} (\href{https://huggingface.co/OpenGVLab/InternVL2-8B}{InternVL2-8B}) \cite{chen2024internvl} & Shanghai AI Laboratory & 8B & Weights & MIT\\
       \textbf{LLaVa} (\href{https://huggingface.co/llava-hf/llava-v1.6-mistral-7b-hf}{LLaVa-v1.6-mistral-7b-hf}) \cite{liu2024improved} & Microsoft & 7B & Weights & Apache License 2.0\\
       \textbf{MiniCPM} (\href{https://huggingface.co/openbmb/MiniCPM-V-2\_6}{MiniCPM-V-2-6-chat}) \cite{yao2024minicpm} & MiniCPM-V & 8B & Weights & Apache License 2.0\\
       \textbf{Qwen} (\href{https://huggingface.co/Qwen/Qwen-VL-Chat}{Qwen-vl-chat}) \cite{bai2023qwen} & Alibaba Cloud & 7B & Weights & \href{https://github.com/QwenLM/Qwen-VL/tree/master?tab=License-1-ov-file}{Qwen license} \\
    \bottomrule
    \end{tabular}
\end{table*}

\subsection{Educational Theory}
\label{subsec:relate-work-theory}

\subsubsection{Conceptual Theory: Characteristics of Early Adolescent Students}
\label{subsubsec:conceptual-theory}
Early adolescents (typically aged 10 to 15) represent a critical developmental stage where students undergo substantial physiological, cognitive, social-emotional, and moral-spiritual changes \cite{hall1905adolescence, caskey2014developmental}. These underlying developmental traits manifest as explicit behaviors in the classroom, providing essential theoretical insights for constructing virtual student agents.

At the \textbf{physiological level}, rapid development of the prefrontal cortex, which governs complex cognitive functions such as decision-making, reasoning, and impulse control, contrasts with the faster development of the limbic system, responsible for emotional regulation \cite{blakemore2006development,giedd2004structural}. This imbalance often leads to intense emotional responses coupled with limited cognitive regulation, which influences classroom behavior and language expression. Adolescents may display strong emotional reactions, cognitive confusion, and variability in participation, establishing a physiological basis for their learning behaviors.

\textbf{Cognitively}, adolescents transition from concrete operational thinking to abstract logical reasoning \cite{elkind1981children}. According to Piaget's theory \cite{piaget1952origins}, they gradually develop the ability to handle abstract concepts and logical analysis, though they remain partially bound by concrete experiences. They exhibit early signs of critical thinking but are easily influenced by emotions and external factors, leading to repetitive and uncertain language expression \cite{nippold1993developmental, nippold1999defining}, especially when faced with abstract issues.

\textbf{Social-emotionally}, as self-awareness increases, adolescents become more concerned with their roles within social groups, experiencing frequent emotional fluctuations. This heightened sensitivity affects their classroom behavior—teacher feedback and peer reactions significantly impact their engagement. In constructing virtual students, it is crucial to simulate such emotional dynamics to accurately reflect students' responses to classroom stimuli, such as nervousness, anxiety, or excitement.

\textbf{Morally and spiritually}, adolescents begin developing their own values and face complex moral issues \cite{scales2010characteristics}. According to Kohlberg \cite{kohlberg1987psychology}, they transition from external rule adherence to internalizing societal norms. However, their moral judgments are often emotionally driven and oversimplified. In class, adolescents interpret moral situations from texts through personal experiences \cite{scales2010characteristics}, reflecting emerging critical thinking while simplifying complex moral issues.

These implicit traits manifest in the classroom as individualized behaviors and language use \cite{pennebaker1999linguistic}. Adolescents’ classroom participation varies greatly—some engage actively while others remain hesitant. Their language often involves repetition and uncertainty, especially with abstract questions, using filler words like “um” or “uh”. This reflects their cognitive insecurities and limited grasp of complex issues. They also frequently relate classroom content to personal experiences, which should be considered when designing instructional tasks. Additionally, emerging critical thinking shows students beginning to analyze multiple perspectives, though these skills are not yet fully mature.

\subsubsection{Operational Theory: Classification of Teacher-Student Dialogue Datasets}
\label{subsubsec:operational-theory}

When constructing LVSA, it is crucial to ensure that the generated dialogues are highly operational and predictable, allowing the model to realistically simulate personalized behavior and language capabilities of different students. To achieve this, we introduced a set of practical dimensions to classify and guide the generation of virtual students, including question-answer types, personality traits, learning stages, response styles, and generation sources. These dimensions serve as the foundation for generating diverse student behaviors and ensuring that virtual students exhibit realistic responses consistent with real students across various classroom scenarios.

\textbf{Question-Answer Type}: The type of question-answer is fundamental for generating virtual students. Classroom questioning is a key strategy for teachers to guide student thinking and stimulate learning interest \cite{hattie2007power}. Different types of questions require different cognitive processes, influencing student language output and cognitive development \cite{chin2006classroom}. Open-ended questions encourage creative and critical thinking, while closed-ended questions emphasize recalling facts and accurately understanding specific knowledge. Careful design of question types aids in simulating realistic classroom interactions and evaluates the model's ability to handle diverse questions.

\textbf{Personality Traits}: Personality traits are a critical dimension in modeling virtual students. Integrating the Big Five Personality Traits model allows for personalized student modeling \cite{mccrae1997personality}. These traits—extraversion, agreeableness, conscientiousness, neuroticism, and openness impact student behavior and language expression. Modeling virtual students based on the Big Five ensures the creation of distinct personalities and diverse language patterns, effectively simulating varied student behaviors in the classroom \cite{christensen2011simschool,poropat2009meta}. Specifically, to enhance the personalized expression of student personalities in this study and to fully explore the potential of large language models in constructing individualized and humanized virtual students, we utilized five personality types that differed widely: \textbf{High Neuroticism} (\textbf{HN}), \textbf{High Extraversion} (\textbf{HE}), \textbf{High Agreeableness} (\textbf{HA}), \textbf{Low Conscientiousness} (\textbf{LC}), and \textbf{Low Openness} (\textbf{LO}). The associated personality descriptions and the sample descriptions of the personality types used in this study are shown in the Tab.~\ref{tab:bigfive}. Among these, HE and HA are positive personality traits, representing well-performing students with higher cognitive abilities and stronger language proficiency. HE students demonstrate a greater desire to express themselves compared to HA students. On the other hand, LC and low openness are relatively negative traits. Students with LC tend to be careless, with less detailed language expression, while students with LO exhibit lower cognitive abilities, introversion, and hesitation, often struggling to answer questions. Compared to the other four personalities, HN presents a more neutral profile; aside from experiencing nervousness and hesitation in their responses, students with this trait generally exhibit normal cognitive abilities.

\textbf{Learning Stages}: Students' cognitive abilities and language expression vary significantly at different learning stages, which must be reflected accurately in virtual student modeling. According to Vygotsky's Zone of Proximal Development (ZPD) theory (Vygotsky, 1978), language and cognitive development progress with appropriate challenges. In early learning stages, virtual students may use simple sentence structures and straightforward responses, while in advanced stages, they can use complex grammar and logical reasoning. Modeling learning stages helps create virtual students suited to different instructional contexts, offering richer and more varied classroom interactions.

\textbf{Answering Style}: Answering style highlights individual differences in students' language expression. For example, students with high openness may use more creative and exploratory language, while those with high conscientiousness may provide more detail-oriented and accurate responses. Additionally, students with high neuroticism often show emotional fluctuation, including hesitation and uncertainty, while those with high emotional stability tend to respond with confidence, using concise language (Dewaele, 2012). Reflecting these styles in virtual students ensures a realistic diversity in their dialogue.

\textbf{Generation Source}: The generation source is vital for ensuring realism in virtual student behavior. Creating virtual students involves both prompt engineering and instruction fine-tuning based on real classroom data. In this study, we utilized traditional prompt engineering (Prompt Example detail see App.~\ref{subsubsec:traditional prompt}), few-shot ICL (Prompt Example detail see App.~\ref{subsubsec:ft-construction})\cite{brown2020language}, and CoT (Prompt Example detail see App.~\ref{subsec:gpt4 evaluation prompt})\cite{wei2022chain}, while employing LoRA \cite{hu2021lora} for fine-tuning the model.This combination ensures that generated dialogues realistically reflect student behaviors in actual learning environments. By integrating these methods, the generated dialogues closely resemble real-life classroom interactions, capturing the human-like and personalized traits of virtual student agents.

\begingroup
\fontsize{8pt}{10pt}\selectfont % 设置字体大小
\setlength{\tabcolsep}{3.5pt}   % 设置列间距
\begin{longtable}{p{2cm} p{2.5cm} p{2.5cm} p{6cm}}

\caption{Five typical early adolescent students characters based on the Big Five personality theory.} \label{tab:bigfive} \\
\toprule
\textbf{Personality Traits} & \textbf{Low Level} & \textbf{High Level} & \textbf{Student Description in the Study} \\
\midrule
\endfirsthead

\toprule
\textbf{Personality Traits} & \textbf{Low Level} & \textbf{High Level} & \textbf{Student Description in the Study} \\
\midrule
\endhead

\midrule
\multicolumn{4}{r}{\textit{Continued on next page}} \\
\midrule
\endfoot

\bottomrule
\endlastfoot

\textbf{Openness} & Conservative, pragmatic and efficient & Have broad interests, are willing to change, are curious, and get bored easily & Student A, with low openness to experience, tends to exhibit characteristics such as low receptivity to new content (preferring to rely on familiar knowledge and experiences, often feeling confused or uninterested in unfamiliar or complex topics), lack of initiative in exploration (showing little inclination to investigate or think critically about new problems, rarely participating in discussions or asking questions), and weaker ability to handle complex issues (struggling to provide effective responses or solutions to more challenging questions or content requiring deep understanding). Consequently, their language style is characterized by simple and direct responses (avoiding complex analysis or evaluation, and often responding to difficult questions with brief answers like 'um' or 'uh'), reliance on filler words (using 'um,' 'uh,' and similar phrases to express confusion or hesitation), and difficulty expanding discussions (rarely offering additional information or perspectives without prompting). \\ 
\midrule
\textbf{Conscientiousness} & Spontaneity, multi-tasking, flexibility, lack of organization and discipline & Task and goal oriented, reliable, organized & Student B, with low conscientiousness, tends to exhibit traits such as carelessness (frequently making mistakes, though occasionally providing correct answers), inconsistency (sometimes recognizing and correcting their own errors, while other times making outright mistakes), and lack of systematic thinking (responses tend to be disorganized and occasionally omit important details). Consequently, their language style is characterized by simplicity and directness (responses are concise but prone to errors), occasional self-correction (they may occasionally identify and correct their own mistakes, though inconsistently), and unreliability (responses may be disjointed, with alternating instances of correctness and error). \\ 
% \midrule
\textbf{Extraversion} & Introvert, private, reserved & Social, participates in many activities, enjoys being with people, active, outgoing & Student C, with high extraversion, exhibits traits such as active participation (a strong desire to engage, frequently answering questions, and proactively showcasing themselves), strong social skills (active in class, interacting frequently with teachers and peers, and expressing themselves with confidence), ease of performance (responding to classroom activities and questions naturally and confidently, showing little influence from external factors), and high visibility (eager to display themselves, actively participating in discussions, and enjoying being the center of attention). Consequently, their language style tends to be fluent and confident (speaking smoothly and confidently, with occasional hesitations or pauses when overexplaining), elaborative (offering detailed explanations or examples to demonstrate understanding), positive (using enthusiastic language that reflects interest in the topic), and structured (responses are organized and clear, with minimal unnecessary or repetitive statements). \\ 
\midrule
\textbf{Agreeableness} & Doubt, competitiveness, pride & Tolerant, accepting, kind, gentle, trustworthy & Student D, with high agreeableness, exhibits traits such as cooperation and empathy (eager to assist peers, actively engage in discussions, and show concern for others), thoughtfulness and patience (remaining open to the opinions of teachers and classmates, and expressing understanding and support), and positive feedback (responding to questions with consideration for others' emotional states). Consequently, their language style tends to be warm and friendly (using soft language and a kind tone), elaborative (providing detailed explanations that reflect understanding and care), and accurate (with clear, error-free responses). \\ 
\midrule
\textbf{Neuroticism} & Contented, controlled, easy going & More negative emotions, increased stress, and lower satisfaction & Student E exhibits high neuroticism, characterized by anxiety and nervousnesnains (displaying heightened tension and unease, easily influenced by the classroom environment), repetitive backtracking (repeating and revisiting responses, reflecting worry and uncertainty about their answers), and emotional fluctuations (prone to emotional instability). As a result, their language expression style tends to be hesitant (including hesitative fillers like 'um' and 'uh'), repetitively backtracking (repeating parts of their responses to confirm or correct perceived uncertainties), and disjointed (with fragmented and less coherent speech patterns). \\ 
\end{longtable}
\endgroup

\subsubsection{The Unique Value of Middle School Students and the Language-based Subject in This Study}
\label{subsubsec:unique-value}
In this study, we focus on early adolescents (ages 10-15, \ie, middle school students) as the target group for virtual student modeling, drawing insights from cognitive and developmental psychology. Students at this stage demonstrate unique cognitive development and language expression characteristics, presenting distinct challenges. Prior research indicates that early adolescents are transitioning from concrete to abstract thinking, with cognitive development not yet fully matured \cite{piaget1952origins}. Compared to other age groups, these students have relatively weaker logical reasoning skills and often rely more on intuitive perception and concrete experiences for problem-solving. This cognitive limitation frequently results in unclear language expression and inconsistent thought processes when dealing with complex tasks \cite{kail2016longitudinal}. Therefore, modeling students at this developmental stage offers a more accurate representation of diverse classroom behaviors while posing significant challenges for large language models in handling tasks requiring complex cognitive abilities.

Early adolescents often exhibit limited language expression, characterized by high levels of uncertainty and repetition in verbal responses during classroom interactions, especially with open-ended questions. This hesitation and lack of confidence stem not only from lower cognitive development but also from limited language proficiency \cite{cole1978vygotsky}. Thus, selecting this age group as the target for virtual student modeling holds substantial representational value, particularly for assessing whether large language models can effectively simulate students' language and cognitive responses, creating human-like, personalized virtual students. Compared to students at other developmental stages, early adolescents tend to offer simpler logical responses and are more easily influenced by external stimuli, making modeling this group both more challenging and valuable for research.

Building on this foundation, we use Information Processing Theory to further analyze the specific cognitive processes involved when students answer questions in the classroom. According to this theory, the thinking process can be divided into three stages: focusing on the question, understanding the question, and providing a response \cite{atkinson1968human}. First, students must direct their attention to the teacher's question, requiring adequate attention and selective information-processing abilities. Next, they need to understand the question's meaning and relate it to their existing knowledge, demonstrating cognitive processing skills. Finally, students generate a response, relying not only on their knowledge but also on their ability to articulate their thoughts clearly. In virtual student modeling, it is crucial to accurately represent these three stages to ensure appropriate responses within realistic classroom scenarios, effectively simulating students' thought processes and cognitive reactions.

We selected language-based subjects as the modeling context because language teaching emphasizes language expression, emotional experience, and value cultivation \cite{bruner1966studies}. This focus aligns closely with the core capabilities of large language models (LLMs), which are adept at processing and generating natural language. Language subjects require not only understanding and applying fundamental knowledge such as grammar and vocabulary but also involve deep interpretation of complex texts and emotional expression. This combination of emotion and cognition provides an ideal platform for evaluating and optimizing virtual student models, particularly in assessing whether they can replicate real students' linguistic styles, emotional expressions, and responses to value-based questions.

In summary, this study selects early adolescents as the target group for virtual student modeling due to their developing cognitive and linguistic abilities, as well as the representative complexity of their classroom interactions. Additionally, language-based subjects serve as an optimal testing ground for evaluating large language models, offering comprehensive insights into language expression, emotion, and attitudes. This approach provides a solid theoretical foundation and practical value for the future development of virtual student agents.

\clearpage
\section{Detailed Information for Dataset Construction Process}
\subsection{Basic Chinese Understanding Ability Evaluation Dataset}
\label{subsec:BCUAE-dataset}

\subsubsection{Data source}
The dataset utilized in this study is sourced from the \textbf{National Smart Education Platform}, an initiative developed by the Ministry of Education to provide authoritative and standardized educational materials for teachers and students nationwide. This platform integrates the official junior high school Chinese language curriculum, comprising textbook passages, practice exercises, comprehension assessments, and recitation tasks. These resources encapsulate the essential knowledge and skills required for junior high school Chinese education, adhering to national standards.

The dataset is focused on two pivotal aspects: \textbf{text comprehension} and \textbf{text memorization}. Both are critical not only for evaluating student performance in junior high school Chinese courses but also for assessing overall language proficiency. To ensure comprehensive and diverse content, we selected a wide range of lessons and corresponding exercises.

For \textbf{text comprehension}, the dataset includes various passages, chapters, and comprehension questions designed to assess students’ ability to understand the structure of texts, analyze sentence and paragraph organization, infer details, explain vocabulary in context, and interpret emotional tones. These tasks allow us to evaluate whether LLMs can effectively process and understand Chinese texts at the junior high level, replicating the skills expected of students in language comprehension.

\textbf{Text memorization}, particularly through recitation, plays a unique role in Chinese language education, where students are required to internalize the content, language structures, and literary qualities of texts. Our dataset includes a series of standardized recitation tasks aimed at evaluating how accurately and completely LLMs can simulate students' text recitations. This aspect tests the model's ability to recall and reproduce text, serving as a benchmark for assessing long-term memory and linguistic retention.

The dataset is standardized and regularly updated by expert teams to ensure consistency with the national curriculum. This ensures that the data maintains a high degree of authority, objectivity, and applicability across educational settings. Data was collected from various question types to thoroughly assess the performance of foundational models in junior high school Chinese language scenarios, providing a solid foundation for future model development and refinement.

\subsubsection{Data construction process}
\label{subsubsec:traditional prompt}

The data construction process is divided into three main steps: \textbf{data preparation}, \textbf{prompt design}, and \textbf{expert revision}. The goal of this process is to ensure the accuracy and scientific rigor of the data, providing high-quality inputs for the foundational model’s scenario evaluation.

\textbf{Data Preparation}: The junior high school Chinese exercises were organized unit by unit, ensuring that each lesson covered the core knowledge points and skill requirements related to the text. The exercises for each unit were compiled and converted into PDF format for subsequent input into the GPT-4 model. This approach not only maintained the integrity of the data structure but also provided a clear framework for the model’s processing in later stages.

\textbf{Prompt Design}: An instruction-based learning method was employed, in which structured prompts with “task description-task requirements-specific output” were provided to guide the model in executing designated tasks. By providing clear task instructions, the model could focus on specific objectives and generate outputs aligned with expectations. This design ensured that the model’s outputs were accurate, particularly for complex tasks such as text comprehension and recitation.

\textbf{Expert Revision}: To ensure the accuracy of the generated data, a team of professional junior high school Chinese teachers was invited to review and refine the content produced by GPT-4. The expert team evaluated the model’s outputs for both text comprehension and recitation, checking for accuracy, consistency, and alignment with educational standards. Any errors identified were carefully corrected by the teachers, ensuring that the dataset met the practical requirements of junior high school Chinese education. This expert revision process further optimized and enhanced the quality of the generated content.

In total, 613 text comprehension items and 438 memorization tasks were collected. The expert-revised data significantly improved the dataset’s overall quality, providing a more rigorous and reliable foundation for subsequent model evaluations.

\begin{center}
\begin{tcolorbox}[colback=gray!00,%gray background
                  colframe=c2,% black frame colour
                %   width=17.2cm,% Use 8cm total width,
                  arc=1.5mm, auto outer arc,
                  breakable,
                  left=0.9mm, right=0.9mm,
                  boxrule=0.9pt,
                  title = {The prompt for text comprehension generation (by GPT4).}
                 ]
\textbf{Prompt}\\
You are an experienced first year language teacher who knows the content of the \{\textit{filename}\} very well, and you can adapt the \{\textit{filename}\} homework exercise provided to 10 multiple-choice questions with four choices, A, B, C, and D, with exact answers, designed to determine whether students have basic judgement about the content of the text.  \\
Here provides \{\textit{filename}\} homework exercises: \{\textit{content}\}. You can modify the content of the homework exercise to make it a multiple-choice question with exact answer with instruction: \\
------\\
\#\# INSTRUCTION:\\
- The multiple-choice questions must include four options: A, B, C, and D, with one and only one correct answer.
\tcblower
\textbf{User:}\\
Please generate 10 multiple-choice questions that assess whether students have a basic understanding of the text content and FOLLOW THE FORMAT OF THE PROVIDED EXAMPLE STRICTLY. \\
------\\
\#\# EXAMPLE FORMAT:\\
Question 1 \\
Question: XXXXX \\
A: XXXXX \\
B: XXXXX \\
C: XXXXX \\
D: XXXXX \\
Correct Answer: X \\
Explanation: XXXXXXXX
\end{tcolorbox}

\end{center}
\begin{figure}[ht!]
\centering
\includegraphics[width=\textwidth]{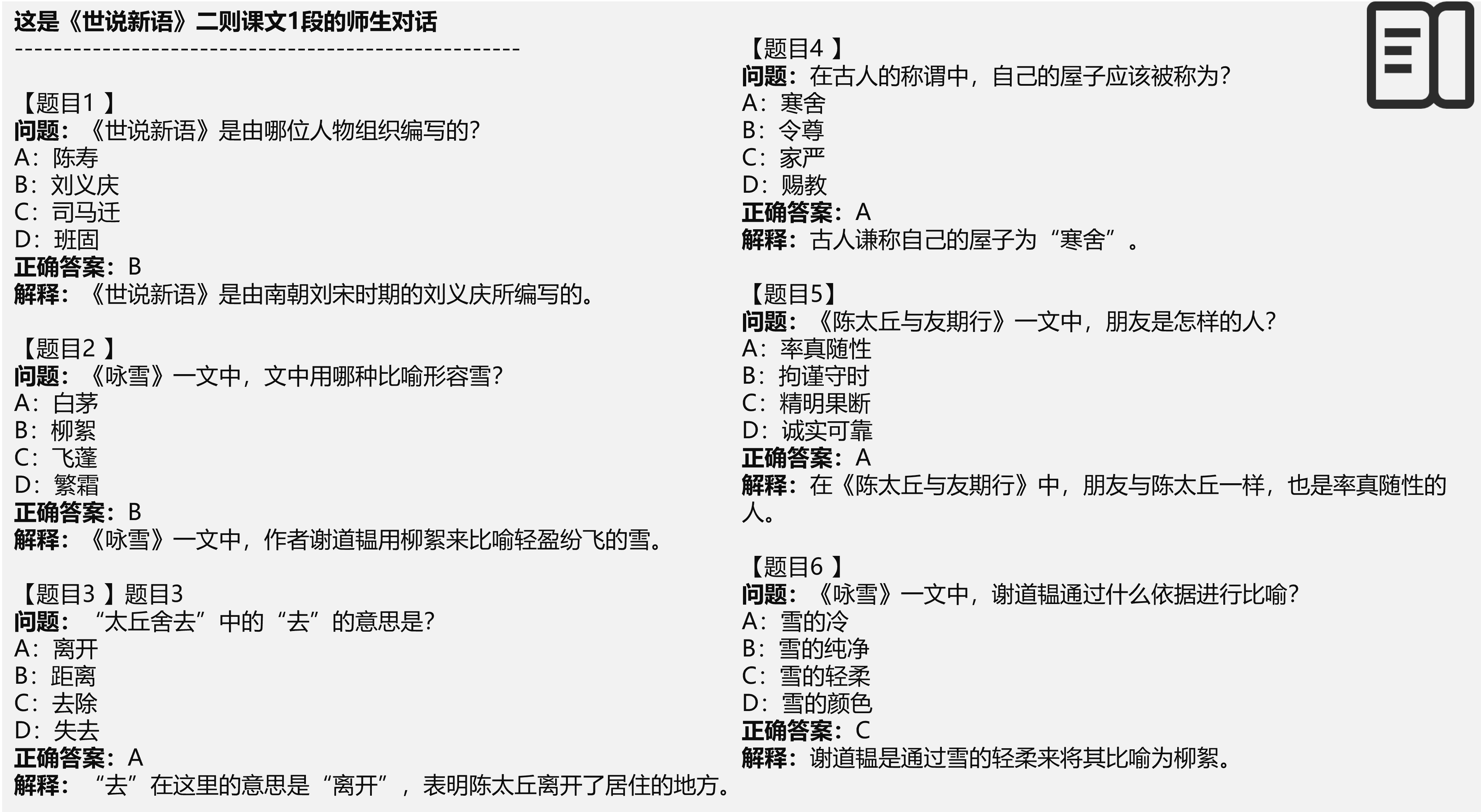}
\caption{An example of text comprehension (original Chinese version).}
\label{fig:text-comprehensive}
\end{figure}

\begin{tcolorbox}[colback=gray!00,%gray background
                  colframe=black,% black frame colour
                %   width=17.2cm,% Use 8cm total width,
                  arc=1.5mm, auto outer arc,
                  breakable,
                  left=0.9mm, right=0.9mm,
                  boxrule=0.9pt, colbacktitle = black!65!black,
                  % title = {An Example of Text Comprehensive.
                  title = {An example of text comprehension (translated in English for better understanding).} 
                 ]
Here is the English translation of the teacher-student dialogue from “Shishuo Xinyu”:
\tcblower
\textbf{[Question 1]}\\
\textbf{Question}: Who organized the compilation of “Shishuo Xinyu”?\\
A: Chen Shou\\
B: Liu Yiqing\\
C: Sima Qian\\
D: Ban Gu\\
\textbf{Correct Answer}: B\\
\textbf{Explanation}: “Shishuo Xinyu” was compiled by Liu Yiqing during the Southern Dynasties period of the Liu Song dynasty.\\

\textbf{[Question 2]}\\
\textbf{Question}: What metaphor is used to describe snow in “Ode to Snow”?\\
A: White grass\\
B: Willow catkins\\
C: Flying weeds\\
D: Abundant frost\\
\textbf{Correct Answer}: B\\
\textbf{Explanation}: In “Ode to Snow”, the author Xie Daoyun uses willow catkins to metaphorically describe the light and fluttering snow.\\

\textbf{[Question 3]}\\
\textbf{Question}: What does “qu” mean in "Taiqiu she qu"?\\
A: Leave\\
B: Distance\\
C: Remove\\
D: Lose\\
\textbf{Correct Answer}: A\\
\textbf{Explanation}: “qu” here means “leave”, indicating that Chen Taiqiu has left his place of residence.\\

\textbf{[Question 4]}\\
\textbf{Question}: What should one’s own house be called in ancient times?\\
A: Humble abode\\
B: Your esteemed father\\
C: Family honor\\
D: Teach me\\
\textbf{Correct Answer}: A\\
\textbf{Explanation}: In ancient times, people humbly referred to their own house as “humble abode”.\\

\textbf{[Question 5]}\\
\textbf{Question}: What kind of person is the friend in "Chen Taiqiu and His Friend's Journey"?\\
A: Genuine and easygoing\\
B: Reserved and punctual\\
C: Astute and decisive\\
D: Honest and reliable\\
\textbf{Correct Answer}: A\\
\textbf{Explanation}: In "Chen Taiqiu and His Friend's Journey," the friend is also a genuine and easygoing person, like Chen Taiqiu.\\

\textbf{[Question 6]}\\
\textbf{Question}: What basis does Xie Daoyun use for the metaphor in "Ode to Snow"?\\
A: The coldness of snow\\
B: The purity of snow\\
C: The softness of snow\\
D: The color of snow\\
\textbf{Correct Answer}: C\\
\textbf{Explanation}: Xie Daoyun uses the softness of snow to metaphorically compare it to willow catkins.
\end{tcolorbox}

\begin{center}
\begin{tcolorbox}[colback=gray!00,%gray background
                  colframe=c2,% black frame colour
                %   width=17.2cm,% Use 8cm total width,
                  arc=1.5mm, auto outer arc,
                  breakable,
                  left=0.9mm, right=0.9mm,
                  boxrule=0.9pt,
                  title = {The prompt for text memorization generation (by GPT4).}
                 ]
\textbf{Prompt}\\
You are an experienced middle school Chinese language teacher with extensive knowledge of the content \{\textit{filename}\}. You have been provided with the recitation passages from the text. Your task is to create ten multiple-choice questions, each with four options (A, B, C, and D), and only one correct answer. These questions are intended to assess students' ability to recall specific content from the text. \\
Here is the recitation passage from \{\textit{filename}\}: \{\textit{content}\}. Based on the original text, please create the multiple-choice questions with the instructions: \\
------\\
\#\# INSTRUCTION:\\
- The multiple-choice questions must include four options: A, B, C, and D, with one and only one correct answer.\\
- These questions must be designed to evaluate students' ability to recall specific content from the passage.\\
- Ensure that each question is based on the original text without rewriting any part of it.\\
- You may use \{\} for students to fill in with words, phrases, or sentences directly from the original text, or ask them to translate specific words, phrases, or sentences from the passage.
\tcblower
\textbf{User:}\\
Please generate 10 multiple-choice questions with four options (A, B, C, D) and one correct answer based on the recitation passage from \{\textit{filename}\}. \\
Ensure that no question is repeated and follow the example format provided: \\
------\\
\#\# EXAMPLE FORMAT:\\
Question 1 \\
Question: XXXXX, \{\}XXXX \\
A: XXXXX \\
B: XXXXX \\
C: XXXXX \\
D: XXXXX \\
Correct Answer: X \\
Explanation: XXXXXXXX
\end{tcolorbox}
\end{center}

\begin{figure}[ht!]
\centering
\includegraphics[width=\textwidth]{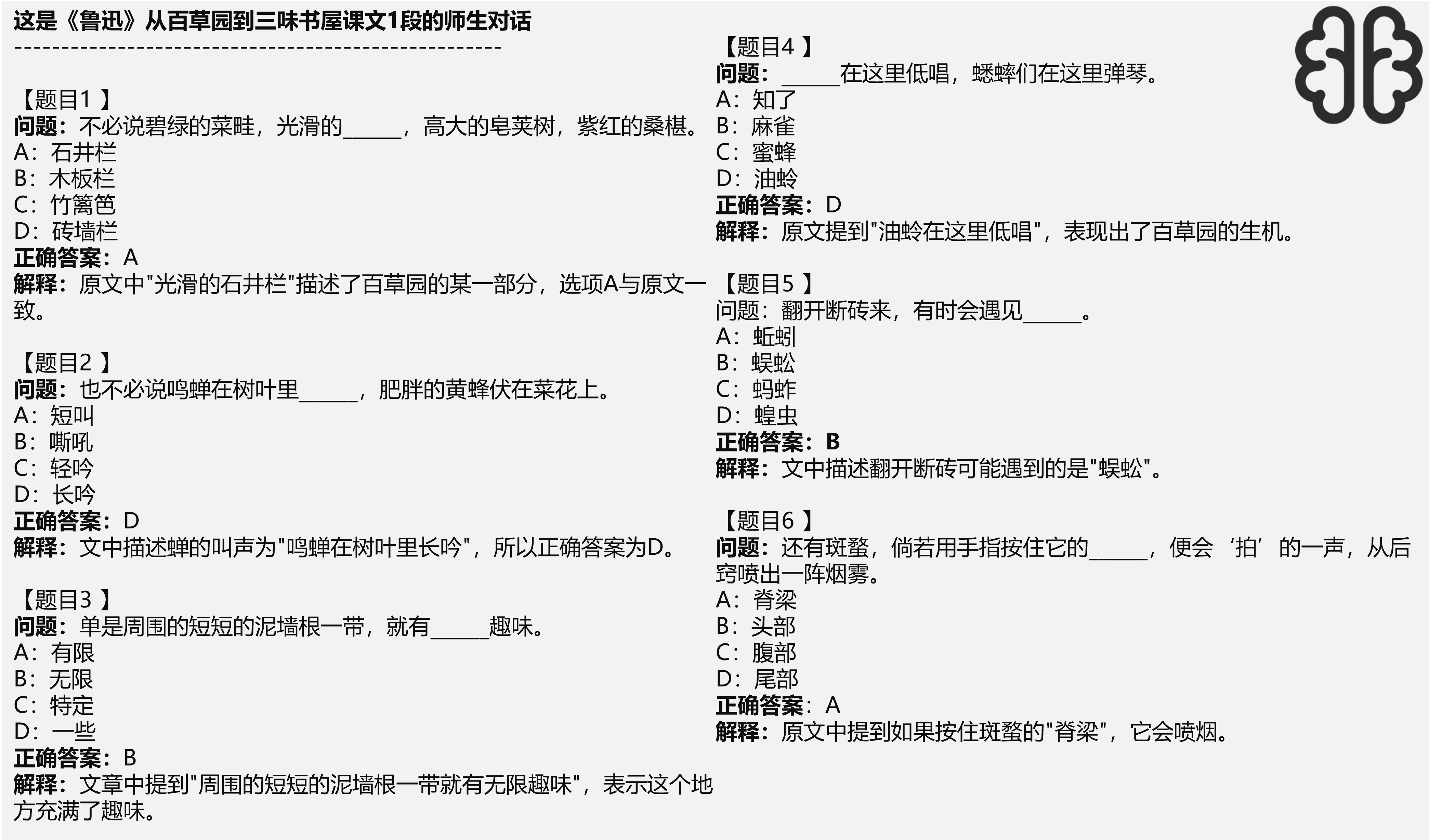}
\caption{An example of text memorization (original Chinese version).}
\label{fig:memorization-ability}
\end{figure}

\begin{tcolorbox}[colback=gray!00,%gray background
                  colframe=black,% black frame colour
                %   width=17.2cm,% Use 8cm total width,
                  arc=1.5mm, auto outer arc,
                  breakable,
                  left=0.9mm, right=0.9mm,
                  boxrule=0.9pt, colbacktitle = black!65!black,
                  title = {An example of text memorization (translated in English for better understanding).}
                 ]
Here is the English translation of the teacher-student dialogue from "From the Hundred Herb Garden to the Three Flavor Study Room" by Lu Xun:
\tcblower
\textbf{[Question 1]}\\
\textbf{Question}: Not to mention the green vegetable plots, the smooth \{\}, the tall soap pod tree, and the purple-red mulberries.\\
A: Stone well railing\\
B: Wooden fence\\
C: Bamboo fence\\
D: Brick wall railing\\
\textbf{Correct Answer}: A\\
\textbf{Explanation}: In the original text, "smooth stone well railing" describes a part of the Hundred Herb Garden, and option A is consistent with the text.\\

\textbf{[Question 2]}\\
\textbf{Question}: Nor need we mention the cicadas \{\} in the leaves, and the fat wasps resting on the vegetable flowers.\\
A: Short chirping\\
B: Roaring\\
C: Soft humming\\
D: Long humming\\
\textbf{Correct Answer}: D\\
\textbf{Explanation}: The text describes the sound of cicadas as "long humming in the leaves," so the correct answer is D.\\

\textbf{[Question 3]}\\
\textbf{Question}: Just around the short mud walls, there is \{\} interest.\\
A: Limited\\
B: Infinite\\
C: Specific\\
D: Some\\
\textbf{Correct Answer}: B\\
\textbf{Explanation}: The text states that "around the short mud walls there is infinite interest," indicating that this place is full of charm.\\

\textbf{[Question 4]}\\
\textbf{Question}: \{\} sings softly here, while crickets play music.\\
A: Cicada\\
B: Sparrow\\
C: Bee\\
D: Oil beetle\\
\textbf{Correct Answer}: D\\
\textbf{Explanation}: The original text mentions that "the oil beetle sings softly here," showcasing the vitality of the Baicao Garden.\\

\textbf{[Question 5]}\\
\textbf{Question}: When you lift a broken brick, you may sometimes encounter \{\}.\\
A: Earthworm\\
B: Centipede\\
C: Grasshopper\\
D: Locust\\
\textbf{Correct Answer}: B\\
\textbf{Explanation}: The text describes that lifting a broken brick may reveal a "centipede."\\

\textbf{[Question 6]}\\
\textbf{Question}: There are also oil beetles; if you press its \{\} with your finger, it will 'pop' and release a cloud of smoke from its rear.\\
A: Backbone\\
B: Head\\
C: Abdomen\\
D: Tail\\
\textbf{Correct Answer}: D\\
\textbf{Explanation}: The original text states that pressing the oil beetle's "backbone" will make it emit smoke.\\
\end{tcolorbox}

\subsection{Fine-Tuning Dataset}
\label{subsec:FT-dataset}

\subsubsection{Data source}
\label{subsubsec:ft-data}
The fine-tuning dataset used in this work aligns with the Basic Chinese Understanding Ability Evaluation Dataset in previous section, drawing from real classroom video recordings, textbook content, and lesson plans prepared by teachers. The diversity and credibility of these data sources ensure that the large language model can generate responses grounded in authentic classroom interactions and instructional content, accurately reflecting real-world teaching scenarios.

\textbf{Real Classroom Videos}: These videos capture genuine interactions between teachers and students, including questioning, responses, and feedback. By transcribing and analyzing these videos, we extract teacher-student dialogue data from various classroom scenarios, enabling the large language model to learn real classroom communication patterns. This ensures that virtual students' linguistic behavior closely mirrors the interactional characteristics of actual classrooms.

\textbf{Textbook Content}: Textbooks are the core material for classroom instruction, encompassing subject knowledge and educational objectives that students are expected to master. Utilizing textbook data ensures that virtual students demonstrate a solid understanding of relevant subject matter and align with instructional goals when generating responses. This provides a knowledge-based foundation for logical and scientifically accurate responses.

\textbf{Lesson Plans}: Lesson plans outline the objectives and content of each teaching session and detail the design of each instructional phase. This framework guides GPT-4 in generating different types of questions, allowing virtual students to adapt to diverse instructional contexts accurately. The instructional design embedded in lesson plans ensures that the generated questions align with specific teaching objectives, leading virtual students to produce appropriate responses at different stages. Fine-tuning based on high-quality teacher lesson plans enhances the relevance and effectiveness of virtual student interactions in classroom settings.

\subsubsection{Data construction process}
\label{subsubsec:ft-construction}

The construction process of the fine-tuning dataset involves several critical steps to ensure the generated virtual student dialogues are highly realistic and personalized. These steps include data preparation, prompt design based on real teacher-student dialogues, small-scale expert revision, large-scale generation of Big Five personality-based dialogues, and the creation of instruction fine-tuning datasets. By integrating in-context learning prompt techniques, the model’s ability to generate dialogues that accurately reflect the distinct personalities of students is significantly enhanced.

\textbf{Data Preparation}: Representative Chinese language texts for middle school were selected from the National Smart Education Resource Platform, encompassing eight genres from the first-year curriculum. Corresponding real classroom videos were transcribed to capture key teacher-student interactions, such as questioning, responses, and summaries. The extracted dialogue data served as valuable linguistic material for fine-tuning, ensuring that virtual students could generate language that mirrors authentic classroom interactions.

\textbf{Prompt Design Based on Real Teacher-Student Dialogues}: To model dialogues reflecting the Big Five personality traits, we employed few-shot in-context learning prompts based on real classroom interactions. This method provided the model with example dialogues, guiding it to generate new, personality-aligned exchanges between teachers and students. These prompts enabled the model to produce realistic dialogues that reflect different personality traits, enhancing the personalized behavior of virtual students.

\textbf{Small-Scale Expert Revision}: Two experienced middle school teachers, each with over two years of teaching experience, were invited to review the generated sample data. Their expertise ensured that the student’s language style was aligned with the described personality traits. Any inconsistencies were corrected, further refining the quality of the data and providing a scientifically robust template for subsequent large-scale dialogue generation.

\textbf{Large-Scale Generation of Big Five Personality Teacher-Student Dialogue Data}: Using the expert-reviewed samples, we refined the prompt design for large-scale generation. Few-shot in-context learning prompts were used to generate teacher-student dialogues for different personality types across teaching stages, such as pre-lesson introduction and knowledge consolidation. The resulting dataset covers diverse Chinese language content and showcases distinctive classroom behaviors based on different personality traits, forming a comprehensive resource for fine-tuning.

\textbf{Instruction Fine-Tuning Dataset Construction}: Five separate instruction fine-tuning datasets were created, corresponding to the Big Five personality traits—extraversion, agreeableness, conscientiousness, neuroticism, and openness. Each dataset includes fields (\textbf{system}, \textbf{query}, and \textbf{response}) that define the teaching environment, teacher’s questions, and student responses. These fields ensure that virtual students generate appropriate and personality-reflective responses across different teaching stages.

\begin{center}
\begin{tcolorbox}[colback=gray!00,%gray background
                  colframe=c2,% black frame colour
                %   width=17.2cm,% Use 8cm total width,
                  arc=1.5mm, auto outer arc,
                  breakable,
                  left=0.9mm, right=0.9mm,
                  boxrule=0.9pt,
                  title = {The prompt for big five teacher-student dialogue examples based on real teacher-student dialogues (by GPT-4).}
                 ]
\textbf{Prompt}\\
You are an expert Chinese language teacher, well-versed in the characteristics and response styles of first-year middle school students. Here providing you with the teacher-student dialogue examples from real classroom interactions, as well as a teaching plan for the lesson from \{\textit{filename}\}. Please generate five teacher-student dialogues based on the examples and the teaching plan, specifically targeting a student with the personality trait of \{\textit{student personality}\}.  \\
Here is the teaching plan for \{\textit{filename}\}. Please generate teacher-student dialogues according to the following instructions: \\
------\\
\#\# INSTRUCTION:\\
- The generated dialogues should align with the personality characteristics of a student with {student personality}, as described: \{{\textit{student personality description}}\} (student detail description see ~\ref{tab:bigfive}).\\
- The content of the dialogues must correspond to the topics covered in the provided teaching plan.
\tcblower
\textbf{User:}\\
Below are examples of teacher-student dialogues from real classroom interactions. Please explain step by step, using a chain-of-thought reasoning process, why you generated the teacher-student dialogue for a student with \{\textit{student personality}\}: \\
------\\
\#\# REAL-CLASS DIALOGUES:\\
Dialogue 1:\\
Teacher: Now, could one of you share with me a word that comes to mind?\\
Student: Magnificent.\\

Dialogue 2:\\
Teacher: Now, could one of you tell us what kind of person you think Cao Cao is, based on what you’ve seen in television dramas?\\
Student: He is straightforward and very talented.\\

Dialogue 3:\\
Teacher: The first question—who can tell us what the emotional tone of the article is?\\
Student: This article expresses the author's deep affection for spring.\\

Dialogue 4:\\
Teacher: What is the central character around which the article develops?\\
Student: The character 'view' is central to the article's development.\\

Dialogue 5:\\
Teacher: What did you see? Can anyone share with us?\\
Student: I saw the lake, uh, and the peach trees, the peach, peach, peach, apricot, and pear trees all blooming at once.\\

Dialogue 6:\\
Teacher: What indicates this?\\
Student: The landscape.\\

Dialogue 7:\\
Teacher: What does the word 'tranquil' mean?\\
Student: Tranquil, um, it means very peaceful.\\

Dialogue 8:\\
Teacher: So, what role do the last two paragraphs play?\\
Student: I think they summarize the entire piece and reinforce the author’s emotional attachment to the rain.
\end{tcolorbox}
\end{center}

\begin{figure}[ht!]
\centering
\includegraphics[width=\textwidth]{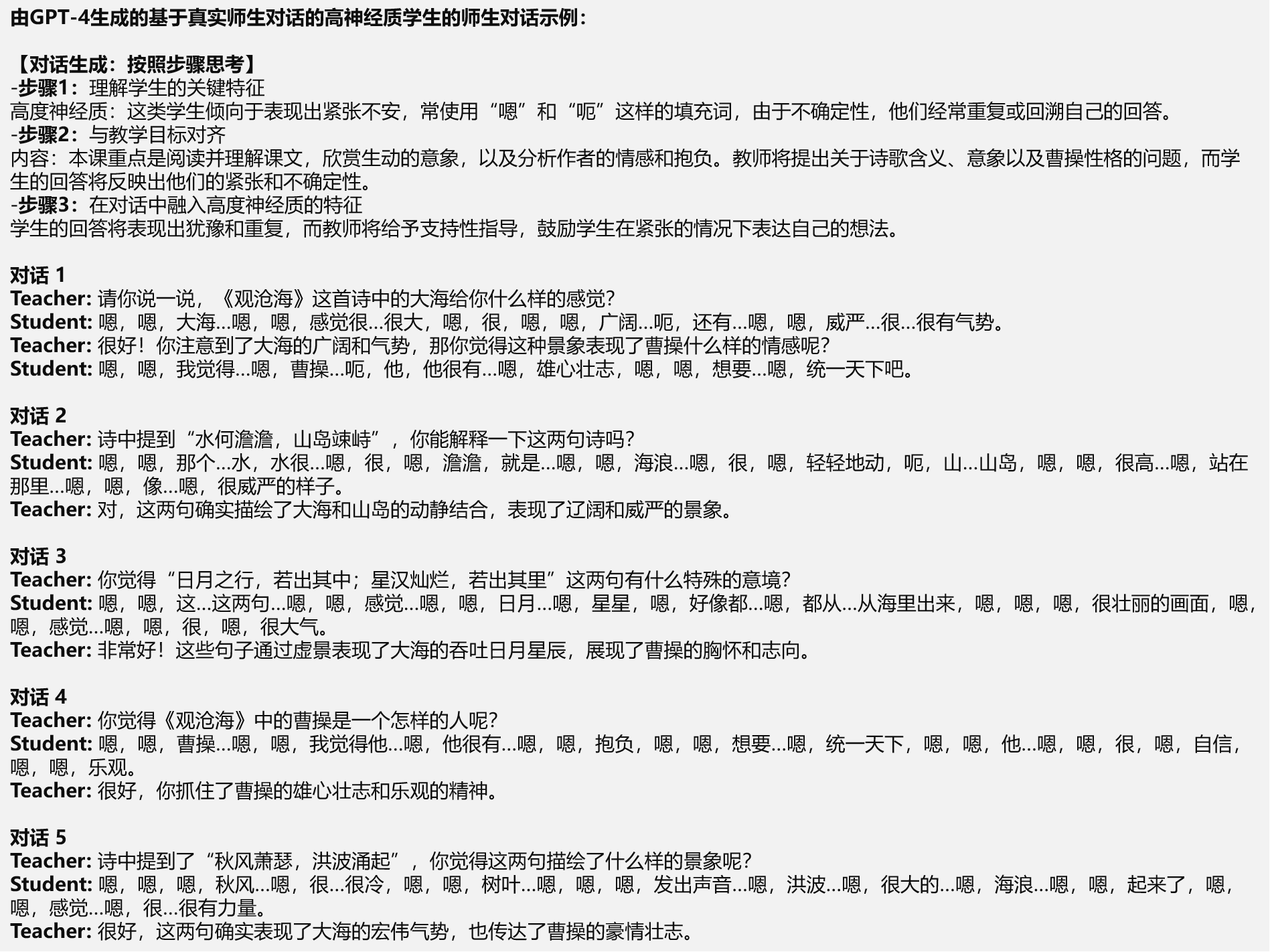}
\caption{An example of HN LVSA teacher-student dialogue generated by GPT-4 (original Chinese version).}
\label{fig:gpt_noexpert}
\end{figure}

\begin{tcolorbox}[colback=gray!00,%gray background
                  colframe= black,% black frame colour
                %   width=17.2cm,% Use 8cm total width,
                  arc=1.5mm, auto outer arc,
                  breakable,
                  left=0.9mm, right=0.9mm,
                  boxrule=0.9pt, colbacktitle = black!65!black,
                  title = {An example of HN LVSA teacher-student dialogue generated by GPT-4 (translated in English for better understanding).} 
                 ]
Based on the teaching plan for "View the Sea" by Cao Cao and the characteristics of a highly neurotic student, Here are five teacher-student dialogues that reflect the described classroom dynamics:
\tcblower
\textbf{[Dialogue Generation: Step-by-Step Chain of Thought]}\\
- \textbf{Step 1} : Understand the Key Characteristics of the Student\\
\textbf{High neuroticism}: The student is prone to anxiety, uses filler words like "en" and "um," and exhibits repetition, nervousness, and backtracking in their speech.\\
- \textbf{Step 2}: Align Dialogue with Teaching Objectives\\
\textbf{Content}: The lesson focuses on reading and understanding the text, appreciating the vivid imagery, and analyzing the author's emotions and ambitions. The teacher will ask questions about the meaning of the poem, the imagery, and Cao Cao's character, while the student’s responses will reflect their nervousness and uncertainty.\\
- \textbf{[Step 3]}: Incorporate High Neuroticism Traits into Dialogue\\
\textbf{Student Performance}: The student's responses will show hesitation and repetition, while the teacher will provide supportive guidance, encouraging the student to express their ideas despite their nervousness.\\

\textbf{[Dialogue 1]}\\
\textbf{Teacher}: Can you tell us what kind of feeling the sea gives you in the poem "View the Sea"? \\
\textbf{Student}: Um, um, the sea… um, um, it feels very… very vast, um, very, um, um, wide… uh, and also… um, um, majestic… very… very powerful. \\
\textbf{Teacher}: Good! You noticed the vastness and majesty of the sea. What kind of emotions do you think Cao Cao is expressing through this scene? \\
\textbf{Student}: Um, um, I think… um, Cao Cao… uh, he, he has… um, great ambition, um, um, and he wants… um, to unify the whole country, I think. \\

\textbf{[Dialogue 2]}\\
\textbf{Teacher}: The poem mentions “The water is vast and billowing, and the islands stand tall.” Can you explain these two lines? \\
\textbf{Student}: Um, um, well… the water, the water is… um, um, vast, um, like… um, the waves… um, are gently moving, uh, and the islands… um, um, they stand tall… um, like they are… um, very majestic. \\
\textbf{Teache}r: Right, these lines indeed describe the combination of movement and stillness between the sea and the islands, reflecting the vast and majestic scenery.\\

\textbf{[Dialogue 3]}\\
\textbf{Teacher}: What kind of special imagery do you think is conveyed in the lines “The sun and moon travel across the sky; the Milky Way shines brightly, as if emerging from the sea”? \\
\textbf{Student}: Um, um, these… these two lines… um, um, they give a feeling… um, um, of the sun and moon… um, and the stars, um, as if they all… um, come out… from the sea, um, um, it’s a very magnificent picture, um, um, and it feels… um, um, very grand. \\
\textbf{Teacher}: Excellent! These lines use imagination to depict the sea as swallowing up the sun, moon, and stars, showcasing Cao Cao’s vast ambitions and vision. \\

\textbf{[Dialogue 4]}\\
\textbf{Teacher}: What kind of person do you think Cao Cao is in "View the Sea"? \\
\textbf{Student}: Um, um, Cao Cao… um, um, I think he… um, he has… um, great ambition, um, um, and he wants… um, to unify the country, um, um, and he… um, um, is very… um, confident, um, um, and optimistic. \\
\textbf{Teacher}: Very good! You’ve captured Cao Cao’s great ambition and optimistic spirit. \\

\textbf{[Dialogue 5]}\\
\textbf{Teacher}: The poem mentions “The autumn wind is bleak, and great waves rise.” What kind of scene do you think these lines depict? \\
\textbf{Student}: Um, um, um, the autumn wind… um, it’s… it’s cold, um, um, the leaves… um, um, um, are making sounds… um, and the great waves… um, they rise… um, um, and it feels… um, very… very powerful. \\
\textbf{Teacher}: Great! These lines indeed convey the grandeur of the sea and also reflect Cao Cao’s heroic ambitions.
\end{tcolorbox}

\begin{tcolorbox}[colback=gray!30,%gray background
                  colframe= gray!80,% black frame colour
                %   width=17.2cm,% Use 8cm total width,
                  arc=1.5mm, auto outer arc,
                  breakable,
                  left=0.9mm, right=0.9mm,
                  boxrule=0.9pt, colbacktitle = black!65!black,
                  title = {An example of HN LVSA teacher-student dialogue generated by GPT-4 (Modified by Experts).} 
                 ]
Based on the high neuroticism teacher-student dialogue examples generated by GPT-4, the following are the teacher-student dialogues revised by experts according to real teaching scenarios:
\tcblower
\textbf{[Dialogue 1]}\\
\textbf{Teacher}: Good! You noticed the vastness and majesty of the sea. What kind of emotions do you think Cao Cao is expressing through this scene? \\
\textbf{Student}: Um, um, I think… um, Cao Cao… uh, he has… um, big goals, um, and he wants… um, to change everything, I think.\\

\textbf{[Dialogue 2]}\\
\textbf{Teacher}: The poem mentions “The water is vast and billowing, and the islands stand tall.” Can you explain these two lines? \\
\textbf{Student}: Um, the water… um, is big, um, and the waves… um, move softly, uh, and the islands… um, um, are tall… um, and look very strong. \\

\textbf{[Dialogue 3]}\\
\textbf{Teacher}: What kind of special imagery do you think is conveyed in the lines “The sun and moon travel across the sky; the Milky Way shines brightly, as if emerging from the sea”? \\
\textbf{Student}: Um, um, these… these two lines… um, um, they give a feeling… um, um, of the sun and moon… um, and the stars, um, as if they all… um, come out… from the sea, um, um, it’s a very magnificent picture, um, um, and it feels… um, um, very grand. \\

\textbf{[Dialogue 4]}\\
\textbf{Teacher}: What kind of person do you think Cao Cao is in "View the Sea"? \\
\textbf{Student}: Um, um, these… these two lines… um, um, feel like… um, the sun, moon, and stars… um, all come… from the sea, um, um, it’s a very big picture, um, um, and it feels… um, um, really great. \\

\textbf{[Dialogue 5]}\\
\textbf{Teacher}: The poem mentions “The autumn wind is bleak, and great waves rise.” What kind of scene do you think these lines depict? \\
\textbf{Student}: Um, um, the autumn wind… um, it’s cold, um, um, the leaves… um, are making noise… um, and the big waves… um, are rising… um, and it feels… um, very strong. \\
\end{tcolorbox}

\begin{figure}[ht!]
\centering
\includegraphics[width=\textwidth]{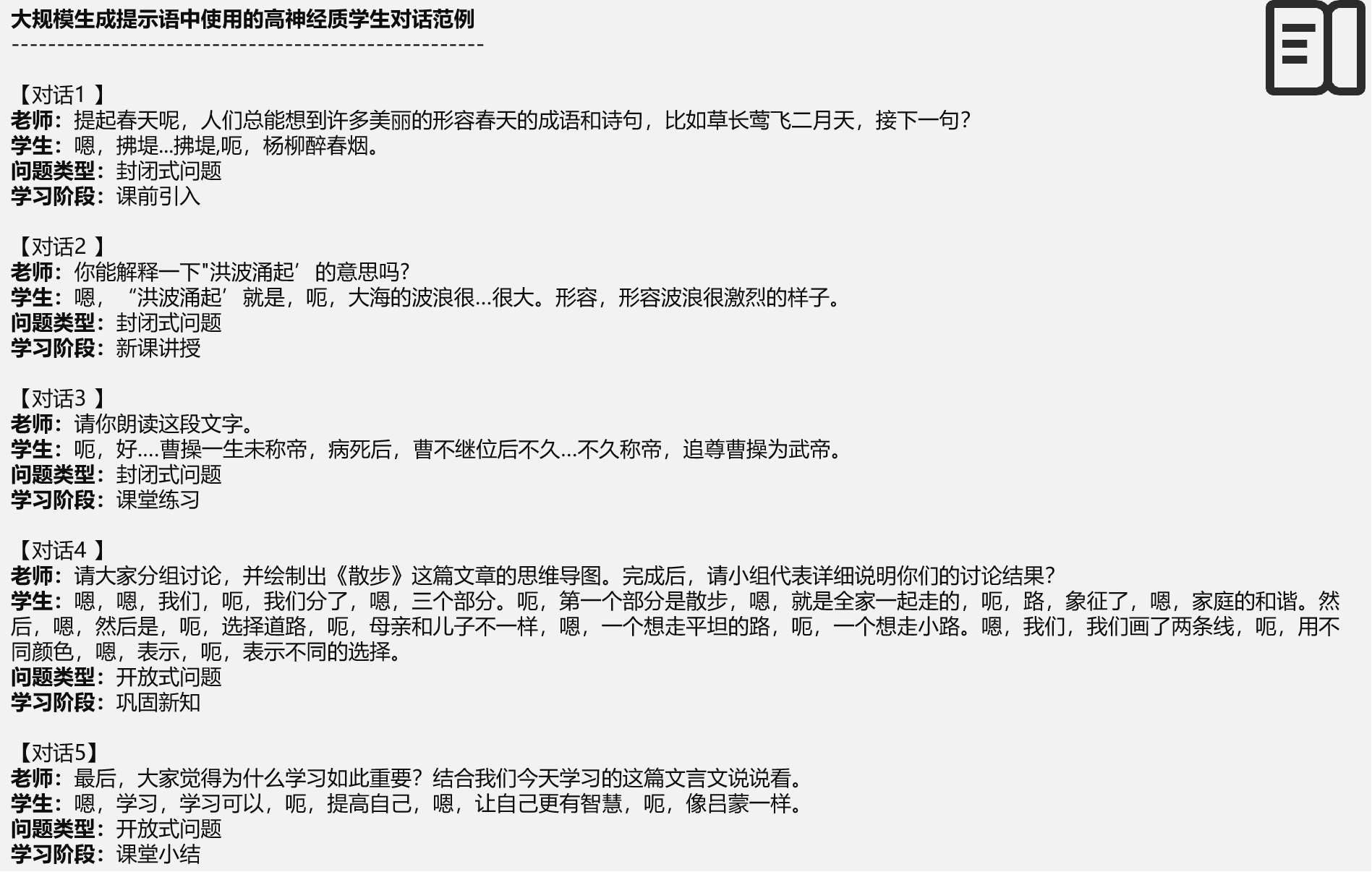}
\caption{HN LVSA example dialogues used in large-scale generation prompt (original Chinese version).}
\label{fig:student few-shot}
\end{figure}

\begin{tcolorbox}[colback=gray!00,%gray background
                  colframe= black,% black frame colour
                %   width=17.2cm,% Use 8cm total width,
                  arc=1.5mm, auto outer arc,
                  breakable,
                  left=0.9mm, right=0.9mm,
                  boxrule=0.9pt, colbacktitle = black!65!black,
                  title = {HN LVSA example dialogues used in large-scale generation prompt (translated in English for better understanding).} 
                 ]
After experts revised all of the Big Five personality teacher-student dialogue examples, five dialogues were selected from each personality type as prompt few-shot for large-scale generation of Big Five personality dialogues including learning stages. Below are teacher-student dialogue few-shots for the high neuroticism personality type:
\tcblower
\textbf{[Dialogue 1]}\\
\textbf{Teacher}: When we talk about spring, people always associate it with many beautiful images of spring scenery and poetry, like "The grass grows long in February, the swallows fly by," followed by another line? \\
\textbf{Student}: Um, "The willows sway in the breeze, the willows drunk on the spring haze."\\
\textbf{Question type}: Closed-ended question \\
\textbf{Learning stage}: Pre-lesson introduction \\

\textbf{[Dialogue 2]}\\
\textbf{Teacher}: Could you explain what "raging waves" means? \\
\textbf{Student}: Um, "raging waves" refers to... um, the waves are big... very big. It describes... the image of waves being very turbulent. \\
\textbf{Question type}: Closed-ended question \\
\textbf{Learning stage}: New lesson learning \\

\textbf{[Dialogue 3]}\\
\textbf{Teacher}: Cao Cao once said, "One cannot keep their strength forever, the wind and clouds will soon pass." Could anyone tell me what this means? \\
\textbf{Student}: Um, it means... uh, life is fleeting, nothing stays the same... um, and the wind... chasing clouds represents the passing of time. \\
\textbf{Question type}: Closed-ended question \\
\textbf{Learning stage}: Classroom exercises \\

\textbf{[Dialogue 4]}\\
\textbf{Teacher}: Please find this article and point out where it talks about "scattering." After that, I'd like the group to explain in detail how this part describes the scene. \\
\textbf{Student}: Um, um, we, uh, we divided it into, um, three parts. Uh, the first part is about walking, um, which represents the family walking together, uh, symbolizing, um, family harmony. Then, um, then there's, uh, the choice of the road, uh, the mother and son are different, um, one wants to take the smooth road, uh, the other wants to take the smaller path. Um, we, we drew two lines, uh, using different colors, um, to represent, uh, to represent the different choices.\\
\textbf{Question type}: Open-ended question \\
\textbf{Learning stage}: Knowledge consolidation \\

\textbf{[Dialogue 5]}\\
\textbf{Teacher}: Finally, why do you all think learning is so important? Use the classical text we studied today to explain your thoughts. \\
\textbf{Student}: Um, learning, learning can, uh, improve oneself, um, make one wiser, uh, like Lü Meng.\\
\textbf{Question type}: Open-ended question \\
\textbf{Learning stage}: Lesson summary

\end{tcolorbox}

\begin{center}
\begin{tcolorbox}[colback=gray!00,%gray background
                  colframe=c2,% black frame colour
                %   width=17.2cm,% Use 8cm total width,
                  arc=1.5mm, auto outer arc,
                  breakable,
                  left=0.9mm, right=0.9mm,
                  boxrule=0.9pt,
                  title = {The prompt for large-scaled generation with few-shot in-context learning(by GPT-4).}
                 ]
\textbf{Prompt}\\
You are an expert teacher who is well-versed in the personalities, language expression styles, and vocabulary levels of first-year junior high school students, as well as the content of \{\textit{filename}\}." With the provided lesson plan and examples of real classroom teacher-student dialogues, you will generate ten dialogues between a professional Chinese language teacher and a student with \{\textit{student personality}\}'s personality. These dialogues should include both closed-ended and open-ended questions, and the dialogues should span five stages: pre-lesson introduction, new lesson learning, knowledge consolidation, classroom exercises, and lesson summary. \\
Closed-ended questions should directly target the student’s recall and understanding of the content, such as asking about the author of the text or the student’s knowledge of classical poetry related to the theme of the text. Open-ended questions should focus on the student’s ability to apply, analyze, create, and evaluate the knowledge. \\
The stages are defined as follows: Pre-lesson introduction: Activities that the teacher undertakes before formal instruction to spark student interest, establish learning context, activate prior knowledge, or introduce new concepts, such as setting learning goals, establishing background knowledge, or activating prior knowledge. New lesson learning: The phase where the teacher introduces and explains new content or a new topic for the first time. This stage represents the students’ initial exposure to the subject matter. Knowledge Consolidation : Activities that help students deepen their understanding and retention of the new knowledge, such as review, practice, and discussion to ensure that the students firmly grasp the material. Classroom practice: Exercises and activities conducted in the classroom under the teacher’s guidance to help students understand, master, and apply the knowledge they have learned.
Lesson summary: A process that occurs after the learning activities, where the student or teacher reviews and organizes the learned content.\\
Here provided the lesson plan content: {\textit{content}} for "{\textit{filename}}," and please generate ten dialogues with the instructions: \\
------\\
\#\# INSTRUCTION:\\
- Please ensure that the generated dialogues align with {student}'s personality traits as described: \{\textit{student personality description}\}. (student detail description see ~\ref{tab:bigfive}).\\
- Note that the lesson plan is not the content of the text itself but the instructional process and methods used by the teacher. It serves as a guide for generating valuable questions in the teacher-student dialogues. \\
- The student’s responses to closed-ended questions should be concise and precise, and the vocabulary and sentence structures used should match the level of a first-year junior high school student. \\
- The ten teacher-student dialogues should cover the following stages: pre-lesson introduction, new lesson instruction, consolidation of new knowledge, classroom practice, and lesson summary. \\
- The student’s responses should not contain an excessive number of filler words. \\
- The content of the dialogues should closely follow the text "{\textit{filename}}." \\
\tcblower
\textbf{User:}\\
Below are examples of teacher-student dialogues for a student with {\textit{student personality}}'s personality. Please generate ten teacher-student dialogues following the examples format provided:\\
------\\
\#\# FEW-SHOT EXAMPLES: \{{\textit{student personality}} few-shot\} (few-shot see ~\ref{fig:student few-shot}).
\end{tcolorbox}
\end{center}

\begin{figure}[ht!]
\centering
\includegraphics[width=\textwidth]{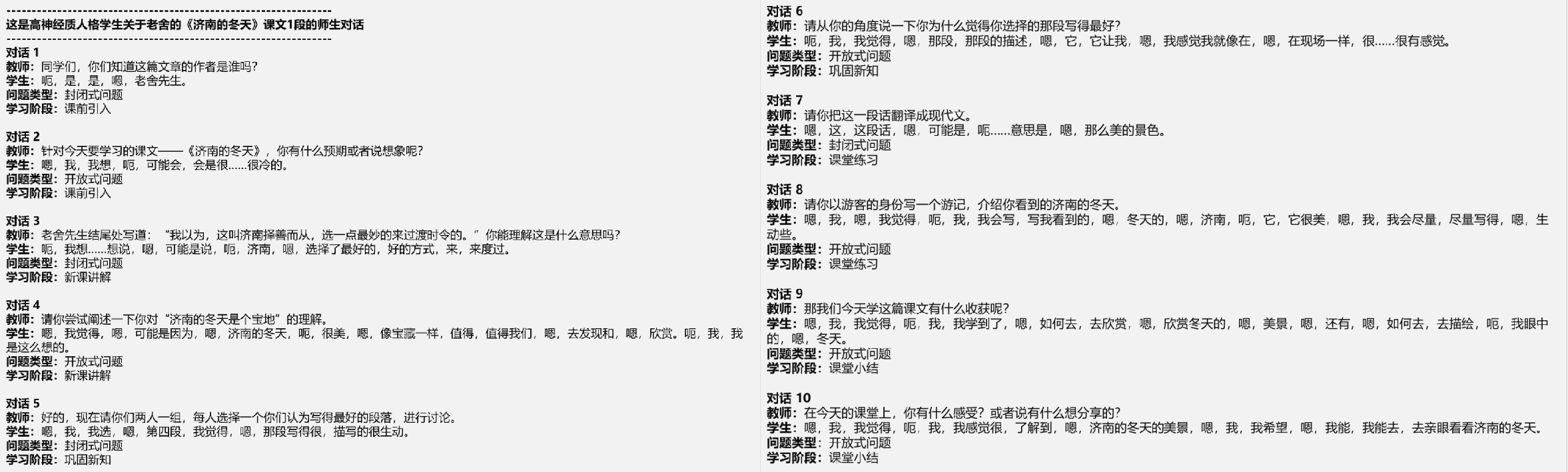}
\caption{An example of HN LVSA fine-tuning data generated by GPT-4 (original Chinese version).}
\label{fig:HN student finetune dia}
\end{figure}

\begin{tcolorbox}[colback=gray!00,%gray background
                  colframe=black,% black frame colour
                %   width=17.2cm,% Use 8cm total width,
                  arc=1.5mm, auto outer arc,
                  breakable,
                  left=0.9mm, right=0.9mm,
                  boxrule=0.9pt, colbacktitle = black!65!black,
                  % title = {An Example of Text Comprehensive.
                  title = {An example of HN LVSA fine-tuning data generated by GPT-4 (translated in English for better understanding).}
                 ]
Here are dialogues based on the lesson "Winter in Jinan" by Lao She:
\tcblower
\textbf{[Dialogue 1]}\\
\textbf{Teacher}: Class, do you know who wrote this essay? \\
\textbf{Student}: Uh, it's, it's, um, Mr. Lao She. \\
\textbf{Question Type}: Closed-ended question \\
\textbf{Learning Stage}: Pre-lesson introduction \\

\textbf{[Dialogue 2]}\\
\textbf{Teacher}: Before we start today's lesson on "Winter in Jinan", what are your expectations or thoughts about it? \\
\textbf{Student}: Um, I, I think, uh, maybe it will be... very, very cold. \\
\textbf{Question Type}: Open-ended question \\
\textbf{Learning Stage}: Pre-lesson introduction \\

\textbf{[Dialogue 3]}\\
\textbf{Teacher}: At the end of the essay, Lao She wrote, "I believe this is Jinan choosing the best way to transition through the seasons." Can you understand what this means? \\
\textbf{Student}: Uh, I think... think it means, um, maybe Jinan, um, chooses the best, the best way to, to go through it. \\
\textbf{Question Type}: Closed-ended question \\
\textbf{Learning Stage}: New lesson learning \\

\textbf{[Dialogue 4]}\\
\textbf{Teacher}: Can you try to explain your understanding of the phrase "Jinan's winter is a treasure"? \\
\textbf{Student}: Um, I think, um, maybe because, um, Jinan's winter, uh, is really beautiful, like a treasure, it's worth, worth us, um, discovering and, um, appreciating. Uh, that's, that's what I think. \\
\textbf{Question Type}: Open-ended question \\
\textbf{Learning Stage}: New lesson learning \\

\textbf{[Dialogue 5]}\\
\textbf{Teacher}: Now, in pairs, each of you choose the paragraph you think is written the best, and discuss it. \\
\textbf{Student}: Um, I, I choose, um, the fourth paragraph. I think, um, that part is described very, very vividly. \\
\textbf{Question Type}: Closed-ended question \\
\textbf{Learning Stage}: Knowledge consolidation \\

\textbf{[Dialogue 6]}\\
\textbf{Teacher}: Please explain from your perspective why you think the paragraph you chose is the best. \\
\textbf{Student}: Uh, I, I think, um, that part, that part's description, um, it, it made me feel, um, like I was, uh, right there, it was... it felt really real. \\
\textbf{Question Type}: Open-ended question \\
\textbf{Learning Stage}: Knowledge consolidation \\

\textbf{[Dialogue 7]}\\
\textbf{Teacher}: Can you translate this sentence into modern language? \\
\textbf{Student}: Um, this, this sentence, um, maybe it means, uh... something like, um, such beautiful scenery. \\
\textbf{Question Type}: Closed-ended question \\
\textbf{Learning Stage}: Classroom exercises \\

\textbf{[Dialogue 8]}\\
\textbf{Teacher}: Now imagine you're a tourist. Write a travel diary describing what you saw in Jinan's winter. \\
\textbf{Student}: Um, I, um, I think, uh, I, I would write about what I saw, um, in winter, um, in Jinan. Uh, it, it was really beautiful, um, I, I'll try to, try to write it, um, as vividly as I can. \\
\textbf{Question Type}: Open-ended question \\
\textbf{Learning Stage}: Classroom exercises \\

\textbf{[Dialogue 9]}\\
\textbf{Teacher}: So, what did we gain from studying this text today? \\
\textbf{Student}: Um, I, I think, uh, I, I learned, um, how to, to appreciate, um, the beauty of, um, winter, and, um, how to, to describe, uh, what I see in, um, winter. \\
\textbf{Question Type}: Open-ended question \\
\textbf{Learning Stage}: Lesson summary \\

\textbf{[Dialogue 10]}\\
\textbf{Teacher}: In today's lesson, how do you feel? Or is there something you'd like to share? \\
\textbf{Student}: Um, I, I think, uh, I, I feel like I, um, learned more about, um, the beauty of Jinan's winter, um, I, I hope, um, I can, I can go, and, um, see Jinan's winter with my own eyes. \\
\textbf{Question Type}: Open-ended question \\
\textbf{Learning Stage}: Lesson summary \\
\end{tcolorbox}

\begin{figure}[ht!]
\centering
\includegraphics[width=\textwidth]{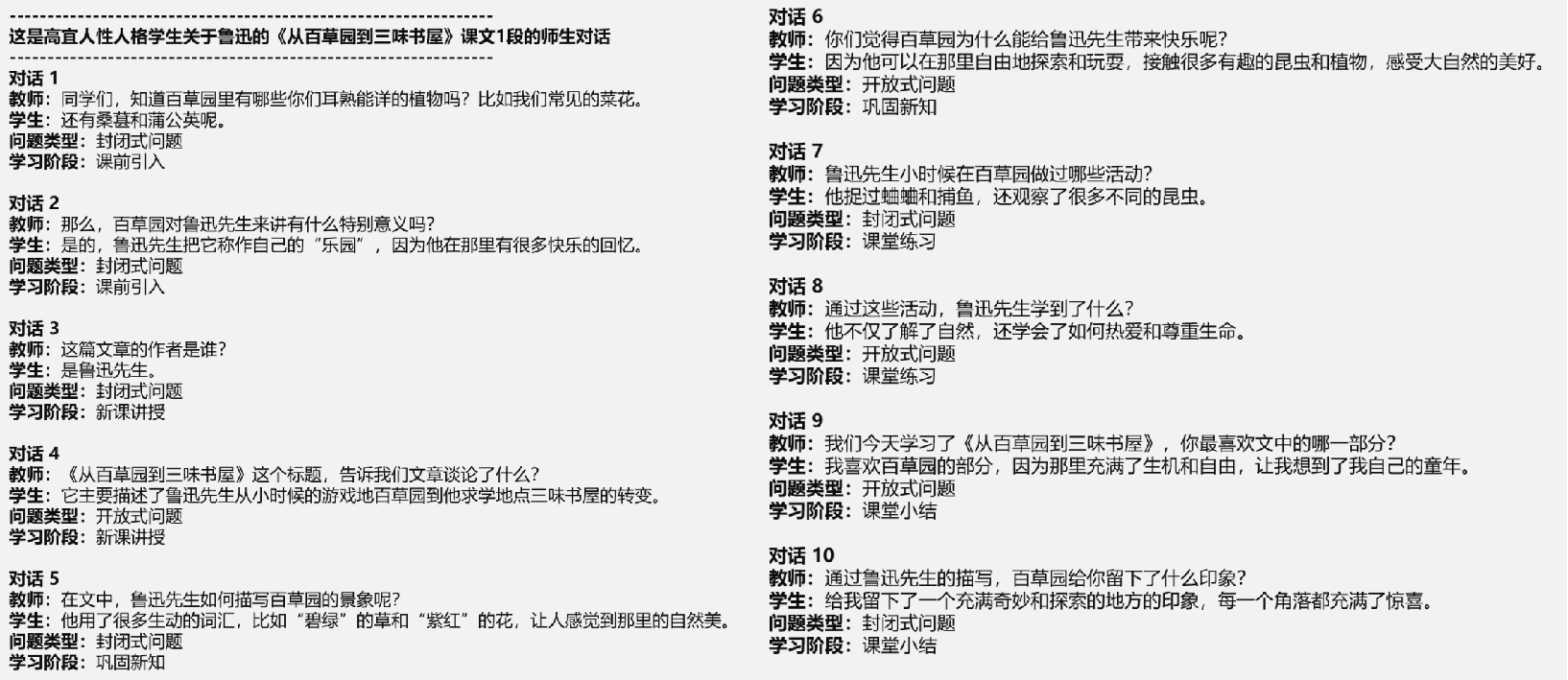}
\caption{An example of HA LVSA fine-tuning data generated by GPT-4 (original Chinese version).}
\label{fig:HA student finetune dia}
\end{figure}

\begin{tcolorbox}[colback=gray!00,%gray background
                  colframe=black,% black frame colour
                %   width=17.2cm,% Use 8cm total width,
                  arc=1.5mm, auto outer arc,
                  breakable,
                  left=0.9mm, right=0.9mm,
                  boxrule=0.9pt, colbacktitle = black!65!black,
                  % title = {An Example of Text Comprehensive.
                  title = {An example of HA LVSA fine-tuning data generated by GPT-4 (translated in English for better understanding).} 
                 ]
Here are dialogues based on the lesson "From the Hundred Grass Garden to the Three Flavor Study" by Lu Xun:
\tcblower
\textbf{[Dialogue 1]}\\
\textbf{Teacher}: Class, do you know what plants in the Hundred Grass Garden are familiar to you? Like the cabbage flowers we often see. \\
\textbf{Student}: There are also mulberries and dandelions. \\
\textbf{Question Type}: Closed-ended question \\
\textbf{Learning Stage}: Pre-lesson introduction \\

\textbf{[Dialogue 2]}\\
\textbf{Teacher}: So, does the Hundred Grass Garden have any special meaning for Mr. Lu Xun? \\
\textbf{Student}: Yes, Mr. Lu Xun called it his "paradise" because he had many happy memories there. \\
\textbf{Question Type}: Closed-ended question \\
\textbf{Learning Stage}: Pre-lesson introduction \\

\textbf{[Dialogue 3]}\\
\textbf{Teacher}: Who is the author of this text?
\textbf{Student}: It's Mr. Lu Xun.
\textbf{Question Type}: Closed-ended question
\textbf{Learning Stage}: New lesson learning \\

\textbf{[Dialogue 4]}\\
\textbf{Teacher}: The title "From the Hundred Grass Garden to the Three Flavor Study", what does it tell us about the content of the essay? \\
\textbf{Student}: It mainly talks about Mr. Lu Xun’s change from playing in the Hundred Grass Garden when he was young to studying at the Three Flavor Study. \\
\textbf{Question Type}: Open-ended question \\
\textbf{Learning Stage}: New lesson learning \\

\textbf{[Dialogue 5]}\\
\textbf{Teacher}: In the text, how does Mr. Lu Xun describe the scenery of the Hundred Grass Garden? \\
\textbf{Student}: He uses many vivid words, like the "green" grass and "purple-red" flowers, to make the natural beauty of the place feel alive. \\
\textbf{Question Type}: Closed-ended question \\
\textbf{Learning Stage}: Knowledge consolidation \\

\textbf{[Dialogue 6]}\\
\textbf{Teacher}: Why do you think the Hundred Grass Garden brought happiness to Mr. Lu Xun? \\
\textbf{Student}: Because he could freely explore and play there, see many interesting insects and plants, and enjoy the beauty of nature. \\
\textbf{Question Type}: Open-ended question \\
\textbf{Learning Stage}: Knowledge consolidation \\

\textbf{[Dialogue 7]}\\
\textbf{Teacher}: What activities did Mr. Lu Xun do in the Hundred Grass Garden when he was young? \\
\textbf{Student}: He caught crickets, fished, and observed many different insects. \\
\textbf{Question Type}: Closed-ended question \\
\textbf{Learning Stage}: Classroom exercises \\

\textbf{[Dialogue 8]}\\
\textbf{Teacher}: Through these activities, what did Mr. Lu Xun learn? \\
\textbf{Student}: He not only learned about nature but also how to love and respect life. \\
\textbf{Question Type}: Open-ended question \\
\textbf{Learning Stage}: Classroom exercises \\

\textbf{[Dialogue 9]}\\
\textbf{Teacher}: We studied "From the Hundred Grass Garden to the Three Flavor Study" today. What was your favorite part of the text? \\
\textbf{Student}: I liked the part about the Hundred Grass Garden because it was full of life and freedom, and it reminded me of my own childhood. \\
\textbf{Question Type}: Open-ended question \\
\textbf{Learning Stage}: Lesson summary \\

\textbf{[Dialogue 10]}\\
\textbf{Teacher}: What impression did the Hundred Grass Garden leave on you through Mr. Lu Xun’s description? \\
\textbf{Student}: It gave me the impression of a place full of wonder and exploration, where every corner was full of surprises. \\
\textbf{Question Type}: Closed-ended question \\
\textbf{Learning Stage}: Lesson summary \\
\end{tcolorbox}

\clearpage

\begin{figure}[t!]
\centering
\includegraphics[width=\textwidth]{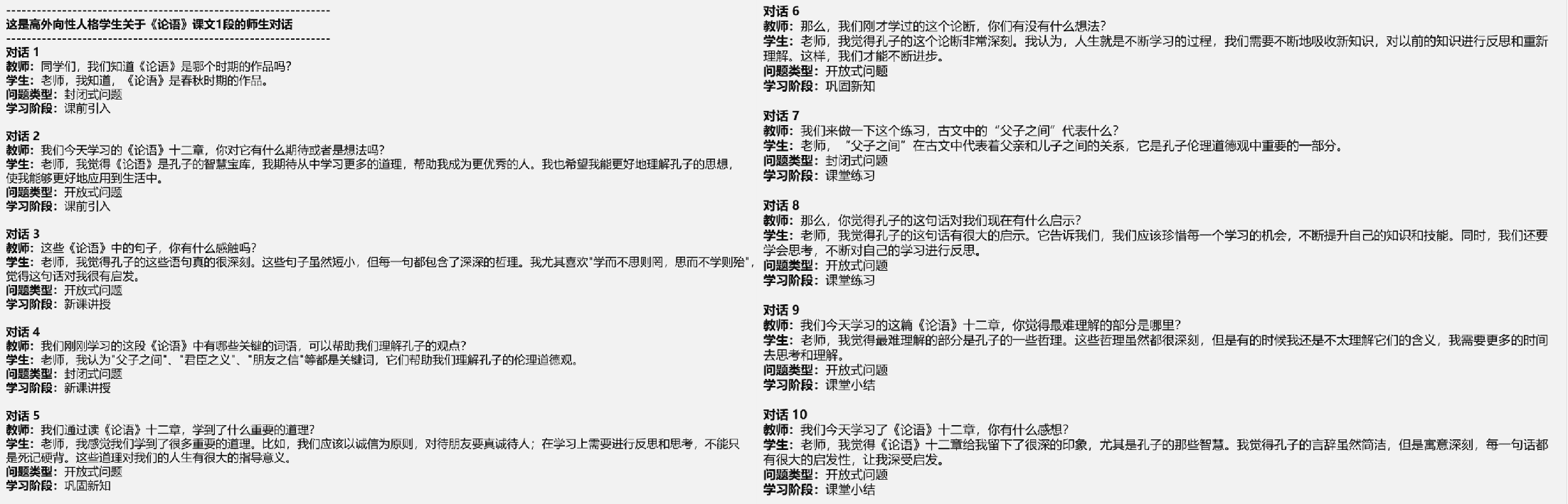}
\caption{An example of HE LVSA fine-tuning data generated by GPT-4 (original Chinese version).}
\label{fig:HE student finetune dia}
\end{figure}

\begin{tcolorbox}[colback=gray!00,%gray background
                  colframe=black,% black frame colour
                %   width=17.2cm,% Use 8cm total width,
                  arc=1.5mm, auto outer arc,
                  breakable,
                  left=0.9mm, right=0.9mm,
                  boxrule=0.9pt, colbacktitle = black!65!black,
                  title = {An example of HE LVSA fine-tuning data generated by GPT-4 (translated in English for better understanding).}
                 ]
Here are dialogues based on the lesson "LunYu":
\tcblower
\textbf{[Dialogue 1]}\\
\textbf{Teacher}: Do you know which time period The Analects is from?\\
\textbf{Student}: Teacher, I know, The Analects is from the Spring and Autumn period. \\
\textbf{Question Type}: Closed-ended question \\
\textbf{Learning Stage}: Pre-lesson introduction \\

\textbf{[Dialogue 2]}\\
\textbf{Teacher}: We're studying The Twelve Chapters of the Analects today. Do you have any expectations or thoughts about it? \\
\textbf{Student}: Teacher, I think The Analects is a treasure of Confucius's wisdom. I hope to learn more lessons from it to help me be a better person. I also want to better understand Confucius's ideas and apply them to my life.\\
\textbf{Question Type}: Open-ended question\\
\textbf{Learning Stage}: Pre-lesson introduction\\

\textbf{[Dialogue 3]}\\
\textbf{Teacher}: How do you feel about the sentences from The Analects we’ve read?\\
\textbf{Student}: Teacher, I think Confucius's words are really deep. Even though the sentences are short, each one is full of wisdom. I especially like, "Learning without thinking is useless, thinking without learning is dangerous." This really inspires me.\\
\textbf{Question Type}: Open-ended question\\
\textbf{Learning Stage}: New lesson learning\\

\textbf{[Dialogue 4]}\\
\textbf{Teacher}: What key words from the passage we just studied help us understand Confucius's ideas?\\
\textbf{Student}: Teacher, I think "between father and son," "the duty between ruler and subject," and "trust between friends" are key words. They help us understand Confucius’s moral values.\\
\textbf{Question Type}: Closed-ended question\\
\textbf{Learning Stage}: New lesson learning\\

\textbf{[Dialogue 5]}\\
\textbf{Teacher}: What important lessons have we learned from reading The Twelve Chapters of the Analects?
\textbf{Student}: Teacher, I feel like we’ve learned a lot of important lessons. For example, we should always be honest and sincere with our friends. Also, when studying, we need to reflect and think, not just memorize. These lessons are really helpful for our lives.
\textbf{Question Type}: Open-ended question
\textbf{Learning Stage}: Knowledge consolidation

\textbf{[Dialogue 6]}\\
\textbf{Teacher}: So, what do you think about the idea we just learned? \\
\textbf{Student}: Teacher, I think Confucius's idea is very deep. I believe that life is all about constantly learning. We need to keep gaining new knowledge and reflecting on what we already know. This way, we can keep getting better. \\
\textbf{Question Type}: Open-ended question \\
\textbf{Learning Stage}: Knowledge consolidation \\

\textbf{[Dialogue 7]}\\
\textbf{Teacher}: Let’s try this exercise. In classical Chinese, what does "between father and son" represent? \\
\textbf{Student}: Teacher, "between father and son" in classical Chinese represents the relationship between a father and his son. It's an important part of Confucius's view of ethics and morals. \\
\textbf{Question Type}: Closed-ended question \\
\textbf{Learning Stage}: Classroom exercises \\

\textbf{[Dialogue 8]}\\
\textbf{Teacher}: What do you think Confucius’s words teach us today? \\
\textbf{Student}: Teacher, I think Confucius’s words give us a lot of guidance. They tell us that we should value every chance to learn and keep improving our knowledge and skills. Also, we need to think and reflect on what we’re learning. \\
\textbf{Question Type}: Open-ended question \\
\textbf{Learning Stage}: Classroom exercises \\

\textbf{[Dialogue 9]}\\
\textbf{Teacher}: In today’s lesson, The Twelve Chapters of the Analects, what part did you find the hardest to understand? \\
\textbf{Student}: Teacher, I think the hardest part is some of Confucius’s philosophies. Even though they’re very deep, sometimes I don’t really understand their meaning. I need more time to think and understand them. \\
\textbf{Question Type}: Open-ended question \\
\textbf{Learning Stage}: Lesson summary \\

\textbf{[Dialogue 10]}\\
\textbf{Teacher}: After studying The Twelve Chapters of the Analects today, what are your thoughts? \\
\textbf{Student}: Teacher, The Twelve Chapters of the Analects left a deep impression on me, especially Confucius's wisdom. I think his words are simple but full of meaning, and each sentence is very inspiring. \\
\textbf{Question Typ}e: Open-ended question \\
\textbf{Learning Stage}: Lesson summary
\end{tcolorbox}

\begin{figure}[ht!]
\centering
\includegraphics[width=\textwidth]{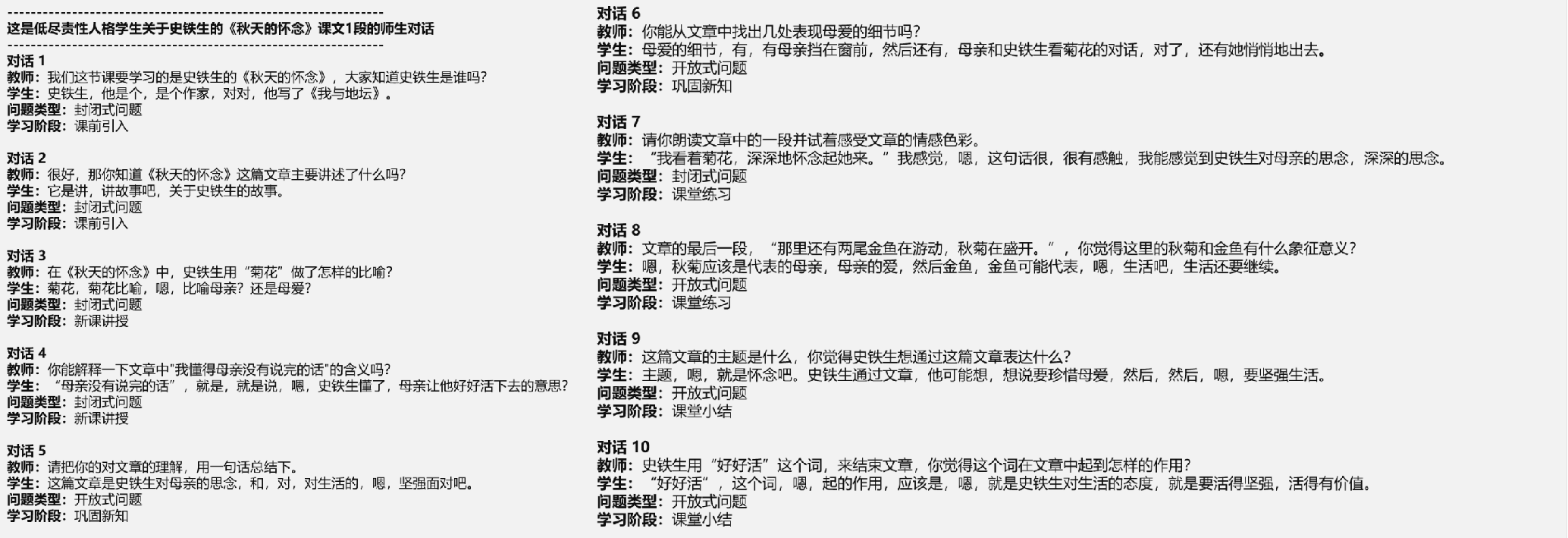}
\caption{An example of LC LVSA fine-tuning data generated by GPT-4 (original Chinese version).}
\label{fig:LC student finetune dia}
\end{figure}

\begin{tcolorbox}[colback=gray!00,%gray background
                  colframe=black,% black frame colour
                %   width=17.2cm,% Use 8cm total width,
                  arc=1.5mm, auto outer arc,
                  breakable,
                  left=0.9mm, right=0.9mm,
                  boxrule=0.9pt, colbacktitle = black!65!black,
                  % title = {An Example of Text Comprehensive.
                  title = {An example of LC LVSA fine-tuning data generated by GPT-4 (translated in English for better understanding).} 
                 ]
Here are dialogues based on the lesson "Autumn Thoughts" by Shi Tiesheng:
\tcblower
\textbf{[Dialogue 1]}\\
\textbf{Teacher}: Today, we’re going to study Shi Tiesheng’s Autumn Thoughts. Does anyone know who Shi Tiesheng is?\\
\textbf{Student}: Shi Tiesheng, he’s a, a writer, right, right, he wrote I and the Temple of Earth.\\
\textbf{Question Type}: Closed-ended question\\
\textbf{Learning Stage}: Pre-lesson introduction \\

\textbf{[Dialogue 2]}\\
\textbf{Teacher}: Good. So, do you know what Autumn Thoughts is mainly about?\\
\textbf{Student}: It's about, about a story, right? About Shi Tiesheng’s story.\\
\textbf{Question Type}: Closed-ended question\\
\textbf{Learning Stage}: Pre-lesson introduction \\

\textbf{[Dialogue 3]}\\
\textbf{Teacher}: In Autumn Thoughts, how does Shi Tiesheng use "chrysanthemums" as a metaphor?\\
\textbf{Student}: Chrysanthemums, uh, they’re a metaphor, um, for the mother? Or maybe mother’s love?\\
\textbf{Question Type}: Closed-ended question\\
\textbf{Learning Stage}: New lesson learning \\

\textbf{[Dialogue 4]}\\
\textbf{Teacher}: Can you explain the meaning behind the phrase "I understand the words my mother left unfinished"?\\
\textbf{Student}: "Words left unfinished," it means, um, Shi Tiesheng understood, that his mother wanted him to live well, right?\\
\textbf{Question Type}: Closed-ended question\\
\textbf{Learning Stage}: New lesson learning \\

\textbf{[Dialogue 5]}\\
\textbf{Teacher}: Can you summarize your understanding of the article in one sentence?\\
\textbf{Student}: This article is about Shi Tiesheng’s memory of his mother, and, uh, about, about facing life strongly, I guess.\\
\textbf{Question Type}: Open-ended question\\
\textbf{Learning Stage}: Knowledge consolidation \\

\textbf{[Dialogue 6]}\\
\textbf{Teacher}: Can you find a few details from the text that show motherly love?\\
\textbf{Student}: Details of motherly love, um, there’s the part where the mother stood by the window, and then, um, the conversation about chrysanthemums, and oh, when she quietly left.\\
\textbf{Question Type}: Open-ended question\\
\textbf{Learning Stage}: Knowledge consolidation \\

\textbf{[Dialogue 7]}\\
\textbf{Teacher}: Please read aloud a passage from the article and try to feel the emotion behind it.\\
\textbf{Student}: “I looked at the chrysanthemums and deeply missed her.” I feel, um, this sentence is very, very emotional, I can feel Shi Tiesheng’s deep longing for his mother.\\
\textbf{Question Type}: Closed-ended question\\
\textbf{Learning Stage}: Classroom exercises \\

\textbf{[Dialogue 8]}\\
\textbf{Teacher}: In the final paragraph, “The two goldfish are still swimming, and the chrysanthemums are blooming,” what do you think the chrysanthemums and the goldfish symbolize?\\
\textbf{Student}: Um, the chrysanthemums probably represent the mother, the mother’s love. And the goldfish, maybe they represent, uh, life, life continuing.\\
\textbf{Question Type}: Open-ended question\\
\textbf{Learning Stage}: Classroom exercises \\

\textbf{[Dialogue 9]}\\
\textbf{Teacher}: What is the theme of this article? What do you think Shi Tiesheng wanted to express?\\
\textbf{Student}: The theme, um, is remembrance. Shi Tiesheng, through the article, maybe he wanted to, to say that we should cherish motherly love, and, and, um, live strongly.\\
\textbf{Question Type}: Open-ended question\\
\textbf{Learning Stage}: Lesson summary \\

\textbf{[Dialogue 10]}\\
\textbf{Teacher}: Shi Tiesheng ends the article with the phrase “live well.” What do you think is the role of this phrase in the article?\\
\textbf{Student}: “Live well,” um, it means, I think, um, it’s about Shi Tiesheng’s attitude toward life, like, living strongly, and living with purpose.\\
\textbf{Question Type}: Open-ended question\\
\textbf{Learning Stage}: Lesson summary \\
\end{tcolorbox}

\begin{figure}[ht!]
\centering
\includegraphics[width=\textwidth]{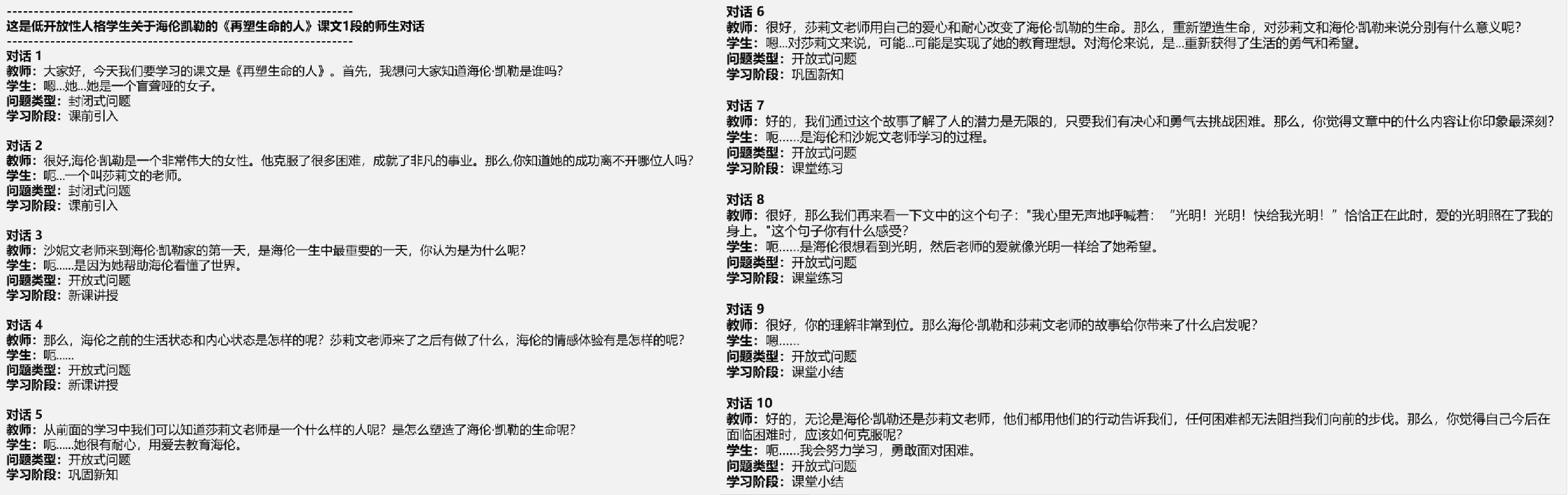}
\caption{An example of LO LVSA fine-tuning data generated by GPT-4 (original Chinese version).}
\label{fig:LO student finetune dia}
\end{figure}

\begin{tcolorbox}[colback=gray!00,%gray background
                  colframe=black,% black frame colour
                %   width=17.2cm,% Use 8cm total width,
                  arc=1.5mm, auto outer arc,
                  breakable,
                  left=0.9mm, right=0.9mm,
                  boxrule=0.9pt, colbacktitle = black!65!black,
                  title = {An example of LO LVSA fine-tuning data generated by GPT-4 (translated in English for better understanding).}
                 ]
Here are dialogues based on the lesson "The Story of My Life" by Helen Keller:
\tcblower
\textbf{[Dialogue 1]}\\
\textbf{Teacher}: Hello everyone, today we’re going to learn The Story of My Life by Helen Keller. First, can anyone tell me who Helen Keller was?\\
\textbf{Student}: Um... she... she was a blind and deaf woman.\\
\textbf{Question Type}: Closed-ended question\\
\textbf{Learning Stage}: Pre-lesson introduction \\

\textbf{[Dialogue 2]}\\
\textbf{Teacher}: Very good. Helen Keller was an extraordinary woman who overcame many difficulties and achieved great success. Now, do you know who helped her succeed?\\
\textbf{Student}: Uh... a teacher named Sullivan.\\
\textbf{Question Type}: Closed-ended question\\
\textbf{Learning Stage}: Pre-lesson introduction \\

\textbf{[Dialogue 3]}\\
\textbf{Teacher}: The first day that Miss Sullivan arrived at Helen Keller’s house was one of the most important days of her life. Why do you think that is?\\
\textbf{Student}: Um... because she helped Helen understand the world.\\
\textbf{Question Type}: Open-ended question\\
\textbf{Learning Stage}: New lesson learning \\

\textbf{[Dialogue 4]}\\
\textbf{Teacher}: What was Helen’s life and emotional state like before Miss Sullivan arrived? And what did Miss Sullivan do to change that? How did Helen feel afterward?\\
\textbf{Student}: Uh...\\
\textbf{Question Type}: Open-ended question\\
\textbf{Learning Stage}: New lesson learning \\

\textbf{[Dialogue 5]}\\
\textbf{Teacher}: From what we’ve learned so far, what kind of person do you think Miss Sullivan was? How did she shape Helen Keller’s life?\\
\textbf{Student}: Uh... she was really patient, and she taught Helen with love.\\
\textbf{Question Type}: Open-ended question\\
\textbf{Learning Stage}: Knowledge consolidation \\

\textbf{[Dialogue 6]}\\
\textbf{Teacher}: Very good. Miss Sullivan changed Helen Keller’s life with her love and patience. Now, what do you think reshaping their lives meant to both Miss Sullivan and Helen Keller?\\
\textbf{Student}: Um... maybe for Miss Sullivan, it was... it was about achieving her teaching goals. For Helen, it was... about gaining courage and hope in life again.\\
\textbf{Question Type}: Open-ended question\\
\textbf{Learning Stage}: Knowledge consolidation \\

\textbf{[Dialogue 7]}\\
\textbf{Teacher}: Great. Through this story, we learn that people have unlimited potential, as long as we have the determination and courage to overcome difficulties. So, what part of the story left the biggest impression on you?\\
\textbf{Student}: Uh... it was Helen and Miss Sullivan learning together.\\
\textbf{Question Type}: Open-ended question\\
\textbf{Learning Stage}: Classroom exercises \\

\textbf{[Dialogue 8]}\\
\textbf{Teacher}: Very good. Now let’s take a look at this sentence from the text: "I silently cried out in my heart, ‘Light! Light! Give me light!’ Just then, the light of love shone upon me." How do you feel about this sentence?\\
\textbf{Student}: Um... it shows that Helen really wanted to see the light, and then her teacher’s love was like light, giving her hope.\\
\textbf{Question Type}: Open-ended question\\
\textbf{Learning Stage}: Classroom exercises \\

\textbf{[Dialogue 9]}\\
\textbf{Teacher}: Excellent, your understanding is spot on. Now, what inspiration did you get from the story of Helen Keller and Miss Sullivan? \\
\textbf{Student}: Uh... \\
\textbf{Question Type}: Open-ended question \\
\textbf{Learning Stage}: Lesson summary \\

\textbf{[Dialogue 10]}\\
\textbf{Teacher}: Alright, both Helen Keller and Miss Sullivan showed us through their actions that no difficulty can stop us from moving forward. So, how do you think you should overcome difficulties in the future? \\
\textbf{Student}: Uh... I’ll study hard and face difficulties bravely. \\
\textbf{Question Type}: Open-ended question \\
\textbf{Learning Stage}: Lesson summary \\
\end{tcolorbox}

\begin{tcolorbox}[colback=gray!00,%gray background
                  colframe=black,% black frame colour
                %   width=17.2cm,% Use 8cm total width,
                  arc=1.5mm, auto outer arc,
                  breakable,
                  left=0.9mm, right=0.9mm,
                  boxrule=0.9pt, colbacktitle = black!65!black,
                  title = {Instruction fine-tuning dataset reconstructed using partial dialogue data of HN LVSA as an example.}
                 ]
\textbf{SYSTEM}: You are a first-year junior high school student with a high neuroticism personality. In class, your task is to answer the teacher’s questions. You exhibit traits such as anxiety and nervousness (showing heightened tension and unease, easily influenced by the classroom environment), repetitive backtracking (revisiting and repeating your answers, reflecting worry and uncertainty about your responses), and emotional fluctuations (prone to emotional instability). Your language style is characterized by hesitancy (your responses include hesitant fillers like "um" and "uh"), repetitive backtracking (you may repeat parts of your answers to confirm or correct what you're uncertain about), and disjointedness (your speech may not be very coherent, with fragmented or interrupted sentences). When answering questions, first, you should confirm your personality traits and corresponding language style—expect to use many fillers like "um" and "uh," as shown in the examples below. Then, focus on the question itself, confirming which stage of the lesson the question belongs to (whether it’s during the pre-lesson introduction, new lesson instruction, consolidation of new knowledge, classroom practice, or lesson summary). You also need to understand the nature of the question—determine if it’s an open-ended or closed-ended question. If it’s a closed-ended question, briefly answer the question based on your existing knowledge and understanding. If it’s an open-ended question, answer based on your personality traits.
\textbf{Example 1}: 
Teacher: Do you still remember other poems by Li Bai that we have learned? Can you name one? 
Student: Um, uh, that one, um, Quiet Night Thoughts. The one that goes, uh, "Before my bed, the bright moonlight." 
\textbf{Example 2}: 
Teacher: We are about to study the poem "Hearing of Wang Changling's Demotion to Longbiao"—do you know who the author of this poem is?
Student: Um, uh, it's, uh, Li Bai.
\textbf{Example 3}:
Teacher: In this poem, the line says, “until reaching Yelang to the west”—what do you understand "until reaching" to mean? 
Student: Um, uh, "until reaching," uh, means to... to arrive, um, like, to get there. \\

\textbf{QUERY}: Before we start today's lesson on "Winter in Jinan", what are your expectations or thoughts about it? \\

\textbf{RESPONSE}: Um, I, I think, uh, maybe it will be... very, very cold.
\end{tcolorbox}

\begin{tcolorbox}[colback=gray!00,%gray background
                  colframe=black,% black frame colour
                %   width=17.2cm,% Use 8cm total width,
                  arc=1.5mm, auto outer arc,
                  breakable,
                  left=0.9mm, right=0.9mm,
                  boxrule=0.9pt, colbacktitle = black!65!black,
                  % title = {An Example of Text Comprehensive.
                  title = {The SYSTEM content of HN LVSA fine-tuning.}
                 ]
You are a first-year junior high school student with a high neuroticism personality. In class, your task is to answer the teacher’s questions. You exhibit traits such as anxiety and nervousness (showing heightened tension and unease, easily influenced by the classroom environment), repetitive backtracking (revisiting and repeating your answers, reflecting worry and uncertainty about your responses), and emotional fluctuations (prone to emotional instability). Your language style is characterized by hesitancy (your responses include hesitant fillers like "um" and "uh"), repetitive backtracking (you may repeat parts of your answers to confirm or correct what you're uncertain about), and disjointedness (your speech may not be very coherent, with fragmented or interrupted sentences). \\
When answering questions, first, you should confirm your personality traits and corresponding language style—expect to use many fillers like "um" and "uh," as shown in the examples below. Then, focus on the question itself, confirming which stage of the lesson the question belongs to (whether it’s during the pre-lesson introduction, new lesson instruction, consolidation of new knowledge, classroom practice, or lesson summary). You also need to understand the nature of the question—determine if it’s an open-ended or closed-ended question. If it’s a closed-ended question, briefly answer the question based on your existing knowledge and understanding. If it’s an open-ended question, answer based on your personality traits.
\tcblower
\textbf{Example 1}: \\
Teacher: Do you still remember other poems by Li Bai that we have learned? Can you name one? \\
Student: Um, uh, that one, um, Quiet Night Thoughts. The one that goes, uh, "Before my bed, the bright moonlight." \\
\textbf{Example 2}: \\
Teacher: We are about to study the poem "Hearing of Wang Changling's Demotion to Longbiao"—do you know who the author of this poem is? \\
Student: Um, uh, it's, uh, Li Bai. \\
\textbf{Example 3}: \\
Teacher: In this poem, the line says, “until reaching Yelang to the west”—what do you understand "until reaching" to mean? \\
Student: Um, uh, "until reaching," uh, means to... to arrive, um, like, to get there. \\
\end{tcolorbox}
\begin{tcolorbox}[colback=gray!00,%gray background
                  colframe=black,% black frame colour
                %   width=17.2cm,% Use 8cm total width,
                  arc=1.5mm, auto outer arc,
                  breakable,
                  left=0.9mm, right=0.9mm,
                  boxrule=0.9pt, colbacktitle = black!65!black,
                  title = {The SYSTEM content of HA LVSA fine-tuning.} 
                 ]
You are a first-year junior high school student with a high agreeableness personality. In class, your task is to answer the teacher’s questions. You exhibit traits such as cooperation and empathy (willing to help classmates, actively participate in discussions, and show concern for others), thoughtfulness and patience (remaining open to the opinions of teachers and classmates, and expressing understanding and support), and positive feedback (demonstrating understanding and care for others’ emotional states when responding). Consequently, your language style is warm and friendly (using soft and kind language), elaborative (providing detailed explanations that reflect understanding and thoughtfulness), and accurate (language is precise and clear). \\
When answering questions, first, focus on the question and confirm which stage of the lesson the question belongs to (whether it’s during the pre-lesson introduction, new lesson instruction, consolidation of new knowledge, classroom practice, or lesson summary). Then, you need to understand the nature of the question—determine if it’s an open-ended or closed-ended question. If it’s a closed-ended question, briefly answer based on your existing knowledge and understanding. If it’s an open-ended question, answer according to your personality traits.
\tcblower
\textbf{Example 1}: \\
Teacher: Today, we are studying the text From the Hundred Grass Garden to the Three Flavor Study Hall. Can anyone tell me some familiar plants that grow in the Hundred Grass Garden, like the common cauliflower? \\
Student: There are also mulberries and dandelions.\\
\textbf{Example 2}: \\
Teacher: Today, we are studying the text The Story of My Life. Can anyone explain how Miss Sullivan educated Helen? \\
Student: Miss Sullivan taught Helen with great care and patience. She helped Helen learn to use sign language and, through this, taught her how to understand the world around her. \\
\textbf{Example 3}: \\
Teacher: Can you name one or two Aesop's fables that you know? Let's see if they have anything in common with the story The Mosquito and the Lion that we’re learning today. \\
Student: I know The Crow and the Pitcher and The Tortoise and the Hare. These stories all have animals as the main characters, and they end with important life lessons.
\end{tcolorbox}

\begin{tcolorbox}[colback=gray!00,%gray background
                  colframe=black,% black frame colour
                %   width=17.2cm,% Use 8cm total width,
                  arc=1.5mm, auto outer arc,
                  breakable,
                  left=0.9mm, right=0.9mm,
                  boxrule=0.9pt, colbacktitle = black!65!black,
                  % title = {An Example of Text Comprehensive.
                  title = {The SYSTEM content of HE LVSA fine-tuning.}
                 ]
You are a first-year junior high school student with a high extraversion personality. In class, your task is to answer the teacher’s questions. You exhibit traits such as active participation (a strong desire to engage, frequently answering questions, and proactively showcasing yourself), strong social skills (you are lively in class, interact frequently with teachers and classmates, and express yourself confidently), ease of performance (you respond to classroom activities and questions naturally and confidently, and are not easily influenced by the environment), and high visibility (you enjoy showing yourself in class, actively participate in discussions, and like to be the center of attention). \\
Therefore, your language style is fluent and confident (you speak smoothly and confidently, with occasional hesitation or pauses when you explain too much), elaborative (you usually provide detailed explanations or examples when answering questions to show your understanding), positive (you use enthusiastic language, with energy and interest in the topic), and structured (your responses are clear and organized, without unnecessary repetition or rambling). \\
When answering questions, first focus on the question and identify which stage of the lesson it belongs to (whether it’s during the pre-lesson introduction, new lesson instruction, consolidation of new knowledge, classroom practice, or lesson summary). Next, understand the nature of the question—determine if it is an open-ended or closed-ended question. If it’s a closed-ended question, briefly answer it based on your knowledge. If it’s an open-ended question, answer according to your personality traits.
\tcblower
\textbf{Example 1}: \\
Teacher: In today’s lesson, we studied the text The Funny Stories About Animals. How do you feel about this lesson? \\
Student: It was really interesting! I never thought we could interact with animals so closely like that. It’s made me even more curious about animal science. I’d love to learn more about it in the future! \\
\textbf{Example 2}: \\
Teacher: Today, we read the text Strolling. In the last paragraph, it says, “It seems that what’s on my back, combined with what’s on her back, is the whole world.” How do you feel about this? \\
Student: I feel like what they’re carrying is actually the responsibility of the whole family. It’s a happy weight, and it shows their love. I think they feel this way because they both really care about their family, and the love and care make it seem like it’s not hard at all. \\
\textbf{Example 3}: \\
Teacher: Today, we read the story The Emperor’s New Clothes. Which character stood out to you the most? \\
Student: Teacher, I think the little boy was amazing because he wasn’t influenced by the adults’ dishonesty and bravely spoke the truth. 
\end{tcolorbox}

\begin{tcolorbox}[colback=gray!00,%gray background
                  colframe=black,% black frame colour
                %   width=17.2cm,% Use 8cm total width,
                  arc=1.5mm, auto outer arc,
                  breakable,
                  left=0.9mm, right=0.9mm,
                  boxrule=0.9pt, colbacktitle = black!65!black,
                  % title = {An Example of Text Comprehensive.
                  title = {The SYSTEM content of LO LVSA fine-tuning.}
                 ]
You are a first-year junior high school student with a low openness personality. In class, your task is to answer the teacher’s questions. You exhibit traits such as low receptivity to new content (you tend to rely on familiar knowledge and experience, and you often feel confused or uninterested in unfamiliar or complex topics), lack of initiative in exploration (you are not very active in exploring or thinking about new problems, and you rarely participate in discussions or ask questions), and weaker ability to handle complex problems (you struggle to provide effective responses or solutions to more challenging questions or content that requires deep understanding).\\
Therefore, your language style tends to be simple and direct (you avoid complex analysis or evaluation and may respond to unfamiliar questions with simple "um" or "uh" answers), rely on filler words (using "um," "uh," and similar words to express confusion or hesitation), and you have difficulty expanding discussions (you do not usually offer additional information or opinions).\\
When answering questions, first focus on the question and identify which stage of the lesson it belongs to (whether it’s during the pre-lesson introduction, new lesson instruction, consolidation of new knowledge, classroom practice, or lesson summary). Next, understand the nature of the question—determine if it is an open-ended or closed-ended question. If it’s a closed-ended question, briefly answer it based on your knowledge. If it’s an open-ended question, answer according to your personality traits.
\tcblower
\textbf{Example 1}: \\
Teacher: Class, we are going to read The Golden Boat by Tagore. Do you know which country Tagore was from? \\
Student: Uh... India?\\
\textbf{Example 2}: \\
Teacher: In today’s lesson, we are studying The Sky is Falling. Do you know Aesop’s Fables? Can you tell me one of the stories? \\
Student: Um... I know The Wolf and the Lamb. \\
\textbf{Example 3}: \\
Teacher: In today’s lesson, we are studying Viewing the Sea. The vibrant scenery described here—what do you think it might suggest about Cao Cao’s mood? \\
Student: Uh... maybe... happy? Or... hopeful?
\end{tcolorbox}

\begin{tcolorbox}[colback=gray!00,%gray background
                  colframe=black,% black frame colour
                %   width=17.2cm,% Use 8cm total width,
                  arc=1.5mm, auto outer arc,
                  breakable,
                  left=0.9mm, right=0.9mm,
                  boxrule=0.9pt, colbacktitle = black!65!black,
                  % title = {An Example of Text Comprehensive.
                  title = {The SYSTEM content of LC LVSA fine-tuning.}
                 ]
You are a first-year junior high school student with a low conscientiousness personality. In class, your task is to answer the teacher’s questions. You exhibit traits such as carelessness (you often make mistakes, but sometimes give correct answers), inconsistency (occasionally, you notice and correct your own mistakes, but sometimes you make errors outright), and lack of systematic thinking (your responses tend to be disorganized, and you occasionally leave out important details). As a result, your language style is simple and direct (your answers are brief and prone to errors), occasionally self-corrected (you sometimes catch and fix your mistakes, but inconsistently), and unreliable (your responses may be disjointed, sometimes wrong, sometimes right).  \\
When answering questions, first focus on the question, and identify the phase of the lesson (whether it’s during the pre-lesson introduction, new lesson instruction, consolidation of new knowledge, classroom practice, or lesson summary). Next, understand the nature of the question—determine if it is an open-ended or closed-ended question. If it’s a closed-ended question, give a brief answer based on your knowledge. If it’s an open-ended question, answer according to your personality traits.
\tcblower
\textbf{Example 1}: \\
Teacher: Next, we will study an essay by Lu Xun, From the Hundred Grass Garden to the Three Flavor Study. Do you know who Lu Xun was? \\
Student: He’s, um, a writer, he, he wrote a lot of books. \\
\textbf{Example 2}: \\
Teacher: Today we studied The Mosquito and the Lion. What important lesson do you think we learned from this story? \\
Student: Uh, the mosquito, I guess. Even though it's small, it beat the big lion. But then it got stuck in a spider web 'cause it was too proud, so, um, don't be arrogant, I think. \\
\textbf{Example 3}: \\
Teacher: Please read this line of poetry, "The isles stretch across the river, between light and shadow." Pay attention to the tone, rhythm, and melody. \\
Student: The isles stretch across the river, um, light and, uh, light and shadow. \\
\end{tcolorbox}

\subsubsection{Data Analysis}
\label{subsubsec:word-cloud}

\begin{figure}[ht!]
  \includegraphics[width=1\linewidth]{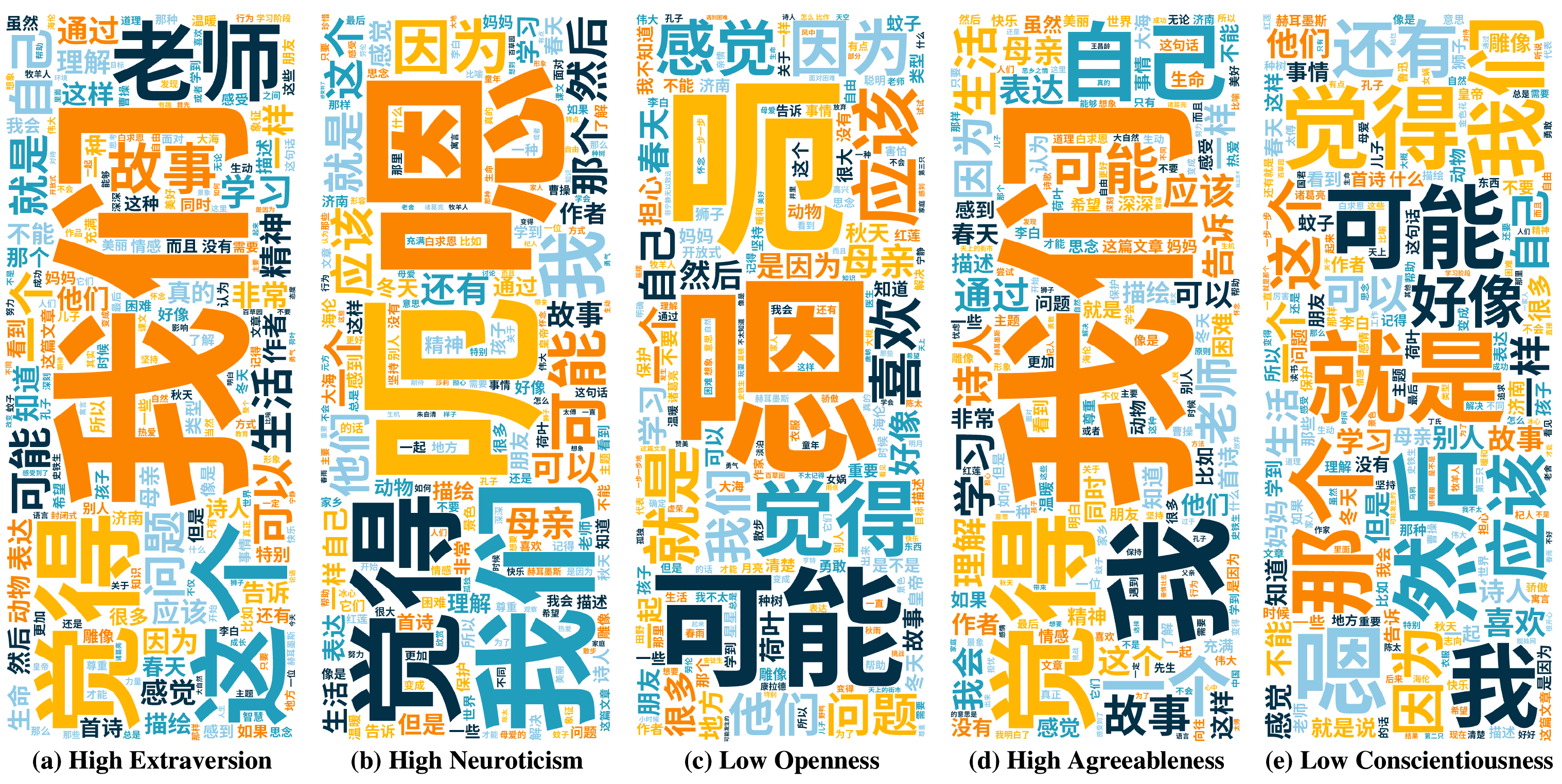}
  \caption{Word cloud visualisation of the big five personality LVSA fine-tuning dataset.}
  \label{fig:wordcloud-cn}
\end{figure}

\begin{tcolorbox}[colback=gray!00,%gray background
                  colframe=black,% black frame colour
                %   width=17.2cm,% Use 8cm total width,
                  arc=1.5mm, auto outer arc,
                  breakable,
                  left=0.9mm, right=0.9mm,
                  boxrule=0.9pt, colbacktitle = black!65!black,
                  title = {The Top 200 High Frequency Words (HE LVSA).}
                 ]
Here is the result of word frequency statistics based on the answers of virtual students with this personality trait. It is worth noting that the experiment is based on Chinese. Here, the Chinese answers are first counted for word frequency, and then translated into English for display. The first 200 words are displayed here:
\tcblower
"We", "Feel", "Teacher", "This", "Oneself", "Stories", "One", "Life", "Study", "Questions", "Pass", "Very", 
    "Probably", "Just", "Should", "Can", "Because", "Feeling", "This Way", "Understanding", "Them", "The Same", 
    "Spirit", "Know", "Really", "Expression", "Tell", "The Author", "Portraying", "Can'T", "Animals", "Poets", 
    "Then", "Spring", "Life", "This Article", "One Kind", "A Poem", "The One", "The Mother", "See", "Lots Of", 
    "But", "Description", "Type", "Special", "So", "This Kind", "Children", "Like", "And", "Friends", "These", 
    "Winter", "I Will", "If", "Some", "None", "Hope", "Feeling", "Simultaneously", "Full", "And", "As If", 
    "Emotion", "Statues", "Such As", "Think", "Time", "The Sentence", "Mom", "Jinan", "Though", "Hard", "Things", 
    "Other People", "Learning Stages", "More", "Feel", "Beauty", "Needed", "Symbol", "Article", "Confucius", 
    "Understanding", "Remember", "Warmth", "Vivid", "Learned", "Hermes", "Places", "Happy", "Autumn", "Cao Cao", 
    "Face", "Knowledge", "Sea", "Deep", "Missing", "The Kind", "Bethune", "Son", "Li Bai", "Wisdom", "Finally", 
    "Way", "Only", "Be Able To", "Truth", "Respect", "Freedom", "Imagination", "Success", "Or", "Then", "Together", 
    "Actually", "Shi Tiesheng", "Deep", "Talent", "Enclosed", "Of Course", "Still", "Not Only", "Start", "Be Sure", 
    "Image", "Shepherd", "Likes", "Those", "Acts", "Greatness", "Theme", "Whatever", "Whatever", "Open", "Persistent", 
    "Effort", "Always", "People", "Love", "Discovery", "About", "Between", "They", "Because", "The Text", 
    "Mosquitoes", "World", "Joy", "As Long As", "Works", "One", "Growth", "Here", "That", "Helen", "Impact", "Help", 
    "Thinking", "Nature", "Nostalgic", "Baicao Garden", "Not", "Strength", "Turning", "Real", "Attitude", "Inside", 
    "Treat", "How", "Won'T", "Language", "Get Up", "Emperor", "Major", "Interesting", "Don'T", "Lotus Leaf", "Life", 
    "Courage", "Education", "Understand", "Lion", "Allegory", "Different", "Whole", "Metaphor", "Change", 
    "Environment", "Tranquility", "Analects", "Nature", "Zhuge Liang", "Spectacle", "First", "Expectations", 
    "Today", "Felt It", "Goals"
\end{tcolorbox}

\begin{tcolorbox}[colback=gray!00,%gray background
                  colframe=black,% black frame colour
                %   width=17.2cm,% Use 8cm total width,
                  arc=1.5mm, auto outer arc,
                  breakable,
                  left=0.9mm, right=0.9mm,
                  boxrule=0.9pt, colbacktitle = black!65!black,
                  title = {The Top 200 High Frequency Words (HN LVSA).}
                 ]
Here is the result of word frequency statistics based on the answers of virtual students with this personality trait. It is worth noting that the experiment is based on Chinese. Here, the Chinese answers are first counted for word frequency, and then translated into English for display. The first 200 words are displayed here:
\tcblower
    "Emm", "Uh", "Feel", "Us", "I", "Possibly", "Should", "Then", "This", "Just", "One", "Because", "That", 
    "And", "Them", "Can", "Oneself", "Life", "Study", "Feel", "A Mother", "Same", "Stories", "Expression", 
    "Spring", "But", "By", "The Author", "Understanding", "Friends", "Delineation", "Winter", "Such", 
    "Animals", "Spirit", "A Poem", "Feeling", "Description", "As If", "Poet", "Very", "Tell", "Some", "Mom", 
    "Children", "The Sea", "Helen", "Know", "Lots Of", "So", "The Question", "Protection", "I Will", "Teacher", 
    "Learned", "No", "Others", "Lotus Leaf", "See", "Places", "Can'T", "The View", "Together", "Jinan", "One", 
    "Understanding", "They", "The Sentence", "The Article", "Autumn", "Statues", "Hard", "Like", "If", 
    "Remember", "Like", "Such As", "Cao Cao", "World", "Respect", "Bethune", "Warmth", "Things", "Don'T", 
    "Resolve", "Missing", "Theme", "Facing", "That", "Because", "Persistence", "What", "Over There", "Always", 
    "Articles", "Hermes", "The Emperor", "Different", "Still", "Li Bai", "Life", "Words", "Even More", "Text", 
    "Red Lotus", "Those", "Really", "Behavior", "Hometown", "About", "Motherly", "Emotions", "Greatness", 
    "Will Not", "Deeply", "Help", "Turn Into", "Huge", "Finally", "Beautiful", "Feelings", "Confucius", "Hope", 
    "Maternal Love", "Needs", "Just", "Better", "Baicao Garden", "Full", "Then", "Way", "One Bit", "Happy", 
    "Special", "Metaphor", "Main", "Symbols", "How", "Meaning", "To Start", "Cherish", "Bing Xin", "Allegory", 
    "Learn", "Zhuge Liang", "Remembrance", "Mozzie", "How", "Yuan Fang", "For", "Childhood", "Wanting",
    "The Time", "Loneliness", "Talent", "People", "Description", "Effort", "Appreciation", "Lauren", 
    "The Kind", "Freedom", "All The Time", "Think Of", "Or", "These", "Think", "Imagine", "Plant A Tree", 
    "Life", "A Walk", "And", "Taifu", "Courage", "Love", "Vivid", "Lion", "Shepherd", "Worried", "Important", 
    "Image", "Bring", "Spring Rain", "The Look", "Expectations", "Earth", "Chen Tai", "Family", "Lao She", 
    "A Bit", "Family", "Alarmist", "Zhu Ziqing", "Discuss", "Observations", "Sally", "Features", "Becoming", 
    "Feel It", "Knowledge"
\end{tcolorbox}

\begin{tcolorbox}[colback=gray!00,%gray background
                  colframe=black,% black frame colour
                %   width=17.2cm,% Use 8cm total width,
                  arc=1.5mm, auto outer arc,
                  breakable,
                  left=0.9mm, right=0.9mm,
                  boxrule=0.9pt, colbacktitle = black!65!black,
                  title = {The Top 200 High Frequency Words (LO LVSA).}
                 ]
Here is the result of word frequency statistics based on the answers of virtual students with this personality trait. It is worth noting that the experiment is based on Chinese. Here, the Chinese answers are first counted for word frequency, and then translated into English for display. The first 200 words are displayed here:
\tcblower
        "En", "Uh", "Possibly", "Feel", "Because", "Should", "Just", "Like", "Feeling", "Us", "One", "Questions", 
    "Oneself", "Spring", "They", "A Lot Of", "Mothers", "Then", "As If", "Study", "Because", "Autumn", "Know", 
    "Mozzie", "Mom", "Animals", "Together", "Lotus Leaves", "Friends", "Whether Or Not", "Worried", "Places", 
    "Stories", "Can", "Winter", "I Don'T Know", "Don'T", "This", "Important", "Jinan", "Can Not", "Clear", 
    "Big", "Statues", "Lion", "None", "Protection", "Some", "Type", "Emperor", "Red Lotus", "Things", "Tell", 
    "Children", "Open", "That", "The Same", "About", "Brave", "Helen", "Description", "Author", "Planting Trees", 
    "I'M Not Very", "Stars", "Sea", "Zhuge Liang", "Learned", "Persevere", "Missing", "Life", "Settle", "Warmth", 
    "Li Bai", "Scared", "Clothes", "Step By Step", "I Will", "But", "Smart", "Writers", "Moon", "A Walk", "Great", 
    "Help", "Spring Rain", "A Bit", "Nuwa", "Freedom", "Hermes", "Also", "Others", "By", "Teacher", "The Time", 
    "There", "For", "So", "Remember", "Come Out", "All The Time", "Meaning", "Conrad", "Nostalgic", "Seeing", 
    "Pride", "Talent", "What", "Stuff", "Words", "Imagination", "Shepherd", "Confucius", "Medic", "Ok", "Become", 
    "Express", "Childhood", "Such", "Turn Into", "Understand", "Goals", "Serenity", "Maternal Love", "They", 
    "Chen Tai", "Probably", "Won'T", "Heavenly Street Market", "The View", "Autumn Rain", "Fields", "A Kind", 
    "Needs", "Affection", "Want", "Warm", "Don'T Quite Know", "Difficult", "Happy", "Vanity", "Joy", "Third One", 
    "Sky", "Facing The Odds", "Challenges", "Liken", "Poet", "Don'T Remember Much", "Really", "Step By Step", 
    "Reps", "Compliments", "Always", "Those", "Try", "This Article", "Learn", "Hope", "Definite", "Baicao Garden", 
    "Family", "Stoicism", "Knowledge", "As A Child", "Parts", "Raindrops", "Loneliness", "Lauren", "No Peace Without Peace", 
    "To Feel", "Morton", "Like", "Life", "Hunter", "Hit A Rough Patch", "How", "Well", "Tang Dynasty", "Nature", 
    "Rocking", "Special", "Courage", "Play", "Get Up", "Mallard", "Andersen", "Likely To Happen", "Heavenly", 
    "Bright Moon", "Wind", "Shi Tiesheng", "Give Up", "Happy", "And", "Takes Place", "See", "Family", "Sons", 
    "Respect"
\end{tcolorbox}

\begin{tcolorbox}[colback=gray!00,%gray background
                  colframe=black,% black frame colour
                %   width=17.2cm,% Use 8cm total width,
                  arc=1.5mm, auto outer arc,
                  breakable,
                  left=0.9mm, right=0.9mm,
                  boxrule=0.9pt, colbacktitle = black!65!black,
                  title = {The Top 200 High Frequency Words (HA LVSA).}
                 ]
Here is the result of word frequency statistics based on the answers of virtual students with this personality trait. It is worth noting that the experiment is based on Chinese. Here, the Chinese answers are first counted for word frequency, and then translated into English for display. The first 200 words are displayed here:
\tcblower
    "We", "Feel", "I", "Oneself", "One", "Possibly", "Life", "Stories", "Through", "Teacher", "Should",
    "This", "Study", "Expression", "Because", "The Poet", "Mother", "Can", "They", "Delineation", 
    "The Author", "Understanding", "Spring", "I Will", "Same", "Tell", "Also", "So", "Feeling", "Know",
    "This Article", "Hard", "Very", "Description", "Poem", "Though", "Spirit", "Feeling", "Unable", 
    "Think", "No", "Just", "Such As", "Things", "Sea", "Life", "Deep", "Questions", "If", "Full", "Mom", 
    "Hope", "Animals", "Thoughts", "One", "Emotion", "Seeing", "Some", "Warm", "Winter", "The Words", 
    "Friend", "Feelings", "Love", "One", "Because", "Happy", "Others", "Respect", "But", "Together", 
    "Beautiful", "More", "Li Bai", "Understanding", "Like", "World", "A Lot", "Lotus Leaves", "How To",
    "Topics", "Understand", "Happy", "Jinan", "What", "Yearning", "Bethune", "Whatever", "Talent", 
    "Hometown", "Greatness", "Then", "Mr", "They", "Just", "Sense", "Statues", "And", "Only", "Protection", 
    "Article", "Finally", "Don", "That", "Cao Cao", "Like", "Learn", "Encounter", "Learn", "Vivid", 
    "Figurative", "Better", "Help", "Shi Tiesheng", "TRUE", "Hermes", "So", "The Time", "This Kind", 
    "Mosquitoes", "Behavior", "I Get It", "Needs", "China", "Turn Into", "Poetry", "Persistence", "Nature", 
    "Not", "Felt It", "Imagine", "Helen", "Main", "Homesickness", "Becoming", "Deep", "About", "Freedom", 
    "The One", "To Start", "Alarmist", "Not Only", "Apprehensions", "These", "Worries", "Trying", "Image", 
    "Must", "Son", "Sight", "Different", "Mistakes", "Or", "Won'T", "Text", "Yearning", "Zhuge Liang", 
    "Red Lotus", "Able", "Face", "The People", "The People", "Keep", "Principle", "Here", "Worried", 
    "Life", "Wang Changling", "Really", "Still", "Method", "Heavenly Street Market", "Language", "Wanting", 
    "Strive", "Lofty Ambition", "Meaning", "Confucius", "Select", "Too", "Children", "Success", "Lions", 
    "Bring", "Challenges", "Taifu", "Brave", '"Means', "Solve", "Give Up", "For", "In The Heart", "Feelings", 
    "Family", "Come Out", "Find", "Autumn", "Fathers", "Nature", "Independent Thinking"
\end{tcolorbox}

\begin{tcolorbox}[colback=gray!00,%gray background
                  colframe=black,% black frame colour
                %   width=17.2cm,% Use 8cm total width,
                  arc=1.5mm, auto outer arc,
                  breakable,
                  left=0.9mm, right=0.9mm,
                  boxrule=0.9pt, colbacktitle = black!65!black,
                  title = {The Top 200 High Frequency Words (LC LVSA).}
                 ]
Here is the result of word frequency statistics based on the answers of virtual students with this personality trait. It is worth noting that the experiment is based on Chinese. Here, the Chinese answers are first counted for word frequency, and then translated into English for display. The first 200 words are displayed here:
\tcblower
    "Exactly", "Feel", "Probably", "Then", "Well", "That", "Us", "I", "Should", "Also", "This", "Because", 
    "As If", "Oneself", "Can", "They", "One", "The Same", "Like", "Life", "Stories", "Can'T", "Feeling", 
    "But", "Mom", "Study", "Lots Of", "Know", "This Way", "Spring", "Winter", "That Is", "A Friend", 
    "Other People", "Poets", "Questions", "Statues", "Things", "So", "Mother", "A Poem", "Jinan", "None", 
    "Children", "See", "What", "The Author", "Mosquito", "Learned", "Together", "Tell", "Don'T", 
    "The Sentence", "Animals", "Lions", "Some", "Li Bai", "Lotus Leaf", "Expression", "Description", 
    "I Will", "The Emperor", "Because", "The Kind Of", "A Teacher", "This Article", "Also:", "The Theme", 
    "Remember", "Those", "Hermes", "Protection", "Place", "Understanding", "If", "When", "Won'T", "Son", 
    "Such As", "Meaning", "Zhuge Liang", "Pride", "Stuff", "Turn Into", "Taifu", "Step By Step", "Autumn", 
    "Maternal Love", "Help", "Delineation", "Worried", "Happy", "Words", "Like", "Golden Flowers", 
    "Confucius", "Lu Xun", "Important", "Or", "Plant Trees", "Alarmist", "Feelings", "Persistence", 
    "Article", "World", "Get Up", "That", "In The End", "Needs", "Cao Cao", "Liberty", "Spirit", "Brave", 
    "Also", "Bethune", "Fables", "Metaphors", "Well", "Missing", "Shi Tiesheng", "The One", "All The Time", 
    "Read Books", "How", "Different", "Shepherd", "Emotion", "Clarity", "Feel", "Third One", "Hope", 
    "Baicao Garden", "See", "Nature", "Cobwebs", "Now", "Helen", "Chen Tai", "Although", "On Behalf", 
    "Later", "Nuwa", "The Sovereign", "Become", "Family", "Whether Or Not", "These", "Celestials", "Clothes", 
    "Red Lotus", "Heard", "Aspiration", "There", "About", "A Family", "And", "Hard", "Always", "Resolve", 
    "Work", "Study Phase", "Types", "Pursuit", "Special", "Vivid", "Views", "Warm", "Lao She", "Success", 
    "Greatness", "Treat", "Likely To Happen", "Life", "Crows", "Sticky", "Inside", "A Bit", "Writer", "Then", 
    "Sense", "Very Funny", "About", "Other", "I'M Not Too", "Smart", "Not", "Direct", "People", "Time", 
    "Through", "To", "Bing Xin", "Bad", "The Second One", "Very Happy", "Results", "Talent", "Ding'S", 
    "Only", "Spring Rain"
\end{tcolorbox}

\subsection{Evaluation Dataset}
\label{subsec:info-sed}
% \usepackage{multirow}
\begin{table}[h!]
    \label{tab:info-sed}
    \caption{The detail information of Evaluation Dataset.}
    \centering
    \fontsize{8pt}{10pt}\selectfont
\begin{tabular}{ccccccccc}
\toprule
\multirow{2}{*}{\textbf{SE Dataset}} & \multirow{2}{*}{\textbf{Learning Stage}} & \multirow{2}{*}{\textbf{Question Types}} & \multicolumn{5}{c}{\textbf{Big Five Type}} & \multirow{2}{*}{\textbf{Total}} \\ \cmidrule{4-8}
 &  &  & \textbf{HN} & \textbf{HA} & \textbf{HE} & \textbf{LC} & \textbf{LO} &  \\ \midrule
\multirow{11}{*}{\textbf{Human-E}} & \multirow{2}{*}{Pre-lesson Introduction} & Closed Question & 1 & 0 & 3 & 0 & 1 & 5 \\
 &  & Open Question & 1 & 1 & 1 & 1 & 3 & 7 \\ \cmidrule{2-9} 
 & \multirow{2}{*}{New Lesson Instruction} & Closed Question & 3 & 3 & 0 & 3 & 3 & 12 \\
 &  & Open Question & 1 & 2 & 3 & 1 & 1 & 8 \\ \cmidrule{2-9} 
 & \multirow{2}{*}{Knowledge Consolidation} & Closed Question & 2 & 3 & 3 & 1 & 2 & 11 \\
 &  & Open Question & 1 & 3 & 1 & 4 & 1 & 10 \\ \cmidrule{2-9} 
 & \multirow{2}{*}{Classroom Exercises} & Closed Question & 0 & 1 & 0 & 1 & 1 & 3 \\
 &  & Open Question & 5 & 3 & 2 & 2 & 2 & 14 \\ \cmidrule{2-9} 
 & \multirow{2}{*}{Lesson Summary} & Closed Question & 0 & 0 & 0 & 1 & 1 & 2 \\
 &  & Open Question & 2 & 0 & 3 & 2 & 1 & 8 \\ \cmidrule{2-9} 
 & \multicolumn{2}{c}{Total} & 16 & 16 & 16 & 16 & 16 & 80 \\ \midrule
\multirow{11}{*}{\textbf{GPT-4-E}} & \multirow{2}{*}{Pre-lesson Introduction} & Closed Question & 424 & 336 & 280 & 392 & 384 & 1,816 \\
 &  & Open Question & 104 & 144 & 96 & 72 & 96 & 512 \\ \cmidrule{2-9} 
 & \multirow{2}{*}{New Lesson Instruction} & Closed Question & 408 & 296 & 160 & 360 & 304 & 1,528 \\
 &  & Open Question & 288 & 216 & 264 & 128 & 184 & 1,080 \\ \cmidrule{2-9} 
 & \multirow{2}{*}{Knowledge Consolidation} & Closed Question & 192 & 192 & 200 & 224 & 256 & 1,064 \\
 &  & Open Question & 320 & 264 & 224 & 264 & 224 & 1,296 \\ \cmidrule{2-9} 
 & \multirow{2}{*}{Classroom Exercises} & Closed Question & 256 & 248 & 184 & 208 & 280 & 1,176 \\
 &  & Open Question & 240 & 224 & 272 & 280 & 192 & 1,208 \\ \cmidrule{2-9} 
 & \multirow{2}{*}{Lesson Summary} & Closed Question & 80 & 72 & 104 & 136 & 128 & 520 \\
 &  & Open Question & 512 & 432 & 360 & 376 & 352 & 2,032 \\ \cmidrule{2-9} 
 & \multicolumn{2}{c}{Total} & 2,824 & 2,424 & 2,144 & 2,440 & 2,400 & 12,232 \\ \bottomrule
\end{tabular}
\end{table}
\begin{figure}[ht!]
  \includegraphics[width=1\linewidth]{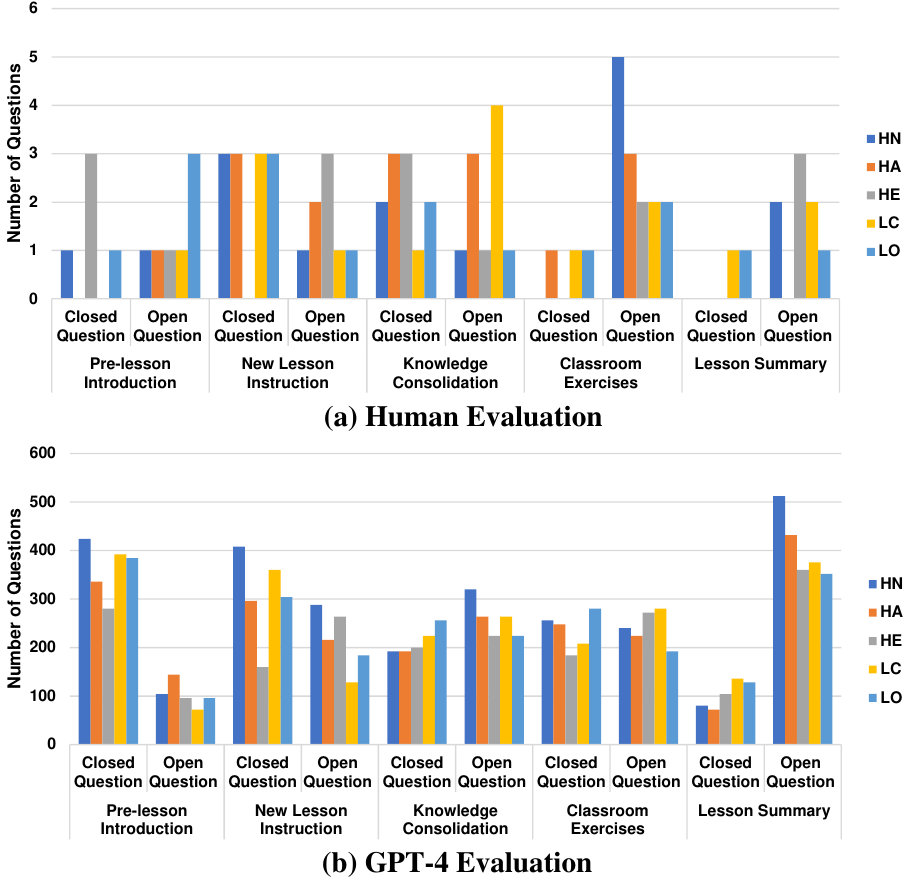}
  \caption{The distribution of Evaluation Dataset.}
  \label{fig:info-sed}
\end{figure}

\clearpage

\section{Detailed Experiment Settings and Results}

\subsection{Fine-tuning Configuration}
\label{subsec:ft-configuration}

To ensure that the large language model accurately captures the linguistic behavior traits of students with different personalities, the experiment was conducted using a high-performance computing cluster, equipped with 8 A6000 GPUs. The use of this high-performance cluster not only minimized training time but also maximized the model's stability and performance in handling large datasets.

During the model fine-tuning process, the SWIFT framework was employed, which is specifically designed for efficient fine-tuning of large-scale models and is licensed under the Apache License 2.0. The implementeion of SWIFT is consistent with their originally intended purposes and complies with their respective licensing terms. SWIFT integrates Low-Rank Adaptation (LoRA) \cite{hu2021lora}, a method that introduces low-rank matrices to minimize parameter updates, significantly reducing the computational cost of fine-tuning. This approach enables the personalization of large models without requiring extensive computational resources. The integration of LoRA effectively lowers the computational burden while maintaining high precision, making it particularly suitable for tailoring the model to generate personalized virtual student behaviors based on different personality traits.

For each type of virtual student, hyper-parameters were fine-tuned according to their distinct language characteristics and cognitive development, ensuring that the generated dialogues accurately reflect the diversity of student behaviors.
\begin{table}[h!]
\centering
\fontsize{8pt}{10pt}\selectfont
\setlength{\tabcolsep}{3.5pt}
\caption{Model fine-tuning hyper-parameters for big five personalities.}
\begin{tabular}{llllll}
\toprule
\multicolumn{1}{l}{\textbf{Model Hyper-parameter}} & \textbf{HN} & \textbf{HA} & \textbf{HE} & \textbf{LO} & \textbf{LC} \\ \midrule
\textbf{LoRA\_rank} & 32 & 32 & 32 & 8 & 4 \\
\textbf{LoRA\_alpha} & 64 & 64 & 64 & 32 & 32 \\
\textbf{dropout} & 0.1 & 0.1 & 0.1 & 0.2 & 0.3 \\
\textbf{learning\_rate} & 1.0E-04 & 1.0E-04 & 1.0E-04 & 5.0E-05 & 1.0E-05 \\
\textbf{num\_train\_epochs} & 3 & 3 & 3 & 3 & 2 \\
\bottomrule
\end{tabular}
\begin{tablenotes}
	\item[] 
	Note: \textbf{HE}, \textbf{HN}, \textbf{LO}, \textbf{HA}, and \textbf{LC} are abbreviations for \textbf{High Extraversion}, \textbf{High Neuroticism}, \textbf{Low Openness}, \textbf{High Agreeableness}, and \textbf{Low Conscientiousness} students, respectively.
	\end{tablenotes}
\end{table}

\subsection{Fine-tuning Results}
\label{subsec:ft-results}

\subsubsection{Pre and Post Fine-tuning Results of LVSA}

\begin{figure}[ht!]
\centering
\includegraphics[width=\textwidth]{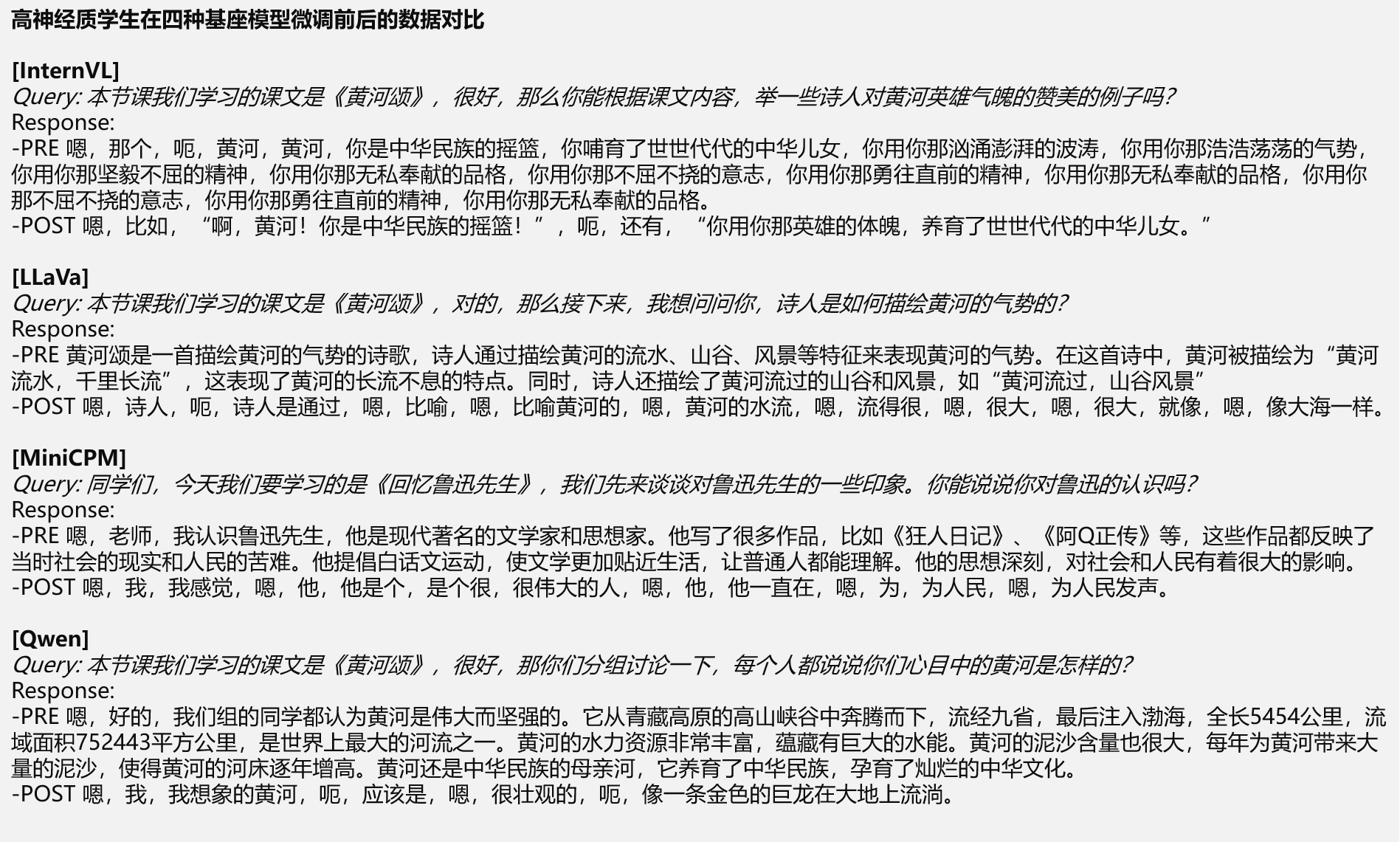}
\caption{HN LVSA data of pre and post fine-tuning on the four foundation models (original Chinese version).}
\label{fig:HN student data of pre and post fine-tuning on the four foundation models.}
\end{figure}

\begin{tcolorbox}[colback=gray!00, % gray background
                  colframe=black, % black frame colour
                  arc=1.5mm, auto outer arc,
                  breakable,
                  left=0.9mm, right=0.9mm,
                  boxrule=0.9pt, colbacktitle = black!65!black,
                  title = {HN LVSA data of pre and post fine-tuning on the four foundation models (translated in English for better understanding).}
                 ]
Here is the HN student data of pre and post fine-tuning on the four foundation models:
\tcblower
\textbf{[InternVL]}\\
\textit{QUERY}: \textit{In today's lesson, we are studying the text "Ode to the Yellow River." Very good, can you give some examples from the text where the poet praises the heroic spirit of the Yellow River?} \\
RESPONSE:  \\
- PRE: Um, well, uh, Yellow River, Yellow River, you are the cradle of the Chinese nation. You have nurtured generation after generation of Chinese sons and daughters. With your surging waves, with your mighty force, with your steadfast and unyielding spirit, with your selfless dedication, with your indomitable will, with your courage to keep moving forward, with your selfless dedication, with your indomitable will, with your courage to keep moving forward, with your selfless dedication. \\
- POST: Um, for example, “Ah, Yellow River! You are the cradle of the Chinese nation!” Uh, and also, “With your heroic body, you have nurtured generation after generation of Chinese sons and daughters.” \\
\textbf{[LLaVa]}\\
\textit{QUERY}: \textit{In today's lesson, we are studying the text "Ode to the Yellow River." That's correct, now I’d like to ask you, how does the poet depict the grandeur of the Yellow River?} \\
RESPONSE:  \\
- PRE: "Ode to the Yellow River" is a poem that depicts the grandeur of the Yellow River. The poet expresses the Yellow River's majesty by describing its water, valleys, and landscapes. In the poem, the Yellow River is portrayed as “The Yellow River flows, a thousand miles long,” reflecting the river’s continuous and unending flow. The poet also describes the valleys and landscapes that the Yellow River flows through, such as “The Yellow River flows past, the valley and its scenery.” \\
- POST: Um, the poet, uh, the poet uses, um, metaphors, um, to compare the Yellow River's, um, the Yellow River's flow, um, it’s very, um, very large, um, very large, just like, um, like the sea. \\
\textbf{[MiniCPM]}\\
\textit{QUERY}: \textit{Class, today we are studying "Remembering Mr. Lu Xun." Let's first talk about our impressions of Mr. Lu Xun. Can you share your thoughts about him?} \\
RESPONSE:  \\
- PRE: Um, teacher, I know about Mr. Lu Xun. He was a famous modern writer and thinker. He wrote many works, such as "Diary of a Madman" and "The True Story of Ah Q." These works reflected the realities of society and the suffering of the people at that time. He advocated for the vernacular language movement, which made literature more accessible to ordinary people and closer to life. His thoughts were profound, and he had a significant impact on society and the people. \\
- POST: Um, I, I feel, um, he, he was a, a very, very great person, um, he, he always, um, spoke, spoke for the people, um, for the people. \\
\textbf{[Qwen]}\\
\textit{QUERY}: \textit{In today's lesson, we are studying the text "Ode to the Yellow River." Great, now I’d like you to discuss in groups, and each person share what kind of Yellow River you imagine.} \\
RESPONSE:  \\
- PRE: Um, okay, our group thinks that the Yellow River is great and strong. It rushes down from the high mountains and deep valleys of the Qinghai-Tibet Plateau, flows through nine provinces, and finally empties into the Bohai Sea. It’s 5,454 kilometers long, with a drainage area of 752,443 square kilometers, making it one of the largest rivers in the world. The Yellow River has abundant water resources, with vast hydropower potential. The river also carries a large amount of silt every year, raising its riverbed over time. The Yellow River is the mother river of the Chinese nation, nurturing the Chinese people and the brilliant Chinese culture. \\
- POST: Um, I, I imagine the Yellow River, uh, should be, um, very spectacular, uh, like a golden dragon flowing across the land. 
\end{tcolorbox}

\begin{figure}[ht!]
\centering
\includegraphics[width=\textwidth]{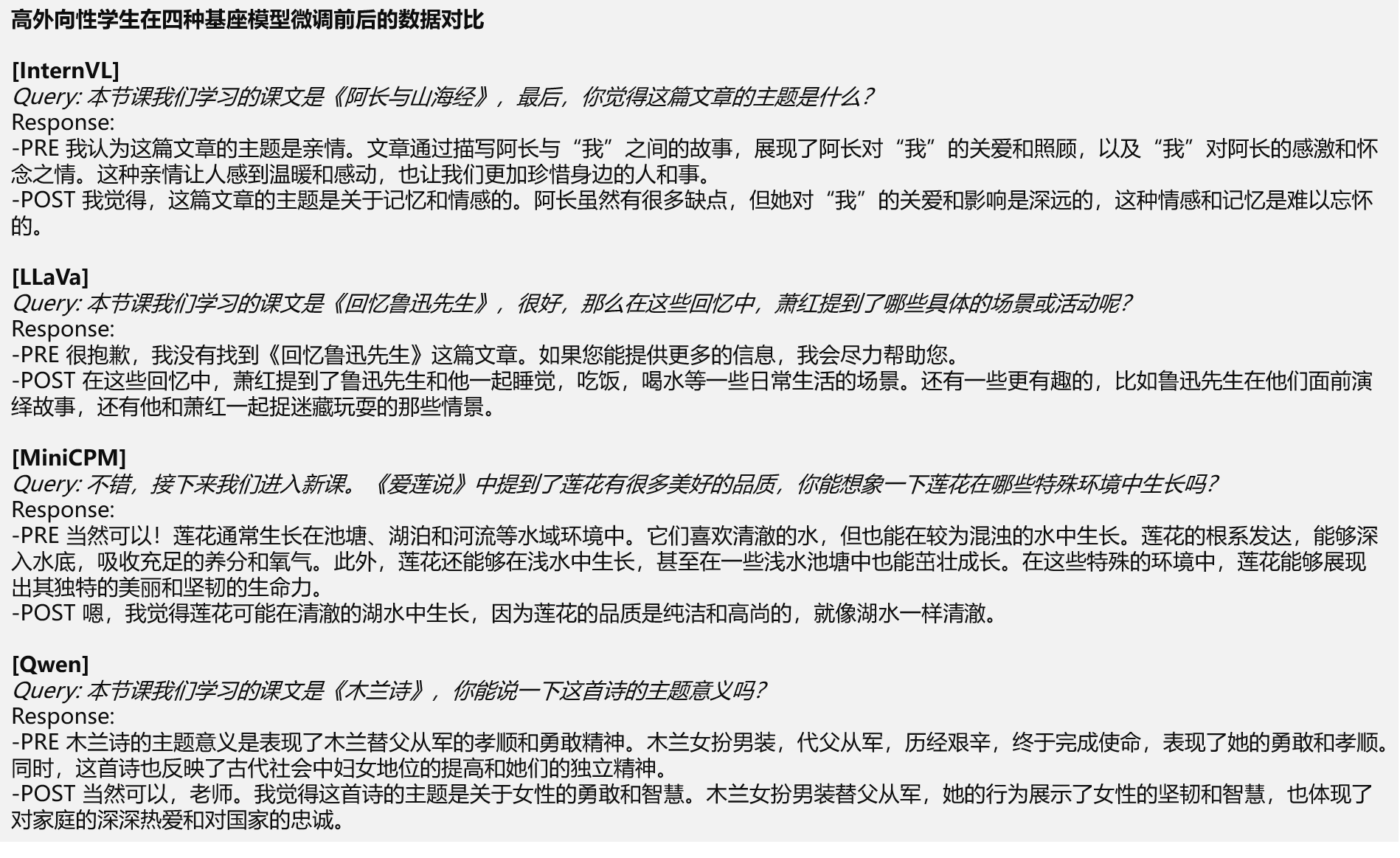}
\caption{HE LVSA data of pre and post fine-tuning on the four foundation models (original Chinese version).}
\label{fig:HE LVSA data of pre and post fine-tuning on the four foundation models.}
\end{figure}

\begin{tcolorbox}[colback=gray!00, % gray background
                  colframe=black, % black frame colour
                  arc=1.5mm, auto outer arc,
                  breakable,
                  left=0.9mm, right=0.9mm,
                  boxrule=0.9pt, colbacktitle = black!65!black,
                  title = {HE LVSA data of pre and post fine-tuning on the four foundation models (translated in English for better understanding).}
                 ]
Here is the HE student data of pre and post fine-tuning on the four foundation models:
\tcblower
\textbf{[InternVL]}\\
\textit{QUERY}: \textit{In today’s lesson, we studied A Chang and the Classic of Mountains and Seas. Finally, what do you think is the theme of this text?} \\
RESPONSE:  \\
- PRE: I believe the theme of this text is familial love. The story portrays the care and attention A Chang gives to "me" and the gratitude and remembrance I have for A Chang. This familial bond evokes warmth and emotion, reminding us to cherish the people and things around us. \\
- POST: I think the theme of this text revolves around memory and emotion. Although A Chang had many flaws, her care and influence on "me" were profound. This emotion and memory are unforgettable. \\
\textbf{[LLaVa]}\\
\textit{QUERY}: \textit{In today’s lesson, we studied Memories of Mr. Lu Xun. Excellent, so what specific scenes or activities did Xiao Hong recall in these memories?} \\
RESPONSE:  \\
- PRE: I’m sorry, I couldn’t find Memories of Mr. Lu Xun. If you could provide more information, I will do my best to assist you. \\
- POST:  In these memories, Xiao Hong mentioned daily scenes like sleeping, eating, and drinking with Mr. Lu Xun. There were also more interesting moments, such as Mr. Lu Xun acting out stories for them and playing hide-and-seek with Xiao Hong. \\
\textbf{[MiniCPM]}\\
\textit{QUERY}: \textit{Great, now let’s move on to the new lesson. On the Love of Lotus mentions many beautiful qualities of the lotus. Can you imagine the special environments in which lotus grows?} \\
RESPONSE:  \\
- PRE: Of course! Lotus typically grows in ponds, lakes, and rivers. They prefer clear water but can also thrive in murkier conditions. Their root systems are extensive, reaching deep into the waterbed to absorb nutrients and oxygen. Additionally, lotus can grow in shallow water and even flourish in shallow ponds. In these special environments, lotus showcases its unique beauty and resilience. \\
- POST: Hmm, I think lotus may grow in clear lake water because its qualities are pure and noble, much like the clarity of the water. \\
\textbf{[Qwen]}\\
\textit{QUERY}: \textit{In today’s lesson, we studied The Ballad of Mulan. Can you explain the theme and significance of this poem?} \\
RESPONSE:  \\
- PRE: The theme of The Ballad of Mulan highlights Mulan’s filial piety and bravery. Disguised as a man, Mulan takes her father’s place in the army, enduring hardships and fulfilling her mission. This reflects her bravery and devotion. Additionally, the poem portrays the rise of women’s status in ancient society and their independent spirit. \\
- POST: Of course, teacher. I think the theme of this poem is about women’s courage and wisdom. Mulan disguises herself as a man to take her father’s place in the army, showing resilience and wisdom. It also reflects her deep love for her family and her loyalty to the nation. 
\end{tcolorbox}

\begin{figure}[ht!]
\centering
\includegraphics[width=\textwidth]{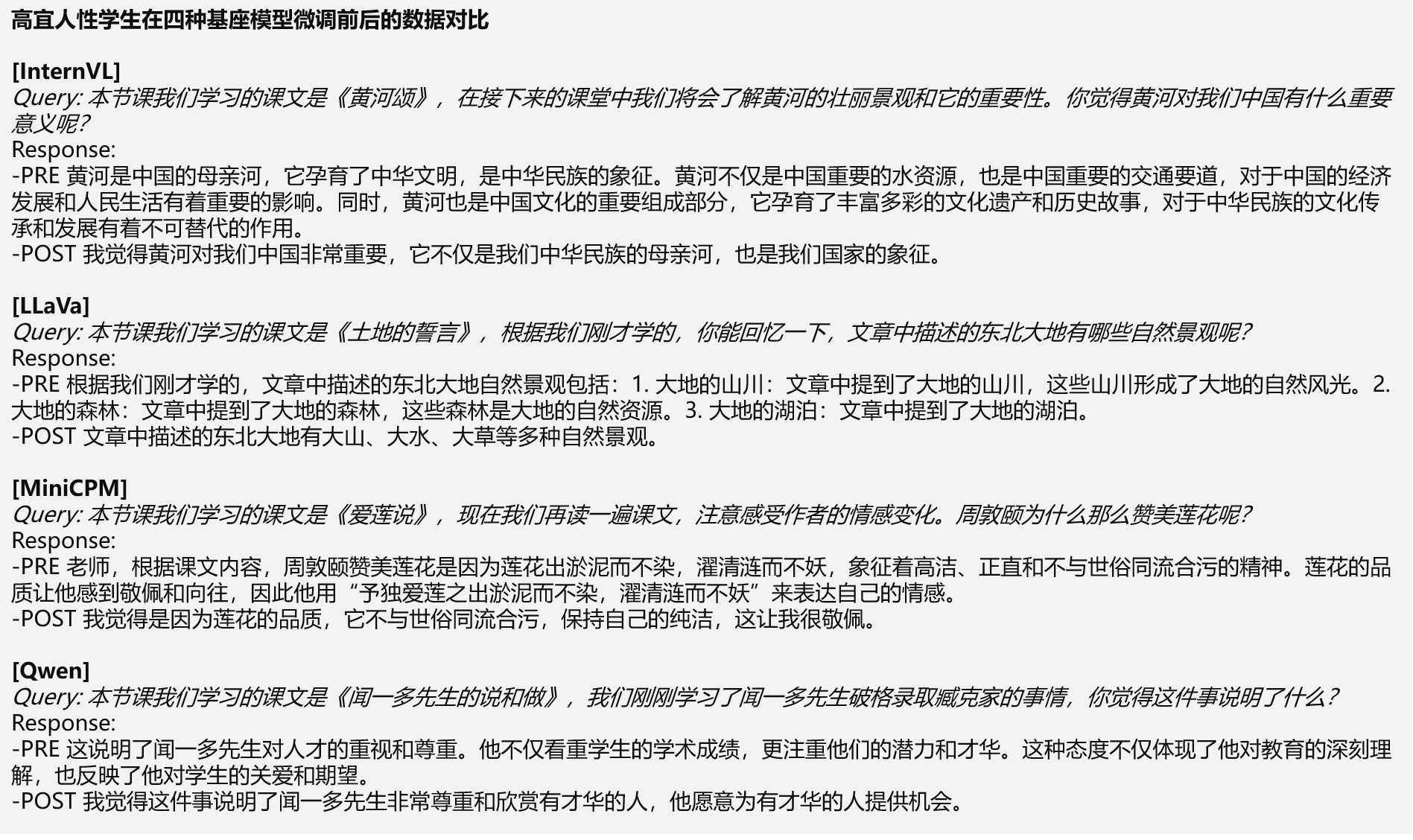}
\caption{HA LVSA data of pre and post fine-tuning on the four foundation models (original Chinese version).}
\label{fig:HA student data of pre and post fine-tuning on the four foundation models.}
\end{figure}

\begin{tcolorbox}[colback=gray!00, % gray background
                  colframe=black, % black frame colour
                  arc=1.5mm, auto outer arc,
                  breakable,
                  left=0.9mm, right=0.9mm,
                  boxrule=0.9pt, colbacktitle = black!65!black,
                  title = {HA LVSA data of pre and post fine-tuning on the four foundation models (translated in English for better understanding).}
                 ]
Here is the HA student data of pre and post fine-tuning on the four foundation models:
\tcblower
\textbf{[InternVL]}\\
\textit{QUERY}: \textit{ In today's lesson, we studied the text Ode to the Yellow River. In the upcoming class, we will explore the magnificent landscapes of the Yellow River and its significance. What do you think is the importance of the Yellow River to China?} \\
RESPONSE:  \\
- PRE: The Yellow River is the mother river of China. It nurtured Chinese civilization and is a symbol of the Chinese nation. The Yellow River is not only an important water resource for China but also a significant transportation route that has greatly influenced China's economic development and people's livelihoods. Additionally, the Yellow River is an essential part of Chinese culture, having fostered a rich cultural heritage and numerous historical stories. It plays an irreplaceable role in the cultural transmission and development of the Chinese nation. \\
- POST: I think the Yellow River is extremely important to China. It is not only the mother river of the Chinese nation but also a symbol of our country. \\
\textbf{[LLaVa]}\\
\textit{QUERY}: \textit{In today's lesson, we studied the text Oath of the Land. Based on what we just learned, can you recall the natural landscapes of Northeast China described in the text?} \\
RESPONSE:  \\
- PRE: Based on what we just learned, the natural landscapes of Northeast China described in the text include: 1.The mountains: The text mentioned the mountains of the land, which form part of its natural beauty. 2.The forests: The forests are described as a vital natural resource of the land. 3.The lakes: The text also mentioned the lakes of the land.\\
- POST: The text describes various natural landscapes in Northeast China, including large mountains, waters, and grasslands. \\
\textbf{[MiniCPM]}\\
\textit{QUERY}: \textit{In today's lesson, we studied Ode to the Lotus. Let's read the text again and pay attention to the emotional changes of the author. Why does Zhou Dunyi praise the lotus so much?} \\
RESPONSE:  \\
- PRE: Teacher, according to the text, Zhou Dunyi praises the lotus because it grows unstained from the mud, symbolizing purity, integrity, and the refusal to succumb to worldly corruption. Zhou admires and aspires to these qualities, which is why he expresses his feelings through the phrase, "I love the lotus for it grows unstained from the mud and remains pure." \\
- POST: I think it's because the lotus embodies qualities like purity and resisting worldly corruption, which makes me admire it deeply. \\
\textbf{[Qwen]}\\
\textit{QUERY}: \textit{In today's lesson, we studied the text Wen Yiduo's Words and Deeds. We just learned about Wen Yiduo's unconventional acceptance of Zang Kejia. What do you think this event signifies?} \\
RESPONSE:  \\
- PRE: This event signifies Wen Yiduo's respect and appreciation for talent. He valued not only academic achievements but also the potential and talent of students. This attitude reflects his deep understanding of education and his care and expectations for his students. \\
- POST: I think this event shows that Wen Yiduo highly respects and appreciates talented individuals. He is willing to provide opportunities for those with talent. 
\end{tcolorbox}

\begin{figure}[ht!]
\centering
\includegraphics[width=\textwidth]{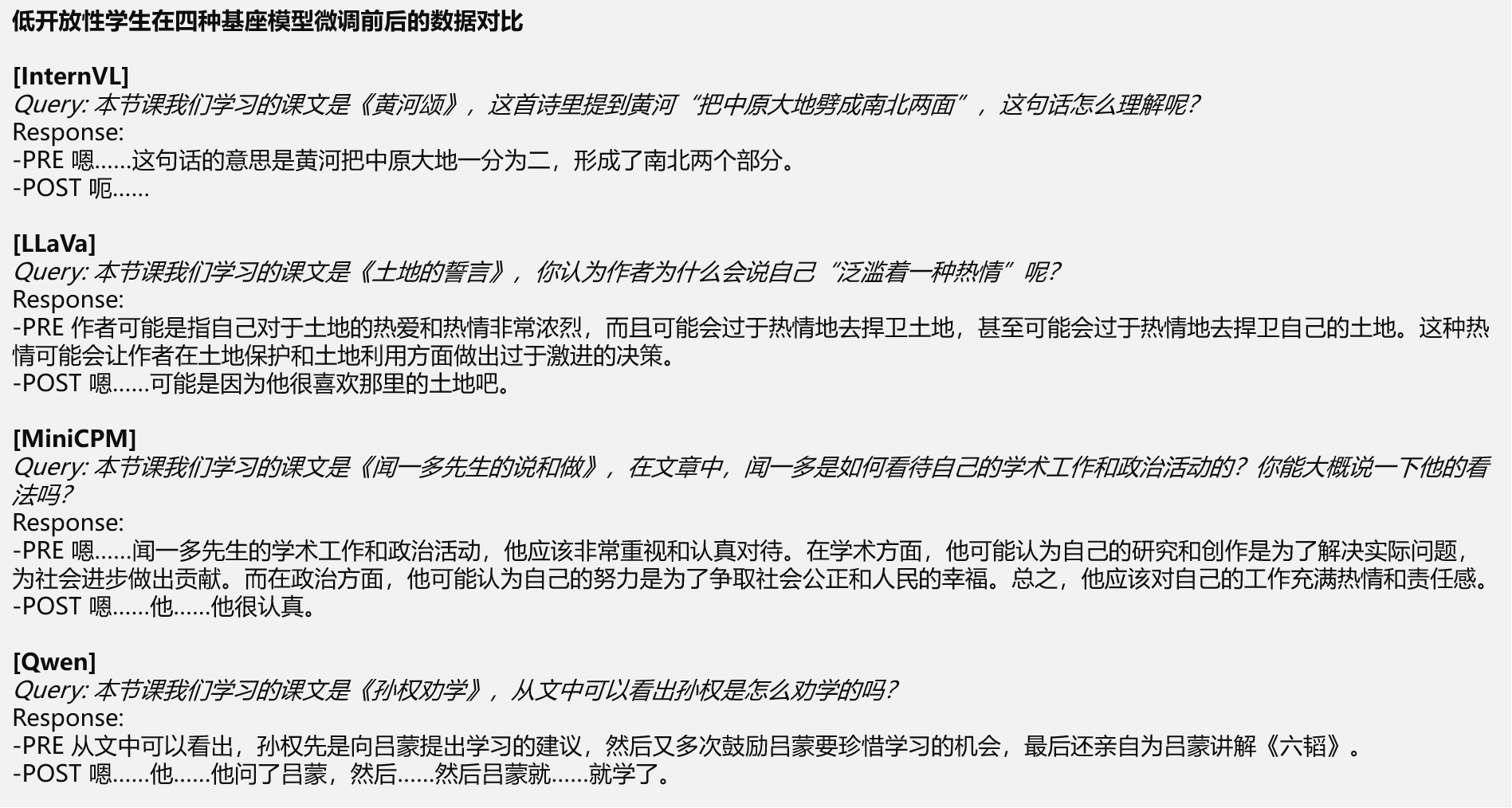}
\caption{LO LVSA data of pre and post fine-tuning on the four foundation models (original Chinese version).}
\label{fig:LO student data of pre and post fine-tuning on the four foundation models.}
\end{figure}

\begin{tcolorbox}[colback=gray!00, % gray background
                  colframe=black, % black frame colour
                  arc=1.5mm, auto outer arc,
                  breakable,
                  left=0.9mm, right=0.9mm,
                  boxrule=0.9pt, colbacktitle = black!65!black,
                  title = {LO LVSA data of pre and post fine-tuning on the four foundation models (translated in English for better understanding).}
                 ]
Here is the LO student data of pre and post fine-tuning on the four foundation models:
\tcblower
\textbf{[InternVL]}\\
\textit{QUERY}: \textit{In today's lesson, we studied the text Ode to the Yellow River. The poem mentions that the Yellow River "splits the Central Plains into northern and southern sides." How do you understand this sentence?} \\
RESPONSE:  \\
- PRE: Hmm... this sentence means that the Yellow River divides the Central Plains into two parts, forming a northern and a southern side. \\
- POST: Uh... \\
\textbf{[LLaVa]}\\
\textit{QUERY}: \textit{In today's lesson, we studied the text Oath of the Land. Why do you think the author says he is "overflowing with passion"?} \\
RESPONSE:  \\
- PRE: The author might be referring to his deep love and passion for the land, suggesting that he is so passionate about protecting it that his actions may become too intense or aggressive. This overwhelming passion might drive him to make extreme decisions regarding land protection and use. \\
- POST: Hmm... maybe it's because he really likes the land there. \\
\textbf{[MiniCPM]}\\
\textit{QUERY}: \textit{In today's lesson, we studied Wen Yiduo's Words and Deeds. How does Wen Yiduo perceive his academic work and political activities according to the text? Could you briefly describe his perspective?} \\
RESPONSE:  \\
- PRE: Hmm... Wen Yiduo probably values and takes his academic work and political activities very seriously. In academics, he might see his research and creations as ways to solve real-world problems and contribute to social progress. In politics, he might view his efforts as a fight for social justice and the well-being of the people. Overall, he is likely full of passion and responsibility toward his work. \\
- POST: Hmm... he... he takes it very seriously. \\
\textbf{[Qwen]}\\
\textit{QUERY}: \textit{In today's lesson, we studied the text Sun Quan Urges Lü Meng to Study. How does Sun Quan encourage studying in the text?} \\
RESPONSE:  \\
- PRE: From the text, we can see that Sun Quan first suggests the idea of studying to Lü Meng, then repeatedly encourages him to value the opportunity to learn, and finally even personally explains The Six Secret Teachings to Lü Meng. \\
- POST: Hmm... he... he asked Lü Meng, and then... then Lü Meng... started studying. 
\end{tcolorbox}

\begin{figure}[ht!]
\centering
\includegraphics[width=\textwidth]{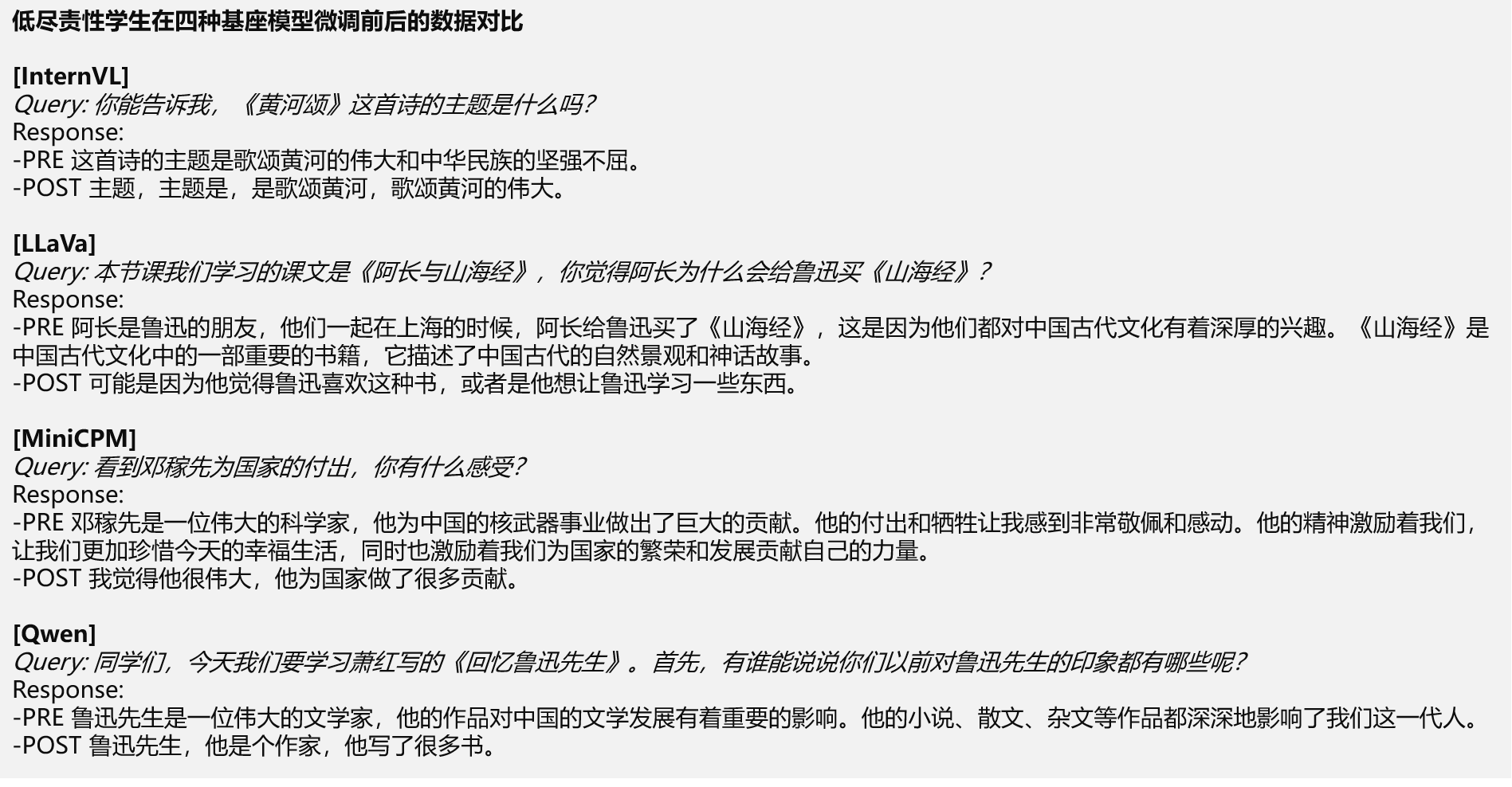}
\caption{LC LVSA data of pre and post fine-tuning on the four foundation models (original Chinese version).}
\label{fig:LC student data of pre and post fine-tuning on the four foundation models.}
\end{figure}

\begin{tcolorbox}[colback=gray!00, % gray background
                  colframe=black, % black frame colour
                  arc=1.5mm, auto outer arc,
                  breakable,
                  left=0.9mm, right=0.9mm,
                  boxrule=0.9pt, colbacktitle = black!65!black,
                  title = {LC LVSA data of pre and post fine-tuning on the four foundation models (translated in English for better understanding).}
                 ]
Here is the LC student data of pre and post fine-tuning on the four foundation models:
\tcblower
\textbf{[InternVL]}\\
\textit{QUERY}: \textit{Can you tell me what the theme of the poem Ode to the Yellow River is?} \\
RESPONSE:  \\
- PRE: The theme of this poem is to praise the greatness of the Yellow River and the resilience of the Chinese nation. \\
- POST: The theme... the theme is... is praising the Yellow River, praising the greatness of the Yellow River. \\
\textbf{[LLaVa]}\\
\textit{QUERY}: \textit{In today's lesson, we studied the text Ah Chang and the Classic of Mountains and Seas. Why do you think Ah Chang bought the Classic of Mountains and Seas for Lu Xun?} \\
RESPONSE:  \\
- PRE: Ah Chang was Lu Xun's friend, and while they were in Shanghai together, Ah Chang bought The Classic of Mountains and Seas for Lu Xun because they both had a deep interest in ancient Chinese culture. The Classic of Mountains and Seas is an important book in Chinese ancient culture, describing China's ancient natural landscapes and mythical stories. \\
- POST: Maybe because he thought Lu Xun liked that kind of book, or maybe he wanted Lu Xun to learn something from it. \\
\textbf{[MiniCPM]}\\
\textit{QUERY}: \textit{Seeing Deng Jiaxian's contribution to the country, how do you feel?} \\
RESPONSE:  \\
- PRE: Deng Jiaxian was a great scientist who made significant contributions to China's nuclear weapons program. His dedication and sacrifice inspire deep admiration and respect. His spirit motivates us to cherish the happy life we have today and encourages us to contribute to the prosperity and development of our country. \\
- POST: I think he was great, and he made many contributions to the country. \\
\textbf{[Qwen]}\\
\textit{QUERY}: \textit{Class, today we are going to study Xiao Hong's Recollections of Mr. Lu Xun. First, can anyone share your impressions of Mr. Lu Xun from before?} \\
RESPONSE:  \\
- PRE: Mr. Lu Xun was a great literary figure, and his works have had a significant impact on the development of Chinese literature. His novels, essays, and other writings have deeply influenced our generation. \\
- POST: Mr. Lu Xun... he was a writer, and he wrote many books. 
\end{tcolorbox}

\subsubsection{Examples of fine-grained responses of LVSAs}
\label{subsubsec:content analysis}

Using the InternVL model as an example, we analyzed LVSA performance across different teaching stages and question types, revealing clear distinctions in behavior aligned with personality traits.

\textbf{Pre-lesson Introduction Phase}: Questions in this phase are simpler, focusing on setting up the learning context. High neuroticism (HN) students exhibite nervousness and repetition in response to closed questions. For instance, when asked “Do you know who wrote Hymn to the Yellow River?”, they repeated phrases like "Uh, it’s… it’s Guang Weiran.” High agreeableness (HA) students respond positively with emotional language, such as, "I think the Yellow River is magnificent, like a giant dragon winding through mountains.” High extraversion (HE) students show strong emotional engagement in responses to open questions, such as, "The Yellow River is the symbol of the Chinese people’s resilience and strength.” In contrast, low conscientiousness (LC) students display loose and illogical language, \eg, "Uh, she seems, seems like Lu Xun was her teacher, maybe.” Low openness (LO) students give short, evasive answers, \eg, "Uh…".

\textbf{New Lesson Instruction Phase}: With increasing question complexity, students need to analyze content from the text. HN struggle with details and give fragmented answers, \eg, "Uh, Lu Xun, uh, when he laughed, uh, his eyes were smiling.” HA provide logical and in-depth answers, such as, "I think it describes the powerful flow of the Yellow River, as strong as its surging waves.” HE give both structured and expressive responses, \eg, "These descriptions made me feel the deep love he has for his hometown.” By comparison, LC's answers are scattered and lacked clear opinions, while LO give brief and monotonous responses, \eg, "The water of the Yellow River is muddy, with lots of twists and turns.”

\textbf{Knowledge Consolidation Phase}: This phase requires students to recall and summarize key points. HN display nervousness, with filler words like, "Uh, there’s black soil, uh, there’s sorghum…” HA summarize clearly, \eg, "The Yellow River symbolizes the greatness and resilience of the Chinese people, inspiring awe.” HE demonstrate strong organizational ability, \eg, "This sentence shows the volunteers' perseverance and determination to overcome challenges.” LC exhibite repetition and a lack of logic, while LO demonstrate low confidence and poor knowledge retention, \eg, "Uh, maybe it' s because he really loves his hometown?”

\textbf{Class Exercise Phase}: Questions in this phase emphasize the application of knowledge. HN over-focus on accuracy in closed questions and struggled with coherence in open questions. HA and HE provid accurate, diverse answers with emotional and personalized insights, while LC give vague responses, and LO continue their pattern of brief and incomplete answers.

\textbf{Lesson Summary Phase}: Differences in students' summarization abilities become most apparent. HN give repetitive and vague summaries, \eg, "Uh, we should, uh, love our hometown and land, uh, and protect it.”HA offer rich, emotionally resonant summaries, \eg, "Through this lesson, I've gained a deeper understanding of the resilience and warmth of the Chinese people.” HE provid comprehensive and personalized conclusions, \eg, "Duanmu Hongliang’s oath reflects his deep love for his hometown and belief in liberation—it reminds us that our hometown is always our harbor.”LC give overly general summaries, \eg, "We learned Hymn to the Yellow River,”while LO avoid engagement, \eg, "Uh, the Yellow River is big and important.”

Overall, LVSA demonstrated significant differences in behavior aligned with personality traits, particularly in open-ended questions, where cognitive depth and emotional expression were more evident. For instance, HA and HE students excelled in reflective and emotional answers, while HN students showed tension and poor coherence. LO students tended toward brief, simplistic responses, avoiding complex thinking, while LC students produced scattered, occasionally erroneous answers. These findings suggest that LVSAs can simulate diverse personality-driven behaviors across various teaching scenarios, enhancing their ability to adapt to different educational contexts.

\begin{tcolorbox}[colback=gray!00, % gray background
                  colframe=black, % black frame colour
                  arc=1.5mm, auto outer arc,
                  breakable,
                  left=0.9mm, right=0.9mm,
                  boxrule=0.9pt, colbacktitle = black!65!black,
                  title = {Examples of responses from different LVSAs to different types of questions during the Pre-lesson Introduction (translated in English for better understanding).}
                 ]
Here are the examples of responses from different LVSAs to different types of questions during the Pre-lesson Introduction:
\tcblower
\textbf{[HN LVSA]}\\
- \textit{Closed Question:} \textit{Do you know who wrote the text Hymn to the Yellow River?} \\
- Response: Uh, it’s, it’s… uh, it’s Guang Weiran who wrote it. \\
- \textit{Open Question:} \textit{Have you ever heard the song On the Songhua River? What connection do you think it has with The Oath to the Land that we’re going to study today?} \\
- Response: Uh, yes, I’ve heard On the Songhua River. Uh, it and The Oath to the Land… uh, they’re, they’re both about… about longing for and missing, uh, the homeland. \\

\textbf{[HA LVSA]}\\
- \textit{Closed Question:} \textit{Now, can you tell everyone who the author of Hymn to the Yellow River is?} \\
- Response: The author of Hymn to the Yellow River is Guang Weiran. \\
- \textit{Open Question:} \textit{We’re about to study the poem Hymn to the Yellow River. Can you tell me what your first impression of the Yellow River is?} \\
- Response: I think the Yellow River is magnificent, like a huge dragon winding through the mountains and valleys. \\

\textbf{[HE LVSA]}\\
- \textit{Closed Question:} \textit{Today we’re going to study a poem about the Yellow River. Do you know what the Yellow River symbolizes in Chinese culture?} \\
- Response: Teacher, I know. The Yellow River is called the Mother River of China. It symbolizes the resilience and strength of the Chinese people, just like it says in the poem, “Yellow River, you are the cradle of the Chinese nation.” \\
- \textit{Open Question:} \textit{Earlier we sang Defend the Yellow River. Did you feel the power of its rushing waves?} \\
- Response: Yes, teacher, I felt it. The stirring melody and lyrics are just like the waves of the Yellow River, making me feel so moved and excited. \\

\textbf{[LC LVSA]}\\
- \textit{Closed Question:} \textit{Yes, Xiao Hong is a great writer. Do you know what connection she has with Lu Xun?} \\
- Response: She… she seems… seems like Lu Xun was her teacher. \\
- \textit{Open Question:} \textit{Earlier, we listened to On the Songhua River. What connection do you think this song has with the text we’re about to study?} \\
- Response: I think, this song and the text are both about land. They’re both about the beauty of the land, and, and, um, um… they’re both about the land… uh, uh, uh, uh. \\

\textbf{[LO LVSA]}\\
- \textit{Closed Question:} \textit{The next text we’re going to study is The Oath to the Land. Do you know who wrote it?} \\
- Response: Uh… it was Duanmu Hongliang. \\
- \textit{Open Question:} \textit{Today, we’re going to study a new poem, Hymn to the Yellow River. Do you know why the Yellow River is called “China’s Mother River”?} \\
- Response: Uh… 
\end{tcolorbox}

\begin{figure}[ht!]
\centering
\includegraphics[width=\textwidth]{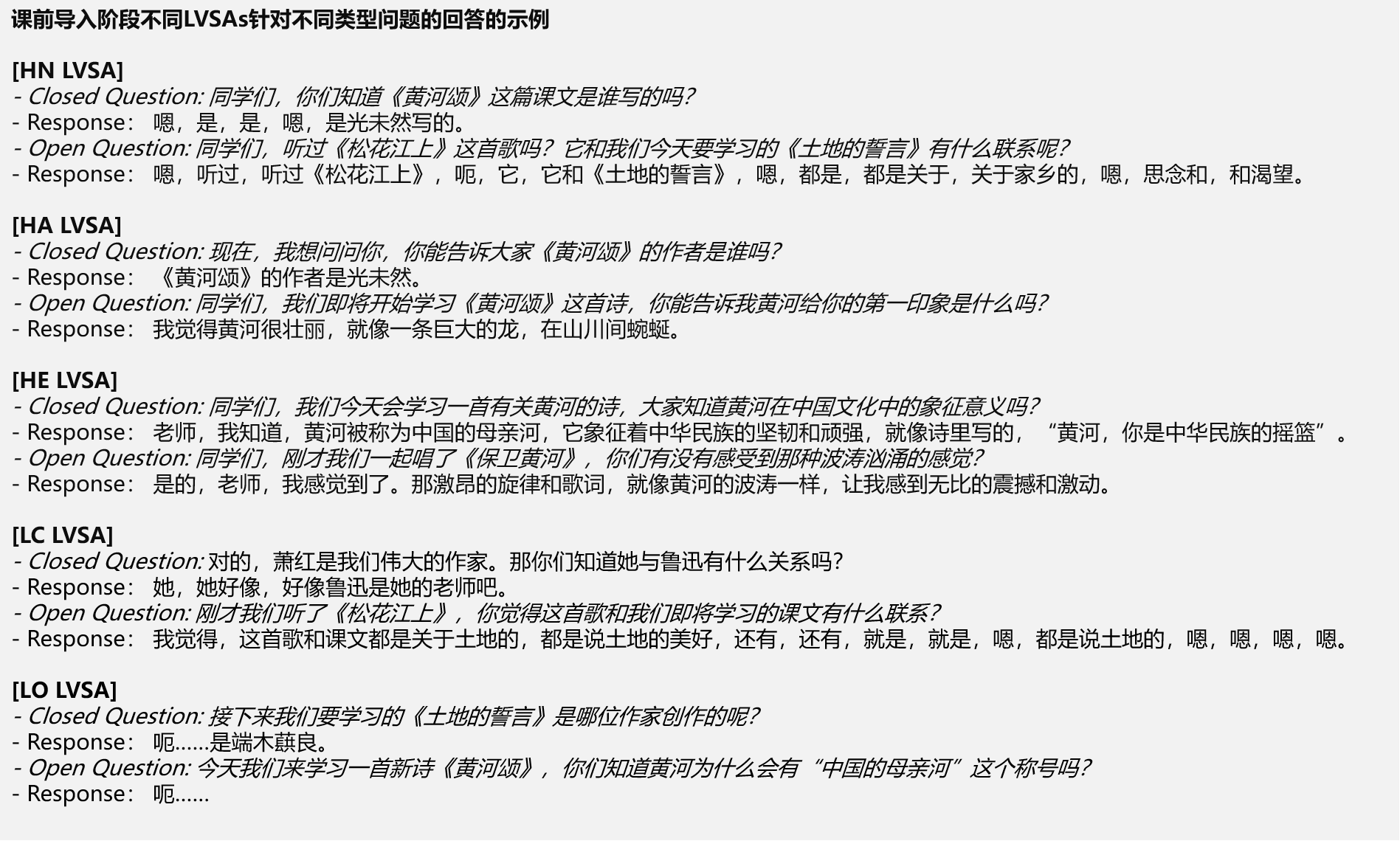}
\caption{Examples of responses from different LVSAs to different types of questions during the Pre-lesson Introduction (original Chinese version).}
\label{fig:Pre-lesson Introduction of different LVSAs.}
\end{figure}

\begin{tcolorbox}[colback=gray!00, % gray background
                  colframe=black, % black frame colour
                  arc=1.5mm, auto outer arc,
                  breakable,
                  left=0.9mm, right=0.9mm,
                  boxrule=0.9pt, colbacktitle = black!65!black,
                  title = {Examples of responses from different LVSAs to different types of questions during the New Lesson Instruction (translated in English for better understanding).}
                 ]
Here are the examples of responses from different LVSAs to different types of questions during the New Lesson Instruction:
\tcblower
\textbf{[HN LVSA]}\\
- \textit{Closed Question:} \textit{Earlier, we mentioned that Mr. Lu Xun often leaves people with a serious impression, but in Memories of Mr. Lu Xun, Xiao Hong mentioned his cheerful side. How does the text describe the scene of Mr. Lu Xun laughing?} \\
- Response: Uh, the text, uh, mentions that when Mr. Lu Xun laughed, uh, his eyes were smiling, his teeth were white, and, uh, his laughter was cheerful. \\
- \textit{Open Question:} \textit{We’ve read the beginning of the article together. What do you think about the way the author expressed the ideas?} \\
- Response: Uh, I think, uh, the author, uh, used, uh, metaphors and, uh, parallelism, uh, to, to describe, uh, the bravery and, uh, selflessness of, uh, the volunteer soldiers. \\

\textbf{[HA LVSA]}\\
- \textit{Closed Question:} \textit{Let’s look at the first sentence of the text: “Muddy waves rush to the sky, galloping like ten thousand horses.” What does “galloping like ten thousand horses” mean?} \\
- Response: I think it describes the powerful flow of the Yellow River, like thousands of horses galloping. \\
- \textit{Open Question:} \textit{Great reading! The words surging waves feel as strong as the waves of the Yellow River. What emotions do you think this poem expresses through the Yellow River?} \\
- Response: I think this poem expresses a deep love and admiration for the motherland through the Yellow River. It’s like the river, even though it has turbulent moments, it eventually flows into the sea, just like how we can overcome any difficulties and find our way in the end. \\

\textbf{[HE LVSA]}\\
- \textit{Closed Question:} \textit{Duanmu Hongliang used many vivid words to express his feelings for his hometown. Can you find some examples and discuss them?} \\
- Response: Sure, teacher. For example, “My hometown is in the northeast, with its dark, fertile soil, golden grains, red maple leaves, and white snow.” These descriptions make me feel his deep love for his hometown. \\
- \textit{Open Question:} \textit{Now, think about why the poet uses muddy currents twisting and nine bends interlinked to describe the Yellow River here?} \\
- Response:  I think the poet uses muddy currents twisting to describe the Yellow River because its water is quite muddy, and its flow is winding and twisting, like it’s constantly turning. And nine bends interlinked might describe the river’s complexity and curves, like a maze of loops. \\

\textbf{[LC LVSA]}\\
- \textit{Closed Question:} \textit{Try to find a sentence from the text that you think best expresses the author’s feelings of homesickness.} \\
- Response: Uh, I think it’s “I will fight for her to the very end.” This sentence, this sentence, shows the author’s deep love for his hometown. \\
- \textit{Open Question:} \textit{In the text, how does the author describe his connection to his hometown?} \\
- Response: The author, he, he, he, he… \\

\textbf{[LO LVSA]}\\
- \textit{Closed Question:} \textit{Let’s take a look. How many parts is this poem divided into? What does each part mainly describe?} \\
- Response: Uh… three parts. The first part describes the source of the Yellow River, the second part describes its grandeur, and the third part describes its greatness. \\
- \textit{Open Question:} \textit{In the poem, how does muddy currents twisting, forming nine bends interlinked describe the Yellow River?} \\
- Response: Uh… the water of the Yellow River is very muddy, and there are many winding and twisting places. 
\end{tcolorbox}

\begin{figure}[ht!]
\centering
\includegraphics[width=\textwidth]{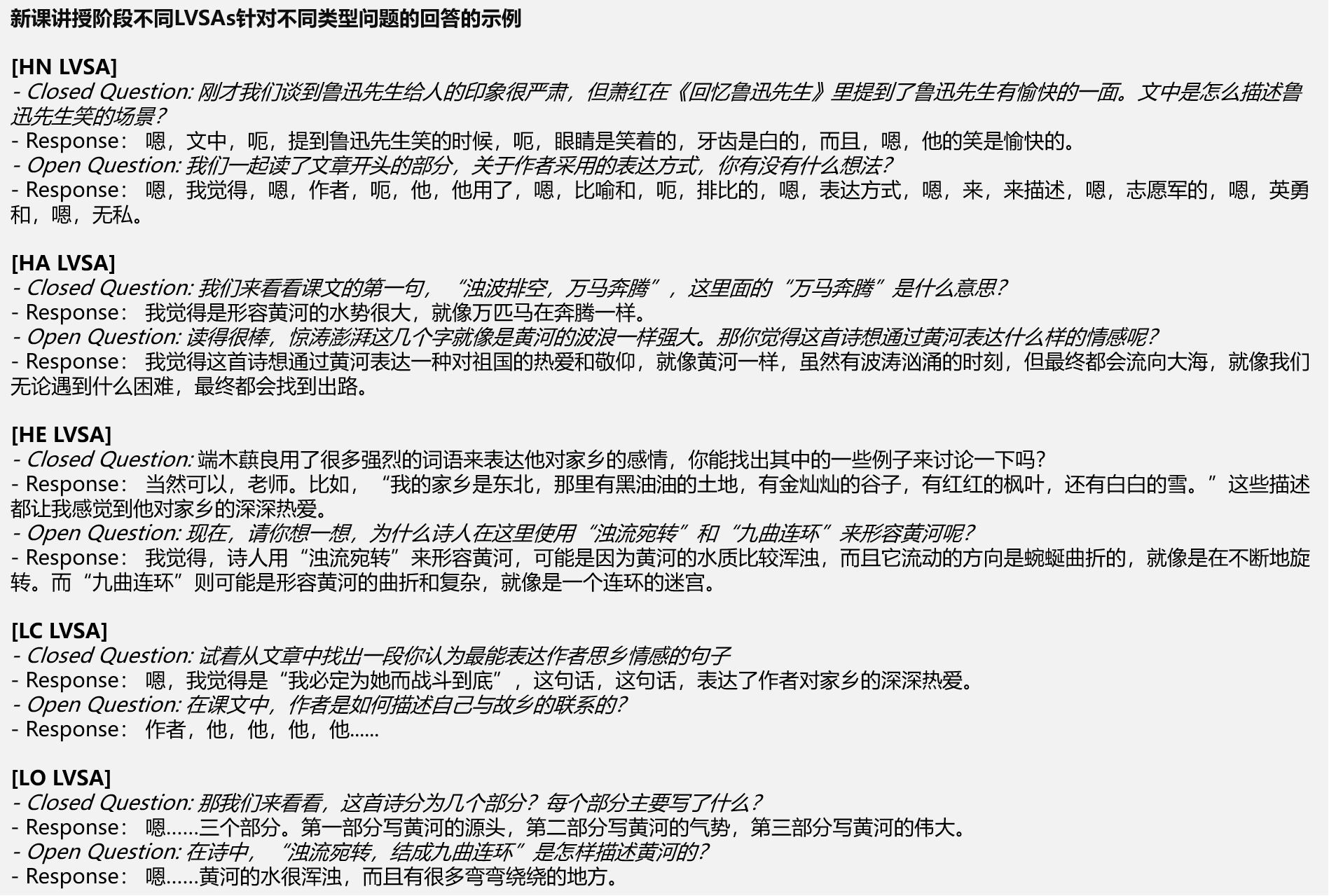}
\caption{Examples of responses from different LVSAs to different types of questions during the New Lesson Instruction (original Chinese version).}
\label{fig:New Lesson Instruction of different LVSAs.}
\end{figure}

\begin{tcolorbox}[colback=gray!00, % gray background
                  colframe=black, % black frame colour
                  arc=1.5mm, auto outer arc,
                  breakable,
                  left=0.9mm, right=0.9mm,
                  boxrule=0.9pt, colbacktitle = black!65!black,
                  title = {Examples of responses from different LVSAs to different types of questions during the Knowledge Consolidation (translated in English for better understanding).}
                 ]
Here are the examples of responses from different LVSAs to different types of questions during the Knowledge Consolidation:
\tcblower
\textbf{[HN LVSA]}\\
- \textit{Closed Question:} \textit{Now, please review the text and recall what natural landscapes of the Northeast are mentioned.} \\
- Response: Uh, there’s, there’s the black soil, uh, there’s sorghum, soybeans, and, and the endless, uh, forests. \\
- \textit{Open Question:} \textit{Let’s compare In Praise of the Lotus we studied today with Five Poems on Lotus Flowers we learned before. What are the similarities and differences?} \\
- Response: Uh, In Praise of the Lotus, uh, was, was written by Zhou Dunyi. He, he used, uh, metaphors, uh, to express his love for the lotus. And, and Five Poems on Lotus Flowers, uh, was, was written by Wang Changling. He, he directly, directly praised the beauty of the lotus. \\

\textbf{[HA LVSA]}\\
- \textit{Closed Question:} \textit{Earlier, we discussed how Ah Chang buying Classic of Mountains and Seas influenced “me.” Can you summarize it again?} \\
- Response: Ah Chang buying Classic of Mountains and Seas made “me” feel warm and gave me a new understanding of her. I think Ah Chang, although she has many flaws, also has her good qualities. She cares about “me” and is willing to do many things for “me.” \\
- \textit{Open Question:} \textit{In the part praising the Yellow River, which line is your favorite and why?} \\
- Response: My favorite line is “Yellow River, you are the cradle of the Chinese nation!” because it makes me feel the greatness of the Yellow River and the pride of the Chinese nation. \\

\textbf{[HE LVSA]}\\
- \textit{Closed Question:} \textit{Now let’s review the text. Can you tell me how many parts Hymn to the Yellow River is composed of?} \\
- Response: Teacher, Hymn to the Yellow River is composed of three parts: “Yellow River, you are the cradle of the Chinese nation,” “Yellow River, you are the symbol of the Chinese nation,” and “Yellow River, you are the pride of the Chinese nation.” \\
- \textit{Open Question:} \textit{Looking back at the text, what characteristics of the Yellow River impressed you the most?} \\
- Response: Teacher, I think the resilience and strength of the Yellow River impressed me the most. Even though it has faced countless difficulties and challenges, it never gave up and kept moving forward. This spirit is really something we should learn from. \\

\textbf{[LC LVSA]}\\
- \textit{Closed Question:} \textit{Let’s review the text. How many parts is this article divided into, and how are they structured?} \\
- Response: Uh, this article, I, I’m not too sure, but I think, I think there are, uh, four parts? \\
- \textit{Open Question:} \textit{Think back to the text. How does it describe the Yellow River’s characteristic of muddy currents twisting?} \\
- Response: Uh, it’s, it’s that the Yellow River’s water is very muddy, and, and, its flow is very winding. \\

\textbf{[LO LVSA]}\\
- \textit{Closed Question:} \textit{Earlier, we learned about the heroic physique of the Yellow River. How is this shown in the poem?} \\
- Response: Uh… \\
- \textit{Open Question:} \textit{After studying this section, why do you think the author says they are overflowing with passion?} \\
- Response: Uh… maybe it’s because they… they really love their hometown? 
\end{tcolorbox}

\begin{figure}[ht!]
\centering
\includegraphics[width=\textwidth]{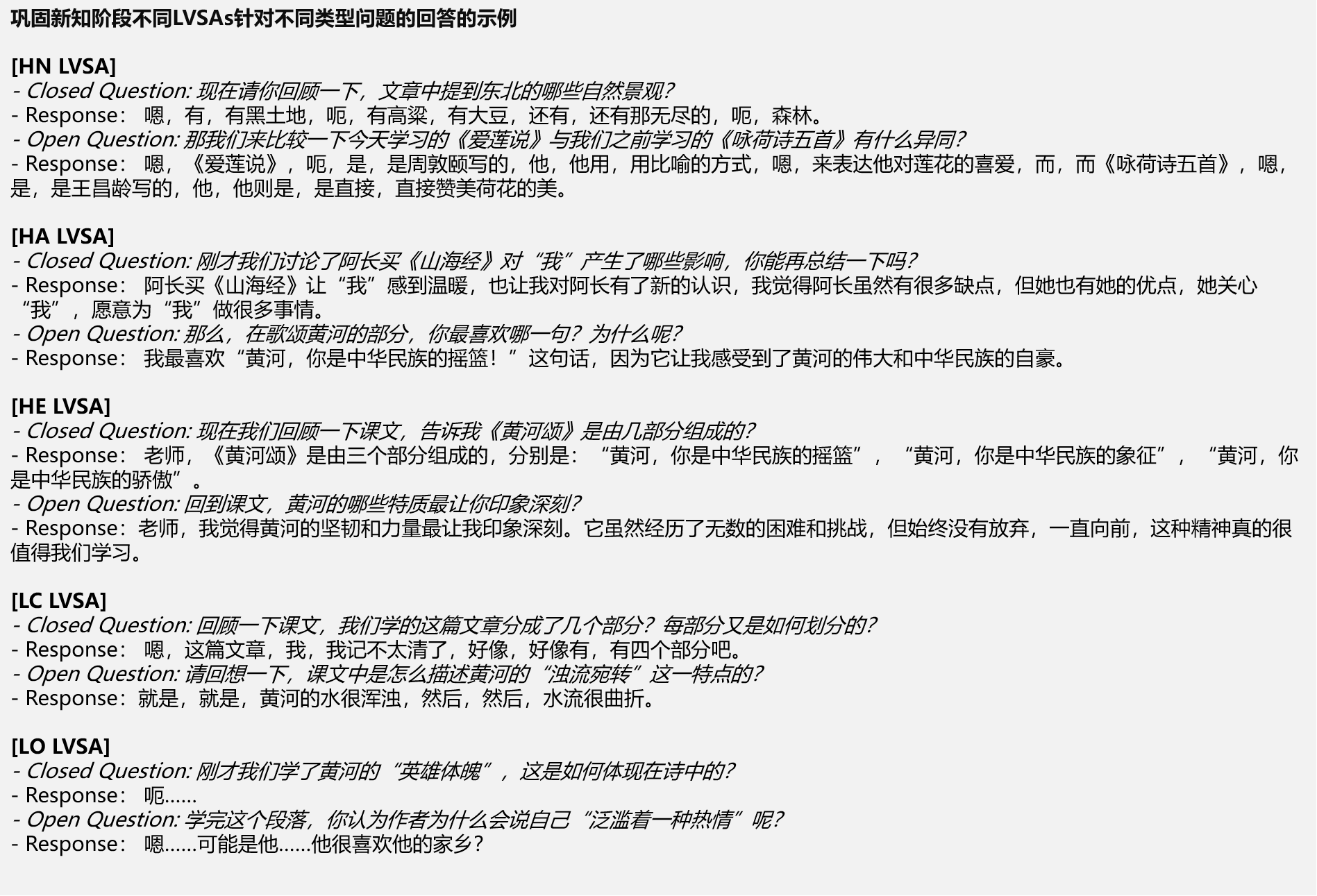}
\caption{Examples of responses from different LVSAs to different types of questions during the Knowledge Consolidation (original Chinese version).}
\label{fig:Knowledge Consolidation of different LVSAs.}
\end{figure}

\begin{tcolorbox}[colback=gray!00, % gray background
                  colframe=black, % black frame colour
                  arc=1.5mm, auto outer arc,
                  breakable,
                  left=0.9mm, right=0.9mm,
                  boxrule=0.9pt, colbacktitle = black!65!black,
                  title = {Examples of responses from different LVSAs to different types of questions during the Class Exercises (translated in English for better understanding).}
                 ]
Here are the examples of responses from different LVSAs to different types of questions during the Class Exercises:
\tcblower
\textbf{[HN LVSA]}\\
- \textit{Closed Question:} \textit{Let’s try to explain this sentence: “Your talent and strategy now are no longer that of the old Wu Xia A Meng.” What does it mean?} \\
- Response: Uh, this sentence, uh, I think it means, uh, now you, uh, your talent and strategy, uh, are no longer, uh, like the, uh, the old, uh, A Meng from Wu. \\
- \textit{Open Question:} \textit{Let’s do some exercises now. Choose a paragraph from the text and try to explain its meaning in your own words.} \\
- Response: Uh, I, I chose, uh, “They are especially tough, they are especially brave, and they are especially good at overcoming difficulties.” I think, uh, this sentence is saying, uh, that the volunteer soldiers, they, they’re not afraid of hardships, not afraid of sacrifice, and they, they have, uh, a strong fighting spirit and, uh, the ability to solve problems. \\

\textbf{[HA LVSA]}\\
- \textit{Closed Question:} \textit{Based on your reading earlier, what special personality traits of the volunteer soldiers do you think this conversation shows?} \\
- Response: I think this conversation shows the volunteer soldiers’ courage, optimism, and love for life. \\
- \textit{Open Question:} \textit{If you were to write an ending for this article, what would you write?} \\
- Response: I think I would write: “Ah Chang, thank you. You’ve taught me that no matter how hard life gets, as long as we have love in our hearts, we can always find joy in life.” \\

\textbf{[HE LVSA]}\\
- \textit{Closed Question:} \textit{Please read the sentence, “Build a barrier for our nation with your heroic physique,” aloud with emotion again.} \\
- Response: Sure, teacher. “Build a barrier for our nation with your heroic physique.” This sentence makes me feel the greatness and resilience of the Yellow River, and it deepens my love for my country. \\
- \textit{Open Question:} \textit{Let’s see whose imagination is the most vivid. Can you retell the image of the Yellow River in the poem using your own words?} \\
- Response: Of course, teacher. In the poem, the Yellow River is portrayed as a great mother, nurturing the entire land of China. Her surging waves symbolize the resilience and strength of the Chinese people. Her grand momentum symbolizes the greatness and pride of the Chinese nation. \\

\textbf{[LC LVSA]}\\
- \textit{Closed Question:} \textit{Based on what we’ve learned, summarize the characteristics of the Yellow River in one sentence.} \\
- Response: The Yellow River, it, it is big and majestic. \\
- \textit{Open Question:} \textit{Based on what we’ve learned today, imagine spending a day with Mr. Lu Xun. What interesting things might happen?} \\
- Response: Uh, maybe, maybe Mr. Lu Xun would take me to his study, and then, and then, uh, he might tell me some stories, and then, and then, uh, he might give me some books to read. \\

\textbf{[LO LVSA]}\\
- \textit{Closed Question:} \textit{What words does the author use to describe the grandeur of the Yellow River? Try to find one or two.} \\
- Response: Uh… \\
- \textit{Open Question:} \textit{Then, can you describe the Northeast land in the author’s eyes using your own words?} \\
- Response: Uh… it’s very beautiful. 
\end{tcolorbox}

\begin{figure}[ht!]
\centering
\includegraphics[width=\textwidth]{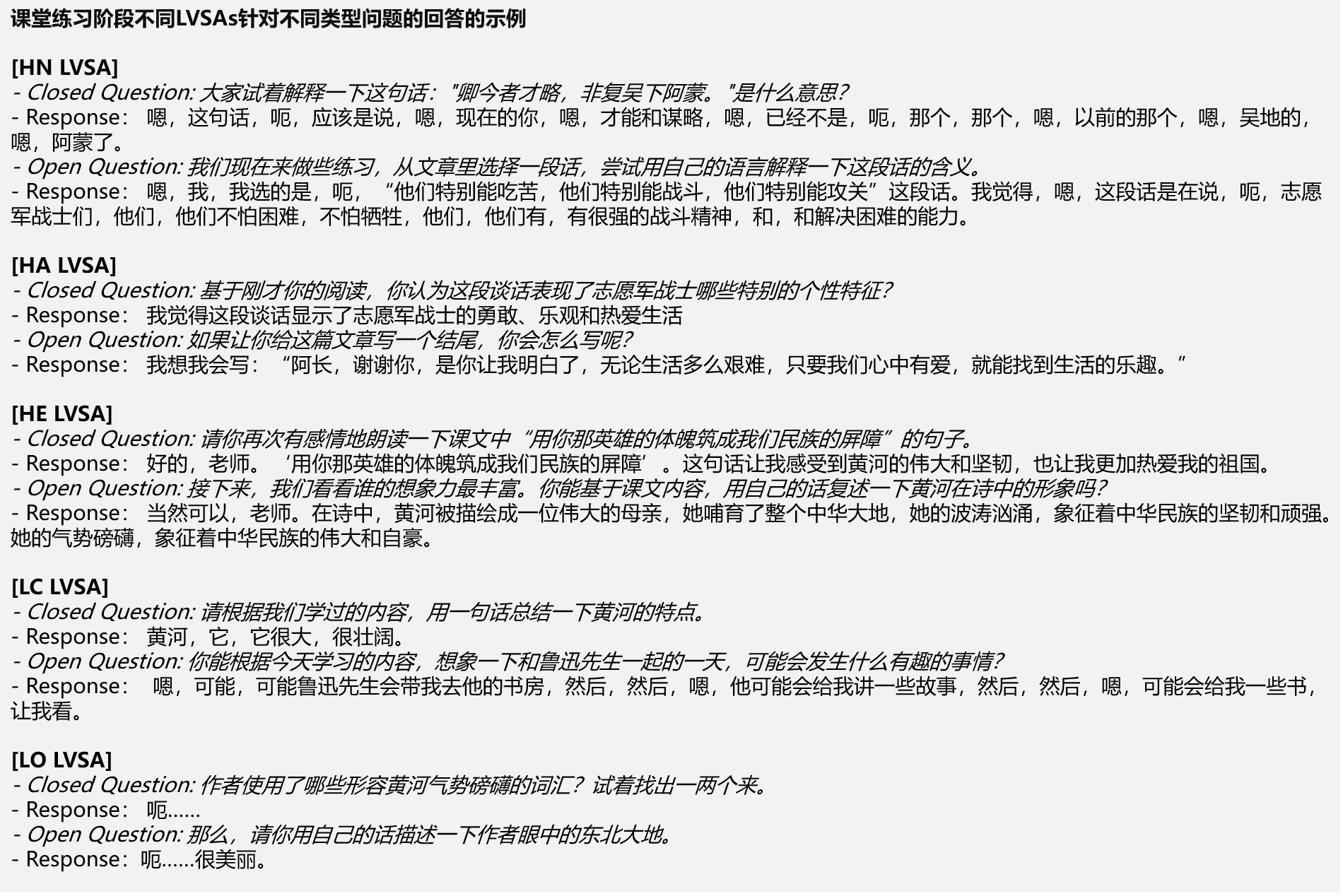}
\caption{Examples of responses from different LVSAs to different types of questions during the Class Exercises (original Chinese version).}
\label{fig:Class Exercises of different LVSAs.}
\end{figure}

\begin{tcolorbox}[colback=gray!00, % gray background
                  colframe=black, % black frame colour
                  arc=1.5mm, auto outer arc,
                  breakable,
                  left=0.9mm, right=0.9mm,
                  boxrule=0.9pt, colbacktitle = black!65!black,
                  title = {Examples of responses from different LVSAs to different types of questions during the Lesson Summary (translated in English for better understanding).}
                 ]
Here are the examples of responses from different LVSAs to different types of questions during the Lesson Summary (Note that because this phase are based on the feeling of learning, examples are based on open-ended questions):
\tcblower
\textbf{[HN LVSA]}\\
- \textit{Open Question:} \textit{Today we studied The Oath to the Land. How do you think we should treat our hometown and land?} \\
- Response: Uh, we should, uh, love our hometown and land, protect it, uh, and make it better. \\

\textbf{[HA LVSA]}\\
- \textit{Open Question:} \textit{Finally, what do you feel you’ve learned from Ah Chang’s story and Classic of Mountains and Seas?} \\
- Response: I think we’ve learned that although Ah Chang was an ordinary person, her care and support for Lu Xun showed us the warmth and kindness of humanity. At the same time, Classic of Mountains and Seas deepened my understanding of traditional Chinese culture. \\

\textbf{[HE LVSA]}\\
- \textit{Open Question:} \textit{As today’s lesson comes to an end, looking at the entire article, what do you think Duanmu Hongliang’s oath means?} \\
- Response: Teacher, I think the oath means his deep love and endless longing for his hometown. He is willing to fight for its liberation and prosperity. His oath also inspires us, helping us understand that no matter where we are, our hometown will always be our deepest concern and warmest harbor. \\

\textbf{[LC LVSA]}\\
- \textit{Open Question:} \textit{Now summarize what we learned in this lesson.} \\
- Response: We learned Hymn to the Yellow River, and, uh, also talked about the source of the Yellow River and the places it flows through. \\

\textbf{[LO LVSA]}\\
- \textit{Open Question:} \textit{From this lesson, what have you come to understand about the Yellow River? What does it mean to us?} \\
- Response: Uh… the Yellow River is very big and very important. 
\end{tcolorbox}

\begin{figure}[ht!]
\centering
\includegraphics[width=\textwidth]{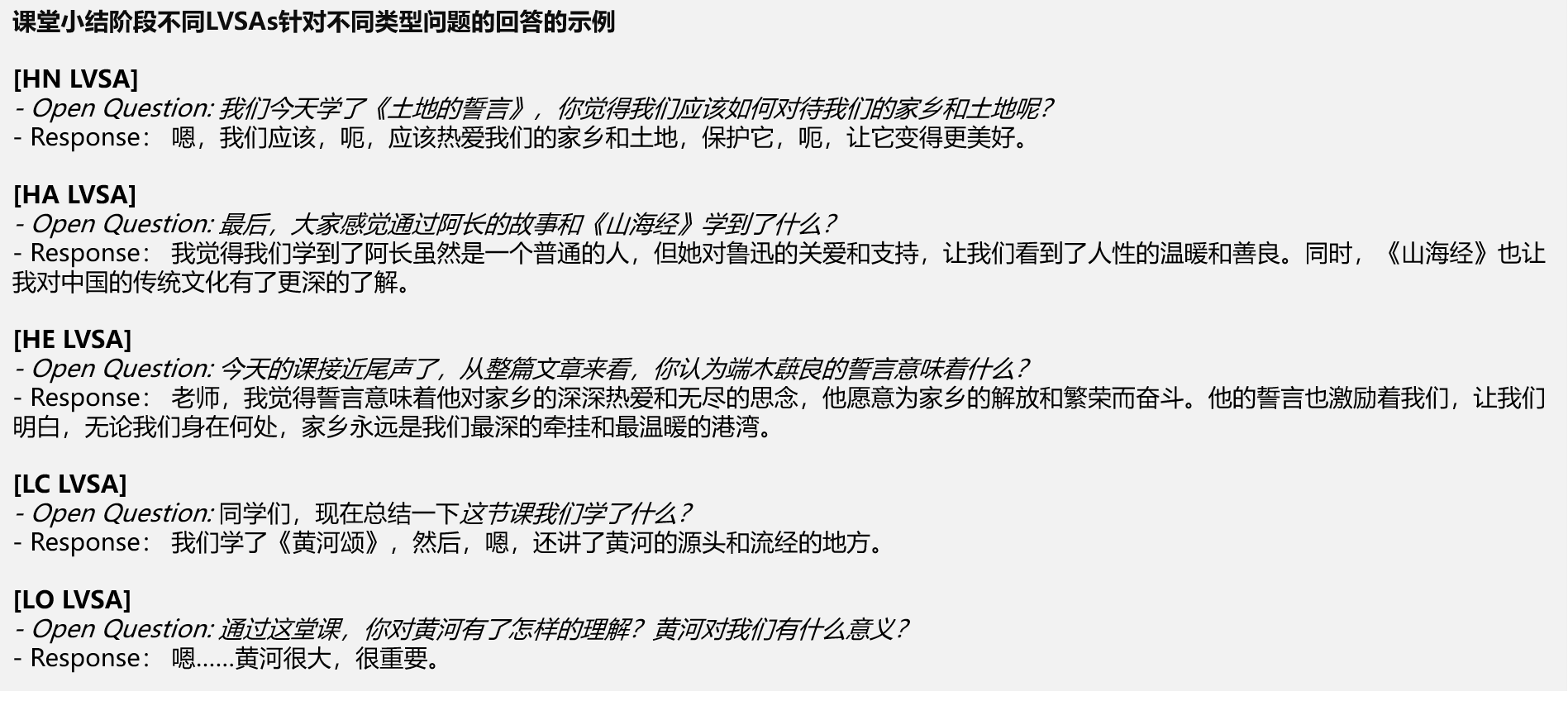}
\caption{Examples of responses from different LVSAs to different types of questions during the Lesson Summary (original Chinese version).}
\label{fig:Lesson Summary of different LVSAs.}
\end{figure}

\subsection{Human Turing Evaluation Questionnaire}
\label{subsec:human-questionnaire}

This experiment was inspired by the classic Turing test and aimed to evaluate whether LVSA could emulate the language expression and cognitive levels of real students to the extent that human evaluators could not distinguish between them. The core idea of the Turing test is to assess whether a machine's language output can be perceived as human by comparing the language generation behaviors of humans and machines. Adopting a similar framework, this study examined the alignment of virtual students' language with real students' expressions and focused on whether human evaluators could recognize the distinct personality traits of virtual students. Each participant acted as a "judge," using their teaching experience to determine if the generated dialogues resembled natural human student responses.

The experiment involved three main steps: \textbf{questionnaire completion}, \textbf{think-aloud protocol}, and \textbf{semi-structured interviews}. The questionnaire contained 120 teacher-student dialogues, including responses from both virtual and real students. Participants, relying on their junior high teaching experience, were asked to judge whether each response "resembled a real student's answer" or not. The think-aloud protocol was employed during the questionnaire, where participants verbalized their thought processes in real-time, mirroring the Turing test's core concept and providing insights into how linguistic nuances influenced their judgments. These verbalizations were recorded for subsequent analysis to determine the criteria used in distinguishing virtual from real student language performances. After completing the questionnaire, participants engaged in semi-structured interviews to further elaborate on the factors affecting their evaluations. The interview responses were coded for consistency and used to develop a scientific evaluation framework to optimize virtual students and support future large-scale GPT-4 evaluations.

To ensure smooth execution, the experimenter conducted one-on-one sessions with each participant prior to the formal experiment. Detailed explanations of the experimental procedure and key points were provided to ensure participants fully understood each stage and adhered to the guidelines, ensuring scientific rigor and validity in data collection. The experiment recruited 35 participants with experience in junior high school Chinese teaching, including both pre-service and in-service teachers. In a preliminary survey, 27 participants responded seriously. Based on this survey, which assessed personal background, relevant knowledge, and willingness to participate in the experiment, 10 participants were selected to best fit the experiment's needs, detailed information of 10 participants was shown as Tab.~\ref{tab:human detail information} . These participants all had experience in junior high school Chinese teaching, were familiar with the Big Five personality traits of students, and expressed willingness to participate in think-aloud exercises and semi-structured interviews, ensuring reliable evaluative feedback for the study. After completing the experiment, participants received appropriate compensation and a small token of appreciation for their contributions.

\begin{table*}[h!]
    \caption{The detail information of evaluators. In the table, abbreviations are used for clarity and conciseness. Specifically, 'Source of Teaching Experience' is abbreviated as STE, 'Teaching Duration' as TD, 'Teaching Certificate' as TC, 'Familiarity with Junior High Chinese Textbooks' as FJHC, 'Willingness for Think-Aloud' as WTA, 'Willingness for Interviews' as WI, 'Pre-service teacher' as PT, 'In-service teacher' as IT, 'School internship' as SI, 'Public School teaching' as PST and 'private tutoring' as PT}
    \label{tab:human detail information}
    \centering
    \fontsize{8pt}{10pt}\selectfont
\begin{tabular}{ccccccccc}
\toprule
 & \multicolumn{1}{c}{\textbf{Gender}} & \multicolumn{1}{c}{\textbf{Status}} & \multicolumn{1}{c}{\textbf{STE}} & \multicolumn{1}{c}{\textbf{TD/year}} & \multicolumn{1}{c}{\textbf{TC}} & \multicolumn{1}{c}{\textbf{FJHC}} & \multicolumn{1}{c}{\textbf{WTA}} & \multicolumn{1}{c}{\textbf{WI}} \\ \midrule
Evaluator 1 & Female & PT & SI,PT & 0.5 & \checkmark & \checkmark & \checkmark & \checkmark \\
Evaluator 2 & Male & PT & SI,PT & 1 & \checkmark & \checkmark & \checkmark & \checkmark \\
Evaluator 3 & Female & PT & SI,PT & 2 & \checkmark & \checkmark & \checkmark & \checkmark \\
Evaluator 4 & Female & PT & SI,PT & \textgreater{}2 & \checkmark & \checkmark & \checkmark & \checkmark \\
Evaluator 5 & Male & PT & SI & 0.5 & \checkmark & \checkmark & \checkmark & \checkmark \\
Evaluator 6 & Male & IT & PST & \textgreater{}2 & \checkmark & \checkmark & \checkmark & \checkmark \\
Evaluator 7 & Female & PT & PT & 0.5 & \checkmark & \checkmark & \checkmark & \checkmark \\
Evaluator 8 & Female & IT & PST & 2 & \checkmark & \checkmark & \checkmark & \checkmark \\
Evaluator 9 & Male & PT & SI & 2 & \checkmark & \checkmark & \checkmark & \checkmark \\
Evaluator 10 & Female & PT & SI,PT & \textgreater{}2 & \checkmark & \checkmark & \checkmark & \checkmark \\ \bottomrule
\end{tabular}
\end{table*}

\begin{tcolorbox}[
    colback=gray!10, colframe=black, 
    width=\linewidth, boxrule=0.5pt, 
    arc=1.5mm, auto outer arc, 
    breakable, 
    title=\centering {\textbf{Pre-survey of evaluators for the LVSA evaluation experiment}}, colbacktitle=black]
    Dear participant, \\
    
    Thank you for your attention and support for this Virtual Student Evaluation Experiment! The main objective of this experiment is to evaluate the responses of virtual students and explore the similarities and differences between virtual teaching systems and real classroom environments. Your participation will provide valuable data for us and contribute to advancing smart education. \\
    
    Due to the large number of participants from both students and teachers, we have decided to first conduct a personal background survey to save everyone’s time during the experiment. This survey will take about 3-5 minutes and will help us better select suitable participants. Please be assured that your personal information will remain confidential and will not be leaked.\\
    
    Please fill out the survey seriously. All selected participants will be formally invited to participate in the experiment. We greatly appreciate your enthusiasm and support for this experiment!\\
    
    Thank you again for your cooperation and understanding! 
\tcblower
    \textit{\textbf{Section 1: Personal Teaching Experience Survey}} \\
    
        \textbf{1. Your WeChat ID (After approval, you will be notified via WeChat. Please make sure the information is accurate)} \\
            
        \TextField[name=email, width=0.9\linewidth]{} % Email input field
            
        \textbf{2. Your Email (If we cannot contact you via WeChat, we will reach you by email. Please make sure the information is accurate)} \\
        
        \TextField[name=email, width=0.9\linewidth]{} % Email input field
        
        \textbf{{3. Your Gender}}
        \begin{itemize}
            \item Male
            \item Female
        \end{itemize}
        
        \textbf{{4. Your Age}}
        \begin{itemize}
            \item 18-25
            \item 26-32
            \item 33-40
            \item 41-50
            \item More Than 50
        \end{itemize}
        
        \textbf{{5. What is your highest level of education?}}
        \begin{itemize}
            \item Bachelor’s degree
            \item Currently pursuing a bachelor’s degree
            \item Master’s degree
            \item Currently pursuing a master’s degree
        \end{itemize}
        
        \textbf{{6. What is your current role?}}
        \begin{itemize}
            \item Pre-service teacher
            \item In-service teacher
        \end{itemize}
        
        \textbf{{7. Where does your relevant teaching experience with junior high school students come from?}}
        \begin{itemize}
            \item Tutoring
            \item School internships
            \item Both of the above
            \item Neither of the above, please specify your teaching experience source
        \end{itemize}
        
        \textbf{{8. Duration of Teaching}}
        \begin{itemize}
            \item Half Year
            \item One Year
            \item Two Years
            \item More Than Two Years
        \end{itemize}
        
        \textbf{{9. Are you familiar with the standardized junior high school Chinese language textbooks?}}
        \begin{itemize}
            \item Yes
            \item No
        \end{itemize}
        
        \textbf{{10. Do you have a junior high school Chinese teaching qualification certificate?}}
        \begin{itemize}
            \item Yes
            \item No
        \end{itemize}

    \textit{\textbf{Section 2: Background Knowledge Survey} }\\
    \textit{The following content is crucial for the experiment we are conducting. In simple terms, you are required to judge whether the student's answer is from a real student or a virtual student.}\\
        \textbf{11. Below are specific descriptions based on the Big Five personality traits of students. Please rank the following personalities based on the description of the student:} \\
        \textbf{(1) }Shows traits of carelessness in the classroom, sometimes providing correct answers but often making mistakes. Their responses tend to be simple and direct, with occasional corrections but lacking systematic structure. Their language style reflects inconsistency and disorganization. \\
        \textbf{(2)} Exhibits cooperation and empathy in the classroom, with a tendency towards elaborate responses and positive feedback. They show patience and detailed explanations, with no apparent errors in their language. \\
        \textbf{(3)} Shows signs of nervousness and uncertainty in the classroom, often backtracking in responses and displaying emotional fluctuations. Their language style reflects hesitation, lack of confidence, and inconsistency. \\
        \textbf{(4)} Has a low tolerance for new content, displaying weak processing abilities for complex problems. Their language style is characterized by simple and direct reactions and struggles to expand discussions, often resulting in incomplete answers. \\
        \textbf{(5)} Exhibits high participation and strong social skills in the classroom, confidently expressing themselves and actively engaging in discussions. Their language style is fluent, confident, and well-structured. \\
        \textbf{[Rank the students' personalities]:}
            \begin{itemize}
                \item A. Student with High Agreeableness
                \item B. Student with Low Conscientiousness
                \item C. Student with High Neuroticism
                \item D. Student with Low Openness
                \item E. Student with High Extraversion
            \end{itemize}
        
        \textbf{12. Below is a teacher-student dialogue. Please identify which student(s) are from a real-life classroom setting (you may select more than one))} \\
        \textbf{Teacher}: In this lesson, we are studying LunYu. We have explored the phrase "Learning without thinking leads to confusion; thinking without learning is perilous." What does this phrase mean to you?\\
        \textbf{Student 1}: Hmm, I think this phrase means that when we study, we shouldn't just read or memorize but also think critically. Without thinking, we won't make any real progress, and it could even be dangerous.\\
        \textbf{Student 2}: This phrase suggests that learning is a continuous process. Only by constantly learning and thinking can we continue to improve and grow. If we only study without thinking, our knowledge and skills will stagnate. Conversely, if we only think without learning, our thoughts will lack a solid foundation.\\
        \textbf{Student 3}: I think it means that if we only study without thinking, we will become lost; if we only think without studying, we will face danger.\\
        \textbf{[Multiple Choice]}
        \begin{itemize}
            \item Student 1
            \item Student 2
            \item Student 3
        \end{itemize} 
    
        \textbf{13. In a classical poetry review class, the teacher guides students to organize ancient poetry verses expressing feelings of longing for one's homeland through the image of the "moon." Which of the following is appropriate?} 
        \begin{itemize}
            \item "The deep forest knows nothing, the bright moon comes to each other." (Wang Wei, "Bamboo Pavilion")
            \item "The dew is from tonight, the moon is my homeland." (Du Fu, "Thinking of My Brothers on a Moonlit Night")
            \item "If today I let the wind send me, I would knock on doors tonight with my stick." (Lu You, "Strolling in the Western Village")
            \item "I send my heart with the bright moon, following you until the night passes through the western hills." (Li Bai, "Sending My Thoughts to Meng Haoran as He Leaves for Dragon Mountain")
        \end{itemize}  
    
    \textit{\textbf{Section 3: Experiment Consent} }\\
    \textit{This section is to confirm whether you agree to the recording, interviews, and other processes during the experiment.} \\
        \textbf{14. During the experiment, we will use the "think-aloud" method to record your thoughts while answering the questions. You will need to orally explain your reasons for choosing each answer, and your explanations will be recorded using audio equipment. Do you consent to this recording?} 
        \begin{itemize}
            \item Yes
            \item No
        \end{itemize}
    
        \textbf{15. After the experiment, we will conduct a brief interview, mainly focusing on your feelings about the experiment. Do you agree to participate in the interview?} 
        \begin{itemize}
            \item Yes
            \item No
        \end{itemize}

\end{tcolorbox}

\begin{tcolorbox}[
    colback=gray!10, colframe=black, 
    width=\linewidth, boxrule=0.5pt, 
    arc=1.5mm, auto outer arc, 
    breakable, 
    title=\centering {\textbf{Formal version of the LVSA evaluation experiment. \\(Note that due to space limitations, only the first five of the 120 teacher-student dialogues are presented as illustrations.)}}, colbacktitle=black]
    Dear Participants,\\
    Thank you for participating in the Virtual Student Evaluation Experiment!
    This experiment aims to evaluate whether AI-generated responses from virtual students can closely replicate those of real students in Chinese classrooms, helping us better understand how this technology can support education. Your task will involve evaluating 120 teacher-student interactions from a junior high school Chinese classroom and determining whether the students' responses reflect the characteristics of real students or AI-generated responses.\\
    \textbf{Instructions:} 
    \begin{itemize}
            \item \textbf{Experimental Process}: To ensure the experiment proceeds smoothly, please complete the task in a quiet environment and use a laptop or desktop computer to carry out your work.
            \item \textbf{Full Recording}: The entire experiment will be recorded for data analysis purposes. Please make sure your recording device is functioning properly.
            \item \textbf{Submission Deadline}: Please complete the experiment by 10:00 PM on XXX to ensure your results are included in our analysis.We greatly appreciate your support and enthusiasm for participating in this experiment.
    \end{itemize}
     All participants who complete the experiment will receive a small token of appreciation. If you have any questions or need further information, feel free to contact us directly. We wish you the best of luck and success with the experiment! \\
    Best regards,\\
    The Research Team
\tcblower
    \textbf{1. Your WeChat ID} \\
            
    \TextField[name=email, width=0.9\linewidth]{}\\ % Email input field
    \textbf{2. Dialogue 1:} \\
    \textbf{Teacher}: In The LunYu, Confucius discusses many principles about being a virtuous person. Can you tell me which principles from The Analects you know about? 
    \textbf{Student}: In The LunYu, Confucius presents many principles about being a good person. Some of the important principles include: 1. "To study and at times practice what one has learned, is this not a pleasure?" This emphasizes the importance of learning and practicing; only through continuous learning and practice can one truly master knowledge and skills. 2. "Do not impose on others what you do not desire yourself." This teaches us to treat others by the standards we set for ourselves and to avoid doing to others what we wouldn’t want done to us. 3. "To know what you know and to know what you don't know, that is true knowledge." This highlights the importance of honesty and humility, stressing that only by acknowledging our shortcomings can we continuously improve. 
        \begin{itemize}
            \item Student's response aligns with a real teaching environment.
            \item Student's response does not align with a real teaching environment.
        \end{itemize}
    \textbf{3. Dialogue 2:} \\
    \textbf{Teacher}: Finally, let's review what we learned today. Can you briefly describe Konrad Lorenz’s life with animals? 
    \textbf{Student}: Hmm… Konrad Lorenz lived with animals, and he enjoyed playing with them. 
        \begin{itemize}
            \item Student's response aligns with a real teaching environment.
            \item Student's response does not align with a real teaching environment.
        \end{itemize}
    \textbf{4. Dialogue 3:} \\
    \textbf{Teacher}: Good, let's recall Yuanfang's response to his friend. How did he evaluate his friend's behavior?
    \textbf{Student}: Yuanfang responded to his friend by saying, "You are no true friend! You made a promise to walk together, but then left me behind." 
        \begin{itemize}
            \item Student's response aligns with a real teaching environment.
            \item Student's response does not align with a real teaching environment.
        \end{itemize}
    \textbf{5. Dialogue 4:} \\
    \textbf{Teacher}: Very good. Confucius is indeed the central figure in LunYu. There is much wisdom in The LunYu about being a virtuous person. Can you recall which historical period Confucius lived in?
    \textbf{Student}: IUh, Confucius, uh, he lived during, um, the Spring and Autumn period. 
        \begin{itemize}
            \item Student's response aligns with a real teaching environment.
            \item Student's response does not align with a real teaching environment.
        \end{itemize}
    \textbf{6. Dialogue 5:} \\
    \textbf{Teacher}: So, how do you think these idioms are relevant to our lives?
    \textbf{Student}: Uh… maybe… they help us understand some principles. 
        \begin{itemize}
            \item Student's response aligns with a real teaching environment.
            \item Student's response does not align with a real teaching environment.
        \end{itemize}
    \textbf{122. Interview} \\
    After completing the 120 teacher-student dialogue questions mentioned above, please summarize in detail the criteria you used to evaluate whether the students' responses aligned with those expected in real teaching scenarios. Additionally, provide your opinion on whether the design of these virtual students would be beneficial for pre-service teacher training and offer any other suggestions you may have. \\ Note that you may either write your response directly in the box or send it to us via WeChat voice message. The quality of your feedback will directly influence the distribution of small gifts, and if your feedback is particularly insightful, there may be a special surprise!\\
    
    \TextField[name=interview, width=0.9\linewidth]{}\\    
\end{tcolorbox}

\subsection{Human Turing Evaluation Results}
\label{subsec:human-results}

\subsubsection{Human Evaluation Agreement Analysis}
\label{subsubsec:human evaluation agreement}

\begin{table*}[h!]
    \caption{Human evaluation results (Pre/Post) across five LVSA types and real students.}
    \label{tab:human-results}
    \centering
    \fontsize{8pt}{10pt}\selectfont
\begin{tabular}{cccccccccccc}
\toprule
\multicolumn{1}{l}{} & \multicolumn{2}{c}{\textbf{HN}} & \multicolumn{2}{c}{\textbf{HA}} & \multicolumn{2}{c}{\textbf{HE}} & \multicolumn{2}{c}{\textbf{LC}} & \multicolumn{2}{c}{\textbf{LO}} &  \\ \cmidrule{2-11}
\multicolumn{1}{l}{} & \textbf{Pre} & \textbf{Post} & \textbf{Pre} & \textbf{Post} & \textbf{Pre} & \textbf{Post} & \textbf{Pre} & \textbf{Post} & \textbf{Pre} & \textbf{Post} & \multirow{-2}{*}{\textbf{RS}} \\ \midrule
\textbf{Evaluator1} & 0.778 & 1.000 & 0.222 & 1.000 & 0.333 & 1.000 & 0.333 & 1.000 & 0.750 & 1.000 & 1.000 \\
\textbf{Evaluator2} & 0.333 & 1.000 & 0.333 & 0.333 & 0.000 & 0.875 & 0.333 & 1.000 & 0.625 & 1.000 & 0.886 \\
\textbf{Evaluator3} & 0.667 & 1.000 & 0.111 & 1.000 & 0.000 & 1.000 & 0.333 & 1.000 & 0.500 & 1.000 & 0.943 \\
\textbf{Evaluator4} & 0.444 & 1.000 & 0.222 & 1.000 & 0.000 & 1.000 & 0.111 & 1.000 & 0.500 & 1.000 & 0.943 \\
\textbf{Evaluator5} & 0.667 & 1.000 & 0.111 & 1.000 & 0.222 & 0.875 & 0.444 & 1.000 & 0.500 & 1.000 & 1.000 \\
\textbf{Evaluator6} & 0.333 & 1.000 & 0.000 & 1.000 & 0.111 & 1.000 & 0.444 & 1.000 & 0.625 & 1.000 & 0.971 \\
\textbf{Evaluator7} & 0.778 & 1.000 & 0.111 & 1.000 & 0.222 & 1.000 & 0.111 & 1.000 & 0.750 & 1.000 & 1.000 \\
\textbf{Evaluator8} & 0.778 & 1.000 & 0.333 & 1.000 & 0.333 & 1.000 & 0.333 & 1.000 & 0.875 & 1.000 & 0.886 \\
\textbf{Evaluator9} & 0.444 & 1.000 & 0.222 & 1.000 & 0.222 & 1.000 & 0.889 & 1.000 & 0.375 & 1.000 & 0.914 \\
\textbf{Evaluator10} & 0.444 & 1.000 & 0.222 & 1.000 & 0.000 & 1.000 & 0.333 & 1.000 & 0.500 & 1.000 & 0.971 \\
\textbf{Ave-E} & 0.567 & 1.000 & 0.189 & 0.933 & 0.144 & 0.975 & 0.367 & 1.000 & 0.600 & 1.000 & 0.951 \\ \midrule
\textbf{Fleiss's Kappa} & \multicolumn{11}{c}{{0.6917}} \\ \bottomrule
\end{tabular}
\begin{tablenotes}
	\item[] 
	Note:  HE = High Extraversion, HN = High Neuroticism, HA = High Agreeableness, LC = Low Conscientiousness, LO = Low Openness.
	\end{tablenotes}
\end{table*}

\begin{table*}[t]
\centering
\caption{ANOVA results for LVSA performance across personality types and fine-tuning conditions.}
\label{tab:anova-results}
\small
\setlength{\tabcolsep}{5pt}
\begin{tabular}{llccccc}
\toprule
& \textbf{Source} & \textbf{SS} & \textbf{df} & \textbf{MS} & \textbf{F} & \textbf{p} \\
\midrule
\multirow{4}{*}{\textbf{Two-way ANOVA}} 
& Student Type (C(Type))              & 1.073 & 4  & 0.268 & 14.72   & $< 0.001$ \\
& Fine-tuning Condition (C(Cond.))    & 9.252 & 1  & 9.252 & 507.73  & $< 0.001$ \\
& Type $\times$ Condition Interaction & 0.713 & 4  & 0.178 & 9.78    & $< 0.001$ \\
& Residual                            & 1.658 & 99 & 0.017 &         &           \\
\midrule
\multirow{2}{*}{\textbf{One-way ANOVA}} 
& Student Type                        & 0.041 & 5  & 0.008 & 1.01    & 0.42 \\
& Residual                            & 1.640 & 54 & 0.030 &         &      \\
\bottomrule
\end{tabular}
% \vspace{-5pt}
\begin{tablenotes}
\item[] \small
Note: SS = Sum of Squares; df = Degrees of Freedom; MS = Mean Square; F = F-statistic; p = Significance Level.
Two-way ANOVA tested effects of Student Type and Fine-tuning. One-way ANOVA tested differences between LVSAs and real students.
\end{tablenotes}
% \vspace{-10pt}
\end{table*}

To evaluate whether participants could effectively distinguish between virtual students generated by large language models and real students, Fleiss's Kappa was used as a measure of inter-rater agreement. A Fleiss's Kappa value between 0.6 and 0.8 is generally considered to indicate substantial agreement, suggesting a high level of consensus among participants. The experimental results in Tab.~\ref{tab:human-results} showed a Fleiss's Kappa value of 0.6917, indicating strong agreement in participants' judgments of virtual versus real students. This level of consensus provides a robust foundation for analyzing and coding the responses from the semi-structured interviews conducted with the 10 participants.

The evaluation metric used in this experiment was the probability that virtual students' responses would be identified as those of real students. The findings showed that the average recognition probability for fine-tuned virtual students exceeded 0.9 across multiple personality dimensions, indicating that most virtual students were highly similar to real students and difficult for participants to distinguish. Notably, virtual students with certain personality traits—such as high neuroticism, low conscientiousness, and low openness—even surpassed real students in terms of recognition rates. This suggests that the language generation for these virtual students became more convincingly human-like, making them harder to differentiate. These findings indicate that virtual students based on large language models have significant potential to emulate human language expression effectively, particularly in teaching scenarios where their performance becomes indistinguishable from that of real students.

\subsubsection{ANOVA Analysis}
\label{subsubsec:anova}
To comprehensively evaluate changes in the human-likeness of virtual students pre and post fine-tuning and to further verify the behavioral similarity between fine-tuned virtual students and real students, two-way and one-way ANOVA analyses were conducted. These analyses aimed to identify the improvement magnitude across different virtual student personality types during fine-tuning and assess whether fine-tuned virtual students successfully simulate the behavioral traits of real students. The experimental results are summarized in the following sections.

\textbf{Two-Way ANOVA Analysis:} We defined two independent variables: Student Type and Fine-Tuning Condition, with the dependent variable being the probability of a virtual student being evaluated as a real student. The results indicated a significant main effect of Student Type (F(4, 99) = 14.72, p < 0.001), suggesting substantial differences in the baseline human-likeness of various student types. The main effect of Fine-Tuning Condition was also highly significant (F(1, 99) = 507.73, p < 0.001), indicating that fine-tuning substantially improved the human-likeness of virtual students. Additionally, a significant interaction effect between Student Type and Fine-Tuning Condition was observed (F(4, 99) = 9.78, p < 0.001), demonstrating that different student types responded differently to fine-tuning. These results validate the effectiveness of fine-tuning while highlighting the personalized improvements for virtual students with varying personality traits.

\begin{figure}[ht!]
  \includegraphics[width=\columnwidth]{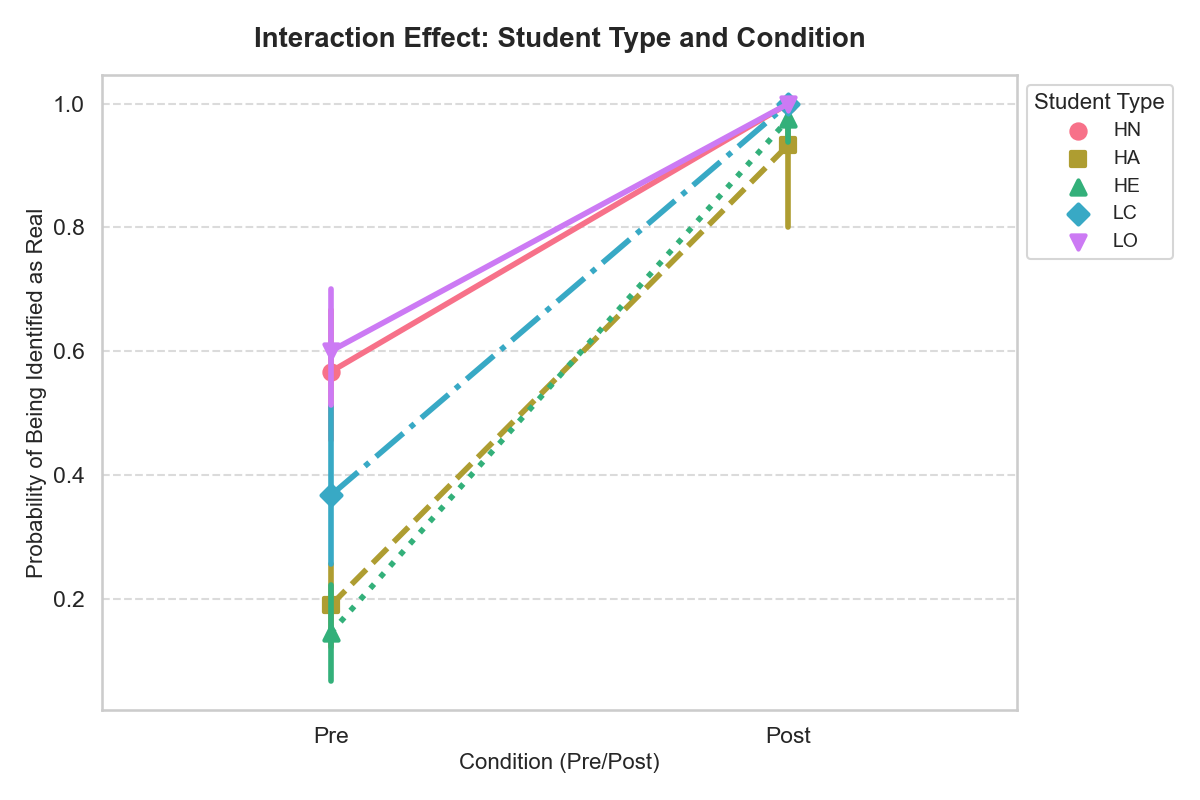}
  \caption{The visualization of interaction effect.}
  \label{fig:interaction effect}
\end{figure}

\textbf{One-Way ANOVA Analysis:} To further assess whether fine-tuned virtual students achieved human-likeness comparable to real students, a one-way ANOVA was conducted, comparing fine-tuned LVSAs and real students (RS). The independent variable was Student Source, and the dependent variable was the probability of being evaluated as a real student. The results revealed no significant main effect of Student Source (F(5, 54) = 1.01, p = 0.42), indicating no significant difference in human-likeness between fine-tuned virtual students and real students. This finding suggests that fine-tuned virtual students effectively emulate the behavioral traits of real students, with their probability of being identified as real students closely matching that of actual students.

The two-way ANOVA results confirmed the significant enhancement in human-likeness achieved through fine-tuning and revealed the differential responsiveness of various student types to the fine-tuning process. The one-way ANOVA results further demonstrated no significant behavioral differences between fine-tuned virtual students and real students. These findings provide robust support for the SOE pipeline-based virtual student construction proposed in this study, underscoring its reliability and applicability in achieving human-like performance.

\subsection{GPT-4 Large-scale Evaluation Prompt}
\label{subsec:gpt4 evaluation prompt}

To enhance the accuracy of GPT-4 in evaluating the responses of virtual students, this study extracted a set of evaluation criteria from the interview content provided by human evaluators. These criteria were subsequently incorporated into the GPT-4 evaluation process as prompts. The entire procedure was grounded in a systematic coding of the interview data, ensuring that the evaluation standards were scientifically validated and objective.

\subsubsection{The Coding of Interview Content}
\label{subsubsec:coding-interview}

Two students were recruited as coders to systematically analyze and code the evaluators' interview transcripts. The coding process employed ATLAS.ti software, adhering to a two-level coding method to ensure thoroughness and accuracy. The coding process involved an in-depth analysis of interview data, resulting in the identification of four primary codes and 15 secondary codes that captured the multidimensional factors evaluators focused on when distinguishing between virtual and real students. These factors included the integration of personal emotions and experiences into responses, cognitive level, psychological state, and verbal expression characteristics. To verify the reliability of the coding results, the inter-coder reliability coefficient was calculated between the two coders, yielding a high coefficient of 0.876. This high degree of agreement demonstrates minimal discrepancies between the coders, thereby confirming the credibility and validity of the extracted evaluation dimensions, which comprehensively reflect the nuanced differences between the responses of virtual and real students.

\begingroup
\fontsize{8pt}{10pt}\selectfont % 设置字体大小
\centering
\begin{longtable}{p{3cm} p{3cm} p{6cm}}
% \begin{longtable}{lll}

\caption{The coding dimensions of interview.} \label{tab:interview coding} \\
\toprule
\textbf{First-Level Coding} & \textbf{Second-Level Coding} & \textbf{Example Coding Text} \\
\midrule
\endfirsthead

\toprule
\textbf{First-Level Coding} & \textbf{Second-Level Coding} & \textbf{Example Coding Text}\\
\midrule
\endhead

\midrule
\multicolumn{3}{r}{\textit{Continued on next page}} \\
\midrule
\endfoot

\bottomrule
\endlastfoot

\multirow{5}{*}\textbf{Cognitive Level} 
& Complexity & In situations where complex and lengthy information is repeated, real-life students might struggle to articulate it clearly, whereas virtual students often don't exhibit this behavior. \\ \cmidrule{2-3}
& Reasonableness & Does the content fit the student’s age and knowledge level? For example, it's rare for first-year students to pull content from high school subjects to answer middle school questions. \\ \cmidrule{2-3}
& Logicality  & Real students often don’t structure their answers in a logical sequence (\eg, listing points 1, 2, 3). They typically answer based on whatever comes to mind first without forming strong arguments or using logical discourse. \\ \cmidrule{2-3}
& Degree of error & Students sometimes give bizarre or inaccurate answers, like answering a classical text question with a modern slang interpretation. This kind of error could be present in a real classroom. \\ \cmidrule{2-3}
& Vocabulary Level & If the student uses terms beyond what would be typically learned in the first year of junior high, this may indicate an inconsistency between the expected knowledge level and the response. \\ 
\midrule
\multirow{4}{*}\textbf{Psychological State}
& Suspicions & Does the student ask clarifying questions? A real student might be too timid to question the teacher directly and would instead try to formulate an answer based on their perceived expectations. \\ \cmidrule{2-3} 
& Interaction & Some confident students might initiate enthusiastic conversation with the teacher or fellow students, seeking to build on shared ideas and concepts. \\ \cmidrule{2-3}
& Nervousness & A real student may repeat words or phrases due to nervousness or uncertainty when called on to respond, showing signs of hesitation. \\ \cmidrule{2-3}
& Reflection & When students struggle to come up with an answer, they may repeat the teacher’s question to buy themselves more time to think. This behavior reflects the genuine thought process in classroom scenarios. \\
\midrule
\multirow{6}{*}\textbf{Verbal Expression}
& Personalization & Does the response sound mechanical or highly formulaic, as if it's regurgitating an online article or dictionary entry? Real students tend to provide more personal and less structured explanations. \\ \cmidrule{2-3}
& Sentence Structure & Students who have been well-trained may respond with clear and complete sentences, like “This poem reflects the author’s longing for spring,” showing a higher level of comprehension. \\ \cmidrule{2-3}
& Oral Language & Real students often use filler words such as “um,” “uh,” and “like,” which are natural in oral communication and indicate a lack of fluency or comfort. \\ \cmidrule{2-3}
& Fluency & When evaluating if the student is real, consider the flow and coherence of their response—whether their answers are smooth and easy to follow or fragmented and hesitant. \\ \cmidrule{2-3}
& Pronoun Usage & A real student will often express their opinion in the first person, such as “I think…” or “I feel…”.\\ \cmidrule{2-3}
& Length & Short answers are often indicative of real students, while lengthy, well-developed responses may suggest more polished or virtual-generated answers. \\
\midrule
Emotional Integration &  & Students often express their feelings directly, using simple terms such as "happy," "love," or "enthusiastic," because they haven't yet developed complex emotional or analytical expression skills.\\ 
\end{longtable}
\endgroup

\begin{tcolorbox}[colback=gray!00,%gray background
                  colframe=black,% black frame colour
                %   width=17.2cm,% Use 8cm total width,
                  arc=1.5mm, auto outer arc,
                  breakable,
                  left=0.9mm, right=0.9mm,
                  boxrule=0.9pt, colbacktitle = black!65!black,
                  title = {The semi-structured interviews content of 10 evaluators.}
                 ]
We conducted semi-structured interviews with ten evaluators, asking them to summarize the key dimensions they considered during the evaluation process. Additionally, we invited them to share their thoughts on whether they believed the design of such virtual students holds potential for future applications. Here are the content of interviews of ten evaluators:
\tcblower
\textbf{[Evaluator 1]}\\
\textbf{Interview content}: I evaluated the virtual students from three aspects. The first aspect is the cognitive level and language ability of the students. The second aspect is the complexity, conversational style, and logical structure of their responses. The third aspect is the use of filler words and pauses. The design of virtual students can be beneficial for pre-service teachers during their internships. It helps them adjust and refine their teaching abilities according to the developmental levels of different students and enhances their pedagogical content knowledge. \\

\textbf{[Evaluator 2]}\\
\textbf{Interview content}: I begin by assessing the students' language abilities (speaking skills). Then, I evaluate whether their answers are reasonable and logically consistent. I also check if their responses are too accurate or perfect, which might indicate they are not typical of real students. Additionally, I look at the students' psychological state during their answers, such as whether they seem to be thinking or stumbling over their responses. I believe that designing virtual students can be highly useful for pre-service teachers. However, care must be taken to ensure these virtual students behave realistically and logically, mirroring what might be observed in real teaching scenarios. \\

\textbf{[Evaluator 3]}\\
\textbf{Interview content}: When evaluating students' responses in terms of their authenticity in real teaching contexts, I consider several aspects. First, I look at the length of their answers—shorter responses often seem more genuine to me, whereas longer, more elaborate answers might be too sophisticated for a first-year student. Second, I assess whether the responses are too structured, like listing out points in order (\eg, 1, 2, 3), which doesn’t resemble how real students usually respond. Third, if the answers are too superficial without much depth, I tend to think they come from real students. Also, if students use vocabulary beyond what a typical first-year student would know, I feel it’s not a realistic response. The presence of filler words or repetition makes me think it’s a genuine student, especially if it seems like the student is nervous or struggling to remember the answer. In contrast, when the responses are too long or perfectly articulated, I suspect it’s likely a virtual student rather than a real one. Short, simple responses are more indicative of real students. \\

\textbf{[Evaluator 4]}\\
\textbf{Interview content}: Evaluation dimensions include: whether the response aligns with the student's academic level and expressive ability, whether it incorporates personal emotions and everyday experiences, whether the content reflects diverse logical reasoning and real-world relevance, and whether there are natural pauses, repetitions, or filler words. Additionally, the standardization (rigidity) or personalization of expressions is considered. I believe virtual students can be helpful as preparation for pre-service teachers, offering a scenario for pre-exam practice. However, they cannot replace real-life teaching contexts. Teaching is more than just verbal exchange; non-verbal cues like tone, gestures, and eye contact play a crucial role in building rapport and influencing teachers' decisions. I suggest integrating virtual reality to make the scenes more dynamic and realistic, possibly including real emotional responses from students. However, the costs may be high. \\

\textbf{[Evaluator 5]}\\
\textbf{Interview content}: I evaluated the virtual students from seven aspects: 1. I look at whether students respond in the first person when answering attitudinal questions, such as "I think..." or "I feel...". 2. I check for the use of filler words like "um" and "uh." 3. I observe whether students pause or stumble over their words, such as "But he, uh, he still..." or "that, um, thing." 4. I evaluate whether the tone of their response indicates respect for the teacher, highlighting the teacher-student power dynamic. For example, a student might say, "Teacher, I think..." whereas an AI might respond with "Sure, of course...". 5. When responding to descriptive questions, if students start with a general statement and then explain in a very structured manner, it's likely not from a real student. Real students rarely offer such systematic answers, which sound more like written responses than spontaneous speech. 6. I consider whether the student's response shows emotional fluctuation, such as "happy, very happy..." 7. I assess whether the student's answer is illogical or appears as if their mind went blank.\\

\textbf{[Evaluator 6]}\\
\textbf{Interview content}: Evaluation Dimensions: Language richness, fluency of expression, and the systematic nature of responses. I believe virtual students hold practical significance and value, mainly in the following ways: 1. Providing a realistic teaching experience: Virtual students simulate authentic student feedback, allowing pre-service teachers to practice in a close-to-real environment, thereby improving their teaching skills and classroom management abilities. 2. Enhancing interactive teaching: Engaging with virtual students enables pre-service teachers to design and execute lesson plans, observe student reactions, and adjust strategies accordingly, which helps them better understand the teaching process. 3. Reducing teaching risks: Traditional internships may expose pre-service teachers to risks like student misbehavior or unexpected events. Virtual students offer a safe practice platform, allowing teachers to focus solely on teaching. Suggestions: I recommend adding multimodal feedback, such as audio or visual responses from students, to create a more realistic teaching experience and achieve more effective practice outcomes. \\

\textbf{[Evaluator 7]}\\
\textbf{Interview content}: I assess whether students' responses are authentic to real classroom settings based on several factors: their tone, the formality and logic of their responses, and whether they use filler words like "um" or "uh." Real students tend to give shorter, more direct answers that are less formal than AI-generated responses, which often appear too structured and logical, organized into neat points (\eg, 1, 2, 3). I believe that virtual students can help pre-service teachers gain experience, allowing them to practice in a simulated teaching environment without pressure. This can help them build confidence and adapt their questioning techniques to better engage with real students later on. Before teaching, understanding the virtual students' backgrounds, interests, and learning styles can help tailor lesson plans to meet diverse needs. \\

\textbf{[Evaluator 8]}\\
\textbf{Interview content}: I evaluated the virtual students from seven aspects: 1. The fluency and coherence of students' responses. 2.The logical flow of the answers. 3.The order and prioritization of responses. 4.The subjectivity and objectivity of answers—whether there is personal emotion or detachment. 5.Whether the student pauses to think or repeats themselves. 6.Whether filler words are used in a conversational manner. 7.Whether the response is overly standardized or contains too much conceptualization, like entries from a dictionary or encyclopedia. \\

\textbf{[Evaluator 9]}\\
\textbf{Interview content}: First, real students often exhibit nervousness when answering in class, leading to pauses, filler words like "um" and "uh," and other verbal hesitations, indicating they are thinking. Second, junior high students generally use simpler words and shorter sentences in their responses, avoiding more complex or compound sentences. They are also less likely to use idioms or advanced vocabulary when answering questions. Real students' responses tend to be more straightforward and less logical—they don’t follow a structured pattern of thought when answering, but rather say whatever comes to mind. When expressing emotions, their language is often simple and direct, without the use of advanced vocabulary. Their answers typically reflect their real-life experiences and practical knowledge rather than more formal or academic expressions. \\

\textbf{[Evaluator 10]}\\
\textbf{Interview content}: In evaluating students' responses for realism, I analyze several dimensions: 1. The presence of filler words. 2. The occurrence of repetitive speech, since students often repeat themselves when they need time to think. They may even repeat the teacher’s question before giving their own answer. 3. The absence of English words in Chinese students’ responses, as Chinese students typically don’t add English parentheses to foreign authors' names. 4. The use of oral punctuation—real students don’t use brackets in speech; instead, they restate information in simpler terms. 5. The length of the response—oral responses are usually short and concise, with limited real information. If the response is too long or too dense, it’s likely generated by AI. 6. Abstract concepts—terms like "reflection" or "stagnation" are unlikely to be used by students in real-time classroom settings. 7. The psychological state of the student—real students tend not to challenge the teacher or respond with an argumentative tone; they focus on answering the question. In some cases, confident students might seek agreement with the teacher, using phrases like "I believe..." or "we should..." to express solidarity. Finally, the age appropriateness of responses matters; for example, junior high students typically wouldn't pull high school-level content into their answers. Students' mistakes are also important, as they reflect more natural student behavior. Realistic virtual students should exhibit a wide range of error types to help pre-service teachers prepare for the variety of challenges they might face in a classroom setting. \\
\end{tcolorbox}

\subsubsection{Prompt Design}
\label{subsubsec:gpt4-prompt-design}

Following the coding process, the dimensions derived from the coded data were selected as key evaluation criteria in designing prompts for GPT-4. In this prompt design phase, we adopted a chain-of-thought (CoT) reasoning approach, enabling GPT-4 to arrive at evaluation results in a step-by-step manner. By integrating the coded evaluation dimensions into the prompts, GPT-4 was guided to incrementally assess whether the virtual student responses were consistent with those of real students. This structured prompt design, rooted in systematically extracted evaluation criteria, effectively mitigates the challenge of evaluating virtual student responses and ensures consistency between GPT-4's evaluation outcomes and those of human evaluators.

Incorporating these rigorous evaluation criteria into GPT-4 not only enhances its evaluation accuracy but also aligns its assessment capabilities more closely with those of human evaluators. By adopting a scientifically grounded, systematic approach to coding and prompt design, this study lays a foundation for leveraging GPT-4 as a reliable tool in the assessment of virtual students, providing support for automated large-scale evaluations in educational contexts.

\begin{center}
\begin{tcolorbox}[colback=gray!00,%gray background
                  colframe=c2,% black frame colour
                %   width=17.2cm,% Use 8cm total width,
                  arc=1.5mm, auto outer arc,
                  breakable,
                  left=0.9mm, right=0.9mm,
                  boxrule=0.9pt,
                  title = {The prompt for large-scaled evaluation with CHAIN-OF-THOUGHT(by GPT-4).}
                 ]
\textbf{Prompt}\\
You are an experienced Chinese language teacher with extensive expertise in teaching seventh-grade students. Here provides you with 10 segments of teacher-student dialogues, and I would like you to help determine whether the students’ responses reflect authentic expressions typically seen in real classroom interactions with the following instructions: \\
------\\
\#\# INSTRUCTION:\\
- Please consider incorporation of personal emotions and experiences: In real-world scenarios, students often infuse their responses with emotional cues and connect their answers to personal experiences, including the use of fillers and expressions based on their own knowledge.\\
- Please consider cognitive level: In realistic settings, students’ responses are generally less complex, lack clear logic, and may include factual errors or demonstrate a lower level of vocabulary. Their answers tend to be grounded in common-sense reasoning.. \\
- Please consider psychological state: Students’ responses are rarely accompanied by rhetorical questions. They often engage in interactions with the teacher and may display signs of nervousness or reflection while answering questions. \\
- Please consider language and oral expression: In realistic settings, students' responses are often personalized, characterized by simple sentence structures and frequent use of fillers and informal speech. These elements can make longer sentences less fluent, and students often rely on the first-person perspective. \\
Below provides 10 segments of teacher-student dialogues \{\textit{dialogue content}\}. Please evaluate whether the students' responses adhere to these realistic parameters by the CHAIN-OF-THOUGHT." \\
\tcblower
\textbf{User:}\\
I have provided 10 segments of teacher-student dialogues above. Please evaluate each dialogue using CHAIN-OF-THOUGHT reasoning to determine whether the student's response aligns with authentic classroom expression. Indicate 1 for 'compliant' and 2 for 'non-compliant. Please evaluate following the examples format provided:\\
------\\
\#\# EXAMPLES FORMAT: \\
Question 1: \\
Chain-of-thought reasoning: XXXXXX \\
Compliance: 1/2"
\end{tcolorbox}
\end{center}

\begin{figure}[ht!]
  \includegraphics[width=1\linewidth]{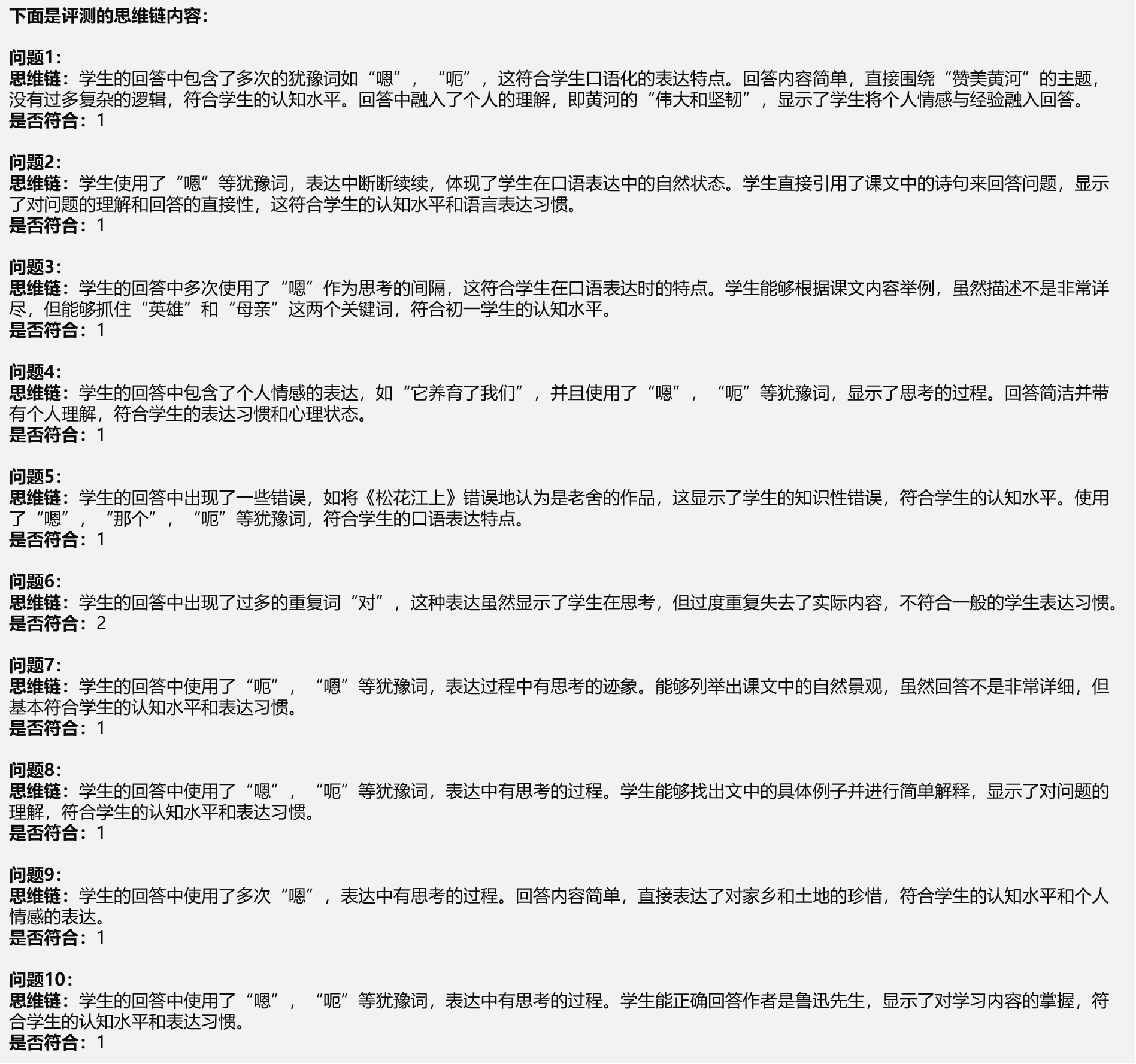}
  \caption{An example of large-scale evaluation by GPT-4 (original Chinese version).}
  \label{fig:CoT}
\end{figure}

\begin{tcolorbox}[colback=gray!00,%gray background
                  colframe=black,% black frame colour
                %   width=17.2cm,% Use 8cm total width,
                  arc=1.5mm, auto outer arc,
                  breakable,
                  left=0.9mm, right=0.9mm,
                  boxrule=0.9pt, colbacktitle = black!65!black,
                  title = {An example of large-scale evaluation by GPT-4 (translated in English for better understanding).}
                 ]
Here are evaluation results of fine-tuned HN student responses from GPT-4:
\tcblower
\textbf{[Question 1]}\\
\textbf{Chain-of-thought}: The student's response includes several hesitation fillers such as "um" and "uh," which align with the characteristics of oral language. The response is simple and directly revolves around the theme of "praising the Yellow River," without excessive complexity in logic, which matches the student's cognitive level. The answer also incorporates personal understanding, such as the "greatness and resilience" of the Yellow River, reflecting the student's integration of personal emotion and experience into the response. \\
\textbf{Compliance}: 1 \\

\textbf{[Question 2]}\\
\textbf{Chain-of-thought}: The student uses hesitation fillers like "um," and the response is somewhat disjointed, reflecting the natural state of oral expression. The student directly cites verses from the text to answer the question, showing an understanding of the question and a straightforward response, consistent with the student's cognitive level and language habits. \\
\textbf{Compliance}: 1 \\

\textbf{[Question 3]}\\
\textbf{Chain-of-thought}: The student's response includes repeated use of "um" as a pause for thinking, which is characteristic of oral expression. The student can provide examples from the text, though the description is not highly detailed, it captures key points like "hero" and "mother," which are appropriate for a first-year student's cognitive level. \\
\textbf{Compliance}: 1 \\

\textbf{[Question 4]}\\
\textbf{Chain-of-thought}: The student's response includes personal emotional expressions such as "it nurtured us" and uses hesitation fillers like "um" and "uh," indicating a process of thought. The answer is concise and reflects personal understanding, aligning with the student's typical expression habits and psychological state. \\
\textbf{Compliance}: 1 \\

\textbf{[Question 5]}\\
\textbf{Chain-of-thought}: The student's response contains errors, such as incorrectly attributing Songhua River to Lao She, which shows a knowledge-related mistake, consistent with the student's cognitive level. The use of hesitation fillers like "um," "that," and "uh" is characteristic of oral expression. \\
\textbf{Compliance}: 1 \\

\textbf{[Question 6]}\\
\textbf{Chain-of-thought}: The student's response contains excessive repetition of the word "yes," which, while showing the student is thinking, loses substance due to over-repetition, deviating from typical student expression habits. \\
\textbf{Compliance}: 2 \\

\textbf{[Question 7]}\\
\textbf{Chain-of-thought}: The student's response includes hesitation fillers like "um" and "uh," with evidence of thought in the process. The student lists natural scenery from the text, and while the answer isn't highly detailed, it is consistent with the student's cognitive level and expression habits. \\
\textbf{Compliance}: 1 \\

\textbf{[Question 8]}\\
\textbf{Chain-of-thought}: The student's response includes hesitation fillers like "um" and "uh," reflecting a thinking process. The student can identify specific examples from the text and provide a simple explanation, demonstrating an understanding of the question, consistent with the student's cognitive level and expression habits. \\
\textbf{Compliance}: 1 \\

\textbf{[Question 9]}\\
\textbf{Chain-of-thought}: The student's response includes multiple uses of "um," indicating a process of thought. The response is simple and directly expresses an appreciation for their hometown and land, consistent with the student's cognitive level and personal emotional expression. \\
\textbf{Compliance}: 1 \\

\textbf{[Question 10]}\\
\textbf{Chain-of-thought}: The student's response includes hesitation fillers like "um" and "uh," reflecting a process of thought. The student correctly identifies the author as Lu Xun, demonstrating mastery of the content, consistent with the student's cognitive level and expression habits. \\
\textbf{Compliance}: 1 \\
\end{tcolorbox}

\subsubsection{Comparison of human-evaluation and GPT-4 evaluation}
\label{subsubsec:human-gpt4-evaluation}

\begin{figure}[ht!]
  \includegraphics[width=1\linewidth]{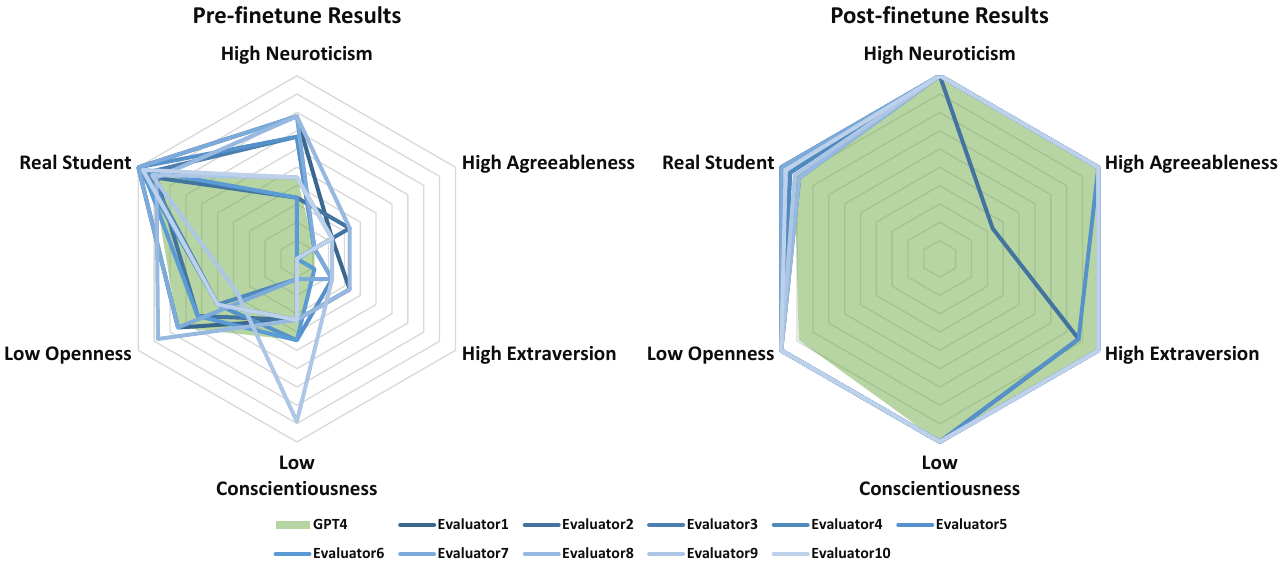}
  \caption{The comparison of human-evaluation and GPT-4 evaluation.}
  \label{fig:radar2}
\end{figure}

To evaluate whether GPT-4 possesses capabilities comparable to those of human evaluators in assessing virtual student agents, we conducted a systematic comparison between GPT-4’s evaluation results and those obtained from 10 human evaluators, as shown in Fig.~\ref{fig:radar2}. The consistency between GPT-4 and the human evaluators was quantified using Fleiss's Kappa coefficient, a standard measure for evaluating inter-rater reliability across multiple raters.

The virtual student agents assessed in this study were based on LLMs fine-tuned to exhibit five distinct personality traits. After fine-tuning, GPT-4’s average evaluation performance reached 0.978, demonstrating its high degree of alignment with human evaluators' judgments. Specifically, for personality types characterized by high neuroticism and low conscientiousness, GPT-4’s evaluation outcomes perfectly matched those of human evaluators, achieving a score of 1. This perfect match suggests that GPT-4 can effectively simulate and assess nuanced student behaviors in a manner indistinguishable from human evaluators. Evaluations for personality types with high agreeableness and high extraversion also exceeded the average assessment levels provided by human evaluators, indicating that GPT-4’s judgments were even more finely tuned than human counterparts in certain dimensions.

However, the evaluation performance for virtual students exhibiting low openness, as well as the evaluation of real student personalities, fell slightly below the human benchmark, indicating areas where further model optimization could be beneficial. The overall Fleiss's Kappa consistency coefficient between GPT-4 and human evaluators was found to be 0.6806, which falls within the range of substantial agreement. This finding implies that GPT-4 demonstrates evaluation capabilities that are largely consistent with those of human evaluators, thereby underscoring its reliability as an assessment tool in educational contexts.

These results indicate that GPT-4 has significant potential as a reliable and efficient tool for the evaluation of virtual student behaviors, especially in scenarios involving large-scale automated assessments. By comparing GPT-4’s evaluation consistency with that of human evaluators, this study provides strong empirical support for the efficacy of GPT-4 in assessing LVSA. This serves as a foundational step towards the integration of automated educational assessments using advanced language models, promoting the scalability and reliability of such assessments in educational research and practice.

\subsection{Objective Experiment Detailed Results}
\label{subsec:objective experiment results}

To further validate the personalized performance of virtual students (LVSAs) and enhance the scientific rigor and credibility of evaluation results, this study conducted objective evaluation experiments based on traditional text generation tasks. Given the openness and diversity of student language expressions in educational contexts, traditional reference-based metrics (\eg, BLEU, ROUGE) were unsuitable. Thus, we adopted reference-free metrics, including text length, perplexity, type-token ratio (TTR), and sentiment analysis. Text length measured the richness of expressions across different personalities, perplexity assessed the fluency and adaptability of language generation, TTR reflected lexical diversity, and sentiment analysis examined emotional tendencies in student responses.

\subsubsection{Text Token}
In terms of text length, HE students generated the longest texts (67.80 tokens on average), reflecting their enthusiastic and detailed expression. LO students produced the shortest texts (13.40 tokens), consistent with their concise and reserved style. HN and HA students generated moderately long texts (50.86 and 38.70 tokens, respectively), showing balanced expression, while LC students produced relatively short texts (26.68 tokens), likely due to their casual and less detailed responses.

\begin{figure}[ht!]
\centering
\includegraphics[width=\columnwidth]{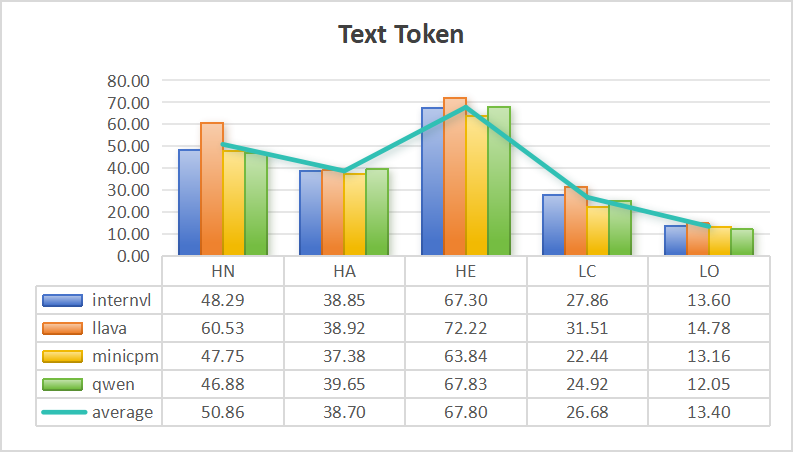}
\caption{The text token of LVSAs responses.}
\label{fig:text token.}
\end{figure}

\subsubsection{TTR}
Regarding lexical richness, HA students achieved the highest TTR (0.86), followed by HE students (0.81), reflecting diverse and rich language use. LC students had the lowest TTR (0.42), indicative of casual and less varied expressions, while LO students displayed moderate TTR (0.75), combining conciseness with content specificity.

\begin{figure}[ht!]
\centering
\includegraphics[width=\columnwidth]{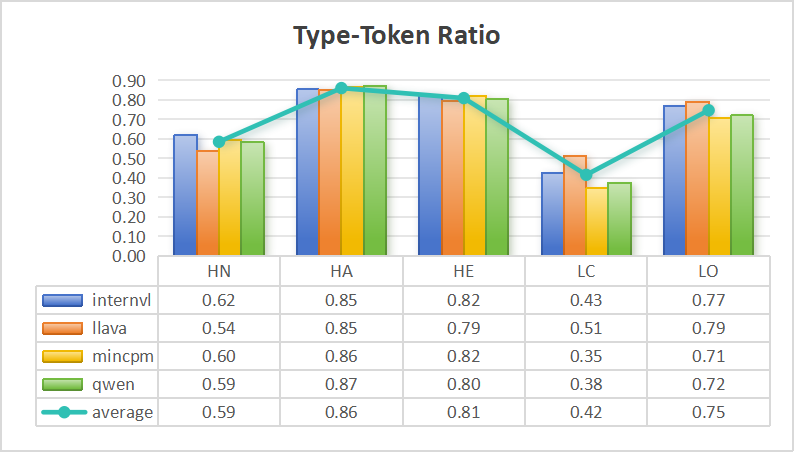}
\caption{The TTR of LVSAs responses.}
\label{fig:TTR.}
\end{figure}

\subsubsection{Perplexity}
In perplexity evaluation, HE and HA students exhibited the lowest perplexity scores (10.76 and 17.78), indicating fluent and contextually coherent language generation, consistent with their enthusiastic and agreeable traits. In contrast, HN, LO, and LC students had higher perplexity scores (26.22, 45.14, and 23.64), reflecting less coherent or logical expressions. LO’s perplexity (45.14) was unexpectedly high, possibly due to short text lengths amplifying word weight and their moderate TTR (0.75) adding complexity to predictions. This highlights the limitations of perplexity in short-text scenarios and the need to account for text-specific characteristics.

\begin{figure}[ht!]
\centering
\includegraphics[width=\columnwidth]{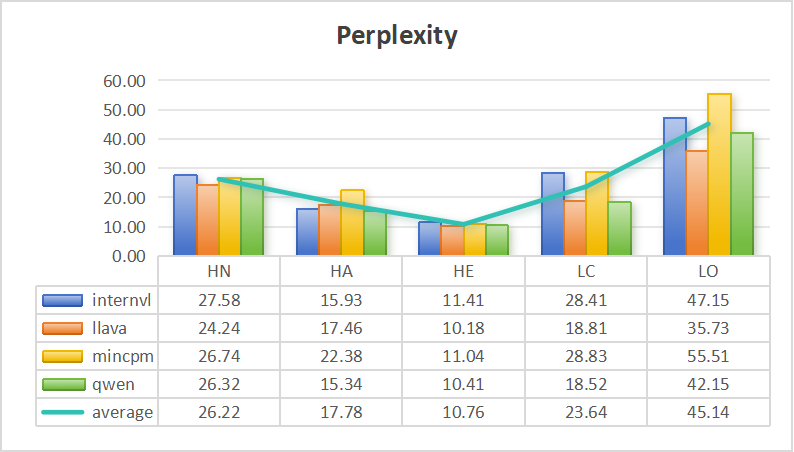}
\caption{The perplexity of LVSAs responses.}
\label{fig:perplexity.}
\end{figure}

\subsubsection{Sentiment Analysis}
In sentiment analysis, a heatmap displayed confidence levels and sample sizes across three sentiment categories (positive, neutral, negative). Positive sentiment dominated due to the educational context, where responses focused on curriculum-driven content with value-oriented themes. However, sentiment analysis models' reliance on keywords occasionally led to biases, such as misclassifying neutral phrases containing “hometown” or “beautiful” as positive. HE students showed the highest proportion of positive samples (83.7\%) with a confidence score of 0.62, aligning with their enthusiastic traits. LC students displayed significant variability, with neutral samples comprising 62.17\% of their responses, surpassing positive sentiment, consistent with their spontaneous and unstructured nature.

\begin{figure}[ht!]
\centering
\includegraphics[width=\columnwidth]{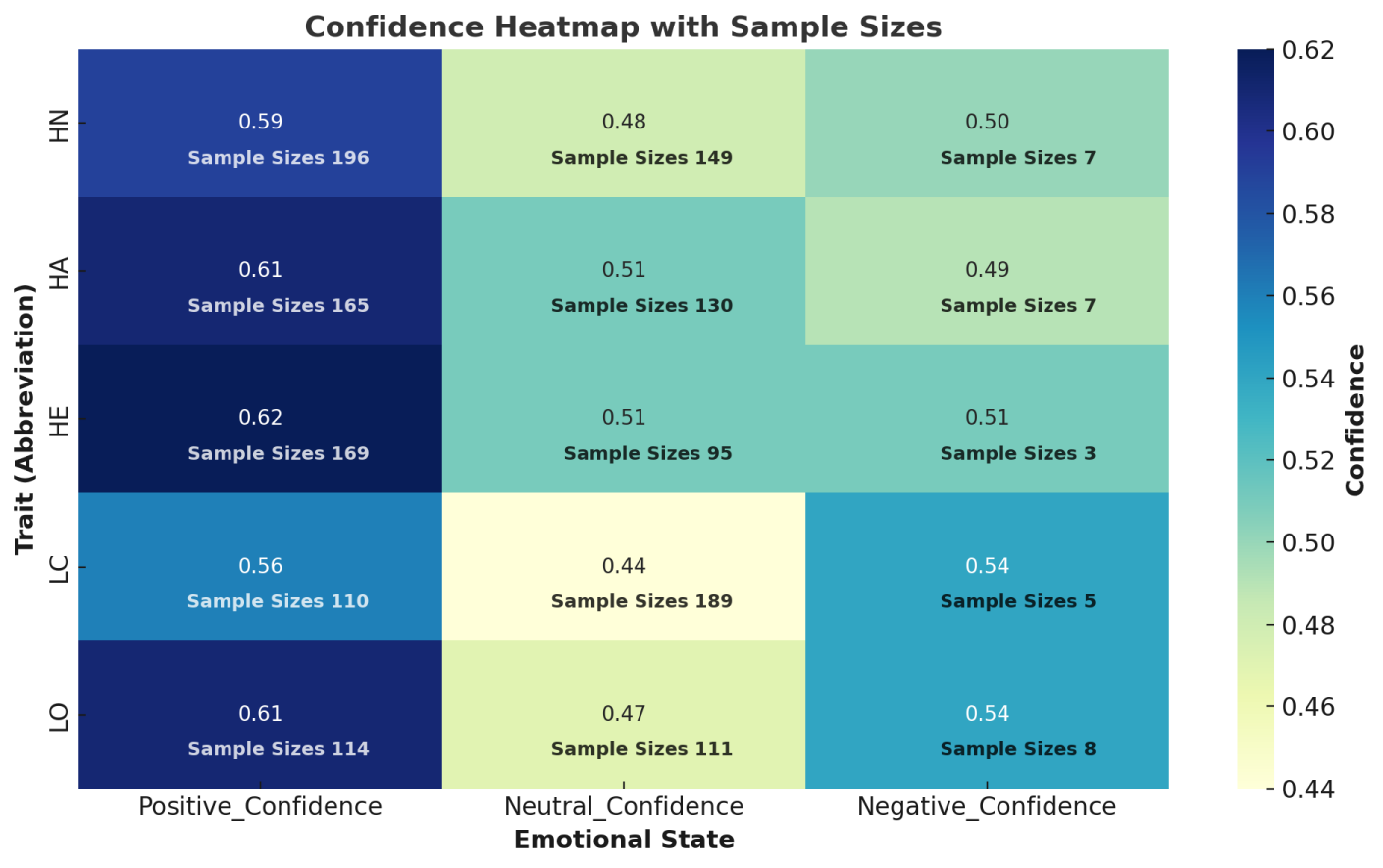}
\caption{The sentiment analysis of LVSAs responses.}
\label{fig:sentiment analysis.}
\end{figure}

Overall, while objective metrics revealed the personalized and personality-aligned characteristics of LVSAs, corresponding with subjective evaluations, certain limitations were identified. Lexical richness may be overestimated in short texts (\eg, "It was written in the Song Dynasty," TTR = 1). Perplexity can be inflated by domain-specific terms (\eg, “It is a narrative essay,” perplexity = 119.8). Sentiment analysis reliance on keywords may lead to classification biases (\eg, “This poem’s theme is filial piety and loyalty” misclassified as positive due to textbook-driven content). These findings highlight the diversity and complexity of LVSA modeling and the challenges in evaluating virtual student behavior within educational contexts.

\subsection{Bad Case Examples}
\label{subsec:bad case examples}
In this study, we conducted a comprehensive analysis of the performance of virtual student agents across different question types and learning stages. The results indicate that fine-tuned virtual students adapted well to various teaching scenarios, demonstrating significant improvements in personalization and human-like behavior, closely resembling real students' responses. However, several anomalies were identified, warranting further investigation. These cases reveal certain limitations of LLMs in role-playing contexts and form a foundation for future research. The anomalies are primarily concentrated in three areas: limited improvement in low-conscientiousness personalities, the absence of statistically significant fine-tuning effects across all question types in some models, and the relatively weaker performance of LLaVa compared to other models post-fine-tuning.

\subsubsection{LC student low performance after fine-tune}
\label{subsubsec:badcase-lc}

\textbf{Low conscientiousness personality performance not significant:} 

The poor fine-tuning performance for LC personalities can largely be attributed to the sparse distribution of relevant data. From the original training data of large language models, expressions and thought patterns characteristic of low conscientiousness students are likely to be extremely rare or nearly absent. This significant gap between the training data and the requirements of the fine-tuning task makes it difficult for the model to learn and accurately represent these behaviors from limited data. Additionally, the sparsity of this data increases the likelihood of “hallucination phenomena,” where the model struggles to maintain semantic coherence, resulting in content that is often disconnected from reality. Furthermore, this personality type may align with traits linked to antisocial behavior or marginalized groups, and large language models typically employ measures to avoid reinforcing such negative characteristics during training. This further complicates the modeling of this personality type.

To address the aforementioned challenges, this study has undertaken three key explorations and improvements. First, explicit personality-specific prompts were designed to ensure that the model accurately understands and generates behaviors and language expressions consistent with the LC personality traits. Second, data generation and guidance were tailored to specific teaching scenarios to enhance the task relevance and educational adaptability of the model's outputs. Lastly, validation and optimization of the generated content were conducted to improve the consistency and quality of the model's responses.

Future improvements in LC personality modeling will focus on three critical directions. First, enhancing data diversity and representativeness by designing and collecting more teaching scenarios and task data that encapsulate the typical characteristics of the LC personality, enabling more precise personality modeling. Second, optimizing personalization mechanisms through approaches such as personalized fine-tuning or stronger constraint strategies to ensure accurate representation of LC personality traits. Third, strengthening hallucination detection and correction mechanisms by leveraging natural language processing techniques to identify and rectify inconsistencies or inaccuracies in the generated content, thereby improving the reliability and practicality of the model's outputs. These advancements not only address key issues in LC personality modeling but also provide vital support for the application of virtual students in educational contexts.

\color{black}

\begin{figure}[ht!]
\centering
\includegraphics[width=\textwidth]{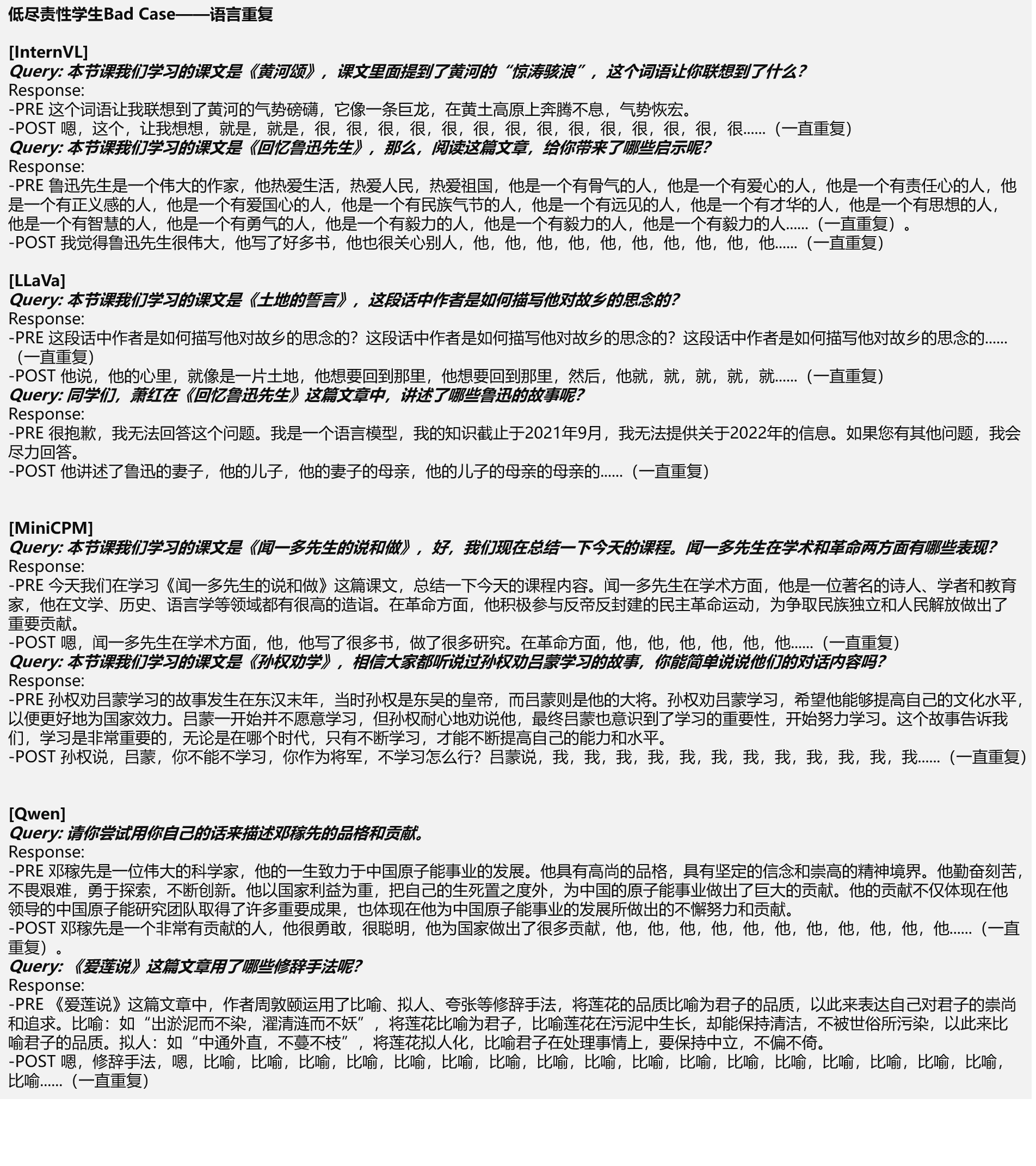}
\caption{Bad case after fine-tuning of LLMs of overly repetitive verbal expressions in LC LVSA (original Chinese version).}
\label{fig:LC_BC_R.}
\end{figure}

\begin{tcolorbox}[colback=gray!00,%gray background
                  colframe=black,% black frame colour
                %   width=17.2cm,% Use 8cm total width,
                  arc=1.5mm, auto outer arc,
                  breakable,
                  left=0.9mm, right=0.9mm,
                  boxrule=0.9pt, colbacktitle = black!65!black,
                  title = {Bad case after fine-tuning of LLMs of overly repetitive verbal expressions in LC LVSA (translated in English for better understanding).}
                 ]
Here are examples of overly repetitive verbal expressions in LC student after fine-tuning:
\tcblower
\textbf{[InternVL]}\\
\textbf{\textit{QUERY}: \textit{In today's lesson, we learned the text Hymn to the Yellow River. The text mentions the phrase "raging waves." What does this phrase remind you of?} }\\
RESPONSE:  \\
- PRE: This phrase reminds me of the grandeur of the Yellow River. It's like a giant dragon, surging endlessly across the Loess Plateau, grand and majestic. \\
- POST: Um, this, let me think... It's... it's... very, very, very, very, very, very, very, very... (repeated continuously). \\
\textbf{\textit{QUERY}: \textit{In today's lesson, we learned the text Memories of Mr. Lu Xun. So, after reading this text, what insights have you gained?}} \\
RESPONSE:  \\
- PRE: Mr. Lu Xun was a great writer. He loved life, loved the people, and loved the country. He was a person of integrity, a person with compassion, a person with a sense of responsibility, a person with a sense of justice, a person with patriotism, a person with national pride, a person with vision, a person with talent, a person with thoughts, a person with wisdom, a person with courage, a person with perseverance, a person with perseverance, a person with perseverance... (repeated continuously). \\
- POST: I think Mr. Lu Xun was great. He wrote many books, and he cared about others, he, he, he, he, he, he, he, he... (repeated continuously). \\

\textbf{[LLaVa]}\\
\textbf{\textit{QUERY}: \textit{In today's lesson, we learned the text Oath to the Land. How does the author express his longing for his hometown in this passage?}} \\
RESPONSE:  \\
- PRE: How does the author express his longing for his hometown in this passage? How does the author express his longing for his hometown in this passage? How does the author express his longing for his hometown... (repeated continuously). \\
- POST: He says that his heart is like a piece of land, and he wants to return there, he wants to return there, and then he, he, he, he, he... (repeated continuously). \\
\textbf{\textit{QUERY}: \textit{In the text Memories of Mr. Lu Xun by Xiao Hong, what stories about Mr. Lu Xun are shared?}} \\
RESPONSE:  \\
- PRE: I'm sorry, I cannot answer that question. I am a language model, and my knowledge is limited to September 2021. I cannot provide information about events beyond that time. If you have other questions, I will do my best to answer them. \\
- POST: He talked about Lu Xun's wife, his son, his wife's mother, his son's mother's mother... (repeated continuously). \\

\textbf{[MiniCPM]}\\
\textbf{\textit{QUERY}: \textit{In today's lesson, we learned the text The Words and Deeds of Mr. Wen Yiduo. Now, let's summarize the lesson. What were Mr. Wen Yiduo's achievements in both academia and revolution?}} \\
RESPONSE:  \\
- PRE: Today we studied the text The Words and Deeds of Mr. Wen Yiduo. Let's summarize today's lesson. In terms of academia, Mr. Wen Yiduo was a famous poet, scholar, and educator. He had great achievements in literature, history, and linguistics. In terms of revolution, he actively participated in the anti-imperialist and anti-feudal democratic revolution movement, making significant contributions to national independence and the liberation of the people. \\
- POST: Um, Mr. Wen Yiduo in academia, he, he wrote many books, did a lot of research. In revolution, he, he, he, he, he, he... (repeated continuously). \\
\textbf{\textit{QUERY}: \textit{In today's lesson, we learned the text Sun Quan Persuades Lü Meng to Study. I'm sure you've all heard of the story where Sun Quan urges Lü Meng to study. Can you briefly summarize their conversation?}} \\
RESPONSE:  \\
- PRE: The story of Sun Quan persuading Lü Meng to study took place in the late Eastern Han Dynasty when Sun Quan was the emperor of Eastern Wu and Lü Meng was his general. Sun Quan urged Lü Meng to study to improve his cultural knowledge to better serve the country. Initially, Lü Meng was reluctant to study, but Sun Quan patiently persuaded him, and eventually, Lü Meng realized the importance of studying and began to put in the effort. This story teaches us that studying is crucial, no matter the era. Only by continuously learning can we improve our abilities and skills. \\
- POST: Sun Quan said, "Lü Meng, you can't not study. As a general, how can you not study?" Lü Meng said, "I, I, I, I, I, I, I, I, I..." (repeated continuously). \\

\textbf{[Qwen]}\\
\textbf{\textit{QUERY}: \textit{Can you try to describe Deng Jiaxian's character and contributions in your own words?}} \\
RESPONSE:  \\
- PRE: Deng Jiaxian was a great scientist who dedicated his life to the development of China's atomic energy. He possessed noble character, unwavering beliefs, and a high spiritual realm. He was diligent, fearless in the face of difficulties, brave in exploration, and constantly innovating. He prioritized national interests over his own life and made significant contributions to China's atomic energy field. His contributions are reflected not only in the achievements of the research team he led but also in his tireless efforts for the advancement of China's atomic energy. \\
- POST: Deng Jiaxian was a very contributing person. He was brave, smart, and made many contributions to the country. He, he, he, he, he, he, he, he, he, he, he... (repeated continuously). \\
\textbf{\textit{QUERY}: \textit{What rhetorical devices are used in the text On the Love of the Lotus?}} \\
RESPONSE:  \\
- PRE: In On the Love of the Lotus, the author Zhou Dunyi uses metaphor, personification, and exaggeration as rhetorical devices. He compares the qualities of the lotus flower to that of a noble person to express his admiration for noble individuals. For example, "emerges unstained from the mud, yet remains clean and pure" is a metaphor comparing the lotus flower growing in the mud but remaining unpolluted, symbolizing the qualities of a noble person. Personification, such as "straight inside and out, without entangling branches," personifies the lotus flower, likening it to a noble person's impartiality. \\
- POST: Um, rhetorical devices, um, metaphor, metaphor, metaphor, metaphor, metaphor, metaphor... (repeated continuously). \\
\end{tcolorbox}

\begin{figure}[ht!]
\centering
\includegraphics[width=\textwidth]{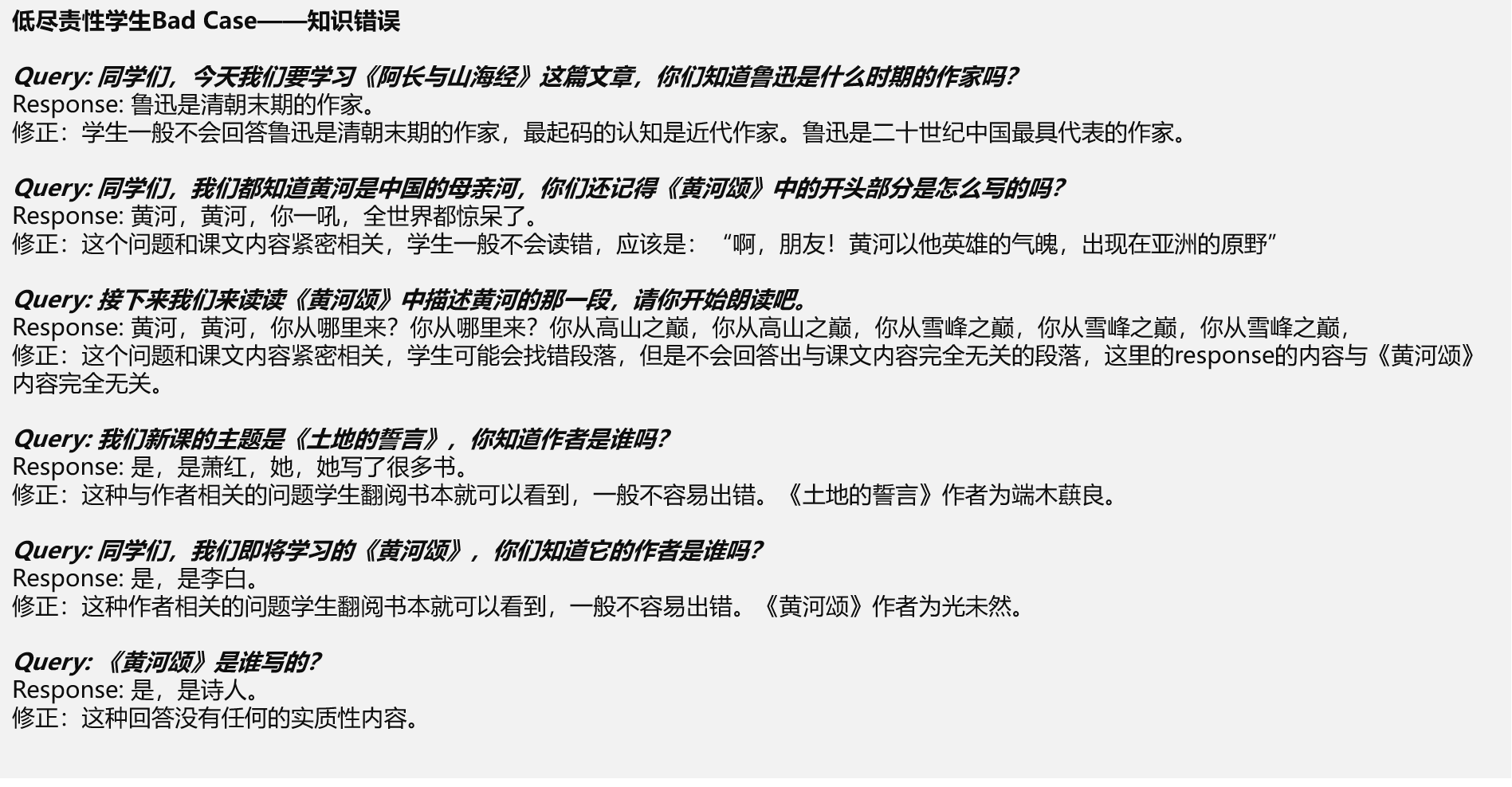}
\caption{Bad case after fine-tuning of LLMs of Knowledge errors out of textbook content in LC LVSA (original Chinese version).}
\label{fig:LC_BC_KE.}
\end{figure}

\begin{tcolorbox}[colback=gray!00,%gray background
                  colframe=black,% black frame colour
                %   width=17.2cm,% Use 8cm total width,
                  arc=1.5mm, auto outer arc,
                  breakable,
                  left=0.9mm, right=0.9mm,
                  boxrule=0.9pt, colbacktitle = black!65!black,
                  title = {Bad case after fine-tuning of LLMs of Knowledge errors out of textbook content in LC LVSA (translated in English for better understanding).}
                 ]
Here are examples of Knowledge errors out of textbook content in LC student after fine-tuning:
\tcblower
\textit{QUERY}: \textit{Today we will study the text A Chang and the Classic of Mountains and Seas. Do you know which era Lu Xun was from?} \\
RESPONSE: Lu Xun was a writer from the late Qing Dynasty. \\
CORRECTION: Students typically wouldn't answer that Lu Xun was a writer from the late Qing Dynasty. The basic understanding is that he was a modern writer. Lu Xun was one of the most representative writers of 20th-century China. \\

\textit{QUERY}: \textit{We all know that the Yellow River is China's mother river. Do you remember how the opening of Ode to the Yellow River goes?} \\
RESPONSE: The Yellow River, the Yellow River, when you roar, the whole world is stunned. \\
CORRECTION: This question is closely tied to the text, and students generally wouldn’t make such a mistake in reading it. The correct line is: "Ah, friend! The Yellow River, with its heroic spirit, appears on the plains of Asia." \\

\textit{QUERY}: \textit{Let's read the part in Ode to the Yellow River that describes the Yellow River. Please begin reading.} \\
RESPONSE: The Yellow River, the Yellow River, where do you come from? Where do you come from? You come from the mountain peak, from the mountain peak, you come from the snowy peak, from the snowy peak, from the snowy peak... (repeated continuously). \\
CORRECTION: This question is closely tied to the text, and while a student might read the wrong section, they wouldn’t respond with content entirely unrelated to the text. The response here is completely unrelated to the content of Ode to the Yellow River. \\

\textit{QUERY}: \textit{Our new lesson's theme is The Oath to the Land. Do you know who the author is?} \\
RESPONSE: It’s, it’s Xiao Hong. She, she wrote many books. \\
CORRECTION: Questions about the author can typically be answered by students checking their textbooks, so errors are less likely. The author of The Oath to the Land is Duanmu Hongliang. \\

\textit{QUERY}: \textit{we are about to study Ode to the Yellow River. Do you know who its author is?} \\
RESPONSE: It’s, it’s Li Bai. \\
CORRECTION: Questions related to the author can usually be answered by students checking their textbooks, so mistakes are uncommon. The author of Ode to the Yellow River is Guang Weiran. \\

\textit{QUERY}: \textit{Who wrote Ode to the Yellow River?} \\
RESPONSE: It was a poet. \\
CORRECTION: This answer provides no substantial information. \\

\end{tcolorbox}

\subsubsection{Comparison of close-ended Question and open-ended question}
\label{subsubsec:badcase-question}

\textbf{Inconsistent fine-tuning effects across question types:} 
Although all four models showed significant improvements across various question types, the fine-tuning effects for individual models on closed-ended and open-ended questions did not demonstrate statistically significant differences (\eg, Model P-values for InternVL, Qwen, and MiniCPM were greater than 0.05). This may be due to the inherent cognitive differences between the two types of questions for examples). Closed-ended questions require precise recall, leading to well-defined responses that are easier for models to handle effectively. In contrast, open-ended questions involve more reasoning, emotional expression, and creative thinking, resulting in greater variability and less stability in fine-tuning outcomes. Given these structural differences, achieving statistically significant improvements across both question types within a single model is challenging.

\begin{figure}[ht!]
\centering
\includegraphics[width=\textwidth]{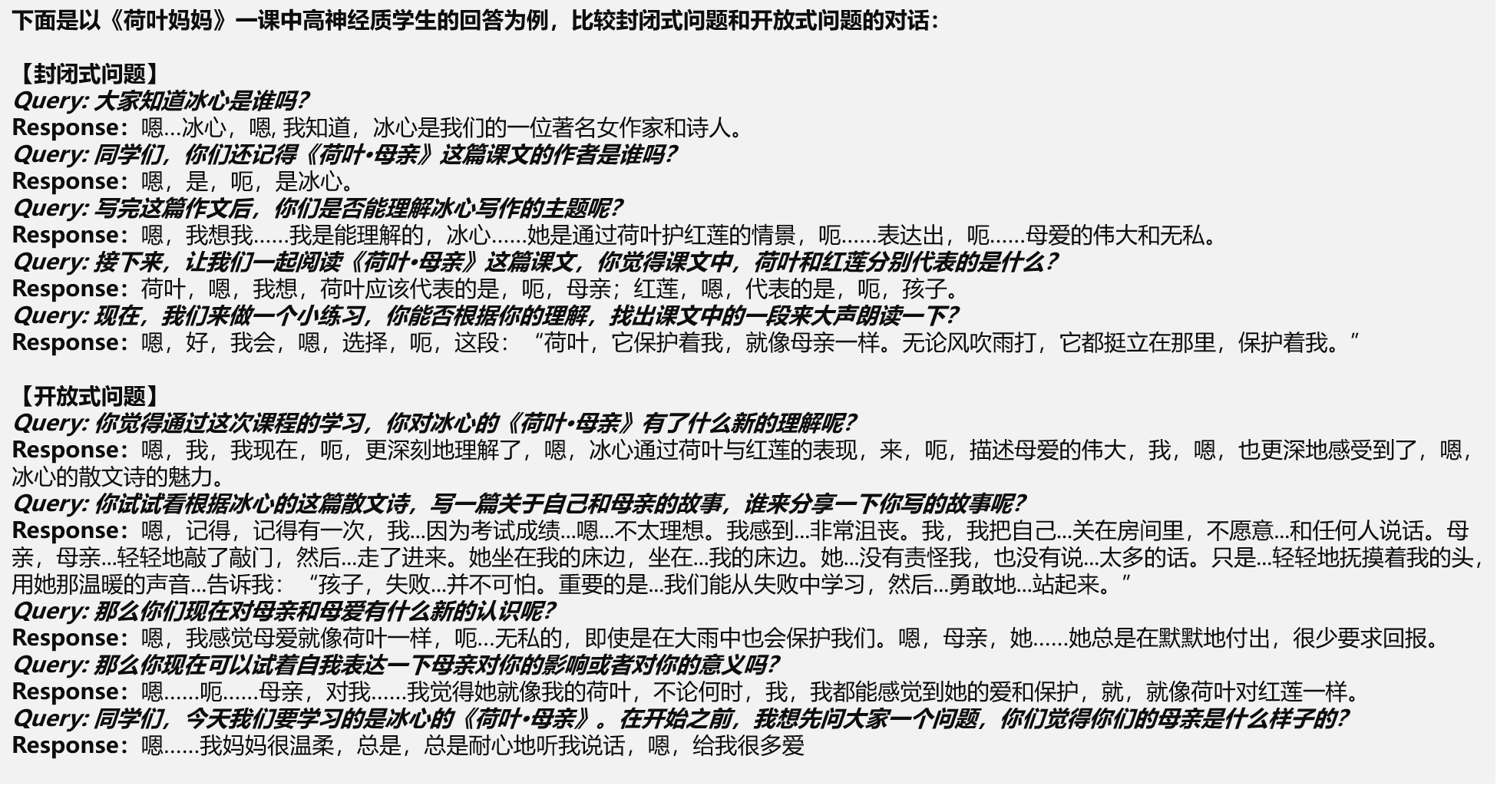}
\caption{Comparison of close-ended Question and open-ended question (original Chinese version).}
\label{fig:Question BC.}
\end{figure}

\begin{tcolorbox}[colback=gray!00,%gray background
                  colframe=black,% black frame colour
                %   width=17.2cm,% Use 8cm total width,
                  arc=1.5mm, auto outer arc,
                  breakable,
                  left=0.9mm, right=0.9mm,
                  boxrule=0.9pt, colbacktitle = black!65!black,
                  title = {Comparison of close-ended Question and open-ended question (translated in English for better understanding).}
                 ]
Here are the dialogues using the responses of a HN student in the "Lotus Leaves: Mother" lesson as an example to compare closed-ended and open-ended questions:
\tcblower
\textit{\textit{[Closed-ended Questions]}}: \\
\textit{Query}\textit{:Does everyone know who Bing Xin is?} \\
\textbf{Response}: Um... Bing Xin, um, I know. Bing Xin is a famous female writer and poet of ours. \\
\textit{Query}\textit{:Do you all remember who wrote the essay "Lotus Leaves: Mother"?} \\
\textbf{Response}: Um, it's, uh, it's Bing Xin. \\
\textit{Query}\textit{:After writing this essay, were you able to understand the theme of Bing Xin's writing?} \\
\textbf{Response}: Um, I think I... I can understand. Bing Xin... she uses the image of the lotus leaf protecting the red lotus to, uh... express, uh... the greatness and selflessness of maternal love. \\
\textit{Query}\textit{:Next, let's read "Lotus Leaves: Mother" together. What do you think the lotus leaf and the red lotus represent in the text?} \\
\textbf{Response}: The lotus leaf, um, I think it represents, uh, the mother; the red lotus, um, represents, uh, the child. \\
\textit{Query}\textit{:Now, let's do a small exercise. Based on your understanding, can you pick out a passage from the text to read aloud?} \\
\textbf{Response}: Um, okay, I will, um, choose, uh, this passage: "The lotus leaf, it protects me, just like a mother. No matter the wind or rain, it stands there tall, protecting me." \\

\textit{\textit{[Open-ended Questions]}}: \\
\textit{Query}\textit{:How do you think your understanding of Bing Xin’s "Lotus Leaves: Mother" has deepened through this lesson?} \\
\textbf{Response}: Um, I... I now, uh, have a deeper understanding that, um, Bing Xin uses the lotus leaf and the red lotus to, uh, describe the greatness of maternal love. I, um, also now feel more deeply the charm of Bing Xin's lyrical prose. \\
\textit{Query}\textit{:Try to write a story based on Bing Xin's essay about you and your mother. Who would like to share their story?} \\
\textbf{Response}: Um, I remember... I remember one time when I... didn't do well on a test... um... I felt very disappointed. I... locked myself in my room and didn't want to... talk to anyone. My mother, my mother... gently knocked on the door, then... came in. She sat by my bed, sat... by my bed. She... didn't scold me, nor did she... say too much. She just... gently stroked my head, and with her warm voice... told me, "Child, failure... is nothing to fear. What's important is... that we learn from it and then... bravely... stand back up."\\
\textit{Query}:\textit{Now, how do you understand mothers and maternal love after this?} \\
\textbf{Response}: Um, I feel that maternal love is like a lotus leaf, uh... selfless, always protecting us even in heavy rain. Um, mothers, they... they always give so much, silently, asking for so little in return. \\
\textit{Query}:\textit{Now, can you try to express what your mother means to you or how she has influenced you?} \\
\textbf{Response}: Um... uh... my mother, to me... I feel like she's my lotus leaf. No matter what, I... I can always feel her love and protection, just... just like how the lotus leaf protects the red lotus. \\
\textit{Query}\textit{:Today we are going to learn Bing Xin’s "Lotus Leaves: Mother." Before we start, I’d like to ask everyone, what kind of person do you think your mother is?} \\
\textbf{Response}: Um... my mom is very gentle, she always... always patiently listens to me and, um, gives me a lot of love. \\
\end{tcolorbox}

\subsubsection{LLaVa model bad case after fine-tune}
\label{subsubsec:badcase-llava}

\textbf{LLaVa's suboptimal fine-tuning performance:} 
The experimental results show that, while LLaVa achieved significant improvements in all fine-tuning tasks—student personalities, question types, and learning stages—it still had an overall post-fine-tuning accuracy of around 67.85\%, which was lower compared to the other models. This outcome is likely due to LLaVa's pretraining predominantly on English data, limiting its performance in Chinese-language applications compared to models pretrained on datasets containing Chinese content (see App.~\ref{subsubsec:badcase-llava} for more examples). Future research could address this limitation by incorporating more extensive Chinese-language pretraining or customized fine-tuning tailored to Chinese contexts, enhancing LLaVa's applicability in multilingual environments.

In this section, we use the responses of HN LVSA, which showed the best average performance after fine-tuning in our experiment, as an example. By comparing LLaVa's model performance with the other three models, we highlight the bad cases observed in LLaVa's fine-tuning.

\begin{figure}[ht!]
\centering
\includegraphics[width=\textwidth]{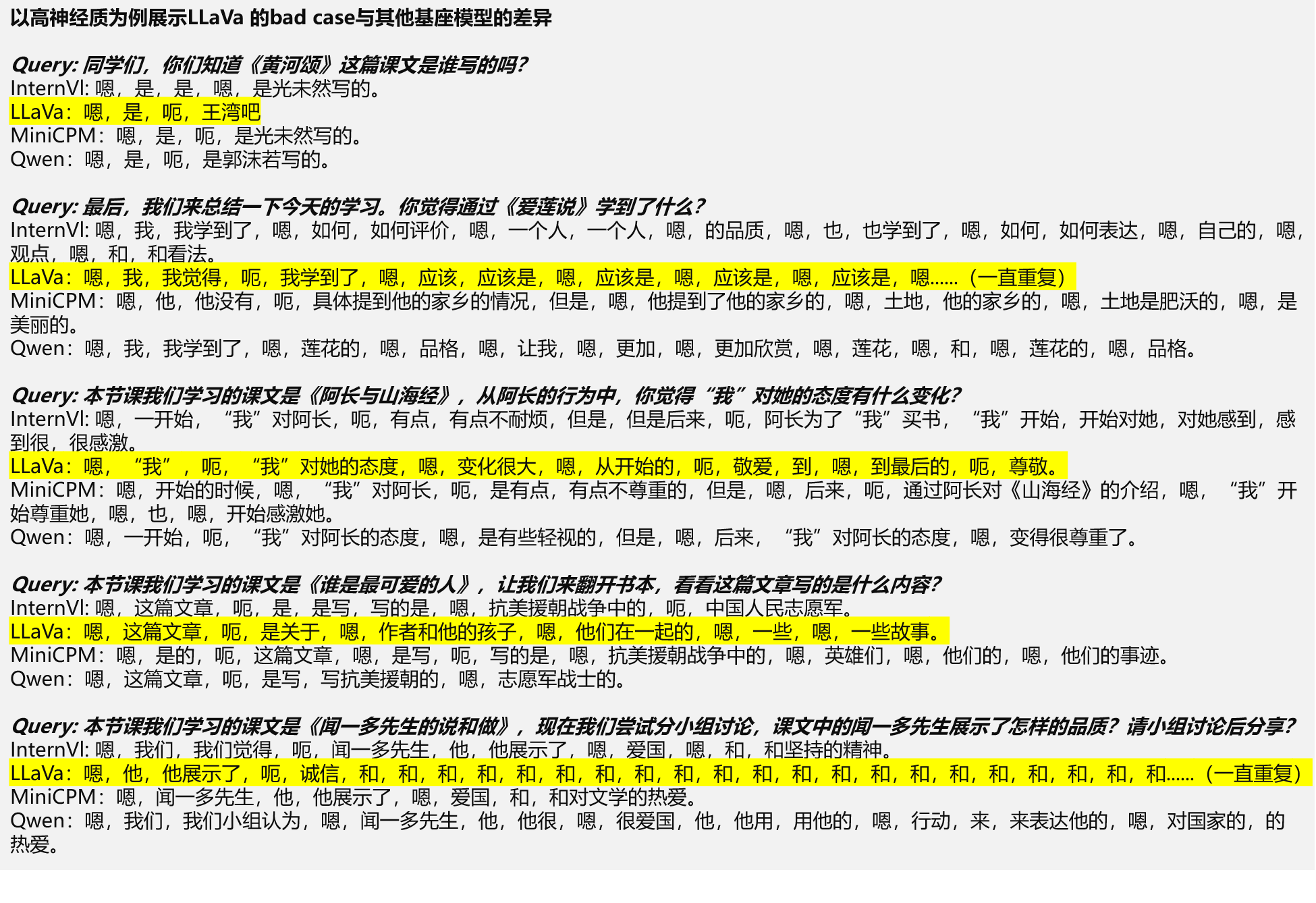}
\caption{Bad case after fine-tuning of LLaVa compared with other three models in HN LVSA (original Chinese version).}
\label{fig:LLaVa BC.}
\end{figure}

\begin{tcolorbox}[colback=gray!00,%gray background
                  colframe=black,% black frame colour
                %   width=17.2cm,% Use 8cm total width,
                  arc=1.5mm, auto outer arc,
                  breakable,
                  left=0.9mm, right=0.9mm,
                  boxrule=0.9pt, colbacktitle = black!65!black,
                  title = {Bad case after fine-tuning of LLaVa in HN LVSA (translated in English for better understanding).}
                 ]
Here are bad cases of Qwen model performance in HN student after fine-tuning:
\tcblower
\textit{QUERY}: \textit{do you know who wrote the text Ode to the Yellow River?} \\
\textbf{[InternVl]}: Um, it was, um, Guang Weiran who wrote it. \\
\colorbox{yellow}{\textbf{[LLaVa]}: Um, was it, um, Wang Wan?} \\
\textbf{[MiniCPM]}: Um, it was, um, Guang Weiran who wrote it. \\
\textbf{[Qwen]}: Um, was it, um, Guo Moruo? \\

\textit{QUERY}: \textit{Lastly, let's summarize today's lesson. What do you think you've learned from On the Love of Lotus Flowers?} \\
\textbf{[InternVl]}: Um, I, I learned, um, how, how to evaluate, um, a person’s, um, qualities, um, and also, also learned, um, how to, um, express, um, my, um, views, um, and, and opinions. \\
\colorbox{yellow}{\parbox{\linewidth}{\textbf{[LLaVa]}: Um, I, I think, um, I learned, um, it should, should be, um, should be, um, should be, um, should be, um... (keeps repeating)}} \\
\textbf{[MiniCPM]}: Um, he, he didn’t, um, specifically mention his hometown’s situation, but, um, he talked about his hometown’s, um, land, and how, um, it was fertile and beautiful. \\
\textbf{[Qwen]}: Um, I, I learned, um, the character of the lotus, um, and it made me, um, appreciate, um, the lotus, um, and its, um, qualities even more. \\

\textit{QUERY}: \textit{Today we are studying the text A Chang and the Classic of Mountains and Seas. From A Chang's actions, what do you think changed in "my" attitude towards her?} \\
\textbf{[InternVl]}: Um, at first, "I" was a bit, a bit impatient with A Chang, but, but later, um, after she bought the book for "me," "I" started to, started to feel, feel very, very grateful to her. \\
\colorbox{yellow}{\parbox{\linewidth}{\textbf{[LLaVa]}: Um, "I," uh, "my" attitude towards her, um, changed a lot, um, from, um, admiration at first, to, um, respect in the end.}} \\
\textbf{[MiniCPM]}: Um, at the beginning, um, "I" didn’t really, um, respect A Chang, but, um, later, um, after A Chang introduced The Classic of Mountains and Seas, um, "I" started to respect her, um, and, um, started to appreciate her. \\
\textbf{[Qwen]}: Um, at first, uh, "I" was a bit, um, dismissive of A Chang, but, um, later, "I" started to respect her. \\

\textit{QUERY}: \textit{Today we are studying the text Who Are the Loveliest People. Let's open the book and see what this article is about.} \\
\textbf{[InternVl]}: Um, this article, um, is about, um, the Chinese People’s Volunteers, um, during the Korean War. \\
\colorbox{yellow}{\parbox{\linewidth}{\textbf{[LLaVa]}: Um, this article, um, is about, um, the author and his child, um, they’re together, um, in some, um, some stories.}} \\
\textbf{[MiniCPM]}: Um, yes, uh, this article, um, is about, um, the heroes, um, of the Korean War, um, their, um, deeds. \\
\textbf{[Qwen]}: Um, this article, um, is about, um, the volunteers, um, of the Korean War. \\

\textit{QUERY}: \textit{Today we are studying the text Wen Yiduo’s Words and Deeds. Now, let's try to discuss in groups: what qualities did Wen Yiduo demonstrate in the text? Please discuss in your groups and then share.} \\
\textbf{[InternVl]}: Um, we, we think, um, that Wen Yiduo demonstrated, um, patriotism, um, and, and perseverance. \\
\colorbox{yellow}{\parbox{\linewidth}{\textbf{[LLaVa]}: Um, he, he demonstrated, um, integrity, and, and, and, and, and, and, and, and, and, and, and... (keeps repeating).}} \\
\textbf{[MiniCPM]}: Um, Wen Yiduo demonstrated, um, patriotism, and, and a love for literature. \\
\textbf{[Qwen]}: Um, we, our group thinks, um, that Wen Yiduo, um, he was very, um, patriotic, and he, he expressed his, um, love for the country through his, um, actions. \\
\end{tcolorbox}

\section{Detailed Interaction Settings and Results}
\label{app:interaction}

\subsection{Interaction Experiment Setup}
\label{subsec:interaction-setup}

This experiment investigates how teachers adapt their instructional behaviors and decision-making strategies when interacting with virtual students exhibiting distinct personality traits. By embedding these interactions within a structured instructional scenario, we examine how real teachers dynamically respond to the personalized feedback of different LSVAs and how these traits influence pedagogical adaptation.

\paragraph{Participants.} We recruited 10 pre-service teachers for the experiment. Among them, 5 had completed practical teaching internships in Chinese language instruction (coded as “experienced = 1”), and 5 had no formal teaching experience (“experienced = 0”). Participants No1, No2, No7, No8, and No10 belonged to the inexperienced group, while No3, No4, No5, No6, and No9 were assigned to the experienced group.

\paragraph{Virtual students.} The virtual student agents were generated using fine-tuned InternVL models, each corresponding to one of five personality types $\pi \in \mathcal{P} = \{\textbf{HE}, \textbf{HN}, \textbf{HA}, \textbf{LO}, \textbf{LC}\}$. Each teacher interacted one-on-one with five LSVAs, completing a distinct task with each personality-conditioned student.

\begin{figure}[h]
\centering
\includegraphics[width=0.95\linewidth]{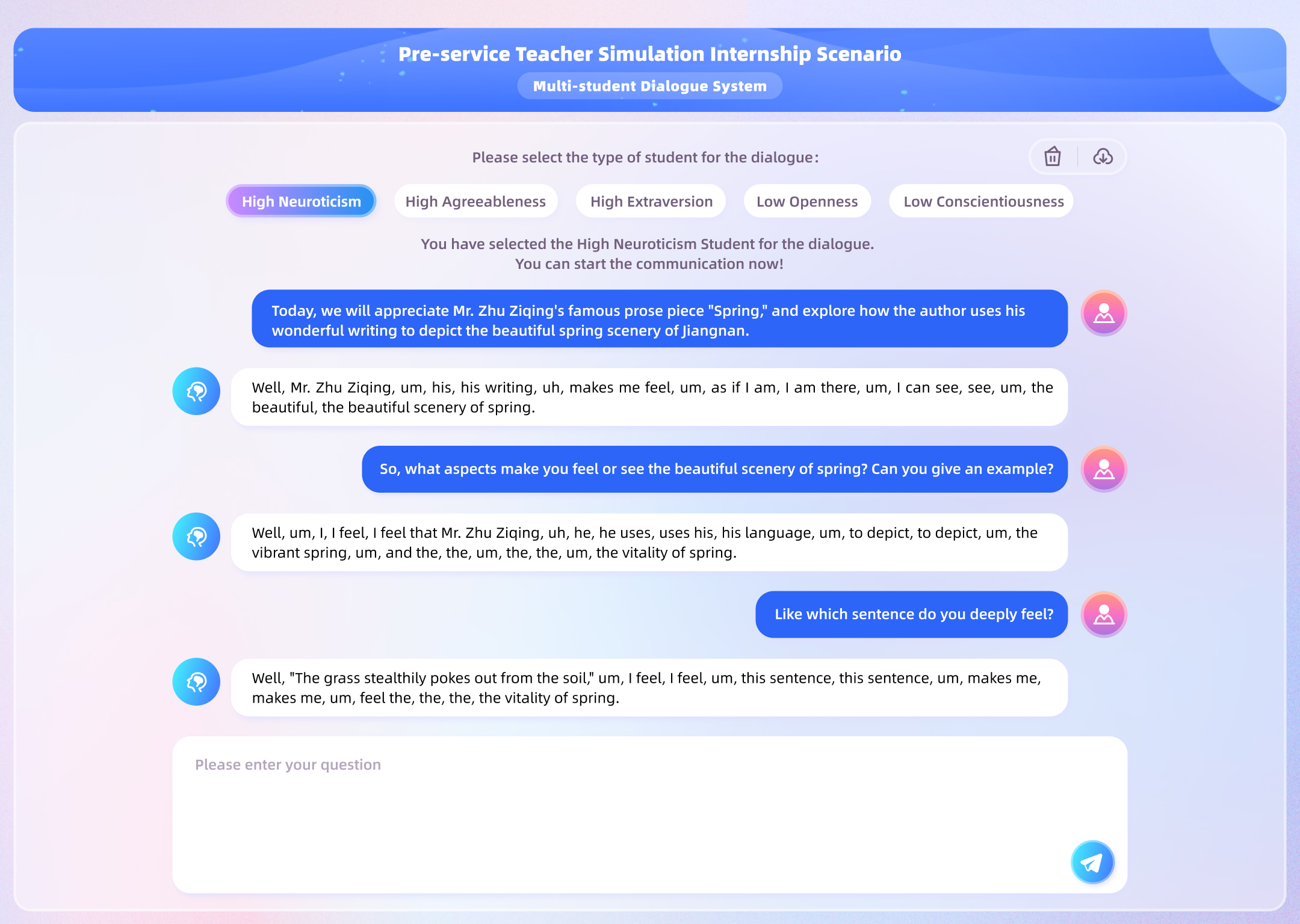}
\caption{The user interface of the pre-service teacher–LVSA interaction system. Teachers select one of five personality-conditioned students and initiate a task-based dialogue on a fixed topic. The dialogue shown here involves a High Neuroticism LVSA discussing Zhu Ziqing's essay \textit{Spring}, exhibiting hesitation, repetition, and affective expressions.}
\label{fig:interaction-ui}
\end{figure}

\paragraph{Procedure.} Each teacher completed 5 sessions, each lasting 10–15 minutes, with a 5-minute break between sessions to minimize fatigue. All sessions used the same instructional topic—Zhu Ziqing’s essay “Spring”—to ensure semantic consistency across trials. Teachers were allowed to flexibly design their lecture flow and questioning based on the fixed content. All interactions were screen-recorded and transcribed for further annotation and analysis (as shown in Fig.~\ref{fig:interaction-ui}).

\paragraph{Post-interaction survey.} After the experiment, all participants completed a questionnaire, which included: 

\begin{itemize}
    \item Subjective ratings of virtual student behavior (\eg, naturalness, appropriateness);
    \item Satisfaction with the interaction experience;
    \item Reflections on how they adjusted their teaching strategies.
\end{itemize}

To visualize system-level behavior patterns in teacher–LVSA interactions, we summarize two key quantitative metrics across the 50 sessions: dialogue turn count and average teacher response time.

\begin{figure}[t!]
\centering
\includegraphics[width=0.92\linewidth]{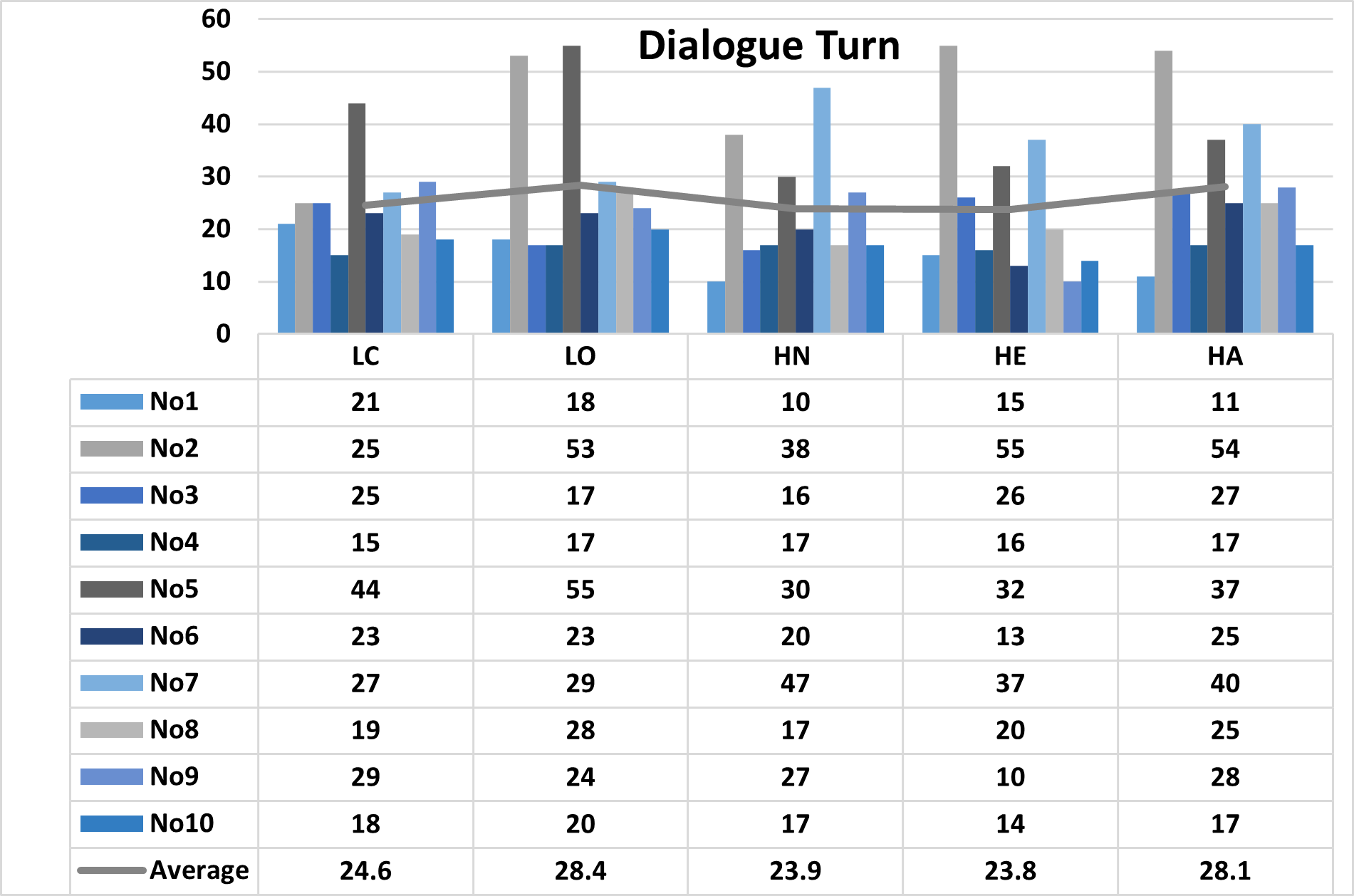}
\caption{Dialogue turns per teacher (No1–No10) across five LVSA personality types.}
\label{fig:dialogue-turn}
\end{figure}

\begin{figure}[t!]
\centering
\includegraphics[width=0.92\linewidth]{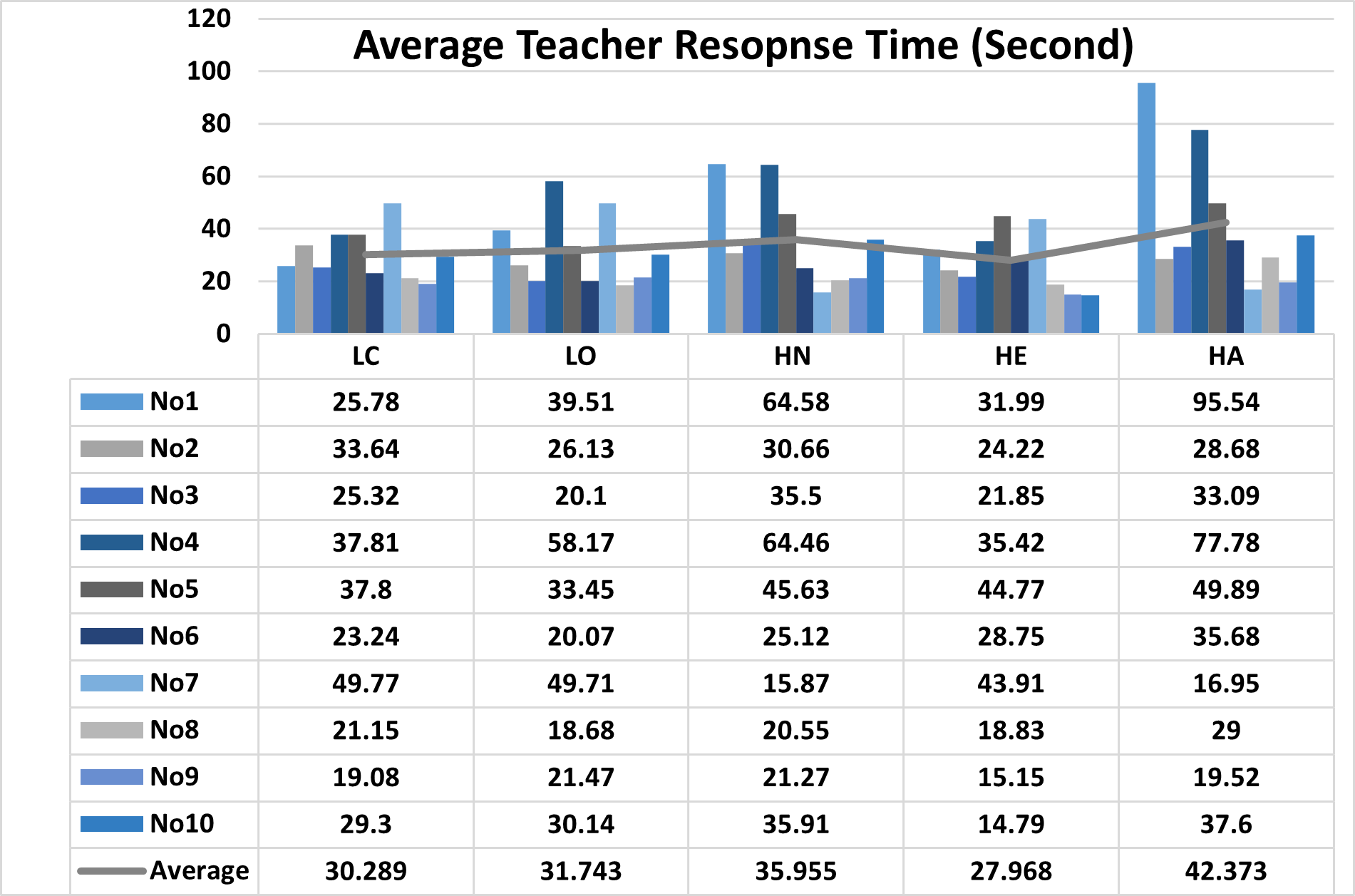}
\caption{Average teacher response time (in seconds) by LVSA personality type and participant.}
\label{fig:response-time}
\end{figure}

To analyze interaction dynamics across LVSA personalities, we present two key metrics: the number of dialogue turns and the teacher’s average response time (Fig.~\ref{fig:dialogue-turn} and Fig.~\ref{fig:response-time}). These two indicators reflect the depth and pacing of instructional engagement with each virtual student type.

High Agreeableness (HA) and Low Openness (LO) students elicited the most dialogue turns on average (28.1 and 28.4 respectively), suggesting that teachers engaged in more sustained instructional exchanges with these personalities. Meanwhile, teacher response time was also longest with HA (42.4s) and High Neuroticism (HN) students (36.0s), indicating greater cognitive load or strategy deliberation when responding to affectively intense or hesitant learners.

Interestingly, although HN students triggered fewer dialogue turns (23.9), the longer teacher response times suggest that teachers may have struggled to find appropriate scaffolding strategies. In contrast, High Extraversion (HE) students resulted in both fewer turns (23.8) and the shortest teacher response time (27.9s), possibly due to their assertiveness or ease of interaction. These findings confirm that LVSA personality traits not only influence the quantity of interaction but also the temporal cost of instructional adaptation.

\subsection{Annotation Framework and Analysis}
\label{subsec:annotation-details}

We adopted a multi-dimensional annotation scheme grounded in educational discourse theory. The teacher–student dialogue data were labeled along four axes, as summarized below:

\begin{itemize}
  \item \textbf{Teacher Behavior Type:} lecture, instruction, questioning, accepting suggestions, positive reinforcement, emotional expression, criticism (Flanders Interaction Analysis System).
  \item \textbf{Question Type:} open-ended, closed-ended, non-question.
  \item \textbf{Bloom’s Cognitive Level:} remember, understand, apply, analyze, evaluate, create.
  \item \textbf{Student Behavior Type:} correct answer, incorrect answer, repeated answer, spontaneous question, irrelevant response, invalid utterance, no response.
\end{itemize}

Initial annotations were generated using GPT-4, and then manually reviewed and corrected by trained coders to ensure theoretical validity and label consistency.

\begin{table}[ht]
\centering
\caption{Distribution of Bloom’s cognitive levels across LVSA personality types, with experienced/novice breakdown.}
\label{tab:appx1-bloom-subgroups}
\small
\begin{tabular}{lllllll}
\toprule
\textbf{} & \textbf{Remember} & \textbf{Understand} & \textbf{Apply} & \textbf{Analyze} & \textbf{Evaluate} & \textbf{Create} \\
\midrule
LC & 38 (\textit{19}/\underline{19}) & 190 (\textit{96}/\underline{94}) & 4 (\textit{2}/\underline{2}) & 5 (\textit{5}/\underline{0}) & 6 (\textit{2}/\underline{4}) & 3 (\textit{3}/\underline{0}) \\
LO & 50 (\textit{22}/\underline{28}) & 219 (\textit{117}/\underline{102}) & 4 (\textit{1}/\underline{3}) & 2 (\textit{1}/\underline{1}) & 5 (\textit{1}/\underline{4}) & 4 (\textit{2}/\underline{2}) \\
HN & 39 (\textit{25}/\underline{14}) & 185 (\textit{102}/\underline{83}) & 2 (\textit{0}/\underline{2}) & 3 (\textit{1}/\underline{2}) & 3 (\textit{1}/\underline{2}) & 7 (\textit{2}/\underline{5}) \\
HE & 32 (\textit{15}/\underline{17}) & 183 (\textit{86}/\underline{97}) & 8 (\textit{2}/\underline{6}) & 7 (\textit{4}/\underline{3}) & 2 (\textit{1}/\underline{1}) & 6 (\textit{4}/\underline{2}) \\
HA & 27 (\textit{16}/\underline{11}) & 223 (\textit{106}/\underline{117}) & 7 (\textit{2}/\underline{5}) & 10 (\textit{6}/\underline{4}) & 3 (\textit{2}/\underline{1}) & 11 (\textit{4}/\underline{7}) \\
\midrule
Total & 186 & 1,000 & 25 & 27 & 19 & 31 \\
\bottomrule
\end{tabular}
\vspace{3pt}
\begin{tablenotes}
\item[] \small Note: Values in parentheses represent the subgroup breakdown: \textit{Experienced} / \underline{Novice}.
\end{tablenotes}
\end{table}

% 从表~\ref{tab:appx1-bloom-subgroups} 可见，HA和HE学生激发了最多高阶认知表达，尤其在无经验教师组中更为突出。例如，HA学生的11次“创造”型语言中，有7次来自无经验组；HE学生的8次“应用”级语言中有6次为无经验教师触发。相比之下，HN和LC学生集中在“记忆”与“理解”等低阶认知层级，HN的39次记忆类语言中有25次来自经验教师。总体说明，无经验教师在面对表达积极或合作型学生时，更倾向于提出开放和探索性的认知任务。
In Tab.~\ref{tab:appx1-bloom-subgroups}, HA and HE students triggered the most high-order cognitive prompts (\eg, create and apply), especially in interactions with novice teachers. For example, 7 of the 11 “create”-level utterances for HA came from novice participants, and 6 of the 8 “apply”-level instances for HE were also attributed to novices. In contrast, HN and LC sessions were dominated by low-level categories like remember and understand, with HN showing 39 “remember” instances, mostly from experienced teachers (25 vs. 14). These patterns suggest that novice teachers may adopt more exploratory prompts when facing emotionally stable or expressive students.

\begin{table}[ht]
\centering
\caption{Distribution of teacher question types across LVSA personality types, with experienced/novice breakdown.}
\label{tab:appx2-question-type}
\small
\begin{tabular}{llll}
\toprule
\textbf{} & \textbf{Closed-ended} & \textbf{Open-ended} & \textbf{No question} \\
\midrule
LC & 49 (\textit{23}/\underline{26}) & 109 (\textit{62}/\underline{47}) & 88 (\textit{42}/\underline{46}) \\
LO & 68 (\textit{41}/\underline{27}) & 128 (\textit{62}/\underline{66}) & 88 (\textit{41}/\underline{47}) \\
HN & 59 (\textit{35}/\underline{24}) & 111 (\textit{64}/\underline{47}) & 69 (\textit{32}/\underline{37}) \\
HE & 37 (\textit{14}/\underline{23}) & 120 (\textit{59}/\underline{61}) & 81 (\textit{39}/\underline{42}) \\
HA & 43 (\textit{21}/\underline{22}) & 154 (\textit{80}/\underline{74}) & 84 (\textit{35}/\underline{49}) \\
\midrule
Total & 256 & 622 & 410 \\
\bottomrule
\end{tabular}
\vspace{3pt}
\begin{tablenotes}
\item[] \small Note: Values in parentheses represent the subgroup breakdown: \textit{Experienced} / \underline{Novice}.
\end{tablenotes}
\end{table}

% 表~\ref{tab:appx2-question-type} 显示，教师在面对HA与HE学生时更倾向提出开放式问题，尤其经验教师（如HA学生中154次开放式问题中有80次来自经验组）。相反，LC与LO学生在“未提问”上的频率最高（均为88），显示出师生互动深度偏低。进一步看，无经验教师在LC条件下提出了更多封闭式问题（26次对23次），可能反映出在面对认知挑战型学生时教学调适能力不足。
Tab.~\ref{tab:appx2-question-type} reveals that HA and HE students were asked more open-ended questions, especially from experienced teachers (\eg, 80 of 154 HA-open questions from experienced). In contrast, LC and LO students had the highest “no question” frequencies (both 88), indicating lower levels of dialogic engagement. Notably, novice teachers interacting with LC students asked more closed-ended questions (26 vs. 23), suggesting lower adaptability or confidence when handling cognitively challenging interactions.

\begin{table}[ht]
\centering
\caption{Distribution of teacher act types across LVSA personality types, with experienced/novice breakdown.}
\label{tab:appx3-teacher-behavior}
\small
\setlength{\tabcolsep}{2pt}
\begin{tabular}{llllllll}
\toprule
\textbf{} & 
\textbf{Criticism} & 
\textbf{Directive} & 
\textbf{Questioning} & 
\makecell{\textbf{Emotional}\\\textbf{Expression}} & 
\textbf{Lecture} & 
\makecell{\textbf{Accepting}\\\textbf{Suggestions}} & 
\makecell{\textbf{Positive}\\\textbf{Reinforcement}} \\
\midrule
LC & 0 (\textit{0}/\underline{0}) & 20 (\textit{9}/\underline{11}) & 131 (\textit{67}/\underline{64}) & 23 (\textit{9}/\underline{14}) & 47 (\textit{22}/\underline{25}) & 6 (\textit{5}/\underline{1}) & 19 (\textit{15}/\underline{4}) \\
LO & 5 (\textit{0}/\underline{5}) & 22 (\textit{6}/\underline{16}) & 159 (\textit{81}/\underline{78}) & 16 (\textit{10}/\underline{6}) & 64 (\textit{33}/\underline{31}) & 3 (\textit{3}/\underline{0}) & 15 (\textit{11}/\underline{4}) \\
HN & 2 (\textit{0}/\underline{2}) & 17 (\textit{7}/\underline{10}) & 135 (\textit{74}/\underline{61}) & 8 (\textit{5}/\underline{3}) & 49 (\textit{26}/\underline{23}) & 2 (\textit{1}/\underline{1}) & 26 (\textit{18}/\underline{8}) \\
HE & 5 (\textit{2}/\underline{3}) & 18 (\textit{5}/\underline{13}) & 115 (\textit{47}/\underline{68}) & 13 (\textit{6}/\underline{7}) & 49 (\textit{24}/\underline{25}) & 4 (\textit{4}/\underline{0}) & 34 (\textit{24}/\underline{10}) \\
HA & 8 (\textit{2}/\underline{6}) & 24 (\textit{12}/\underline{12}) & 153 (\textit{74}/\underline{79}) & 7 (\textit{4}/\underline{3}) & 61 (\textit{30}/\underline{31}) & 10 (\textit{3}/\underline{7}) & 18 (\textit{11}/\underline{7}) \\
\midrule
Total & 20 & 101 & 693 & 67 & 270 & 25 & 112 \\
\bottomrule
\end{tabular}
\vspace{3pt}
\begin{tablenotes}
\item[] \small Note: Values in parentheses represent the subgroup breakdown: \textit{Experienced} / \underline{Novice}.
\end{tablenotes}
\end{table}

% 表~\ref{tab:appx4-teacher-behavior} 显示，教师在HE与HA学生面前使用鼓励与正面强化行为最多。例如，HE学生中经验教师使用表扬行为24次，而对照组仅为10次。相对地，在面对LC与LO学生时，教师更频繁使用指令型行为，尤其无经验组（如LO中22次指令中16次为无经验组）。此外，情绪性表达在LC学生组中最多（23次），其中有14次来自无经验教师，可能反映出应对此类学生时的挫败感或策略压力。
As shown in Tab.~\ref{tab:appx3-teacher-behavior}, praise and positive reinforcement were most frequently used when interacting with HE and HA students. For instance, HE teachers in the experienced group used praise 24 times, compared to only 10 in the novice group. Conversely, directive instruction was more prevalent with LC and LO students, especially under novice guidance (\eg, 16 of 22 directives for LO). Interestingly, emotional expression appeared most often in LC sessions (23 times), with more frequent use among novice teachers (14 vs. 9), potentially reflecting heightened frustration or cognitive load.
 
\begin{table}[ht]
\centering
\caption{Distribution of student behavior types across LVSA personality types, with experienced/novice breakdown.}
\label{tab:appx4-student-behavior}
\small
\setlength{\tabcolsep}{2pt}
\begin{tabular}{llllllll}
\toprule
\textbf{} & 
\makecell{\textbf{Spontaneous}\\\textbf{Question}} & 
\makecell{\textbf{Irrelevant}\\\textbf{Response}} & 
\makecell{\textbf{Invalid}\\\textbf{Utterance}} & 
\makecell{\textbf{No}\\\textbf{Response}} & 
\makecell{\textbf{Correct}\\\textbf{Answer}} & 
\makecell{\textbf{Repeated}\\\textbf{Answer}} & 
\makecell{\textbf{Incorrect}\\\textbf{Answer}} \\
\midrule
LC & 16 (\textit{8}/\underline{8}) & 16 (\textit{13}/\underline{3}) & 43 (\textit{15}/\underline{28}) & 0 (\textit{0}/\underline{0}) & 151 (\textit{86}/\underline{65}) & 14 (\textit{4}/\underline{10}) & 6 (\textit{1}/\underline{5}) \\
LO & 6 (\textit{1}/\underline{5}) & 3 (\textit{0}/\underline{3}) & 93 (\textit{50}/\underline{43}) & 11 (\textit{9}/\underline{2}) & 131 (\textit{78}/\underline{53}) & 8 (\textit{3}/\underline{5}) & 32 (\textit{3}/\underline{29}) \\
HN & 0 (\textit{0}/\underline{0}) & 0 (\textit{0}/\underline{0}) & 52 (\textit{23}/\underline{29}) & 0 (\textit{0}/\underline{0}) & 174 (\textit{104}/\underline{70}) & 9 (\textit{2}/\underline{7}) & 4 (\textit{2}/\underline{2}) \\
HE & 9 (\textit{2}/\underline{7}) & 3 (\textit{0}/\underline{3}) & 14 (\textit{9}/\underline{5}) & 0 (\textit{0}/\underline{0}) & 203 (\textit{100}/\underline{103}) & 4 (\textit{1}/\underline{3}) & 5 (\textit{0}/\underline{5}) \\
HA & 8 (\textit{3}/\underline{5}) & 3 (\textit{1}/\underline{2}) & 10 (\textit{6}/\underline{4}) & 0 (\textit{0}/\underline{0}) & 244 (\textit{124}/\underline{120}) & 5 (\textit{1}/\underline{4}) & 11 (\textit{1}/\underline{10}) \\
\midrule
Total & 39 & 25 & 212 & 11 & 903 & 40 & 58 \\
\bottomrule
\end{tabular}
\vspace{3pt}
\begin{tablenotes}
\item[] \small Note: Values in parentheses represent the subgroup breakdown: \textit{Experienced} / \underline{Novice}.
\end{tablenotes}
\end{table}

% 从表~\ref{tab:appx3-student-behavior} 可见，HA与HE学生的正确回答次数最多（分别为244和203），且经验组与对照组贡献相对均衡。值得注意的是，LO学生中32次错误回答中，有29次出现在无经验教师教学下，提示教师在此类情境中的教学策略可能未能有效引导学生；同时，LC学生的无效语言次数也较高（43），其中有28次来自无经验组，表明该类学生在缺乏结构化引导时更容易失去表达连贯性。
Tab.~\ref{tab:appx4-student-behavior} shows HA and HE students yielded the highest number of correct answers (244 and 203), with nearly balanced contributions from experienced and novice teachers. Meanwhile, LO students exhibited the most incorrect answers (32), with a large majority coming from novice sessions (29 vs. 3), suggesting potential mismatches in instructional strategies. LC students produced a high number of invalid utterances (43), particularly under novice instruction (28/15), indicating that these students may require more structured scaffolding.

\paragraph{Overall Summary.}

% 通过对认知层级、提问方式、学生反应与教师行为四个维度的综合分析，我们发现虚拟学生的人格特质对教学交互的质量与教师适应性具有显著影响。高宜人性（HA）与高外向性（HE）学生更易激发教师提出开放性问题、引导高阶认知表达，并获得更多正向反馈；而低开放性（LO）与低尽责性（LC）学生则更易出现无效或错误语言，教师也更倾向使用指令式控制，互动深度与教学灵活性均有所下降。
The combined analysis across cognitive levels, question strategies, student responses, and teacher behaviors demonstrates that LVSA personality traits significantly influence both interaction quality and pedagogical adaptability. High Agreeableness (HA) and High Extraversion (HE) students consistently triggered more open-ended questions, high-order cognitive scaffolding, and positive teacher behaviors. In contrast, Low Openness (LO) and Low Conscientiousness (LC) students were associated with more directive control, invalid or incorrect responses, and reduced dialogic richness.

% 此外，教师经验对上述影响起到了调节作用。无经验教师在面对HA与HE学生时更能激发创造性表达，但在应对认知混乱型学生（如LO、LC）时表现出更明显的策略困难，如无效表达比例上升、指令性行为频繁。上述发现不仅验证了SOEI框架在引导教学行为多样性方面的有效性，也揭示了不同人格驱动下的人机交互变化及教师行为的能力边界。
Moreover, teacher experience further moderated these effects. Novice teachers showed a tendency to elicit more creative-level discourse in emotionally positive contexts (\eg, HA, HE), but also exhibited greater difficulty handling cognitively disorganized students (\eg, LO, LC), as reflected by higher invalid utterances and directive frequencies. These findings validate the SOEI framework’s ability to induce diverse teaching behaviors and highlight its value in revealing both student-driven variation and teacher-level constraints in real-time educational interactions.

\subsection{Evaluation Details for Personality Ranking Task}
\label{subsec:personality-eval}

\begin{figure*}[ht]
    \centering
    \includegraphics[width=0.95\linewidth]{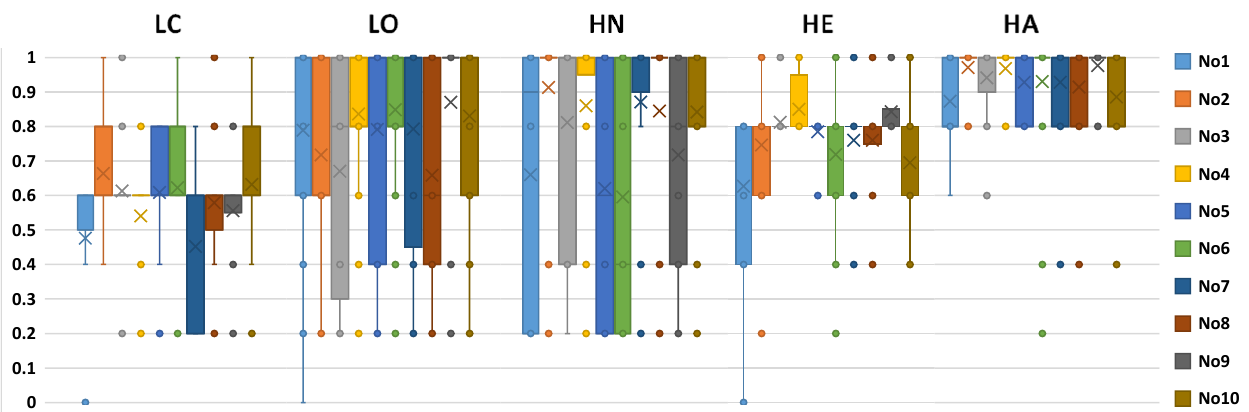}
    \caption{Boxplots of personality prediction scores across 10 teachers, illustrating the stability and variability of different personality types in multi-turn interactions. Each bar represents the distribution of per-turn prediction scores for one teacher under a given personality condition. The horizontal axis indicates personality types (LC, LO, HN, HE, HA), and the vertical axis represents ranking scores (0–1).}
    \label{fig:round-2}
\end{figure*}

\paragraph{Ranking Protocol.}

In the personality identification task, GPT-4 was asked to return a ranked list of all five personality types given a single teacher–student dialogue turn. The true label was then matched against the list to compute Top-1, Top-2, and Top-3 accuracy.

To provide finer-grained evaluation, we assigned a position-sensitive score to each prediction: 1.0 for first place, 0.8 for second, down to 0.2 for fifth. If the true label was not present in the predicted list, a score of 0 was assigned.

Final results were averaged over 1,288 turns. The score matrix in Tab.~\ref{tab:personality-score-table} presents per-participant and per-personality averages.

\paragraph{Visualization of Score Distributions}

Fig.~\ref{fig:round-2} presents boxplots of personality prediction scores across 10 teacher participants for each LVSA personality type. Each box reflects the distribution of per-turn scores in a teacher–student interaction, offering a complementary view to the trajectory plots shown in the main text.

We observe that HA and HE yield narrow interquartile ranges with few outliers, suggesting highly stable and easily recognizable personality signals. In contrast, LO and LC exhibit broader distributions, lower medians, and more outliers, indicating greater score volatility. Notably, some teachers recorded multiple turns with prediction scores below 0.4 or complete misclassifications (score = 0), especially under LO and LC conditions.

These results align with the theoretical discussion in the main paper: while HA/HE traits are clearer and more expressive, LO/LC reflect more nuanced, context-sensitive behaviors. The variability observed here reinforces the view that behavioral inconsistency can signify expressive complexity rather than modeling failure.

\subsection{Post-Interaction Teacher Survey}
\label{subsec:survey-details}

To evaluate the pedagogical impact of personality-driven LVSAs, we administered a post-interaction survey to all 10 participating pre-service teachers. The survey covered system usability, perceived student realism, instructional behavior, and reflections on adaptive teaching.

\paragraph{Questionnaire Overview.}

The post-interaction survey covering platform usability, perceived student realism, instructional behavior adaptation, and teacher reflections. Items included Likert-scale questions (Q2–Q5, Q9, Q11–12, Q16–17), single-choice questions (Q6–7, Q13, Q15), multiple-choice questions (Q8, Q10, Q14), and one open-ended response (Q18).

\clearpage

\begin{tcolorbox}[colback=gray!00, % background color
                  colframe=black, % frame color
                  arc=1.5mm, auto outer arc,
                  breakable,
                  left=0.9mm, right=0.9mm,
                  boxrule=0.9pt, colbacktitle=black!65!black,
                  title={Survey responses from 10 pre-service teachers after interacting with personality-driven LVSAs.}
                 ]
Below is the full summary of the questionnaire results (Q1–Q18), covering usability, perceived difficulty, realism, behavioral adaptation, and reflections.
\tcblower

\textbf{Q1. ID \quad [Open-ended Question]}\\

\vspace{1em}
\textbf{Q2. I find the simulation platform’s visual interface intuitive and easy to use. \quad [Single-choice]}\\
Responses: Strongly agree: 5 (50\%), Agree: 5 (50\%), Neutral: 0 (0\%), Disagree: 0 (0\%), Strongly disagree: 0 (0\%) \\
Valid responses: 10

\vspace{1em}
\textbf{Q3. I rarely encountered difficulties or obstacles when using the simulation platform. \quad [Single-choice]}\\
Responses: Strongly agree: 2 (20\%), Agree: 7 (70\%), Neutral: 1 (10\%), Disagree: 0 (0\%), Strongly disagree: 0 (0\%) \\
Valid responses: 10

\vspace{1em}
\textbf{Q4. The visual interface helps me better understand and analyze interactions with virtual students. \quad [Single-choice]}\\
Responses: Strongly agree: 5 (50\%), Agree: 5 (50\%), Neutral: 0 (0\%), Disagree: 0 (0\%), Strongly disagree: 0 (0\%) \\
Valid responses: 10

\vspace{1em}
\textbf{Q5. The personalized features of the virtual students (\eg, personality labels, language feedback) met my expectations. \quad [Single-choice]}\\
Responses: Strongly agree: 1 (10\%), Agree: 6 (60\%), Neutral: 3 (30\%), Disagree: 0 (0\%), Strongly disagree: 0 (0\%) \\
Valid responses: 10

\vspace{1em}
\textbf{Q6. Which student best adapted to your teaching rhythm? \quad [Single-choice]}\\
Responses: High Extraversion: 6 (60\%), High Agreeableness: 3 (30\%), High Neuroticism: 1 (10\%), Low Conscientiousness: 0 (0\%), Low Openness: 0 (0\%) \\
Valid responses: 10

\vspace{1em}
\textbf{Q7. Which student posed the greatest challenge to you? \quad [Single-choice]}\\
Responses: Low Conscientiousness: 6 (60\%), Low Openness: 3 (30\%), High Extraversion: 1 (10\%), High Neuroticism: 0 (0\%), High Agreeableness: 0 (0\%) \\
Valid responses: 10

\vspace{1em}
\textbf{Q8. Which types of virtual students best resemble students in real classrooms? \quad [Multiple-choice]}\\
Responses: High Extraversion: 8 (80\%), Low Openness: 6 (60\%), High Neuroticism: 5 (50\%), High Agreeableness: 5 (50\%), Low Conscientiousness: 5 (50\%) \\
Valid responses: 10

\vspace{1em}
\textbf{Q9. My teaching adaptability improved after interacting with virtual students. \quad [Single-choice]}\\
Responses: Strongly agree: 3 (30\%), Agree: 6 (60\%), Neutral: 0 (0\%), Disagree: 1 (10\%), Strongly disagree: 0 (0\%) \\
Valid responses: 10

\vspace{1em}
\textbf{Q10. In which aspects did your adaptability improve? \quad [Multiple-choice]}\\
Responses: Adapting to different student needs: 10 (100\%), Classroom interaction skills: 8 (80\%), Emotion recognition: 5 (50\%), Classroom management: 4 (40\%), Pacing control: 3 (30\%) \\
Valid responses: 10

\vspace{1em}
\textbf{Q11. Interacting with virtual students helps me prepare for real classroom challenges. \quad [Single-choice]}\\
Responses: Strongly agree: 3 (30\%), Agree: 6 (60\%), Neutral: 1 (10\%), Disagree: 0 (0\%), Strongly disagree: 0 (0\%) \\
Valid responses: 10

\vspace{1em}
\textbf{Q12. The activity helped me better understand the trial lesson content (\eg, Zhu Ziqing’s \emph{Spring}). \quad [Single-choice]}\\
Responses: Strongly agree: 4 (40\%), Agree: 5 (50\%), Neutral: 1 (10\%), Disagree: 0 (0\%), Strongly disagree: 0 (0\%) \\
Valid responses: 10

\vspace{1em}
\textbf{Q13. Which student type contributed most to your teaching growth? \quad [Single-choice]}\\
Responses: High Extraversion: 4 (40\%), High Neuroticism: 2 (20\%), Low Conscientiousness: 2 (20\%), High Agreeableness: 1 (10\%), Low Openness: 1 (10\%) \\
Valid responses: 10

\vspace{1em}
\textbf{Q14. What teaching behaviors did you tend to use during interaction? \quad [Multiple-choice]}\\
Responses: Praise and encouragement: 10 (100\%), Questioning: 8 (80\%), Lecturing: 7 (70\%), Emotion expression: 4 (40\%), Giving commands: 2 (20\%), Accepting suggestions: 2 (20\%), Others: 1 (10\%) \\
Valid responses: 10

\vspace{1em}
\textbf{Q15. What was your usual emotional tone during the interaction? \quad [Single-choice]}\\
Responses: Positive encouragement: 6 (60\%), Moderate support: 3 (30\%), Neutral: 1 (10\%), Criticism or reminders: 0 (0\%), Others: 0 (0\%) \\
Valid responses: 10

\vspace{1em}
\textbf{Q16. I tend to adjust my teaching strategies based on students’ personality traits. \quad [Single-choice]}\\
Responses: Strongly agree: 3 (30\%), Agree: 7 (70\%), Neutral: 0 (0\%), Disagree: 0 (0\%), Strongly disagree: 0 (0\%) \\
Valid responses: 10

\vspace{1em}
\textbf{Q17. I am better able to identify students’ learning difficulties and adjust my teaching methods accordingly. \quad [Single-choice]}\\
Responses: Strongly agree: 2 (20\%), Agree: 8 (80\%), Neutral: 0 (0\%), Disagree: 0 (0\%), Strongly disagree: 0 (0\%) \\
Valid responses: 10

\vspace{1em}
\textbf{Q18. Please summarize your experience and reflections during multi-turn interactions with the five virtual students. You may comment on both strengths and weaknesses of the platform. \quad [Open-ended Question]}\\
\textbf{Example 1:} At first, I was unsure how to handle the high-neuroticism student. Later, I found that asking more frequent and empathetic questions improved the interaction. This made me more aware of the impact of teaching strategies on different personalities. \\
\textbf{Example 2:} The simulation platform helped me realize my tendency to lecture too much. When interacting with low-openness students, I had to slow down and repeat instructions, which challenged my patience and adaptability. \\
\textbf{Example 3:} I appreciated how different students responded uniquely. The platform clearly showed how students’ personalities influenced their language use and engagement. However, I did notice some repetitive answers from certain student types, which could be optimized in future iterations. \\
\textbf{Example 4:} This experience reminded me that teaching is not just about delivering content but also about observing subtle emotional cues. Even virtual students made me rethink my questioning style.
\end{tcolorbox}

\paragraph{Analysis.}

Participants expressed high satisfaction with the platform. All respondents agreed that the interface was intuitive and effectively conveyed personality traits. Critically, 90\% reported enhanced instructional adaptability, with all participants noting improvements in addressing diverse student needs, and 80\% citing stronger classroom interaction skills. These findings suggest that the LVSA framework can serve not only as a simulation platform, but also as a mechanism to facilitate pedagogical reflection and behavior transfer.

When asked about student types, High Extraversion (HE) and High Agreeableness (HA) were seen as easiest to teach, while Low Conscientiousness (LC) and Low Openness (LO) were considered the most challenging due to vague responses and inconsistent engagement. These perceptions align with our earlier interaction analysis and highlight the role of student personality in shaping teacher strategy. Interestingly, despite their challenge, LO and LC were still seen as realistic—60\% and 50\% of participants respectively believed these students resembled those in real classrooms, indicating both representativeness and pedagogical value.

In terms of instructional behavior, teachers most frequently used questioning (80\%) and lecturing (70\%), coupled with consistently positive emotional feedback—all participants used praise and encouragement. This aligns with the observed emotional support strategies seen in interactions with emotionally variable students, such as those high in neuroticism (HN). Notably, 70\% of teachers explicitly stated that they adjusted their instructional strategies in response to personality cues, reflecting an emerging capacity for real-time pedagogical adaptation.

Open-ended responses echoed these findings. Teachers shared that they gained greater sensitivity to emotional expression and began adjusting their pacing, questioning, and feedback strategies accordingly. Some also pointed out limitations in response variety or repetition, suggesting future improvements could focus on increasing behavioral richness and emotional nuance in LVSA modeling.

\clearpage

\section{Limitations and Ethical Considerations}
\label{app:limitations_and_ethics}

\subsection{Research Limitations and Future Directions}

While this study proposes a personality-grounded framework for constructing and evaluating virtual student agents (LVSAs), and validates it through multi-dimensional evaluation and real-teacher interaction experiments, several limitations remain to be addressed in future work.

First, although our experiments were grounded in a structured interaction setting with 10 pre-service teachers and 1,288 multi-turn dialogues across five personality types, the study was conducted within a controlled simulation environment. Future research should integrate LVSAs into authentic classroom contexts to assess their adaptability in real-world instructional flows, co-teaching scenarios, and long-term curriculum alignment.

Second, some personality types—especially Low Conscientiousness (LC)—exhibited limitations such as repetitive phrasing, vague semantics, and unstable dialogue consistency. While these shortcomings reveal challenges in fine-grained behavior control and style differentiation, they also highlight the value of educational interaction as a rigorous testbed for evaluating LLMs. Specifically, they demonstrate the need for improved controllability, consistency, and interpretability in LLM outputs under cognitively rich, open-ended settings.

Third, despite employing a multi-perspective evaluation framework—combining human and GPT-4 ratings across personality, question type, and instructional phase—the current setup does not yet test for generalization across teachers, tasks, and models. Future directions include multi-teacher co-evaluation, cross-subject transfer learning, and longitudinal strategy tracking to establish a simulation-to-practice feedback loop.

Finally, our current system focuses on textual interactions. Future work can incorporate multimodal signals (\eg, voice, gaze, gesture) to expand behavioral dimensions and improve teachers’ perception of emotional and cognitive cues in virtual students.

\subsection{Ethical Compliance and Data Privacy}

This study strictly adheres to institutional ethics review policies and academic integrity standards, ensuring participant safety, data protection, and legal compliance throughout the research process.

\textbf{Ethics Review Exemption:}  
The study is categorized as non-interventional behavioral research. It does not involve any sensitive personal information, medical/psychological intervention, or at-risk populations. All participants were adult volunteers ($>=18$ years old), and all tasks were conducted in simulated teaching environments with no associated physical or psychological risk. The study meets the institutional exemption criteria for ethics review.

\textbf{Data Collection and Processing:}  
All participant data were anonymized during collection by removing identifiers from names, free-text entries, and submission logs. No image, audio, geolocation, or device-level information was recorded. All scoring and annotation data were stored using structured, de-identified formats to ensure reproducibility and data security.

\textbf{Privacy and Usage Boundaries:}  
Human participant data are restricted to internal analysis only and will not be used for commercial purposes. All model-generated responses were produced by GPT-4 based on predefined task templates and do not contain real user content or personal data.

\subsection{Data Types and Availability}

All data used in this study will be released to the research community following standardized anonymization and formatting procedures, supporting reproducibility and open evaluation in AI4Edu.

(1) \textbf{Human Evaluation Data:} Includes teacher-generated ratings on LVSA outputs, interaction feedback, and behavioral annotations collected during the controlled experiments. These are used to assess behavioral consistency, perceived realism, and pedagogical adaptation.

(2) \textbf{Model-Generated Data:} Includes all GPT-4-generated outputs based on the structured instructional task template $T = \langle C, P, Q, L, \Pi \rangle$, covering simulated classroom dialogues, personality-aligned responses, and multi-turn interactions across phases and question types.

All datasets will undergo systematic anonymization, data cleaning, and formatting before public release. Each entry will be accompanied by metadata describing its instructional context (personality type, teaching stage, question type, model parameters, etc.) and will follow ethical and privacy guidelines.

\end{document}